\documentclass{arxiv}

\usepackage{enumitem}
\usepackage[T1]{fontenc}
\usepackage{lmodern}
\usepackage[sort&compress]{natbib}
\usepackage{amsfonts}
\usepackage{amsmath}
\usepackage{amssymb}
\usepackage{amsthm}
\usepackage{mathtools}
\usepackage{graphicx}
\usepackage{xcolor}
\usepackage{bm}
\usepackage{bbm}
\usepackage{dsfont}
\usepackage{float}
\usepackage{booktabs}
\usepackage{wrapfig}
\usepackage{makecell}
\usepackage{cancel}
\usepackage{url}
\usepackage{caption}
\usepackage[title]{appendix}
\usepackage{cuted}
\usepackage{algorithm}
\usepackage{algpseudocode}
\usepackage{multicol}

\newcommand{\A}{\mathbf{A}} 
\newcommand{\B}{\mathbf{B}} 
\newcommand{\Br}{\text{\raisebox{\depth}{\scalebox{1}[-1]{$\mathbf{P}$}}}}
\newcommand{\X}{\mathbf{X}} 
\newcommand{\Y}{\mathbf{Y}} 
\newcommand{\Z}{\mathbf{Z}}
\newcommand{\K}{\mathbf{K}}

\newcommand{\Lb}{\mathbf{L}}
\newcommand{\C}{\mathbf{C}} 
\newcommand{\U}{\mathbf{U}}
\newcommand{\W}{\mathbf{W}}
\newcommand{\V}{\mathbf{V}}
\newcommand{\I}{\mathbf{I}}
\newcommand{\Q}{\mathbf{Q}}

\newcommand{\ts}{\hspace*{0.1em}}

\DeclareMathAlphabet{\bbold}{U}{bbold}{m}{n}
\newcommand{\id}{\ensuremath{\bbold{1}}}

\numberwithin{equation}{section}

\theoremstyle{plain}
\newtheorem{theorem}{Theorem}[section]

\newtheorem{proposition}[theorem]{Proposition}
\newtheorem{corollary}[theorem]{Corollary}

\theoremstyle{definition}
\newtheorem{definition}[theorem]{Definition}

\theoremstyle{remark}
\newtheorem{remark}[theorem]{Remark}

\begin{document}

\title{Bayesian Transfer Operators\\in Reproducing Kernel Hilbert Spaces}

\author{%
  \parbox{\textwidth}{\centering
    \textbf{Septimus Boshoff$^{\,1}$\,\orcid{0009-0002-5817-3580}, 
            Stefan Klus$^{\,2}$\,\orcid{0000-0002-9672-3806}, 
            Sebastian Peitz$^{\,3}$\,\orcid{0000-0002-3389-793X}} \\[6pt]
    $^1$TU Dortmund, Germany \quad
    $^2$Heriot--Watt University, UK \quad 
    $^3$Paderborn University, Germany \\[6pt]
    septimus.boshoff@uni-paderborn.de \quad
    s.klus@hw.ac.uk \quad
    sebastian.peitz@uni-paderborn.de
  }
}


\begin{abstract}
Transfer operators provide a linear description of nonlinear dynamical systems by propagating observables or probability distributions. In reproducing kernel Hilbert spaces, these operators are represented through conditional mean embeddings, whose regularized estimator coincides with the posterior mean of Gaussian process regression. Building on this correspondence, we formulate Gaussian Process Dynamic Mode Decomposition, a Bayesian transfer-operator method that combines sparse variational Gaussian processes with kernel-based Koopman and Perron--Frobenius operator approximation. The resulting framework provides a principled route to uncertainty quantification, scalable sparse approximations, and data-adaptive dictionary learning for prediction and spectral analysis. Additionally, the model-selection formulation naturally gives rise to a forward--backward operator structure, which establishes a connection between Bayesian transfer-operator learning and time-lagged canonical correlation analysis.
\end{abstract}

\setcounter{tocdepth}{2}
\renewcommand{\contentsname}{ }
\tableofcontents

\section{Introduction}
\label{sec: Introduction}

Modeling and forecasting the behavior of noisy, high-dimensional, and nonlinear systems remains an active area of research. Yet, the governing laws of these systems are often unknown or prohibitively expensive to simulate. With the proliferation of measurement sensors and the advent of efficient computational hardware, our modern world is abundant with multi-fidelity data that enables us to construct models directly from data without relying solely on first principles. Beyond prediction, some algorithms can reveal interpretable spatio-temporal patterns and coherent structures that support the analysis of complex dynamical systems.

Transfer operator theory \citep{Klus2016Numerical, mauroy2020koopman, brunton2022modern} and \emph{reproducing kernel Hilbert spaces} (RKHSs) \citep{aronszajn1950theory, cressie1990origins, scholkopf2002learning, berlinet2011reproducing, shawe2004kernel, williams2006gaussian, hofmann2008kernel} are popular approaches that result in such analytically informative algorithms, while also contributing widely to practical applications \citep{rowley2009spectral, susuki2011nonlinear, susuki2011coherent, budivsic2012applied, berger2015estimation, netto2018robust, Bruder-RSS-2019}. Both Koopman operator and RKHS techniques frame their corresponding problems in a functional analytic setting. Therefore, it stands to reason that by combining methods from both of these two fields, we can design robust algorithms that provide compact and interpretable models for (stochastic) time-series.

\subsection{Transfer Operator Learning}

Transfer operators, such as \emph{Koopman} and \emph{Perron--Frobenius} \emph{operators}, provide an exact linear description of the evolution of observables or probability densities in an infinite-dimensional function space, rather than a local linearization of the state-space dynamics. The operator-theoretic perspective connects well-understood statistical and geometric views with insights into the physical behavior of dynamical systems \citep{Koopman1931hamiltonian, Koopman1932dynamical, lasota1994chaos, mezic2019spectrumKoopmanoperators}. 

The Koopman operator, being an infinite-dimensional object, must in general be empirically approximated before it can be implemented on a computer. This typically involves discretizing the operator and identifying a suitable set of basis functions that span a finite-dimensional subspace wherein the spectral properties of the operator can be represented --- a task that falls under the heading of \emph{dictionary learning} \citep{li2017extended, takeishi2017learning, wehmeyer2018time, yeung2019learning, azencot2020forecasting, eivazi2021recurrent, mardt2018vampnets}.

Finite-dimensional approximations of transfer operators are attractive because they reduce to linear algebra, namely a linear regression in the lifted observable space followed by an eigenvalue or singular value decomposition. After lifting the system to a suitable higher-dimensional observable space, such models describe nonlinear \citep{williams2015data, williams2015kernel} and stochastic systems \citep{klus2020data, wanner2022robust, vcrnjaric2020Koopman, colbrook2024beyond}. As a consequence, techniques developed for linear dynamical systems, such as spectral analysis, multi-step forecasting, numerical simulation, and control \citep{peitz2019Koopman, peitz2020data, proctor2016dynamic, otto2021Koopman}, can be applied to the learned lifted model of a nonlinear system.

One such suite of algorithms that discretizes the Koopman operator is \emph{dynamic mode decomposition} (DMD) \citep{rowley2009spectral, schmid2010dynamic}. DMD may be viewed as combining PCA spatial dimensionality reduction with a Fourier temporal decomposition \citep{brunton2022modern}. In other words, one may think of DMD as rotating the (lower-dimensional) PCA space such that each basis has dynamics oscillating according to a single frequency and a single growth (or decay) rate. \emph{Extended DMD} (EDMD) \citep{williams2015data} generalizes DMD from the state coordinates to a dictionary of nonlinear observables, and recovers DMD when the dictionary contains only the state coordinates. In the large-data limit, EDMD converges to a Galerkin approximation of the Koopman operator \citep{williams2015data, Klus2016Numerical, colbrook2024rigorous, colbrook2024multiverse}.

DMD relies on an eigendecomposition, which requires an operator that maps a function space into itself. The Koopman operator maps functions of the successor to functions of the predecessor. If each function is measured with the distribution of its own state, the operator acts between two Hilbert spaces, which coincide only for stationary data \citep{wu2020variational}. A singular value decomposition, by contrast, can accommodate two such spaces. The step from an eigendecomposition to a singular value decomposition mirrors the step from a Galerkin to a Petrov--Galerkin method, whose test functions differ from those that span the approximation space \citep{Klus2016Numerical}. Time-lagged canonical correlation analysis (TCCA) estimates this decomposition, and its dominant singular functions identify coherent sets \citep{wu2020variational, FROYLAND20131, klus2019kernel}.

\subsection{Bayesian Learning and Kernel Embeddings}

As in many kernel-based transfer operator methods \citep{williams2015kernel, klus2018kernel, degennaro2019scalable, fujii2019dynamic, das2020Koopman, das2021reproducing, ikeda2022Koopman, bevanda2025Koopman, klus2020eigendecompositions, baddoo2022kernel}, the underlying machinery of our transfer operator formulation is a reproducing kernel, which also serves as the covariance function of a Gaussian process. Kernel methods are \emph{non-parametric} and define a kernel-dependent hypothesis space whose geometry and regularity properties are determined by the chosen reproducing kernel. It is the non-parametric nature of kernel algorithms that grants them the intrinsic ability to implicitly construct a functional basis for a model directly from data. Thereby, they provide a natural choice for the hypothesis space: an RKHS. It is these properties that often lead to theoretical performance guarantees \citep{scholkopf2002learning, klus2020kernel, PHILIPP2024101657}.

Kernel mean embeddings extend reproducing kernels from functions to probability distributions. A distribution is represented by its mean element in an RKHS, and a conditional distribution by its conditional mean embedding \citep{Muandet20171, 10.1145/1553374.1553497}. Distances and dependencies between embeddings yield two-sample and independence tests \citep{Muandet20171}, and kernel counterparts of the sum, product, and Bayes' rules support message passing in graphical models, filtering in hidden Markov models, and causal discovery \citep{6530747, fukumizu2013kernel, Muandet20171}. 

Bayesian learning is a philosophy of model design that treats every unknown quantity as a random variable. A prior distribution encodes the assumptions about the unknown, and Bayes' rule conditions it on the data. For example, \emph{Gaussian process (GP) regression} applies this principle to an unknown function, with a reproducing kernel as the prior covariance function \citep{mackay1998introduction, williams2006gaussian, girard2002gaussian, wang2005gaussian, solak2002derivative, doi:10.1137/23M1560252}. Some of the advantages are that Bayesian models yield fully probabilistic predictions from a limited set of noisy targets, and that the model's marginal likelihood provides a well-founded criterion for model selection. We apply the same Bayesian treatment to kernel mean embeddings and transfer operators.

\subsection{Contributions of this work} \label{sec: Contributions of this work}

Learning a transfer operator from data raises three natural questions: how to estimate the operator from finitely many transitions, how to quantify our trust in the model, and how to choose the function space in which the operator is represented. We answer all three questions within a single Bayesian model. We treat the \emph{embedded Perron--Frobenius operator} \citep{klus2020eigendecompositions} as a random variable under a GP prior and show that its posterior mean recovers kernel EDMD \citep{williams2015kernel, williams2015data}. The resulting posterior answers the second question by separating epistemic from aleatoric uncertainty, which we propagate into multi-step state predictions. We also place an inverse-Wishart process prior on the covariance geometry of the embedded successor states given their predecessors. Its posterior mean yields Bayesian versions of the residuals that Residual DMD uses to flag spurious Koopman eigenpairs \citep{colbrook2023residual, colbrook2024beyond}. Bayesian model selection answers the third question. The data-fit term of its hyperparameter objective recovers the VAMP-$2$ score \citep{wu2020variational}. A well-established variational principle for Koopman representations thereby acquires a sound Bayesian foundation.

The remainder of this manuscript is structured as follows. Section~\ref{sec: Problem Formulation} fixes the dynamical setting and the modeling assumptions on the flow map and its lifted representation. Section~\ref{sec: Sparse GP DMD} develops a Bayesian formulation of the embedded Perron--Frobenius operator. It then addresses uncertainty quantification, forecasting, and model selection, and closes with the Bayesian round-trip operators. Section~\ref{sec: Numerical Experiments} demonstrates the resulting algorithms qualitatively on five benchmark systems that span time-homogeneous, reversible, and time-inhomogeneous dynamics. 

For ease of reference, in Appendix~\ref{sec: Gaussian Process Regression} we review (sparse) GP regression and the notation of \citet{williams2006gaussian}, and for completeness we also present an overview of (kernel) transfer operators \citep{klus2020eigendecompositions} and RKHS probability embeddings in Appendices~\ref{sec: Transfer Operator Theory}~and~\ref{sec: Embedding Probability Distributions}.

\section{Problem Formulation} \label{sec: Problem Formulation}

Before constructing the model, we fix the setting in which it operates. Two representations of a dynamical system run through the entire development: a flow map on the state space, and a transfer operator on an RKHS. For each we state the assumptions that hold throughout, and hence the basic limitations that the model inherits. Additional assumptions that qualify only particular readings of the model are stated where those readings are introduced. For example, the assumptions needed for spectral analysis, Bayesian consistency, and multi-step propagation are presented in Sections~\ref{sec: Koopman Mode Decomposition}, \ref{sec: Bayesian Consistency}, and~\ref{sec: Prediction Theory}.

Each representation is modeled in both directions of time, which gives four models, two on the state space and two on the RKHS. The forward models predict the successor from its predecessor, and the backward models predict the predecessor from its successor. At the population level, the two directions are coupled through the joint law of a transition pair. The forward and the backward transition law are the two conditional laws of one joint law of a transition pair. Once the density $p_X$ of the initial state is fixed, and with $p_Y$ the density of its successor, Bayes' rule turns the forward transition law into the backward transition law,
\begin{alignat}{3} \label{eq: Bayes transition laws}
    p_X(\bm{x}) \, \overrightarrow{p}_{\Delta t}(\bm{y} \mid \bm{x})
    &=
    p_{XY}(\bm{x}, \, \bm{y}) 
    &=
    \overleftarrow{p}_{\Delta t}(\bm{x} \mid \bm{y}) \, p_Y(\bm{y}).
\end{alignat}
In operator form, the backward conditional expectation is the adjoint of the forward conditional expectation with respect to the two endpoint laws, as~\eqref{eq: forward-backward pairing} in Appendix~\ref{subsec: Forward-backward Operators} shows. The construction of the four models and the mechanisms coupling them are developed throughout the manuscript.

\vspace{0.25\baselineskip}
\noindent\begin{minipage}{\columnwidth}
    \centering
    \includegraphics[width=0.9\columnwidth]{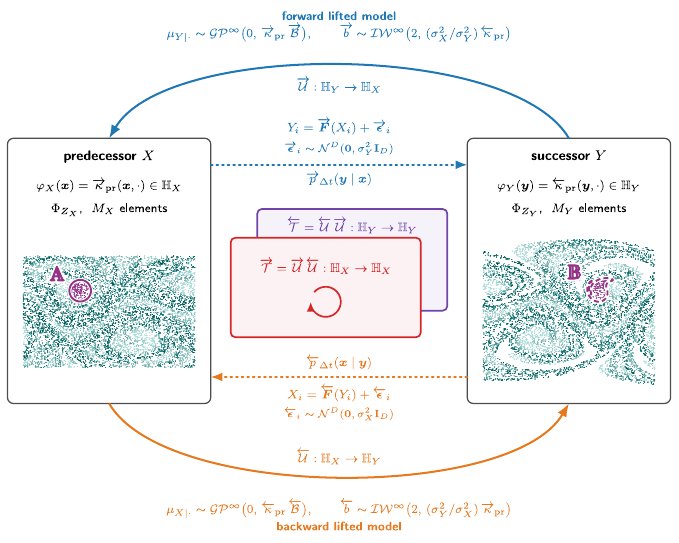}
    \captionsetup{hypcap=false}
    \phantomsection
        \captionof{figure}{\emph{Four coupled models and the Bayesian round trip}. The dotted arrows carry the forward and backward state-space models, the solid arcs the lifted Koopman operators $\overrightarrow{\mathcal{U}}$ and $\overleftarrow{\mathcal{U}}$, and the point clouds illustrate a coherent pair $( \mathbb{A} , \mathbb{B} )$.}
    \label{fig: GP-TCCA mechanism}
    \vspace{0.25\baselineskip}
\end{minipage}

Figure~\ref{fig: GP-TCCA mechanism} shows the four models and how they are coupled. Each state-space model consists of a flow map with Gaussian residuals and the transition law that it represents. Each lifted Koopman operator, $\overrightarrow{\mathcal{U}} : \mathbb{H}_Y \to \mathbb{H}_X$ and $\overleftarrow{\mathcal{U}} : \mathbb{H}_X \to \mathbb{H}_Y$, pulls a function back through its transition, i.e., it returns the expected value of the function at the later state as a function of the earlier state. It therefore maps in the opposite direction to its adjoint, the embedded Perron--Frobenius operator, which pushes the embedded state forward in time. The composition $\overrightarrow{\mathcal{T}} = \overrightarrow{\mathcal{U}} \, \overleftarrow{\mathcal{U}}$ is the round trip on $\mathbb{H}_X$. For a coherent pair $( \mathbb{A} , \mathbb{B} )$, most of the probability mass in $\mathbb{A}$ is transported into $\mathbb{B}$, and the forward--backward chain $X \rightarrow Y \rightarrow X'$ then returns $X' \in \mathbb{A}$ with high probability for $X \in \mathbb{A}$.

\begin{remark}
    An arrow marks an object of one direction, as in $\overrightarrow{\bm{F}}$ and $\overleftarrow{\bm{F}}$, and a subscript $X$ or $Y$ an object of one endpoint, as in $\Z_X$ and $\Z_Y$. An object without either mark is common to both directions. Each backward object follows from its forward counterpart by the exchange $X \leftrightarrow Y$ and the reversal of the arrow, and quantities derived from an arrowed object, such as its eigenvalues, inherit its direction.
\end{remark}

\subsection{Sampling and the (Forward) Flow Map} \label{sec: Sampling and the Flow Map}

Throughout the manuscript we assume that the dynamics are observed in discrete time at a fixed lag $\Delta t$, as one-step pairs $(\bm{x}_i,\bm{y}_i)$ drawn from a time-homogeneous Markov transition law~\footnote{The continuous-time and random-dynamical-system constructions that produce this law are collected in Appendix~\ref{sec: Transfer Operator Theory}.} $\overrightarrow{p}_{\Delta t}(\cdot \mid \cdot)$ on $\mathbb{X} \subseteq \mathbb{R}^D$. We collect the $N$ training pairs in the snapshot matrices $\X \coloneqq [\bm{x}_1, \dots, \bm{x}_N]$ and $\Y \coloneqq [\bm{y}_1, \dots, \bm{y}_N]$.

We further assume an \emph{isotropic} additive Gaussian residual model for the one-step transition,
\[
    Y_i = \overrightarrow{\bm F}(X_i) + \overrightarrow{\bm{\epsilon}}_i,
    \qquad
    \overrightarrow{\bm{\epsilon}}_i\sim\mathcal N^D(\bm 0,\sigma_Y^2\I_D),
\]
where the residuals are conditionally independent across transitions and $\sigma_Y^2$ denotes the per-channel residual variance. Note that the residual acts on the successor only, so the predecessor states $\bm{x}_i$ are treated as noise-free. In Section~\ref{sec: Prediction Theory} we separate the variance $\sigma_Y^2$ into sensor noise and transition variability, which allows both stochastic and deterministic dynamics to be treated.

The main development below uses the additive Gaussian residual model above, with $\overrightarrow{\bm F} : \mathbb{X} \rightarrow \mathbb{X}$ interpreted as the conditional mean of the one-step transition. This leads directly to a homoskedastic GP regression model over the state space $\mathbb X$. For the sake of clarity, allow a temporary simplification and consider the $D\! =\! 1$ case. Let $\overrightarrow{\kappa}_\textrm{pr}, \, \overrightarrow{\kappa}_\textrm{pst}  : \mathbb{X} \times \mathbb{X} \rightarrow \mathbb{R}$ denote the scalar covariance functions defining the GP prior and posterior over this one-dimensional flow map $\overrightarrow{\bm F}$. We take $\overrightarrow{\kappa}_\textrm{pr}$ to be strictly positive definite, i.e., $\overrightarrow{\kappa}_\textrm{pr}(\Z, \Z) \succ 0$ for every finite set $\Z$ of distinct points.

\subsection{Transfer Operators in Reproducing Kernel Hilbert Spaces} \label{sec: Kernel Transfer Operators}

For notational simplicity, the lifted representations reuse the prior covariance functions of the two flow maps as reproducing kernels. Put differently, a lifted representation maps the same one-step pairs into the reproducing kernel Hilbert spaces $\mathbb{H}_X$ and $\mathbb{H}_Y$ with the reproducing kernels $\overrightarrow{\kappa}_\textrm{pr}$ and $\overleftarrow{\kappa}_\textrm{pr}$ and the canonical feature maps $\varphi_X(\bm{x}) = \overrightarrow{\kappa}_\textrm{pr}(\bm{x},\cdot) \in \mathbb{H}_X$ and $\varphi_Y(\bm{y}) = \overleftarrow{\kappa}_\textrm{pr}(\bm{y},\cdot) \in \mathbb{H}_Y$. The associated second-moment operators $\mathcal{C}_{XX} = \mathbb{E}_X[\varphi_X(X) \otimes \varphi_X(X)]$ and $\mathcal{C}_{YX} = \mathbb{E}_{YX}[\varphi_Y(Y) \otimes \varphi_X(X)]$, together with $\mathcal{C}_{XY} = \mathcal{C}_{YX}^{*}$, encode the one-step predecessor--successor relation in the lifted spaces (see Appendix~\ref{sec: Embedding Probability Distributions}).

Our model is formulated in terms of the \emph{embedded} Perron--Frobenius operator \citep{klus2020eigendecompositions}
\[
\overrightarrow{\mathcal{P}}_\varepsilon
=
\mathcal{C}_{YX}
\left(
\mathcal{C}_{XX}
+
\varepsilon \, \mathcal{I}
\right)^{-1}
:
\mathbb{H}_X
\rightarrow
\mathbb{H}_Y.
\]
This operator is the Tikhonov-regularized counterpart of the conditional mean embedding operator $\mathcal{C}_{Y \mid X}$ introduced in Definition~\ref{def: cme}. In this formulation, probability distributions are represented through their kernel mean embeddings, so $\overrightarrow{\mathcal{P}}_\varepsilon$ acts on elements of the RKHSs and not directly on probability densities.

At the Dirac level, the corresponding source and target embeddings are $\varphi_X(\bm{x}_i)$ and $\varphi_Y(\bm{y}_i)$, respectively. Then the conditional mean embedding (CME) of the successor distribution is
\[
    \mu_{Y\mid\bm{x}_i}
    \coloneqq
    \mathbb{E}\!\left[\varphi_Y(Y_i)\mid X_i = \bm{x}_i\right] \approx \overrightarrow{\mathcal{P}}_\varepsilon \, \varphi_X(\bm{x}_i).
\]

Since $\mathcal{C}_{XX}$ is self-adjoint, the adjoint of $\overrightarrow{\mathcal{P}}_\varepsilon$ with respect to the inner products of $\mathbb{H}_X$ and $\mathbb{H}_Y$ is
\begin{alignat*}{3}
    \overrightarrow{\mathcal{P}}_\varepsilon^{*}
    &=
    \left(\mathcal{C}_{XX} + \varepsilon \mathcal{I}\right)^{-1}
    \mathcal{C}_{XY}
    =
    \overrightarrow{\mathcal{U}}_\kappa,
\end{alignat*}
so the embedded Perron--Frobenius operator (PFO) is the adjoint of the \emph{kernel} Koopman operator $\overrightarrow{\mathcal{U}}_\kappa$ of \citet{klus2020eigendecompositions}. Both operators are bounded for any bounded measurable kernel and $\varepsilon > 0$. When the range condition of Proposition~\ref{prop: CME} holds, the Tikhonov-regularized operator recovers the conditional mean embedding (CME) in the limit $\varepsilon \rightarrow 0$.

\section{Bayesian Transfer Operators} \label{sec: Sparse GP DMD}

With the setting fixed, we now construct the model. Because kernel-based DMD algorithms often quickly become impractical for large data sets due to the high training cost, one of our goals is to apply the variational free energy (VFE) method \citep{titsias2009variational} to kernel transfer operators \citep{klus2020eigendecompositions}. For example, in \citet{klus2020eigendecompositions}, to estimate the Koopman matrix we need to take the inverse of a Gramian which has a complexity of $\mathcal{O}(N^3)$. Therefore, we are motivated to find a principled trade-off between the model's generalization error and its computational complexity. In other words, our goal is a non-parametric model that automatically adapts to the complexity of the dynamics, but not to the point where all the training inputs define the basis dictionary.

We will demonstrate how our algorithm is a special case of EDMD \citep{williams2015data}. This means that it inherits some of EDMD's limitations \citep{colbrook2019compute, colbrook2023residual}. This should not be surprising, since it is quite common to derive the same expressions within either a frequentist or Bayesian framework. This duality between ridge-regression and Bayesian models is well-known, and has been pointed out in \citet{hastie2009elements, poggio1990networks, wahba1990spline, williams2006gaussian}. One might then assume that the choice of framework merely reflects an attitude, with each attitude corresponding to a preferred mode of reasoning about how to construct a (data-driven) model. However, we could point out that attitudes matter because they suggest different approaches to improving upon the state-of-the-art.

For example, EDMD does not deliver uncertainty-aware predictions, nor does it account for measurement noise from the outset. Our unification of DMD and GP regression wraps up the entire model inference pipeline into one coherent theory that simultaneously addresses hyperparameter optimization and sparse dictionary learning.

\subsection{Operators as Random Variables} \label{sec: Operators as Random Variables}

\begin{wraptable}{r}{0.5\columnwidth}
    \vspace{-1.5\baselineskip}  
  \centering
  \renewcommand{\arraystretch}{1.1}
  \setlength{\tabcolsep}{4pt}
  \caption{\emph{Nomenclature}. The symbols for some of the explicit and implicit data structures.}
  \label{tab:notation}

  \resizebox{\linewidth}{!}{%
  \begin{tabular}{@{}l l@{}}
    \toprule
    Variable & Symbol \\
    \midrule

    \makecell[l]{Pseudo-inputs}  & $\Z \coloneqq [\bm{z}_1, ..., \bm{z}_{M}]$ \\
    \makecell[l]{Snapshot matrices} & $\X \coloneqq [\bm{x}_1, ..., \bm{x}_N]$ \\
                                    & $\Y \coloneqq [\bm{y}_1, \dots, \bm{y}_N]$ \\
    \makecell[l]{Dictionaries} & $\Phi_{Z} \coloneqq [ \varphi_X(\bm{z}_1), ...,  \varphi_X(\bm{z}_{M}) ]$ \\
                               & $\Phi_X \coloneqq [ \varphi_X(\bm{x}_1), ...,  \varphi_X(\bm{x}_N) ]$ \\
    \makecell[l]{Covariance functions} & $\overrightarrow{\kappa}_\textrm{pr}(\bm{x}, \bm{x}') \coloneqq \langle \varphi_X(\bm{x}), \varphi_X(\bm{x}')\rangle_{\mathbb{H}_X}$ \\
                                       & $\overrightarrow{\xi}_\textrm{pr}(\bm{x}, \bm{x}') = \overrightarrow{\kappa}_\textrm{pr}(\bm{x}, \bm{x}') \, \overrightarrow{\mathcal{B}}$ \\
    \makecell[l]{Covariance matrices} & $\K_{ZZ} \coloneqq \overrightarrow{\kappa}_\textrm{pr}(\Z_X, \Z_X)$ \\
                                      & $\K_{ZX} \coloneqq \overrightarrow{\kappa}_\textrm{pr}(\Z_X, \X)$ \\
    \makecell[l]{Covariance operators} & $\Xi_{XX} \coloneqq \overrightarrow{\kappa}_\textrm{pr}(\X, \X) \otimes \overrightarrow{\mathcal{B}}$ \\
                                       & $\mathcal{C}_{YX} \coloneqq \mathbb{E}_{YX}\left[\varphi_Y(Y) \otimes \varphi_X(X)\right]$ \\
    \makecell[l]{Feature vectors} & $\mathbf{k}_{Z}(\bm{x}) \coloneqq \Phi_{Z_X}^\ast \varphi_X(\bm{x})$ \\
                                  & $\mathbf{\xi}_{X}(\bm{x}) \coloneqq \mathbf{k}_{X}(\bm{x}) \otimes \overrightarrow{\mathcal{B}}$ \\
    \makecell[l]{Mass matrix} & $\C_{XX} \coloneqq \K_{ZX}\, \K_{ZX}^\top$ \\
    \makecell[l]{Stiffness matrix} & $\C_{XY} \coloneqq \K_{ZX}\, \K_{ZY}^\top$ \\
    \makecell[l]{Regularization} & $\tilde{\K}_{XX} \coloneqq {\K}_{XX} + \sigma_{Y}^2 \I_N$ \\
                                 & $\widetilde{\C}_{XX} \coloneqq {\C}_{XX} + \sigma_Y^2 \widetilde{\K}_{ZZ}$ \\
    \bottomrule
  \end{tabular}
  }
\end{wraptable}

In Appendix~\ref{sec: Gaussian Process Regression}, we review GP regression with outputs in $\mathbb{R}$. We now extend this setting by taking the target space to be the RKHS $\mathbb{H}_Y$ of the successors. This is the proverbial ``lifting of the dynamical system'', and presents a challenge of handling not just finitely many outputs, but an infinite number of correlated output channels. To address this, we employ an \emph{operator-valued} kernel, and subsequently derive the posterior GP associated with the embedded PFO.

\begin{definition}\emph{Lifted Observation Model}. \label{def: lifted observation model}
    For a fixed source state $X=\bm{x}$, let $Y$ denote a successor drawn from the corresponding conditional transition law, with conditional mean embedding (CME) $\mu_{Y\mid\bm{x}} \in \mathbb{H}_Y$, which the model represents by $\overrightarrow{\mathcal{P}}_\varepsilon\varphi_X(\bm{x})$. We assume that $\mathbb{H}_X$ and $\mathbb{H}_Y$ are separable and that their kernels are bounded, so that the covariance operator $\overrightarrow{\mathcal{B}}$ introduced below is positive, self-adjoint, and trace class (Proposition~\ref{prop: residual geometry}). We then adopt a homoskedastic working Gaussian likelihood for the sampled lifted target $\varphi_Y(Y)$ conditional on $X=\bm{x}$:
    \begin{alignat*}{3}
        \varphi_Y(Y)\mid X=\bm{x}
        &\approx
        \mu_{Y\mid\bm{x}} + \overrightarrow{\nu},
        \enspace
        \overrightarrow{\nu}
        \sim
        \mathcal{N}_{\mathbb{H}_Y}\!\left(
            0,\,
            \sigma_Y^2 \overrightarrow{\mathcal{B}}
        \right).
    \end{alignat*}
    Intuitively, the covariance operator in the lifted observation model should measure how much of the successor cannot be predicted from the source state. More precisely, for every observable $g \in \mathbb{H}_Y$, we predict $g(Y)$ using functions $h(X) = \langle h, \varphi_X(X) \rangle_{\mathbb{H}_X}$ of the lifted source and take the smallest achievable mean-squared error as the variance in the direction $g$. These residuals define the global conditional covariance operator $\mathcal{C}_{YY \mid X}$ of Appendix~\ref{sec: Output Covariance Geometry}, which is in accordance with covariance-based constructions of separable operator-valued kernels \citep{pmlr-v28-kadri13}.\footnote{Under the range condition of Proposition~\ref{prop: CME}, $\mathcal{C}_{YY \mid X}$ is the uncentered analogue of the conditional covariance operator of \citet{fukumizu2004dimensionality}, defined through centered covariance operators. The pointwise conditional covariance $\mathcal{C}_{YY \mid \bm{x}}$ of Appendix~\ref{subsec: Local Conditional Covariance} specializes the local conditional cross-covariance embedding of \citet{10.1145/1553374.1553497}.} By factoring out the state-space residual variance $\sigma_Y^2$, we define the normalized Bayesian-consistency covariance operator
    \begin{alignat}{3} \label{eq: Bayesian consistency covariance}
        \overrightarrow{\mathcal{B}}
        &\coloneqq
        \sigma_Y^{-2} \, \mathcal{C}_{YY \mid X},
        \qquad
        \big\langle g, \, \mathcal{C}_{YY \mid X} \, g \big\rangle_{\mathbb{H}_Y}
        =
        \inf_{h \in \mathbb{H}_X}
        \mathbb{E}\big[ \big( g(Y) - h(X) \big)^2 \big], \; \forall g \in \mathbb{H}_Y.
    \end{alignat}
    Under the range condition of Proposition~\ref{prop: CME}, the infimum is attained by the conditional expectation $h(\bm{x}) = \langle g, \mu_{Y \mid \bm{x}} \rangle_{\mathbb{H}_Y}$, and $\overrightarrow{\mathcal{B}}$ reduces to the Schur complement $\sigma_Y^{-2} ( \mathcal{C}_{YY} - \mathcal{C}_{YX} \mathcal{C}_{XX}^{-1} \mathcal{C}_{XY} )$. In this case, $\mathcal{C}_{YY \mid X}$ equals the average $\mathbb{E}_{X}[\mathcal{C}_{YY \mid \bm{x}}]$ of the pointwise conditional covariances. However, in general, $\mathcal{C}_{YY \mid X}$ exceeds this average by the second moment of the part of the conditional mean that $\mathbb{H}_X$ cannot approximate (Proposition~\ref{prop: residual geometry}).
\end{definition}

The covariance $\overrightarrow{\mathcal{B}}$ provides the \emph{geometry} needed to relate uncertainty in the lifted RKHS representation back to uncertainty in the original state space. Note that the Gaussianity and point-independent covariance in Definition~\ref{def: lifted observation model} are modeling assumptions: after the nonlinear feature transformation, the true conditional law of $\varphi_Y(Y)$ is generally neither Gaussian nor homoskedastic. The working likelihood therefore retains only second-order information, since $\mathcal{C}_{YY \mid X}$ depends only on the second moments of $(\varphi_Y(Y), \varphi_X(X))$ and is exactly the conditional covariance, independent of the conditioning value, of a zero-mean Gaussian pair with the same second moments (see Appendix~\ref{sec: Output Covariance Geometry}). Whether the covariance of the reconstructed state in $\mathbb{X}$ then matches the covariance induced by the lifted prediction is what we term the \emph{Bayesian-consistency principle} (see Section~\ref{sec: Bayesian Consistency}).

Similar to scalar-valued GP regression, we proceed by placing a GP on the Hilbert-valued function $\mu_{Y \mid \cdot}: \mathbb{X} \rightarrow \mathbb{H}_Y$, conditional on the Bayesian-consistent geometry, $\mu_{Y \mid \cdot} \sim \mathcal{GP}^\infty \left(\overrightarrow{\mathcal{M}}_\textrm{pr}, \, \overrightarrow{\xi}_\textrm{pr} \right)$\footnote{The superscript attached to $\mathcal{GP}$ indicates the dimensionality of the random object associated with the Gaussian process.}. However, here the prior covariance $\overrightarrow{\xi}_\textrm{pr} \colon \mathbb{X} \times \mathbb{X} \to \mathfrak{B}(\mathbb{H}_Y)$ is an \emph{operator-valued} kernel, where $\mathfrak{B}(\mathbb{H}_Y)$ denotes the space of bounded linear operators on $\mathbb{H}_Y$ \citep{pmlr-v28-kadri13,JMLR:v17:11-315}. For simplicity, we choose the prior mean function $\overrightarrow{\mathcal{M}}_\textrm{pr}:\mathbb{X} \rightarrow \mathbb{H}_Y$ to be identically zero. It follows that the prior distribution over the latent evaluations $\mu_{Y \mid \X} \in \mathbb{H}_Y^{1 \times N}$, which the model represents by $\overrightarrow{\mathcal{P}}_\varepsilon \Phi_X$, is
\begin{alignat*}{3}
    p_\textrm{pr}(\mu_{Y \mid \X}) =\mathcal{GP}^N\left(\mu_{Y \mid \X}; \overrightarrow{\mathcal{M}}_\textrm{pr}(\X), \Xi_{XX} \right),
\end{alignat*}
where $\Xi_{XX}$ is the prior covariance evaluated at $\left(\X, \, \X \right)$ and forms an $N \times N$ block operator on $\mathbb{H}_Y^{\oplus N}$.

The \emph{Riesz--Fréchet representation theorem} \citep{hsing2015theoretical, frechet1904operations}, applied to the bounded evaluation functional, guarantees that for each $\bm{x}\! \in\! \mathbb{X}$ and $g \! \in\! \mathbb{H}_Y$ there exists a unique element $\Xi(\bm{x})g\! : \mathbb{X} \rightarrow \mathbb{H}_Y$ that satisfies the reproducing property
\begin{alignat*}{3}
    \langle \mu_{Y \mid \cdot}, \, \Xi(\bm{x})g \rangle_{\mathbb{H}_Y^{\mathbb{X}}} = \langle \mu_{Y \mid \bm{x}}, \,g\rangle_{\mathbb{H}_Y}. 
\end{alignat*}
This implies that the feature map $\Xi(\bm{x})$ is a bounded linear operator, $\Xi(\bm{x}) \in \mathfrak{B}(\mathbb{H}_Y, \mathbb{H}_Y^{\mathbb{X}})$, where $\mathbb{H}_Y^{\mathbb{X}}$ denotes the $\mathbb{H}_Y$-valued RKHS on $\mathbb{X}$ (i.e., $\mu_{Y \mid \cdot} \in \mathbb{H}_Y^{\mathbb{X}}$). Thereby, the relationship between the reproducing kernel $\overrightarrow{\xi}_\textrm{pr}(\cdot, \cdot)$ and its associated feature map $\Xi(\bm{x})$ is
\begin{alignat*}{3}
    \langle \Xi(\bm{x}) \, g, \, \Xi(\bm{x}') \, g'\rangle_{\mathbb{H}_Y^{\mathbb{X}}} = \langle g , \,  \overrightarrow{\xi}_\textrm{pr}(\bm{x}, \bm{x}') \, g'\rangle_{\mathbb{H}_Y}.
\end{alignat*}
Evaluating the embeddings at successor states, the prior covariance between two such evaluations follows from the operator-valued kernel,
\begin{alignat*}{3}
    \mu_{Y \mid \bm{x}}(\bm{y})
    &=
    \langle \varphi_Y(\bm{y}), \, \mu_{Y \mid \bm{x}} \rangle_{\mathbb{H}_Y},
    \\
    \operatorname{Cov} \big[ \mu_{Y \mid \bm{x}}(\bm{y}), \, \mu_{Y \mid \bm{x}'}(\bm{y}') \big]
    &=
    \big\langle \varphi_Y(\bm{y}), \, \overrightarrow{\xi}_\textrm{pr}(\bm{x}, \bm{x}') \, \varphi_Y(\bm{y}') \big\rangle_{\mathbb{H}_Y}.
\end{alignat*}
The operator-valued kernel $\overrightarrow{\xi}_\textrm{pr}$ therefore determines how strongly the responses at $(\bm{x}, \bm{y})$ and $(\bm{x}', \bm{y}')$ are correlated under the chosen feature-space geometry.

To make the prior explicit, we use the Bayesian-consistency geometry introduced in Definition~\ref{def: lifted observation model} as the output-space factor of a separable operator-valued kernel \citep{pmlr-v28-kadri13},
\[
    \overrightarrow{\xi}_\textrm{pr}(\bm{x},\bm{x}')
    =
    \overrightarrow{\kappa}_\textrm{pr}(\bm{x},\bm{x}')\,\overrightarrow{\mathcal{B}}.
\]
At the population level, $\overrightarrow{\mathcal{B}}$ is a fixed property of the joint law of the transitions and the feature map, determined through Definition~\ref{def: lifted observation model}, and its relation to the state-space model is clarified in Section~\ref{sec: Bayesian Consistency}. At this stage, the separable structure allows the covariance operators to be written as Kronecker products between scalar input-space covariances and $\overrightarrow{\mathcal{B}}$, for example,
\[
    \Xi_{XX}
    \coloneqq
    \K_{XX}\otimes\overrightarrow{\mathcal{B}},
    \qquad \textrm{where }
    \K_{XX}
    \coloneqq
    \Phi_X^\ast \Phi_X
    \in\mathbb R^{N\times N}.
\]
Similarly, the residual covariance operator is
\[
    \Sigma
    \coloneqq
    \I_N\otimes\sigma_Y^2\overrightarrow{\mathcal{B}},
\]
and the corresponding operator-valued and scalar kernel feature vectors are
\[
    {\xi}_X(\bm{x})
    \coloneqq
    \mathbf{k}_X(\bm{x})\otimes\overrightarrow{\mathcal{B}},
    \qquad
    \mathbf{k}_X(\bm{x})
    \coloneqq
    \Phi_X^\ast\varphi_X(\bm{x}).
\]
Thus, the construction is an instance of the intrinsic coregionalization model \citep{alvarez2012kernels}.

Since the population geometry $\overrightarrow{\mathcal{B}}$ is unknown in practice, it has to be learned from the same transitions that are used to learn the transfer operator. We therefore treat the output covariance as random as well. Only finitely many evaluations of the lifted targets enter the model, so we describe the output covariance through its covariance function $\overrightarrow{b} \colon \mathbb{X} \times \mathbb{X} \rightarrow \mathbb{R}$, whose population value is $\langle \varphi_Y(\bm{y}), \overrightarrow{\mathcal{B}} \, \varphi_Y(\bm{y}') \rangle_{\mathbb{H}_Y}$. Conditional on $\overrightarrow{b}$, the prior covariance between $\mu_{Y \mid \bm{x}}(\bm{y})$ and $\mu_{Y \mid \bm{x}'}(\bm{y}')$ is therefore $\overrightarrow{\kappa}_\textrm{pr}(\bm{x}, \bm{x}') \, \overrightarrow{b}(\bm{y}, \bm{y}')$, and the residual evaluations $\overrightarrow{\nu}(\bm{y})$ and $\overrightarrow{\nu}(\bm{y}')$ have the covariance $\sigma_Y^2 \, \overrightarrow{b}(\bm{y}, \bm{y}')$. Accordingly, the operator expressions of this section are read at finite target-side sets $\Z_Y \subset \mathbb{X}$ of distinct evaluation points, where the separable kernel turns every output factor $\overrightarrow{\mathcal{B}}$ into the matrix $\overrightarrow{\B} \coloneqq \overrightarrow{b}(\Z_Y, \Z_Y)$.

We place an inverse-Wishart process prior \citep{shah2014student} on $\overrightarrow{b}$, written $\overrightarrow{b} \sim \mathcal{IW}^\infty ( 2 , \, \overrightarrow{b}_\textrm{pr} )$, with index $2$ and base kernel $\overrightarrow{b}_\textrm{pr} \coloneqq (\sigma_X^2 / \sigma_Y^2) \, \overleftarrow{\kappa}_\textrm{pr}$. The base kernel is built on the residual variance $\sigma_X^2$ and the prior covariance function $\overleftarrow{\kappa}_\textrm{pr}$ of the backward flow map. Any positive base kernel would give a valid prior, but only this particular choice relates the lifted models by Bayes' rule (Section~\ref{sec: Bayesian Forward-Backward Operators}) in the same way as the two transition laws~\eqref{eq: Bayes transition laws}. This means that, for every set $\Z_Y$, the matrix $\overrightarrow{\B}$ is inverse-Wishart distributed with index $2$ and scale matrix $\overrightarrow{b}_\textrm{pr}(\Z_Y, \Z_Y)$. With the index of \citet{dawid1981some}, every principal sub-block of an inverse-Wishart matrix is again inverse-Wishart distributed with the same index, so these marginals are consistent and define a single process.

Collecting the features $\varphi_Y(\bm{z})$ of the points $\bm{z} \in \Z_Y$ in $\Phi_{Z_Y}$ gives $\overrightarrow{b}_\textrm{pr}(\Z_Y, \Z_Y) \! = \!(\sigma_X^2 / \sigma_Y^2) {\K}_{ZZ}$,  where ${\K}_{ZZ} = \Phi_{Z_Y}^\ast \Phi_{Z_Y}$ is the target-side pseudo-covariance matrix. The base kernel is therefore the finite-evaluation counterpart of the target-side Tikhonov term $\sigma_X^2 \mathcal{I}$ on $\mathbb{H}_Y$, normalized by $\sigma_Y^2$ as in Definition~\ref{def: lifted observation model}. To obtain the posterior of $\overrightarrow{b}$, we evaluate the lifted targets at $\Z_Y$. By Definition~\ref{def: lifted observation model}, the evaluations $\K_{ZY} \coloneqq \Phi_{Z_Y}^\ast \Phi_Y$ follow the finite observation model
\[
    \operatorname{vec} \K_{ZY} \mid \mu_{Y \mid \X} , \overrightarrow{b}
    \sim
    \mathcal{N} \big( \operatorname{vec} ( \Phi_{Z_Y}^\ast \mu_{Y \mid \X} ) , \, \I_N \otimes \sigma_Y^2 \overrightarrow{\B} \big) ,
\]
while the prior gives $\operatorname{vec} ( \Phi_{Z_Y}^\ast \mu_{Y \mid \X} ) \mid \overrightarrow{b} \sim \mathcal{N} ( \bm{0} , \, \K_{XX} \otimes \overrightarrow{\B} )$. Integrating out the embedded PFO adds the two covariances, and their common factor $\overrightarrow{\B}$ keeps the sum separable,
\[
    \operatorname{vec} \K_{ZY} \mid \overrightarrow{b}
    \sim
    \mathcal{N} \big( \bm{0} , \, \tilde{\K}_{XX} \otimes \overrightarrow{\B} \big) ,
\]
where $\tilde{\K}_{XX} \coloneqq \K_{XX} + \sigma_Y^2 \I_N$ correlates the transitions and $\overrightarrow{\B}$ correlates the evaluation points. As a function of $\overrightarrow{\B}$, this likelihood is proportional to $| \overrightarrow{\B} |^{-N/2} \exp ( - \tfrac{1}{2} \operatorname{Tr} ( \overrightarrow{\B}^{-1} \K_{ZY} \tilde{\K}_{XX}^{-1} \K_{ZY}^\top ) )$. It depends on the data only through the \emph{scatter matrix} $\K_{ZY} \tilde{\K}_{XX}^{-1} \K_{ZY}^\top$ and is therefore conjugate to the inverse-Wishart prior~\citep{dawid1981some}. The posterior of $\overrightarrow{\B}$ is therefore again inverse-Wishart, with the index raised from $2$ to $2 + N$ and the scatter matrix added to the prior scale $\overrightarrow{b}_\textrm{pr}(\Z_Y, \Z_Y)$.

The scale after the Bayesian update mirrors~\eqref{eq: Bayesian consistency covariance}, with the expectation replaced by the average over the transitions $\mathcal{D} = (\X, \Y)$ and with two penalties added. To write this scale, we define the operator $\hat{\mathcal{C}}_{YY \mid X}$ on $\mathbb{H}_Y$ through its quadratic form, for every $g \in \mathbb{H}_Y$,
\begin{alignat}{3} \label{eq: posterior conditional covariance}
    \big\langle g, \, \hat{\mathcal{C}}_{YY \mid X} \, g \big\rangle_{\mathbb{H}_Y}
    \coloneqq
    \min_{h \in \mathbb{H}_X}
    \Big\{
        \frac{1}{N} \sum_{i=1}^{N} \big( g(\bm{y}_i) - h(\bm{x}_i) \big)^2
        +
        \frac{\sigma_Y^2}{N} \, \| h \|_{\mathbb{H}_X}^2
    \Big\}
    +
    \frac{\sigma_X^2}{N} \, \| g \|_{\mathbb{H}_Y}^2 .
\end{alignat}
Equivalently, Proposition~\ref{prop: posterior residual geometry} gives the finite-sample Schur-complement representation
\begin{alignat*}{3}
    \hat{\mathcal{C}}_{YY \mid X}
    &=
    \hat{\mathcal{C}}_{YY}
    +
    \frac{\sigma_X^2}{N}\mathcal{I}
    -
    \hat{\mathcal{C}}_{YX}
    \left(
        \hat{\mathcal{C}}_{XX}
        +
        \frac{\sigma_Y^2}{N}\mathcal{I}
    \right)^{-1}
    \hat{\mathcal{C}}_{XY}.
\end{alignat*}
The source-side penalty is the regularization of the conditional mean embedding (Appendix~\ref{subsec: Conditional Distributions}), and the target-side penalty is its mirror, contributed by the base kernel $\overrightarrow{b}_\textrm{pr}$ of the inverse-Wishart prior.

By Proposition~\ref{prop: posterior residual geometry}, the posterior scale at any finite set $\Z_Y$ equals $N \sigma_Y^{-2} \, \Phi_{Z_Y}^\ast \hat{\mathcal{C}}_{YY \mid X} \Phi_{Z_Y}$, i.e., the same kernel evaluated at whichever points $\Z_Y$ are chosen. The finite-dimensional posteriors are therefore consistent, and the posterior of the output covariance function is again an inverse-Wishart process,
\begin{alignat*}{3}
    \overrightarrow{b} \mid \mathcal{D}
    &\sim
    \mathcal{IW}^\infty ( 2 + N , \, N \, \overrightarrow{b}_\textrm{pst} ) ,
    \\
    \overrightarrow{b}_\textrm{pst}(\bm{y}, \bm{y}')
    &=
    \langle \varphi_Y(\bm{y}), \, \overrightarrow{\mathcal{B}}_\kappa \, \varphi_Y(\bm{y}') \rangle_{\mathbb{H}_Y} ,
    \qquad
    \overrightarrow{\mathcal{B}}_\kappa \coloneqq \sigma_Y^{-2} \, \hat{\mathcal{C}}_{YY \mid X} .
\end{alignat*}
The posterior mean $\overrightarrow{b}_\textrm{pst}$ is thus represented by $\overrightarrow{\mathcal{B}}_\kappa$, the posterior-mean counterpart of the population operator. The index $2$ is chosen for exactly this property, even though it means that the prior itself has no well-defined mean \citep{shah2014student}. The geometry of Definition~\ref{def: lifted observation model} is thus learned from the transitions rather than imposed. Proposition~\ref{prop: posterior residual geometry} in Appendix~\ref{sec: Output Covariance Geometry} verifies the corresponding finite-sample ridge representation and the inverse-Wishart moments used in the posterior-consistency argument.

We now return to the embedded PFO. Conditional on $\overrightarrow{b}$, the posterior of the embedded PFO has the standard operator-valued GP form \citep{alvarez2012kernels, micchelli2004kernels, micchelli2005learning}:
\vspace{-0.5\baselineskip}
\begin{alignat}{3}
    \overrightarrow{\mathcal{M}}_\textrm{pst}(\bm{x})  &\coloneqq  \mathcal{A} \; \xi_{X}({\bm{x}})= \sum_{i = 1}^N \mathcal{A}_i \;\overrightarrow{\xi}_\textrm{pr} (\bm{x}_i, \bm{x}),  \label{eq: GP mean operator}\\
    \overrightarrow{\xi}_\textrm{pst}(\bm{x}, \bm{x}') 
    &\coloneqq \overrightarrow{\xi}_\textrm{pr}(\bm{x}, \bm{x}') - \mathbf{\xi}_{X}^\top(\bm{x}) \text{ } \tilde{\Xi}_{XX}^{-1}\text{ }\mathbf{\xi}_{X}(\bm{x}').
\end{alignat}
We unpack these equations by concatenating the noisy targets `end-to-end' in $\Upsilon^\top \! \coloneqq \! \operatorname{vec}(\Phi_Y)\! \in \!\mathbb{H}_Y^{\oplus N}$. We then denote the weight operator by $\mathcal{A} \coloneqq \Upsilon \; \tilde{\Xi}_{XX}^{-1}$, the noisy covariance operator by $\tilde{\Xi}_{XX} \coloneqq \Xi_{XX} + \Sigma$, and the feature ``vector'' by $\mathbf{\xi}_{X}(\bm{x}) \coloneqq [\overrightarrow{\xi}_\textrm{pr}(\bm{x}_1, \bm{x}), \ldots, \overrightarrow{\xi}_\textrm{pr}(\bm{x}_N, \bm{x})]^\top$.

These expressions are read at a finite target-side set $\Z_Y$ of evaluation points, where $\overrightarrow{\B}$ is almost surely nonsingular under both the prior and the posterior \citep{dawid1981some}. Then by recalling the mixed matrix-vector product property: $(\A \otimes \mathbf{B}) \, \operatorname{vec}(\mathbf{C}) = \operatorname{vec}(\mathbf{B} \, \mathbf{C} \, \A^\top)$, the evaluations of the posterior mean~\eqref{eq: GP mean operator} of the embedded PFO at $\Z_Y$ reduce to
\begin{alignat*}{3}
    \Phi_{Z_Y}^\ast \overrightarrow{\mathcal{M}}_\textrm{pst}(\bm{x})
    &=
    \big( \mathbf{k}_X^\top(\bm{x}) \otimes \overrightarrow{\B} \big)
    \big( \tilde{\K}_{XX} \otimes \overrightarrow{\B} \big)^{-1}
    \operatorname{vec} \K_{ZY}
    \\
    &=
    \Phi_{Z_Y}^\ast \Phi_Y \tilde{\K}_{XX}^{-1} \mathbf{k}_X(\bm{x}) .
\end{alignat*}
The output covariance cancels at every finite set of evaluation points, so for every draw of $\overrightarrow{b}$,
\begin{alignat}{3} \label{eq: embedded PFO}
    \overrightarrow{\mathcal{M}}_\textrm{pst}(\bm{x})
    =
    \Phi_Y \tilde{\K}_{XX}^{-1} \Phi_X^\ast \, \varphi_X(\bm{x}) .
\end{alignat}
Thus, $\overrightarrow{\mathcal{M}}_\textrm{pst}(\bm{x})$ is the posterior mean of the CME $\mu_{Y \mid \bm{x}}$. This recovers the expression for $\overrightarrow{\mathcal{P}}_\varepsilon$ of \citet{klus2020eigendecompositions} and mirrors the frequentist approach of \citet{grunewalder2012conditional}. The resulting prediction $\bm{x} \mapsto \langle g, \overrightarrow{\mathcal{M}}_\textrm{pst}(\bm{x}) \rangle_{\mathbb{H}_Y}$ of $g(Y)$ is the minimizer $h$ in~\eqref{eq: posterior conditional covariance}. The posterior covariance over $\overrightarrow{\mathcal{P}}_\varepsilon \varphi_X(\bm{x})$, by contrast, retains the separable form of the prior with $\overrightarrow{\kappa}_\textrm{pst}$ in place of $\overrightarrow{\kappa}_\textrm{pr}$, i.e., a linear transformation of the covariance over the flow map. In other words, while the mean does not depend on $\overrightarrow{b}$, the posterior covariance between $\mu_{Y \mid \bm{x}}(\bm{y})$ and $\mu_{Y \mid \bm{x}'}(\bm{y}')$ is $\overrightarrow{\kappa}_\textrm{pst}(\bm{x}, \bm{x}') \, \overrightarrow{b}_\textrm{pst}(\bm{y}, \bm{y}')$.

The present construction infers the one-step conditional law $\overrightarrow{p}_{\Delta t}(\bm{y} \mid \bm{x})$ rather than only its conditional mean \citep{klus2020eigendecompositions}. Additionally the posterior quantifies the uncertainty of this estimate. We regard this as a principled approach for stochastic and chaotic systems, whose statistics, rather than individual trajectories, are well defined.

\begin{remark}
    At the population level, under the conditions required for the uncentered conditional mean embedding, $\mu_{Y \mid \bm{x}}$ is the mean embedding of the conditional distribution for almost every $\bm{x}$ with respect to $\mathbb{P}_X$ \citep{klebanov2020rigorous}, and the regularized operator $\overrightarrow{\mathcal{P}}_\varepsilon$ estimates this embedding. If the output kernel is characteristic, the mean embedding is injective on probability measures, and $\mu_{Y \mid \bm{x}}$ then determines the conditional distribution uniquely \citep{sriperumbudur2008injective}. Definition~\ref{def: lifted observation model} is pointwise in the conditioning state, so the likelihood is specified at the Dirac source laws $\delta_{\bm{x}_i}$ of the paired observations $(\bm{x}_i,\bm{y}_i)$. For a general source law $\mathbb{P}$ with mean embedding $\mu_X = \mathbb{E}_{X\sim\mathbb{P}}[\varphi_X(X)]$, the population conditional embedding extends the pointwise conditional means by linearity. Under the range condition of Proposition~\ref{prop: CME},
    \[
        \mathcal{C}_{YX} \, \mathcal{C}_{XX}^{-1} \, \mu_X
        =
        \mathbb{E}_{X\sim\mathbb{P}}[\mu_{Y\mid X}],
    \]
    while $\overrightarrow{\mathcal{P}}_\varepsilon \, \mu_X$ provides its Tikhonov-regularized approximation.
\end{remark}

\begin{remark}
At finite source and target evaluation sets, the conditional Gaussian prior over the embedded PFO and the inverse-Wishart process prior $\overrightarrow{b} \sim \mathcal{IW}^\infty ( 2 , \, \overrightarrow{b}_\textrm{pr} )$ form a matrix-normal--inverse-Wishart hierarchy. Marginalizing over $\overrightarrow{b}$ yields a matrix-variate Student-$t$ posterior for the embedded PFO evaluations, with $N+2$ degrees of freedom \citep{shah2014student}.
\end{remark}

\begin{remark}\label{rem: D > 1 systems}
    For $D > 1$, the flow-map channels may be correlated and their residuals anisotropic. More generally, we may specify
    \[
        \overrightarrow{\bm F}
        \sim
        \mathcal{GP}^D ( \bm{0}, \, \overrightarrow{\boldsymbol{\kappa}}_\textrm{pr} ),
        \qquad
        \overrightarrow{\bm{\epsilon}}_i
        \sim
        \mathcal{N}^D ( \bm{0}, \, \boldsymbol{\Sigma}_Y ),
    \]
    where $\overrightarrow{\boldsymbol{\kappa}}_\textrm{pr}(\bm{x}, \bm{x}') \in \mathbb{R}^{D \times D}$ is a positive-semidefinite matrix-valued kernel and $\boldsymbol{\Sigma}_Y \succ 0$ is the residual covariance matrix. The backward construction is analogous, with $\overleftarrow{\boldsymbol{\kappa}}_\textrm{pr}$ and $\boldsymbol{\Sigma}_X$. The independent-channel and isotropic assumptions used in the present manuscript are recovered by taking $\overrightarrow{\boldsymbol{\kappa}}_\textrm{pr} = \overrightarrow{\kappa}_\textrm{pr} \, \I_D$ and $\boldsymbol{\Sigma}_Y = \sigma_Y^2 \I_D$, with analogous expressions in the backward direction. General matrix-valued flow-map kernels remain admissible, although the separable covariance structure used for the lifted inverse-Wishart conjugacy must be preserved or generalized separately \citep{alvarez2012kernels}.
\end{remark}

\subsection{Sparse Variational Bayesian Transfer Operators} \label{sec: Sparse Transfer Operators}

Approximating the exact posterior in \eqref{eq: embedded PFO} with a dictionary $\Phi_{Z_X} \coloneqq [\varphi_X(\bm{z}_1), \ldots, \varphi_X(\bm{z}_{M_X})] \in \mathbb{H}_X^{1\times M_X}$ at the pseudo-inputs $\Z_X \coloneqq [\bm{z}_1, \ldots, \bm{z}_{M_X}]$, where $M_X<N$, is straightforward and demonstrates how Gaussian process techniques can be applied to kernel transfer operators. Once the problem is cast in Bayesian terms, we can implement the variational free energy (VFE) method of \citet{titsias2009variational} (see Appendix~\ref{sec: VFE}):
\begin{alignat}{3} \label{eq: sparse embedded PFO}
    \begin{split}
        \overrightarrow{\mathcal{M}}_\textrm{pst}(\bm{x})
        &=
        \xi_Z^\top(\bm{x})
        \left(
            \Xi_{ZX} \, \Sigma^{-1} \, \Xi_{XZ}
            +
            \Xi_{ZZ}
        \right)^{-1}
        \Xi_{ZX} \, \Sigma^{-1} \, \Upsilon^\top \\
        &=
        \left(\mathbf{k}^\top_{Z}(\bm{x}) \,\widetilde{\C}_{XX}^{-1} \otimes \mathcal{I} \right) \, \operatorname{vec}\left(\Phi_Y \K_{ZX}^\top \right)\\
        &=
        \Phi_Y \K_{ZX}^\top \widetilde{\C}_{XX}^{-1} \Phi_{Z_X}^\ast \, \varphi_X(\bm{x}),
    \end{split}
\end{alignat}
where the compressed covariance operators are $\Xi_{ZX} \coloneqq \K_{ZX} \otimes \overrightarrow{\mathcal{B}}$ and $\Xi_{ZZ} \coloneqq \widetilde{\K}_{ZZ} \otimes \overrightarrow{\mathcal{B}}$, with the cross-covariance matrix $\K_{ZX} \coloneqq \Phi_{Z_X}^\ast \Phi_X$ and the regularized pseudo-covariance matrix
\begin{alignat}{3} \label{eq: regularized pseudo cov}
    \widetilde{\K}_{ZZ}
    \coloneqq
    \K_{ZZ}
    +
    \sigma_Z^2 \, \I_{M_X},
    \qquad
    \K_{ZZ}
    \coloneqq
    \Phi_{Z_X}^\ast \Phi_{Z_X}.
\end{alignat}
These matrices define the noisy Gramian $\widetilde{\C}_{XX} \coloneqq \K_{ZX} \K_{ZX}^\top + \sigma_Y^2 \widetilde{\K}_{ZZ}$. Note that the pseudo-covariance matrices for both directions share the same symbol ${\K}_{ZZ}$, but are constructed using the kernel and dictionary implied by the context. Lastly, the feature vectors are $\mathbf{k}_Z(\bm{x}) \coloneqq \Phi_{Z_X}^\ast \varphi_X(\bm{x})$ and $\xi_Z(\bm{x}) \coloneqq \mathbf{k}_Z(\bm{x}) \otimes \overrightarrow{\mathcal{B}}$. As in Section~\ref{sec: Operators as Random Variables}, these expressions are read conditionally on $\overrightarrow{b}$ at a target-side set $\Z_Y$.

The variance $\sigma_Z^2$ stems from the inducing variables. Following the generalized inducing-variable construction of \citet{titsias2008modelselection}, these are noisy evaluations of the latent function at the pseudo-inputs,
\[
    \tilde{\bm{f}}_Z
    \coloneqq
    \bm{f}_Z
    +
    \bm{\epsilon}_Z,
    \qquad
    \bm{\epsilon}_Z
    \sim
    \mathcal{N}^{M_X} \! \left( \bm{0}, \, \sigma_Z^2 \, \I_{M_X} \right),
\]
with covariance $\operatorname{Cov}( \tilde{\bm{f}}_Z ) = \widetilde{\K}_{ZZ}$. For the operator-valued GP, the noise on the values of the embedded PFO at the pseudo-inputs has the covariance $\sigma_Z^2 \, \I_{M_X} \otimes \overrightarrow{\mathcal{B}}$, which yields the compressed covariance operator $\Xi_{ZZ}$. A positive $\sigma_Z^2$ bounds the condition number of $\widetilde{\K}_{ZZ}$ and thereby acts as an adaptive jitter \citep{titsias2008modelselection}. Appendix~\ref{sec: VFE} shows how $\sigma_Z$ enters the variational bound.

\subsection{The Finite Output Covariance} \label{sec: Finite Output Factor}

From here until Section~\ref{sec: Bayesian Forward-Backward Operators}, the two directions share one prior covariance function $\kappa_\textrm{pr} \coloneqq \overrightarrow{\kappa}_\textrm{pr} = \overleftarrow{\kappa}_\textrm{pr}$, and hence one RKHS $\mathbb{H} \coloneqq \mathbb{H}_X = \mathbb{H}_Y$ with the feature map $\varphi \coloneqq \varphi_X = \varphi_Y$. The lifted targets are evaluated at the source pseudo-inputs, i.e., we adopt the shared dictionary for the predecessors and the successors $\Z \coloneqq \Z_X = \Z_Y$ with $M \coloneqq M_X = M_Y$ elements, and hence the output covariance reduces to the $M \times M$ matrix $\overrightarrow{\B} = \overrightarrow{b}(\Z, \Z)$. The shared dictionary closes the estimated transfer operator on the single space $\operatorname{span}(\Phi_Z)$, as is required by the Koopman eigenproblem of Section~\ref{sec: Koopman Mode Decomposition}. In the terminology of \citet{alvarez2012kernels}, $\overrightarrow{\B}$ is the coregionalization matrix of these $M$ output channels. It cancels from the posterior mean~\eqref{eq: sparse embedded PFO} but enters both the posterior covariance and the evidence. The posterior of $\overrightarrow{\B}$ follows from the conjugate update of Section~\ref{sec: Operators as Random Variables}, with the likelihood replaced by its sparse counterpart. This gives the closed-form expressions used in the remainder of the paper.

Let $\mathbf{R}_{XX} \coloneqq \K_{ZX}^{\top} \widetilde{\K}_{ZZ}^{-1} \K_{ZX}$ denote the \emph{Nystr\"om approximation} of $\K_{XX}$ and $\widetilde{\mathbf{R}}_{XX} \coloneqq \mathbf{R}_{XX} + \sigma_Y^2 \I_N$ its noisy variant (Appendix~\ref{sec: VFE}). In the sparse model, $\widetilde{\mathbf{R}}_{XX}$ replaces $\tilde{\K}_{XX}$ in the likelihood of Section~\ref{sec: Operators as Random Variables}. This follows from integrating out the inducing variables, i.e., the noisy values of the transfer operator at the pseudo-inputs, under the Nystr\"om approximation. The evaluations $\K_{ZY} \coloneqq \Phi_Z^\ast \Phi_Y$ of the lifted targets at the pseudo-inputs then satisfy 
\begin{alignat}{3} \label{eq: lifted likelihood}
    \operatorname{vec} \K_{ZY} \mid \overrightarrow{\B}
    \sim
    \mathcal{N}^{MN}\!\left(
        \bm{0}, \,
        \widetilde{\mathbf{R}}_{XX} \otimes \overrightarrow{\B}
    \right).
\end{alignat}
By the Woodbury identity $\I_N - \K_{ZX}^\top \, \widetilde{\C}_{XX}^{-1} \, \K_{ZX} = \sigma_Y^2 \, \widetilde{\mathbf{R}}_{XX}^{-1}$, the scatter matrix of $\K_{ZY}$ is the residual of the ridge regression that defines the posterior mean~\eqref{eq: sparse embedded PFO}, and hence
\begin{alignat}{3} \label{eq: lifted residual identity}
    \K_{ZY} \, \widetilde{\mathbf{R}}_{XX}^{-1} \, \K_{ZY}^\top
    =
    \frac{1}{\sigma_Y^2}
    \left(
        \C_{YY}
        -
        \C_{YX} \, \widetilde{\C}_{XX}^{-1} \, \C_{XY}
    \right),
    \qquad
    \C_{YY}
    \coloneqq
    \K_{ZY} \, \K_{ZY}^\top,
    \qquad
    \C_{YX}
    \coloneqq
    \C_{XY}^\top .
\end{alignat}

At $\Z_Y = \Z$, we use the prior of Section~\ref{sec: Operators as Random Variables} with index $2$ and with its base kernel augmented by the inducing noise to $( \sigma_X^2 / \sigma_Y^2 ) \, ( \kappa_\textrm{pr} + \sigma_Z^2 \delta )$, where $\delta$ is the Kronecker delta of the noise kernel. Its marginal scale is $( \sigma_X^2 / \sigma_Y^2 ) \, \widetilde{\K}_{ZZ}$. The inducing noise thus also enters the target side,
\begin{alignat}{3} \label{eq: output covariance prior}
    p_\textrm{pr}(\overrightarrow{\B})
    &=
    \mathcal{IW}^M \! \left( \overrightarrow{\B}; \, 2 , \, \frac{\sigma_X^2}{\sigma_Y^2} \, \widetilde{\K}_{ZZ} \right)
    \propto
    \big| \overrightarrow{\B}\big|^{-(M + 1)}
    \exp\!\left(
        - \frac{\sigma_X^2}{2 \sigma_Y^2}
        \operatorname{Tr}\!\left( \widetilde{\K}_{ZZ} \, \overrightarrow{\B}^{-1} \right)
    \right).
\end{alignat}

As in Section~\ref{sec: Operators as Random Variables}, the update raises the index to $2 + N$ and adds the scatter matrix~\eqref{eq: lifted residual identity} to this scale.\footnote{This posterior is also the variational optimum. Extending the collapsed bound of \citet{titsias2009variational} to the output covariance, with the inducing variables optimized conditionally on $\overrightarrow{\B}$, yields this posterior as the optimal factor for $\overrightarrow{\B}$, since the sparsity term of~\eqref{eq: kZ paired score} does not depend on $\overrightarrow{\B}$ (Section~\ref{sec: Model Selection}).} With the target-side counterpart $\widetilde{\C}_{YY} \coloneqq \C_{YY} + \sigma_X^2 \, \widetilde{\K}_{ZZ}$ of $\widetilde{\C}_{XX}$, the posterior is therefore
\begin{alignat*}{3}
    p_\textrm{pst}(\overrightarrow{\B} \mid \mathcal{D})
    &=
    \mathcal{IW}^M \! \left( \overrightarrow{\B}; \, 2 + N , \, \sigma_Y^{-2} \, \widetilde{\C}_{YY \mid X} \right) ,
\end{alignat*}
where
\begin{alignat}{3}
    \widetilde{\C}_{YY \mid X}
    &\coloneqq
    \widetilde{\C}_{YY}
    -
    \C_{YX} \, \widetilde{\C}_{XX}^{-1} \, \C_{XY}
    =
    \sigma_Y^2 \, \K_{ZY} \, \widetilde{\mathbf{R}}_{XX}^{-1} \, \K_{ZY}^\top
    +
    \sigma_X^2 \, \widetilde{\K}_{ZZ} .
    \label{eq: KYYX}
\end{alignat}
Hence, the matrix $N^{-1} \widetilde{\C}_{YY \mid X}$ is the sparse counterpart of $\hat{\mathcal{C}}_{YY \mid X}$. Its regularized residual characterization is derived in Appendix~\ref{sec: Output Covariance Geometry}. As in Section~\ref{sec: Operators as Random Variables}, the posterior mean of $\overrightarrow{\B}$ is
\[
    \mathbb{E}\big[ \overrightarrow{\B} \mid \mathcal{D} \big]
    =
    \overrightarrow{\B}_\kappa
    \coloneqq
    ( N \sigma_Y^2 )^{-1} \, \widetilde{\C}_{YY \mid X}
\]
and hence $\overrightarrow{\B}_\kappa$ is the finite posterior-mean Bayesian-consistency covariance. Its large-sample limit and its relation to the population operator $\overrightarrow{\mathcal{B}}$ are also given in Appendix~\ref{sec: Output Covariance Geometry}.

\subsection{Koopman Mode Decomposition} \label{sec: Koopman Mode Decomposition}

Taking the adjoint of the embedded PFO in \eqref{eq: sparse embedded PFO} gives the estimated kernel Koopman operator $\overrightarrow{\mathcal{U}}_\kappa$. Since its range is contained in the finite-dimensional dictionary subspace $\mathbb{H}_M \coloneqq \operatorname{span}(\Phi_Z) \subset \mathbb{H}$, its action on the dictionary satisfies $\overrightarrow{\mathcal{U}}_\kappa \Phi_Z = \Phi_Z \overrightarrow{\U}_\kappa$. The resulting \emph{Koopman matrix} therefore provides the exact finite-dimensional representation of the estimated operator on $\mathbb{H}_M$ (see \eqref{eq: koopman matrix} in Appendix~\ref{sec: koopman matrix}):

\begin{alignat}{3}
    \begin{split} \label{eq: CME matrix}
        \overrightarrow{\U}_\kappa &= \widetilde{\C}_{XX}^{-1} \, \C_{XY}\in \mathbb{R}^{M \times M}.
    \end{split}
\end{alignat}

Notice that since $\C_{XY} \coloneqq \K_{ZX} \K_{ZY}^\top$ and $\K_{ZY} \coloneqq \Phi_Z^\ast \, \Phi_Y$, equation \eqref{eq: CME matrix} recovers the standard algorithm for EDMD \citep{Klus2016Numerical}. In other words, the Bayesian approach we have followed produces the same expression as the least-squares approach of EDMD with a Tikhonov parameter, $\sigma_Y^2$. Of course, since EDMD is valid for more general choices of dictionaries, the previous statement is only true for a kernel-based dictionary \citep{williams2015kernel}.

Importantly, representing the estimated kernel Koopman operator $\overrightarrow{\mathcal{U}}_\kappa$ by the matrix $\overrightarrow{\U}_\kappa$ on $\mathbb{H}_M$ does not require $\mathbb{H}_M$ to be invariant under the underlying stochastic Koopman operator $\mathcal{U}$ of Definition~\ref{def: stochastic Koopman operator}. Since adjunction is a bounded linear isometry on the Hilbert--Schmidt class, applying it to the posterior over the embedded PFO conditional on $\overrightarrow{b}$ yields the corresponding posterior over $\overrightarrow{\mathcal{U}}_\kappa$, with the input- and output-space roles of $\overrightarrow{\kappa}_\textrm{pst}$ and $\overrightarrow{b}$ interchanged. However, for the spectral interpretation that follows, we do assume that $\mathbb{H}_M$ is invariant under the underlying Koopman operator $\mathcal{U}$. Under this assumption, the spectral quantities of $\overrightarrow{\mathcal{U}}_\kappa$, namely its eigenvalues, eigenfunctions, and Koopman modes, can be interpreted as estimates of the corresponding quantities of $\mathcal{U}$.  This assumption concerns only the dictionary space and places no condition on the remaining spectrum of $\mathcal{U}$, which may be continuous. We refer to the resulting method as Gaussian process DMD (GP-DMD), and Figure~\ref{fig:gp-dmd-diagram} summarizes its posterior-mean prediction pipeline.

\begin{minipage}{\columnwidth} 
    \centering
    \includegraphics[
        trim={0.75cm 0.5cm 0.75cm 0.5cm},
        clip,
        width=0.75\columnwidth
    ]{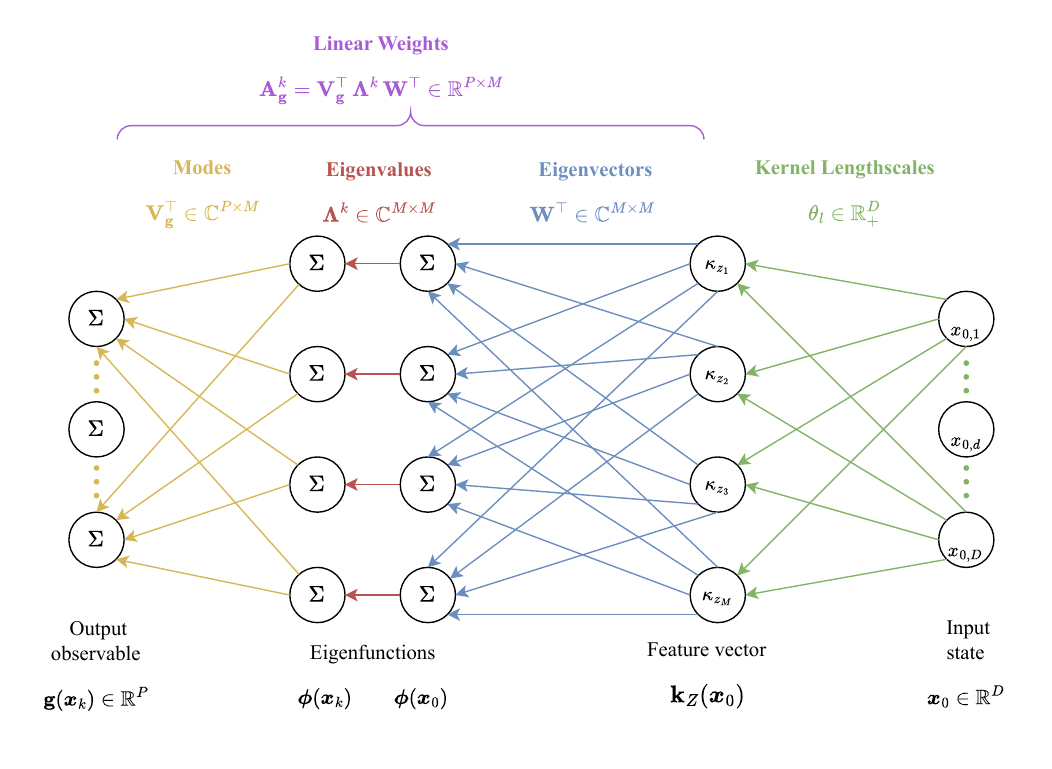}

    \vspace{-1.0\baselineskip}

    \captionsetup{hypcap=false}
    \phantomsection
    {\captionsetup{justification=centering,singlelinecheck=false}\captionof{figure}{Schematic of GP-DMD prediction from the input state to the output observable through kernel features, Koopman eigenfunctions, eigenvalues, and modes. Only the ARD lengthscales are shown; other kernel hyperparameters, such as signal variances, are omitted for clarity.}\label{fig:gp-dmd-diagram}}
    \vspace{0.25\baselineskip}
\end{minipage}

Assuming that $\overrightarrow{\U}_\kappa$ is diagonalizable, the spectral quantities of the estimated kernel Koopman operator $\overrightarrow{\mathcal{U}}_\kappa$ are obtained from the eigendecomposition of its finite-dimensional representation:
\begin{align} \label{eq: eigendecomp}
    \overrightarrow{\U}_\kappa = \W \; \mathbf{\Lambda} \; {\V}_{\kappa},
\end{align}
where $\mathbf{\Lambda}\in \mathbb{C}^{M \times M}$ is a diagonal matrix containing the discrete-time eigenvalues. The right eigenvectors $\W \in \mathbb{C}^{M \times M}$ contain the coordinates of the eigenfunctions in the dictionary basis, whereas the left eigenvectors ${\V}_{\kappa} = \W^{-1}$ correspond to the modes of the observable $\mathbf{k}_Z(\cdot) = [\kappa_\textrm{pr}(\bm{z}_1, \cdot), \, \ldots, \, \kappa_\textrm{pr}(\bm{z}_M, \cdot)]^\top$.

Provided that the retained eigenvalues are nonzero, the \emph{projected} modes with respect to the full-state observables, ${\V_{\bm{F}}} \in \mathbb{C}^{M \times D}$, are estimated via
\begin{alignat}{3} \label{eq: GP DMD modes}
    \begin{split}
        {\V}_{\bm{F}}
        &=
        \mathbf{\Lambda}^{-1}
        \, {\V}_{\kappa}
        \, \widetilde{\C}_{XX}^{-1}
        \, \K_{ZX}\Y^\top.
    \end{split}
\end{alignat}

The implication of \eqref{eq: GP DMD modes} is that for a single-step prediction, the model in \eqref{eq: GP DMD} is also the posterior mean of a GP prior placed on the flow map (using the dynamically evolved states $\Y$ as the targets). The point is that we can perform DMD on a random variable associated with the Koopman operator while maintaining consistency with the nonlinear dynamics on the state manifold. For exact state reconstruction, the underlying assumption is that the full-state observable belongs to the dictionary subspace, i.e., $\mathbf{g}_{\bm{F}} \in \mathbb{H}_M$, with $\mathbf{g}_{\bm{F}}(\bm{x}) \coloneqq \bm{x}$. This requirement is in addition to the invariance assumption used for the spectral interpretation.

The requirement that the full-state observable be representable in the learned dictionary space is a double-edged feature of the Bayesian-consistent construction. On the one hand, the constraint $\mathbf{g}_{\bm{F}} \in \mathbb{H}_{M}$ is restrictive, especially if exact Koopman invariance is also required. For deterministic dynamics, \citet{10.1371/journal.pone.0150171} show that requiring a finite-dimensional Koopman-invariant subspace to contain the original state variables rules out systems with multiple fixed points, periodic orbits, or more complex attractors. On the other hand, the same requirement ties state reconstruction to the model that produces the forward prediction. This relationship allows the lifted model to be compared against the state-space model, as shown in Section~\ref{sec: Bayesian Consistency}.

\subsection{The Bayesian-Consistency Principle} \label{sec: Bayesian Consistency}

The Bayesian-consistency principle stated after Definition~\ref{def: lifted observation model} asks whether the one-step covariance of the lifted model, once reconstructed to the state space, agrees with the state-space covariance of the flow-map model. Reconstruction uses the projected modes~\eqref{eq: GP DMD modes} through the matrix $\A_{\bm{F}}^0 \coloneqq (\W \, {\V}_{\bm{F}})^\top \in \mathbb{R}^{D \times M}$, which acts like a linear decoder from lifted coordinates to the state space \citep{azencot2020forecasting}. We now establish the conditions under which this agreement holds.

The matrix $\sigma_Y^2 \, \overrightarrow{\B}_\kappa = N^{-1} \widetilde{\C}_{YY \mid X}$ is the regularized second moment of the residual when the posterior mean predicts $\mathbf{k}_Z(Y)$ from $\mathbf{k}_Z(X)$. As $N \rightarrow \infty$, the regularization vanishes, and $\sigma_Y^2 \, \overrightarrow{\B}_\kappa$ tends to the least-squares counterpart of $\mathcal{C}_{YY \mid X}$ on the dictionary space $\mathbb{H}_M$ (Appendix~\ref{subsec: Large-sample Limit}). Now suppose that the flow map is represented exactly in the dictionary space, such that $\overrightarrow{\bm{F}} = \mathcal{U} \, \mathbf{g}_{\bm{F}} \in \mathbb{H}_M$ component-wise and $\mathbb{P}_X$-almost everywhere, i.e., $\mathcal{U}$ maps the components of $\mathbf{g}_{\bm{F}} \in \mathbb{H}_M$ back into $\mathbb{H}_M$. Then, with exact reconstruction,
$
     \bm{x} = \A_{\bm{F}}^0 \, \mathbf{k}_Z(\bm{x}),
$
and under the isotropic observation model $\operatorname{Cov}(Y \mid X) = \sigma_Y^2 \I_D$, the limit satisfies
\begin{alignat}{3} \label{eq: consistency check}
    \sigma_Y^2 \,
    \A_{\bm{F}}^0
    \, \overrightarrow{\B}_\kappa \,
    \left(\A_{\bm{F}}^0\right)^\top
    =
    \mathbb{E}_X\!\left[
        \operatorname{Cov}(Y \mid X)
    \right],
    \qquad
    \A_{\bm{F}}^0
    \, \overrightarrow{\B}_\kappa \,
    \left(\A_{\bm{F}}^0\right)^\top
    =
    \I_D .
\end{alignat}

If $\overrightarrow{\bm{F}}$ is not represented exactly in $\mathbb{H}_M$, the decoded covariance in~\eqref{eq: consistency check} exceeds $\I_D$ in the limit by the resulting representation error (Proposition~\ref{prop: residual geometry}\ref{item: residual geometry ii}). More generally, a deviation from $\I_D$ signals that one of the assumptions behind~\eqref{eq: consistency check} fails, i.e., either the dictionary does not close the flow map, the state is not linearly decodable from $\mathbb{H}_M$, or the isotropic state-space model is misspecified. The same comparison can be made at finite $N$ for every observable in $\mathbb{H}_M$.

For $\bm{\mathrm{w}} \in \mathbb{C}^M$, let $g \coloneqq \Phi_Z \bm{\mathrm{w}}$, i.e., $g(\bm{x}) = \bm{\mathrm{w}}^\top \mathbf{k}_Z(\bm{x})$, and let $\overrightarrow{\mathcal{U}}_\kappa g = \Phi_Z \overrightarrow{\U}_\kappa \bm{\mathrm{w}}$ be its one-step prediction by the estimated kernel Koopman operator. Applied to the real and imaginary parts of $\bm{\mathrm{w}}$, the sparse residual identity~\eqref{eq: sparse residual geometry} of Appendix~\ref{sec: Output Covariance Geometry} gives
\begin{alignat}{3} \label{eq: directional lifted residual}
    \bm{\mathrm{w}}^* \, \widetilde{\C}_{YY \mid X} \, \bm{\mathrm{w}}
    &=
    \sum_{j=1}^{N} \big| g(\bm{y}_j) - (\overrightarrow{\mathcal{U}}_\kappa g)(\bm{x}_j) \big|^2
    +
    \gamma(\bm{\mathrm{w}}),
    \\
    \gamma(\bm{\mathrm{w}})
    &\coloneqq
    \sigma_Y^2 \, (\overrightarrow{\U}_\kappa \bm{\mathrm{w}})^* \, \widetilde{\K}_{ZZ} \, \overrightarrow{\U}_\kappa \bm{\mathrm{w}}
    +
    \sigma_X^2 \, \bm{\mathrm{w}}^* \, \widetilde{\K}_{ZZ} \, \bm{\mathrm{w}} .
    \nonumber
\end{alignat}
Divided by $N$, these quadratic forms are the posterior means of the one-step residual variance of the observables in $\mathbb{H}_M$.

For a Koopman eigenpair $\overrightarrow{\U}_\kappa \bm{\mathrm{w}}_i = \lambda_i \bm{\mathrm{w}}_i$ with the associated eigenfunction $\phi_i = \Phi_Z \bm{\mathrm{w}}_i$, we have $(\overrightarrow{\mathcal{U}}_\kappa \phi_i)(\bm{x}) = \lambda_i \phi_i(\bm{x})$, and~\eqref{eq: directional lifted residual}, normalized by $\bm{\mathrm{w}}_i^* \, \widetilde{\C}_{XX} \, \bm{\mathrm{w}}_i$, gives
\begin{alignat}{3} \label{eq: Bayesian ResDMD residual}
    [ \operatorname{res}^{\textrm{var}}_{\textrm{pst}}(\lambda_i, \phi_i) ]^2
    \coloneqq
    \frac{
        \bm{\mathrm{w}}_i^* \, \widetilde{\C}_{YY \mid X} \, \bm{\mathrm{w}}_i
    }{
        \bm{\mathrm{w}}_i^* \, \widetilde{\C}_{XX} \, \bm{\mathrm{w}}_i
    }
    =
    \frac{
        \sum_{j=1}^{N} | \phi_i(\bm{y}_j) - \lambda_i \phi_i(\bm{x}_j) |^2
        +
        \gamma(\bm{\mathrm{w}}_i)
    }{
        \sum_{j=1}^{N} | \phi_i(\bm{x}_j) |^2
        +
        \sigma_Y^2 \, \bm{\mathrm{w}}_i^* \, \widetilde{\K}_{ZZ} \, \bm{\mathrm{w}}_i
    } .
\end{alignat}
Without the prior terms $\gamma(\bm{\mathrm{w}}_i)$ and $\sigma_Y^2 \, \bm{\mathrm{w}}_i^* \, \widetilde{\K}_{ZZ} \, \bm{\mathrm{w}}_i$, this is the variance residual $[ \operatorname{res}^{\textrm{var}}(\lambda_i, \phi_i) ]^2$ of \citet{colbrook2024beyond}, which reduces to the ResDMD residual of \citet{colbrook2023residual} for deterministic dynamics. In that case, a small residual places $\lambda_i$ in a pseudospectrum of the Koopman operator, whereas a large residual flags a spurious eigenpair. For stochastic dynamics, $[ \operatorname{res}^{\textrm{var}}(\lambda_i, \phi_i) ]^2$ additionally contains the integrated conditional variance of $\phi_i(Y)$, which \citet{colbrook2024beyond} separate from the squared Koopman residual $[ \operatorname{res}(\lambda_i, \phi_i) ]^2$ with batched data.

The state-space model of Section~\ref{sec: Sampling and the Flow Map} offers a separation that requires no batched data. It supplies the expected value of the residual in~\eqref{eq: directional lifted residual} under its fitted one-step law $Y \mid X = \bm{x} \sim \mathcal{N}^D ( \A_{\bm{F}}^1 \mathbf{k}_Z(\bm{x}), \, \sigma_Y^2 \I_D )$, where $\A_{\bm{F}}^1 \coloneqq ( \W \, \mathbf{\Lambda} \, {\V}_{\bm{F}} )^\top$ gives the one-step forecast mean. Note that this law carries the residual variance $\sigma_Y^2$ but not the epistemic variance $\overrightarrow{\kappa}_\textrm{pst}$. We write $\mathbb{E}_{\bm{F}}$ for the conditional expectation under this law and define the Hermitian conditional variance
\[
    \operatorname{Var}_{\bm{F}} [ g(Y) \mid X = \bm{x} ]
    \coloneqq
    \mathbb{E}_{\bm{F}}
    \left[
        \left|
            g(Y)
            -
            \mathbb{E}_{\bm{F}}[g(Y)\mid X=\bm{x}]
        \right|^2
        \,\middle|\,
        X=\bm{x}
    \right].
\]
As in the residual--variance decomposition of \citet{colbrook2024beyond}, the predicted residual of an observable $g \in \mathbb{H}_M$ splits into a mean mismatch and an integrated conditional variance,
\begin{alignat}{3} \label{eq: state-space residual split}
    \sum_{j=1}^{N} \mathbb{E}_{\bm{F}} \big[ | g(Y) - (\overrightarrow{\mathcal{U}}_\kappa g)(\bm{x}_j) |^2 \mid X = \bm{x}_j \big]
    &=
    \sum_{j=1}^{N} \big| \mathbb{E}_{\bm{F}} [ g(Y) \mid X = \bm{x}_j ] - (\overrightarrow{\mathcal{U}}_\kappa g)(\bm{x}_j) \big|^2
    \nonumber \\
    &\phantom{=}
    +
    \sum_{j=1}^{N} \operatorname{Var}_{\bm{F}} [ g(Y) \mid X = \bm{x}_j ] .
\end{alignat}
Evaluated under the true transition law, the two terms divided by $N$ converge to the closure error and the averaged conditional variance of Theorem~\ref{thm: output covariance consistency}. Unless $g$ is linear in the state, both terms require the mean and covariance of the kernel features $\mathbf{k}_Z(Y)$ under the fitted law, which in general have no closed form. These moments are finite for a bounded kernel and can be evaluated by Monte Carlo sampling or deterministic quadrature, as discussed in Section~\ref{sec: Prediction Theory}. Since $g$ is linear in the kernel features, a single evaluation of these moments at each training input serves every observable in $\mathbb{H}_M$.

For an eigenpair, the two terms in~\eqref{eq: state-space residual split} separate the posterior variance residual~\eqref{eq: Bayesian ResDMD residual} into a spectral error and a stochastic variance. One difficulty is that this variance, which the fitted law inherits from the residual variance $\sigma_Y^2$, may stem from transition variability, measurement noise, or both (see~\eqref{eq: noise apportionment}). Setting $g = \phi_i$, we define the posterior Koopman residual through the first term of~\eqref{eq: state-space residual split}, with the prior terms and the normalization of~\eqref{eq: Bayesian ResDMD residual},
\begin{alignat}{3} \label{eq: posterior Koopman residual}
    [ \operatorname{res}_{\textrm{pst}}(\lambda_i, \phi_i) ]^2
    \coloneqq
    \frac{
        \sum_{j=1}^{N} \big| \mathbb{E}_{\bm{F}} [ \phi_i(Y) \mid X = \bm{x}_j ] - \lambda_i \phi_i(\bm{x}_j) \big|^2
        +
        \gamma(\bm{\mathrm{w}}_i)
        }{
        \bm{\mathrm{w}}_i^* \, \widetilde{\C}_{XX} \, \bm{\mathrm{w}}_i
    } .
\end{alignat}
By~\eqref{eq: Bayesian ResDMD residual}, the observed residual of $\phi_i$ in~\eqref{eq: directional lifted residual} equals the residual predicted by~\eqref{eq: state-space residual split} exactly when the posterior variance residual decomposes into the spectral error and the integrated conditional variance of $\phi_i(Y)$ predicted by the model,
\begin{alignat}{3} \label{eq: posterior residual decomposition}
    [ \operatorname{res}^{\textrm{var}}_{\textrm{pst}}(\lambda_i, \phi_i) ]^2
    =
    [ \operatorname{res}_{\textrm{pst}}(\lambda_i, \phi_i) ]^2
    +
    \frac{1}{\bm{\mathrm{w}}_i^* \, \widetilde{\C}_{XX} \, \bm{\mathrm{w}}_i}
    \sum_{j=1}^{N} \operatorname{Var}_{\bm{F}} [ \phi_i(Y) \mid X = \bm{x}_j ] .
\end{alignat}
The right-hand side follows from the fitted law alone, for instance by the quadrature discussed above, whereas the posterior variance residual on the left is computed from the posterior mean $\overrightarrow{\B}_\kappa$ of the output covariance, i.e., from the lifted successors $\mathbf{k}_Z(\bm{y}_j)$. The interpretation of the posterior Koopman residual~\eqref{eq: posterior Koopman residual} and the variance term in~\eqref{eq: posterior residual decomposition} as spectral error and stochastic variance therefore relies on the assumptions of the state-space GP. Moreover, the decomposition in~\eqref{eq: posterior residual decomposition} holds exactly only when the split of the residual by the state-space GP into spectral error and stochastic variance is \emph{consistent} with the output covariance inferred by the lifted model.

We now return to the Bayesian-consistency principle, which concerns only the full-state observable, and state it at finite $N$. For clarity we take $D = 1$, since the case $D > 1$ follows by applying the same argument to each component of $\mathbf{g}_{\bm{F}}$. Exact reconstruction gives $\mathbf{g}_{\bm{F}} = \Phi_Z (\A_{\bm{F}}^0)^\top$, i.e., $\bm{\mathrm{w}} = (\A_{\bm{F}}^0)^\top$. Being linear in the state, the full-state observable has conditional mean $\A_{\bm{F}}^1 \mathbf{k}_Z(\bm{x})$ and residual variance $\sigma_Y^2$ under the fitted law. The mean mismatch in~\eqref{eq: state-space residual split} vanishes, because the lifted prediction $(\overrightarrow{\mathcal{U}}_\kappa \mathbf{g}_{\bm{F}})(\bm{x}) = \A_{\bm{F}}^0 \overrightarrow{\U}_\kappa^\top \mathbf{k}_Z(\bm{x}) = \A_{\bm{F}}^1 \mathbf{k}_Z(\bm{x})$ equals this mean, and the predicted residual reduces to $N \sigma_Y^2$ in closed form. By~\eqref{eq: directional lifted residual}, the observed residual of $\mathbf{g}_{\bm{F}}$ equals this prediction exactly when
\begin{alignat}{3} \label{eq: finite consistency check}
    \sigma_Y^2 \, \A_{\bm{F}}^0 \, \overrightarrow{\B}_\kappa \, (\A_{\bm{F}}^0)^\top
    =
    \sigma_Y^2 + \frac{\gamma ( \A_{\bm{F}}^0 )}{N} .
\end{alignat}
The decoded posterior covariance must therefore equal the state-space residual variance up to a prior correction. Under the assumptions of Theorem~\ref{thm: output covariance consistency}, and provided the eigenvalues of $\overrightarrow{\U}_\kappa$ remain bounded away from zero, this correction is of order $1 / N$. As $N \rightarrow \infty$, it vanishes and~\eqref{eq: finite consistency check} becomes~\eqref{eq: consistency check}.

Now observe that both~\eqref{eq: posterior residual decomposition} and~\eqref{eq: finite consistency check} are instances of a single condition, namely that the residual of an observable equals the residual predicted by the state-space model. This condition applies to every $g \in \mathbb{H}_M$, and we quantify it with the \emph{Bayesian-consistency ratio}
\begin{alignat}{3} \label{eq: directional consistency ratio}
    c(\bm{\mathrm{w}})
    \coloneqq
    \frac{
        \bm{\mathrm{w}}^* \, \widetilde{\C}_{YY \mid X} \, \bm{\mathrm{w}}
    }{
        \sum_{j=1}^{N} \mathbb{E}_{\bm{F}} \big[ | g(Y) - (\overrightarrow{\mathcal{U}}_\kappa g)(\bm{x}_j) |^2 \mid X = \bm{x}_j \big]
        +
        \gamma(\bm{\mathrm{w}})
    } .
\end{alignat}
By~\eqref{eq: directional lifted residual}, the same $\gamma(\bm{\mathrm{w}})$ term enters the numerator and the denominator. Hence $c(\bm{\mathrm{w}}) = 1$ exactly when the observed residual of $g$ equals the predicted residual, whereas $c(\bm{\mathrm{w}}) > 1$ indicates an observed residual above the prediction and $c(\bm{\mathrm{w}}) < 1$ an observed residual below the prediction. Since both residuals consist of a mean term and a variance term, $c(\bm{\mathrm{w}})$ departs from one when the state-space model misstates either the conditional mean or the conditional variance of $g(Y)$, and the ratio alone cannot distinguish between the two causes. The Bayesian-consistency principle is thus the full-state, large-sample case of the condition $c(\bm{\mathrm{w}}) = 1$.

\begin{remark} \label{rem: residual variance}
    The posterior variance residual~\eqref{eq: Bayesian ResDMD residual} and the Bayesian-consistency ratio~\eqref{eq: directional consistency ratio} evaluate the output covariance at its posterior mean $\overrightarrow{\B}_\kappa$. Evaluated at $\overrightarrow{\B}$ instead, both become random variables with the posterior means $[ \operatorname{res}^{\textrm{var}}_{\textrm{pst}}(\lambda_i, \phi_i) ]^2$ and $c(\bm{\mathrm{w}})$. Proposition~\ref{prop: residual variance} and the discussion following it in Appendix~\ref{subsec: Posterior Variance of the Residuals} give their posterior variances in closed form, and both coefficients of variation are at most $\sqrt{2 / (N - 2)}$. At an eigenvector, the decomposition~\eqref{eq: posterior residual decomposition} holds exactly when $c(\bm{\mathrm{w}}_i) = 1$. A deviation of $c(\bm{\mathrm{w}}_i)$ from one that is comparable to $c(\bm{\mathrm{w}}_i) \sqrt{2 / (N - 2)}$ is therefore consistent with the posterior epistemic uncertainty of the output covariance, whereas a much larger deviation points to a mismatch between the lifted model and the state-space model, such as a miscalibrated variance model.
\end{remark}

\begin{remark}
    The Bayesian-consistency principle expresses agreement between the lifted and state-space descriptions of the same dynamics, without necessarily requiring identical prior covariance functions. The principle also extends to nonlinear reconstruction, for example through a GP decoder mapping the lifted coordinates back to the state space. In this case, the consistency comparison must account for the nonlinear propagation of the lifted covariance and the additional predictive uncertainty of the decoder. Additionally, the Bayesian-consistency ratio~\eqref{eq: directional consistency ratio} and posterior Koopman residual~\eqref{eq: posterior Koopman residual} extend to more general state-space likelihoods, including heteroskedastic models, through the corresponding conditional moments in~\eqref{eq: state-space residual split}. The closed-form condition~\eqref{eq: finite consistency check} is the instructive `spherical cow' example with a shared prior covariance function, exact linear reconstruction, and an isotropic residual variance.
\end{remark}

\subsection{Forecasting and Uncertainty Propagation} \label{sec: Prediction Theory}

To construct multi-step forecasts and quantify predictive uncertainty, we must differentiate between uncertainty in the learned transition and inherent one-step variability. The posterior kernel $\overrightarrow{\kappa}_\textrm{pst}$ quantifies \emph{epistemic} uncertainty in the learned flow map, which is expected to contract in well-supported regions as information accumulates, whereas the \emph{aleatoric} component $\sigma_Y^2$ represents persistent variability in the one-step transition. Both contributions are represented through the same Bayesian-consistency geometry introduced in Definition~\ref{def: lifted observation model}, and they may therefore differ only in scalar amplitude and in \emph{when} that amplitude is applied. Accordingly, the forecast family is determined by two independent choices: how the residual variance is apportioned between propagated transition variability and measurement noise, and whether the reported covariance describes the latent state or a future measurement.

The first of these choices is not settled by the training likelihood of the lifted observation model. Definition~\ref{def: lifted observation model} places the residual on the lifted target, which is consistent with a continuum of interpretations of the same fitted equation. Two limiting cases are particularly useful. At one extreme, $\overrightarrow{\bm{F}}$ is a deterministic flow map and $\bm{\epsilon}_i$ represents sensor noise incurred when the successor is measured. At the other, $\overrightarrow{\bm{F}}$ is the conditional mean of a stochastic one-step transition and $\bm{\epsilon}_i$ represents transition randomness or process noise. Intermediate interpretations apportion the residual variance between these two limits. Within the working lifted likelihood, all such interpretations are represented by zero-mean residuals on the lifted target with covariance proportional to the output covariance, and the training likelihood identifies only the residual variance $\sigma_Y^2$, but not its decomposition. Since randomness in the dynamics and noise on the observables play different roles in stochastic Koopman approximation \citep{wanner2022robust}, we make this apportionment explicit and treat it as a modeling assumption.

Accordingly, we apportion the fitted residual variance into a process and a measurement component,
\begin{alignat}{3} \label{eq: noise apportionment}
    \sigma_Y^2 = \sigma_\textrm{proc}^2 + \sigma_\textrm{meas}^2,
    \qquad
    \sigma_\textrm{proc}^2, \, \sigma_\textrm{meas}^2 \geq 0.
\end{alignat}
At $\sigma_\textrm{proc}^2 = 0$ the residual is treated as measurement error on a deterministic latent flow, whereas at $\sigma_\textrm{meas}^2 = 0$ it is treated as transition variability, and intermediate splits describe a mixture of the two.

To obtain tractable multi-step forecasts, we approximate the projected dynamics in $\mathbb{R}^M$ by the stochastic difference equation \citep{hansen2013recursive}:
\begin{alignat}{3} \label{eq: LSDE}
    {\mathbf{k}}_{Z,k}
    = \A_\kappa^1 \, {\mathbf{k}}_{Z,k-1}
    + {\Lb}_\textrm{pst}(X_{k-1}) \, \bm{\omega}_{k},
    \qquad
    {\mathbf{k}}_{Z,0} = \mathbf{k}_Z(\bm{x}_0).
\end{alignat}
Here ${\mathbf{k}}_{Z,k} \in \mathbb{R}^M$ denotes the coordinate vector propagated by the linear model, which does not necessarily coincide with an exact kernel evaluation $\mathbf{k}_Z(\bm{x})$ \citep{honeine2011preimage}. The corresponding state is reconstructed with
$
    X_k = \A_{\bm{F}}^0 \, {\mathbf{k}}_{Z,k},
$
where $\A_{\bm{F}}^0$ is the reconstruction matrix of Section~\ref{sec: Bayesian Consistency} and $\bm{\omega}_k \sim \mathcal{N}^M(\bm{0}, \I_M)$. Lastly, the multi-step transition matrix is $\A_\kappa^k \coloneqq \left( \W \,\mathbf{\Lambda}^k \, {\V}_{\kappa} \right)^\top$. 

The recursion in~\eqref{eq: LSDE} rests on two closures, one for the projected dynamics and one for the uncertainty. The Mori--Zwanzig formalism \citep{zwanzig1961memory, mori1965transport} clarifies the first. For a finite-rank Mori projection onto the chosen observables, the Markov term corresponds to the population least-squares Koopman approximation underlying EDMD. The exact projected evolution additionally contains memory and orthogonal-dynamics terms \citep{doi:10.1137/21M1401759, LINLU2021109864}. Consequently, when the lifted space is not invariant, the one-step projected dynamics do not in general reproduce the exact multi-step projected evolution. The propagation $\A_\kappa^k$ in~\eqref{eq: LSDE} therefore imposes a Markov closure. In the ideal population setting, this closure is exact when the lifted space is Koopman-invariant and $\A_\kappa$ exactly represents the restricted Koopman operator. Explicitly modeling the memory terms provides a natural extension when the learned lifted space is not dynamically closed.

The second closure concerns the uncertainty. We adopt another Markov moment-closure approximation, in which a fresh Gaussian innovation at each step represents the one-step predictive covariance. These innovations neglect three sources of dependence across steps. First, they neglect the two-time correlations of the orthogonal dynamics, and the innovations thus form a memoryless stochastic closure rather than the exact Mori--Zwanzig orthogonal dynamics. Second, they neglect temporal correlations in the epistemic error, since every step evaluates the same posterior GP realization. Third, they neglect the dependence induced by the output covariance $\overrightarrow{\B}$ of Section~\ref{sec: Finite Output Factor}, which is shared by all steps of a forecast. Under this closure, the process $\{ \bm{\omega}_{i} \}_{i = 1}^{k}$ is an i.i.d. \emph{martingale difference sequence} adapted to the sequence of information sets $\{ J_i \}_{i = 0}^{k}$. Each \emph{information set} $J_k$ represents all the information available to an agent at time-step $k$, and contains all measurable functions of $\{X_0, \, \bm{\omega}_1, \ldots, \, \bm{\omega}_k \}$.

The matrix ${\Lb}_\textrm{pst}(X_{k-1})$ is chosen as a positive-semidefinite square root of ${\mathbf{\Xi}}_{\textrm{pst}}^{1}(X_{k-1})$, such that
\[
    {\Lb}_\textrm{pst}(X_{k-1})\,{\Lb}_\textrm{pst}^\top(X_{k-1})
    =
    {\mathbf{\Xi}}_{\textrm{pst}}^{1}(X_{k-1}).
\]
For strictly positive-definite covariance matrices this square root may be obtained by a Cholesky factorization. Put differently, $\bm{\omega}_k$ provides a standardized Gaussian representation of the one-step predictive covariance. The covariance matrix ${\mathbf{\Xi}}_{\textrm{pst}}^{1}(X_{k-1})$ thus encodes the one-step-ahead forecasting error of ${\mathbf{k}}_{Z,k}$ conditioned on $J_{k-1}$. To characterize this covariance, recall that the sparse posterior kernel (see \eqref{eq: VFE cov} in Appendix~\ref{sec: VFE}) is
\begin{alignat}{3} \label{eq: posterior kernel VFE}
    \overrightarrow{\kappa}_\textrm{pst}(\bm{x}, \bm{x}') &= \kappa_\textrm{pr}(\bm{x}, \bm{x}') - \mathbf{k}_{Z}^\top(\bm{x}) \text{ }\tilde{\Q}_{XX}\text{ }\mathbf{k}_{Z}(\bm{x}')
\end{alignat}
where $\tilde{\Q}_{XX} \coloneqq \left(\widetilde{\K}_{ZZ}^{-1} -\sigma_Y^2 \widetilde{\C}_{XX}^{-1}\right)$. The scalar $\overrightarrow{\kappa}_\textrm{pst}(\bm{x},\bm{x})$ is the posterior variance of the latent GP evaluation and therefore quantifies the epistemic contribution to the one-step predictive covariance. The corresponding noisy one-step predictive variance is $\overrightarrow{\kappa}_\textrm{pst}(\bm{x},\bm{x}) + \sigma_Y^2$ \citep{williams2006gaussian}, so the two contributions in \eqref{eq: one step combined covariance} are the lifted analogues of the latent and noisy scalar predictive variances. Accordingly, the one-step lifted covariance is
\begin{alignat}{3} \label{eq: one step combined covariance}
    {\mathbf{\Xi}}_{\textrm{pst}}^{1}(X_{k-1})
    = \left( \overrightarrow{\kappa}_\textrm{pst}(X_{k-1},X_{k-1}) + \sigma_\textrm{proc}^2 \right) \overrightarrow{\B}_\kappa,
\end{alignat}
where $\overrightarrow{\B}_\kappa$ is the posterior mean of the output covariance from Section~\ref{sec: Finite Output Factor}. Using $\overrightarrow{\B}_\kappa$ instead of the random output covariance $\overrightarrow{\B}$ loses nothing for the mean and covariance of the one-step prediction. Given $\overrightarrow{\B}$, the predictive mean does not involve $\overrightarrow{\B}$, and the predictive covariance is $\overrightarrow{\B}$ scaled by the factor $\overrightarrow{\kappa}_\textrm{pst}(X_{k-1},X_{k-1}) + \sigma_\textrm{proc}^2$. Averaging this covariance over the posterior of $\overrightarrow{\B}$, as the law of total covariance prescribes, therefore replaces $\overrightarrow{\B}$ by $\overrightarrow{\B}_\kappa$, which makes~\eqref{eq: one step combined covariance} the exact covariance of the one-step predictive law. Marginalizing the inverse-Wishart covariance yields a heavy-tailed multivariate Student-$t$ distribution \citep{dawid1981some}. The Gaussian innovation $\bm{\omega}_k$ used in the Markov moment closure approximates this distribution by matching only its mean and covariance.

Under the conditions of Section~\ref{sec: Bayesian Consistency}, applying \eqref{eq: consistency check} to \eqref{eq: one step combined covariance} gives the reconstructed one-step covariance of the latent state as
$\A_{\bm{F}}^0\, 
 {\mathbf{\Xi}}_{\textrm{pst}}^{1}(\bm{x}) \, 
(\A_{\bm{F}}^0)^\top
=
\left(
\overrightarrow{\kappa}_\textrm{pst}(\bm{x},\bm{x})
+
\sigma_\textrm{proc}^2
\right)\I_D$.
The complementary observation variance $\sigma_\textrm{meas}^2\I_D$ is not propagated by the dynamics. If the reported quantity is a future measurement, it is added once after reconstruction, giving the total predictive covariance
$\left(
\overrightarrow{\kappa}_\textrm{pst}(\bm{x},\bm{x})
+
\sigma_Y^2
\right)\I_D $.

Expanding \eqref{eq: LSDE} into a recursive expression that is a function of $\{X_{k-j-1}, \, \bm{\omega}_{k-j} \}_{j = 0}^{k-1}$ results in
\begin{alignat*}{3}
    \begin{split}   
        {\mathbf{k}}_{Z,k}
        &=
        \A_\kappa^k \, \mathbf{k}_Z(\bm{x}_0)
        +
        \sum_{j=0}^{k-1}
        \A_\kappa^j \,
        {\Lb}_\textrm{pst}(X_{k-j-1}) \,
        \bm{\omega}_{k-j}.
    \end{split}
\end{alignat*}
Therefore, the mean vector of $X_{k}$ given the realization $\bm{x}_{0}$ is
\begin{alignat}{3} \label{eq: GP DMD}
    \begin{split}   
    \hat{\bm{x}}_k &= \mathbb{E}\left[X_{k} \mid X_0 =\bm{x}_0\right] =\A_{\bm{F}}^k \; \mathbf{k}_Z(\bm{x}_{0}),
    \end{split}
\end{alignat}
where the weight matrix is $\A_{\bm{F}}^k \coloneqq \left( \W \,\mathbf{\Lambda}^k \, {\V}_{\bm{F}} \right)^\top \in \mathbb{R}^{D \times M}$. In deriving \eqref{eq: GP DMD} we used \eqref{eq: GP DMD modes}, the relationship $X_k = \A_{\bm{F}}^0 \, {\mathbf{k}}_{Z,k}$, and the martingale-difference property $\mathbb{E}[\bm{\omega}_k \mid J_{k-1}] = \bm{0}$.

Now we are in a position to present the mechanism to propagate the covariance matrix $\mathbf{\Xi}_{\textrm{pst}}^{k}(\bm{x}_0)$. We start from the $k$-step forecast error, i.e., the deviation of the propagated coordinates from their conditional mean:
\begin{alignat*}{3} 
    \zeta_{k}
    \coloneqq
    {\mathbf{k}}_{Z,k}
    -
    \A_\kappa^k \, \mathbf{k}_Z(\bm{x}_0) = \sum_{j=0}^{k-1}\A_\kappa^j \;{\Lb}_\textrm{pst}(X_{k-j-1}) \, \bm{\omega}_{k-j}.
\end{alignat*}
Since $\bm{\omega}_n \sim \mathcal{N}^M(\bm{0},\I_M)$ is independent of the preceding state $X_{n-1}$ and of the other innovations, each term $\Lb_\textrm{pst}(X_{n-1}) \, \bm{\omega}_n$ has zero conditional mean and distinct innovation terms have zero cross-covariance. Hence, when forming $\operatorname{Cov}[\zeta_k \mid X_0 = \bm{x}_0]$, all cross terms vanish. Consequently,
\begin{alignat*}{3}
    \mathbb{E}\!\left[
        \Lb_\textrm{pst}(X_{n-1})
        \bm{\omega}_n \bm{\omega}_n^\top
        \Lb_\textrm{pst}(X_{n-1})^\top
        \,\middle|\,
        X_0 = \bm{x}_0
    \right] 
    & =
    \mathbb{E}\!\left[
        \Lb_\textrm{pst}(X_{n-1})
        \Lb_\textrm{pst}(X_{n-1})^\top
        \,\middle|\,
        X_0 = \bm{x}_0
    \right],
\end{alignat*}
and therefore the propagated covariance is
\begin{alignat}{3} \label{eq: GP DMD cov}
    {\mathbf{\Xi}}_{\textrm{pst}}^{k}(\bm{x}_0) &= \sum_{j=0}^{k-1} \A_\kappa^j \, \mathbb{E} \! \left[ {\mathbf{\Xi}}_{\textrm{pst}}^{1}(X_{k-j-1}) \mid X_0 = \bm{x}_0 \right] \left( \A_\kappa^j \right)^\top .
\end{alignat}
Since we choose to fix $\overrightarrow{\B}$ at $\overrightarrow{\B}_\kappa$ over the forecast trajectory, substituting \eqref{eq: one step combined covariance} into the conditional expectation reduces the expected one-step covariance to a scalar amplitude,
\begin{alignat*}{3}
    \tilde{\kappa}_{k-1}
    \coloneqq
    \bar{\kappa}_{k-1} + \sigma_\textrm{proc}^2,
    \qquad
    \bar{\kappa}_{k-1}
    \coloneqq
    \mathbb{E} \! \left[ \overrightarrow{\kappa}_\textrm{pst}(X_{k-1},X_{k-1}) \mid X_0 = \bm{x}_0 \right].
\end{alignat*}
Equivalently, by \eqref{eq: one step combined covariance}, $\tilde{\kappa}_{k-1}\overrightarrow{\B}_\kappa = \mathbb{E}\!\left[{\mathbf{\Xi}}_{\textrm{pst}}^{1}(X_{k-1}) \mid X_0=\bm{x}_0\right]$, so $\tilde{\kappa}_{k-1}$ is the expected scalar amplitude of the same one-step covariance that enters the stochastic difference equation \eqref{eq: LSDE}.

For multi-step forecasts, \eqref{eq: GP DMD cov} can therefore be evaluated recursively as
\begin{alignat}{3} \label{eq: GP DMD cov recursive}
    {\mathbf{\Xi}}_{\textrm{pst}}^{k}(\bm{x}_0)
    &= \A_\kappa \, {\mathbf{\Xi}}_{\textrm{pst}}^{k-1}(\bm{x}_0) \, \A_\kappa^\top
    + \tilde{\kappa}_{k-1} \, \overrightarrow{\B}_\kappa,
    \qquad {\mathbf{\Xi}}_{\textrm{pst}}^{0} = \bm{0}.
\end{alignat}

The first term in \eqref{eq: GP DMD cov recursive} propagates the accumulated covariance through the learned lifted dynamics, while the second introduces the expected one-step predictive covariance at the uncertain intermediate state. Between reprojections, the forecast mean $\hat{\bm{x}}_k = \A_{\bm{F}}^k \, \mathbf{k}_Z(\bm{x}_0)$ is exact under the learned linear lifted recursion, so only the scalar expectation $\bar{\kappa}_{k-1}$ requires approximation. This expectation is taken over the $D$-dimensional Gaussian approximation $X_{k-1} \sim \mathcal{N}^D(\hat{\bm{x}}_{k-1},\K_\textrm{pst}^{k-1})$, where the state forecast covariance is obtained with $\K_\textrm{pst}^{k-1} = \A_{\bm{F}}^0 \, {\mathbf{\Xi}}_{\textrm{pst}}^{k-1} \, (\A_{\bm{F}}^0)^\top$.

The expectation $\bar{\kappa}_{k-1}$ is in general not analytically tractable. It is, however, always finite, since $0 \leq \overrightarrow{\kappa}_\textrm{pst}(\bm{x},\bm{x}) \leq \kappa_\textrm{pr}(\bm{x},\bm{x})$ and the kernel is bounded, and it can be evaluated under the Gaussian approximation $X_{k-1} \sim \mathcal{N}^D(\hat{\bm{x}}_{k-1}, \K_\textrm{pst}^{k-1})$ by Monte Carlo sampling or deterministic quadrature \citep{girard2002gaussian, girard2002gpuncertaininputs}. While Monte Carlo sampling requires no regularity of the kernel, the Gauss--Hermite rule \citep{ito2000gaussian} is more efficient when $\bm{x} \mapsto \overrightarrow{\kappa}_\textrm{pst}(\bm{x},\bm{x})$ is sufficiently smooth. Its tensor-product form, however, needs a number of nodes that grows exponentially with the state dimension, and sparse-grid variants \citep{jia2012sparse} mitigate this growth.

\begin{remark}
Under persistent covariance injection, bounded multi-step state uncertainty depends jointly on the stability of the lifted dynamics, whether neutral or unstable directions are excited by the covariance injection, and whether those directions are visible under reconstruction to the state observable. Directions associated with eigenvalues inside the unit disk attenuate previously propagated uncertainty, whereas neutral or unstable directions can cause uncertainty to persist or grow when they are both excited and visible under state reconstruction.
\end{remark}

\begin{remark}
    The lifted covariance~\eqref{eq: GP DMD cov recursive} also yields the uncertainty of any other observable represented linearly in the lifted space, through the same kind of projection that gives the state covariance $\K_\textrm{pst}^{k} = \A_{\bm{F}}^0 \, \mathbf{\Xi}_{\textrm{pst}}^{k} \, (\A_{\bm{F}}^0)^\top$. For example, since the eigenfunction vector is given by $\boldsymbol{\phi} (\bm{x}) = \W^\top \, \mathbf{k}_Z (\bm{x})$, the $k$-step covariance matrix of the eigenfunctions is $\W^\top \, \mathbf{\Xi}_{\textrm{pst}}^{k} ( \bm{x}_0 ) \, ( \W^\top )^*$, where $^*$ denotes the Hermitian adjoint. This ability to compute the multi-step posterior for eigenfunctions is showcased in Figure~\ref{fig: quadruple well eigenfunctions}, while the propagated state uncertainty is illustrated in Figures~\ref{fig: Duffing trajectory} and~\ref{fig: vdP trajectory}.
\end{remark}

\subsection{Model Selection} \label{sec: Model Selection}

The quality of a finite-dimensional transfer operator approximation depends critically on the feature space in which the operator is represented. This has motivated data-driven approaches that learn the representation together with the operator approximation, for example by minimizing lifted linear-evolution residuals or prediction and reconstruction losses \citep{li2017extended, takeishi2017learning, yeung2019learning, azencot2020forecasting}. For kernel-based methods, the reproducing kernel determines the observable space in which the operator is represented, while the domain and properties of the operator depend on how this function space interacts with the dynamics. The kernel function should therefore be adapted to suit the dynamics at hand \citep{ikeda2022Koopman}.

One possible criterion for selecting the representation is the flow-map \emph{variational free energy} (VFE) $\mathcal{F}_{\bm{F}}$, i.e., the sum over output channels of the scalar objective in Appendix~\ref{sec: VFE}. This criterion suffices when the system admits a finite-dimensional Koopman-invariant subspace containing the full-state observable, but in general an accurate model of the flow map does not guarantee an accurate operator model \citep{doi:10.1137/24M1696974}. We therefore perform operator model selection directly in the lifted space by evaluating the one-step evolution of the lifted observables $\K_{ZY}$.\footnote{As in Section~\ref{sec: Finite Output Factor}, the predecessors and the successors share one kernel and one dictionary for notational simplicity. The derivation below carries over to the separate kernels of Section~\ref{sec: Problem Formulation} and to separate pseudo-input sets.}

Model selection then rests on the evidence of the lifted observables, i.e., their marginal likelihood with the transfer operator and its output covariance integrated out. The challenge is that the lifted targets $\K_{ZY}$ also depend on the kernel hyperparameters and the pseudo-inputs. Changing the representation therefore changes the very data that the evidence scores, and evidences computed for different data cannot be compared. Like any probability density, an evidence depends on the coordinates of its data, and changing the representation acts like a change of units. The situation resembles statistical mechanics, where the absolute free energy depends on an arbitrary reference, and hence only free-energy differences are meaningful \citep{gibbs1902elementary}. 

We therefore compare two models of the lifted observables that share the inverse-Wishart process prior of Section~\ref{sec: Operators as Random Variables} and whose marginal at the pseudo-inputs is~\eqref{eq: output covariance prior}. The conditional or \emph{paired} model is the model of Section~\ref{sec: Finite Output Factor}, which predicts each successor from its predecessor. The marginal or \emph{reference} model, by contrast, ignores the predecessors. The ratio of their evidences is the \emph{Bayes factor} of the paired model against the reference model. Since the two models share the same targets and prior, but only the paired model uses the predecessor states to explain the successor targets, the Bayes factor measures the evidence provided by the predecessor--successor pairing.

Formally, the reference model is simply the paired model with the degenerate operator-valued prior $\overrightarrow{\xi}_\textrm{pr} \equiv 0$. The noisy prior covariance then reduces to $\sigma_Y^2 \I_N$, and each lifted target is modeled as $\mathbf{k}_Z(\bm{y}_i) \mid \overrightarrow{\Br}\sim \mathcal{N}^M(\bm{0}, \sigma_Y^2 \overrightarrow{\Br})$. The calculation of Section~\ref{sec: Finite Output Factor}, with the scatter $\sigma_Y^{-2} \C_{YY}$ in place of~\eqref{eq: lifted residual identity}, gives the posterior distribution $\mathcal{IW}^M ( \overrightarrow{\Br}; 2 + N , N \overrightarrow{\Br}_\kappa )$, whose mean is the \emph{pre-pairing} output covariance
\begin{alignat}{3} \label{eq: K0}
    \overrightarrow{\Br}_\kappa
    &\coloneqq ( N \sigma_Y^2 )^{-1} \, \widetilde{\C}_{YY},
    \qquad
    \widetilde{\C}_{YY}
    = \C_{YY} + \sigma_X^2 \, \widetilde{\K}_{ZZ} .
\end{alignat}
The matrix $\overrightarrow{\Br}_\kappa$ is the finite inducing-space counterpart of $\sigma_Y^{-2} \mathcal{C}_{YY}$. Hence $\sigma_Y^2 \overrightarrow{\Br}_\kappa = N^{-1} \widetilde{\C}_{YY}$ is the posterior-mean marginal covariance of the lifted successors, whereas $\sigma_Y^2 \overrightarrow{\B}_\kappa = N^{-1} \widetilde{\C}_{YY \mid X}$ is their posterior-mean residual covariance after conditioning on the predecessors.

For a fixed representation $( \theta , \Z )$, and with the transfer operator integrated out as in~\eqref{eq: lifted likelihood} but without sparsification, the Bayes factor of the paired model against the reference model is
\begin{alignat}{3}
    \beta_\mathcal{F}
    &\coloneqq
    \frac{
        \int \mathcal{N}^{MN} \! \left( \operatorname{vec} \K_{ZY} ; \bm{0} , \tilde{\K}_{XX} \otimes \overrightarrow{\Br}\right) p_\textrm{pr} ( \overrightarrow{\Br}) \, \mathrm{d} \overrightarrow{\Br}
    }{
        \int \mathcal{N}^{MN} \! \left( \operatorname{vec} \K_{ZY} ; \bm{0} , \sigma_Y^2 \I_N \otimes \overrightarrow{\Br}\right) p_\textrm{pr} ( \overrightarrow{\Br}) \, \mathrm{d} \overrightarrow{\Br}
    }
    =
    \mathbb{E} \! \left[
        \frac{
            \mathcal{N}^{MN} \! \left( \operatorname{vec} \K_{ZY} ; \bm{0} , \tilde{\K}_{XX} \otimes \overrightarrow{\Br}\right)
        }{
            \mathcal{N}^{MN} \! \left( \operatorname{vec} \K_{ZY} ; \bm{0} , \sigma_Y^2 \I_N \otimes \overrightarrow{\Br}\right)
        }
    \right] ,
    \label{eq: Bayes factor}
\end{alignat}
where $\mathbb{E}$, here and in the rest of this section, denotes the expectation with respect to the posterior of $\overrightarrow{\Br}$ under the reference model \citep{meng1996simulating}. Because the prior is shared, it cancels from the ratio of the two integrands, and therefore the Bayes factor is the average of the likelihood ratio at a \emph{common} output covariance over the reference posterior $\mathcal{IW}^M ( \overrightarrow{\Br}; 2 + N , N \overrightarrow{\Br}_\kappa )$.

Taking the logarithm of~\eqref{eq: Bayes factor} and bounding it from below twice gives
\begin{alignat}{3}
    \operatorname{ln} \beta_\mathcal{F}
    &\geq
    \mathbb{E} \! \left[
        \operatorname{ln}
        \frac{
            \mathcal{N}^{MN} \! \left( \operatorname{vec} \K_{ZY} ; \bm{0} , \tilde{\K}_{XX} \otimes \overrightarrow{\Br}\right)
        }{
            \mathcal{N}^{MN} \! \left( \operatorname{vec} \K_{ZY} ; \bm{0} , \sigma_Y^2 \I_N \otimes \overrightarrow{\Br}\right)
        }
    \right]
    \geq
    \mathbb{E} \! \left[
        \mathcal{F}_{\mathbf{k}_Z}^{\textrm{pair}}(\theta,\Z)
        -
        \mathcal{F}_{\mathbf{k}_Z}^{\textrm{ref}}(\theta,\Z)
    \right] ,
    \label{eq: Bayes factor bound}
\end{alignat}
where $\mathcal{F}_{\mathbf{k}_Z}^{\textrm{pair}}$ and $\mathcal{F}_{\mathbf{k}_Z}^{\textrm{ref}}$ are the VFEs of the paired and the reference model. The first inequality is Jensen's inequality, since the logarithm is concave.\footnote{In statistical mechanics, the reference and paired models play the roles of an unperturbed and a perturbed system, and $\operatorname{ln} \beta_\mathcal{F}$ is minus their free-energy difference \citep{meng1996simulating}. The Jensen step is then the Gibbs--Bogoliubov inequality, the first-order perturbative estimate of this difference \citep{kuzemsky2015variational}.} The second inequality applies the variational bound of \citet{titsias2009variational} to each output channel of the paired model.

We therefore define the transfer operator objective function as the reference-posterior average of the difference between the paired and the reference VFEs,
\begin{alignat}{3} \label{eq: kZ score contrast}
    \overrightarrow{\mathcal{L}}_{\mathbf{k}_Z}(\theta,\Z)
    &\coloneqq
    -2 \, \mathbb{E} \! \left[
        \mathcal{F}_{\mathbf{k}_Z}^{\textrm{pair}}(\theta,\Z)
        -
        \mathcal{F}_{\mathbf{k}_Z}^{\textrm{ref}}(\theta,\Z)
    \right].
\end{alignat}

At a common output covariance $\overrightarrow{\Br}$ and for fixed $(\theta,\Z)$, the VFE of the paired model is
\begin{alignat}{3} \label{eq: kZ paired score}
    \mathcal{F}_{\mathbf{k}_Z}^{\textrm{pair}}(\theta,\Z)
    &\coloneqq
    \operatorname{ln}\!\left[
        \mathcal{N}^{MN}\!\left(
            \operatorname{vec}\K_{ZY};
            \bm{0},
            \widetilde{\mathbf{R}}_{XX} \otimes \overrightarrow{\Br}
        \right)
    \right]
    -
    \frac{M}{2\sigma_Y^2}
    \operatorname{Tr}\!\left(
        \K_{XX} - \mathbf{R}_{XX}
    \right).
\end{alignat}
The trace term is free of the output covariance, which scales both the prior and the residual covariance and therefore cancels. For the reference model, the degenerate prior $\overrightarrow{\xi}_\textrm{pr} \equiv 0$ removes the Nystr\"om sparsity error, so the variational bound equals the exact log-likelihood,
\begin{alignat}{3} \label{eq: kZ null VFE}
    \mathcal{F}_{\mathbf{k}_Z}^{\textrm{ref}}(\theta,\Z)
    &\coloneqq
    \operatorname{ln}\!\left[
        \mathcal{N}^{MN}\!\left(
            \operatorname{vec}\K_{ZY};
            \bm{0},
            \sigma_Y^2 \I_N \otimes \overrightarrow{\Br}
        \right)
    \right].
\end{alignat}

Writing the matrix-normal terms explicitly gives
\begin{alignat*}{3}
    -2 \, \mathcal{F}_{\mathbf{k}_Z}^{\textrm{pair}}
    &=
    MN \operatorname{ln}(2\pi)
    + M \operatorname{ln}\!\left| \widetilde{\mathbf{R}}_{XX} \right|
    + N \operatorname{ln}\!\left| \overrightarrow{\Br}\right|
    + \operatorname{Tr}\!\left(
            \overrightarrow{\Br}^{-1} \, \K_{ZY} \, \widetilde{\mathbf{R}}_{XX}^{-1} \, \K_{ZY}^\top
    \right)
    + \frac{M}{\sigma_Y^2}\operatorname{Tr}\!\left(
        \K_{XX} - \mathbf{R}_{XX}
    \right),
\end{alignat*}
\begin{alignat*}{3}
    \textrm{whereas} \quad -2 \, \mathcal{F}_{\mathbf{k}_Z}^{\textrm{ref}}
    &=
    MN \operatorname{ln}(2\pi)
    + MN \operatorname{ln}(\sigma_Y^2)
    + N \operatorname{ln}\!\left| \overrightarrow{\Br}\right|
    + \frac{1}{\sigma_Y^2}
    \operatorname{Tr}\!\left(
            \overrightarrow{\Br}^{-1} \, \C_{YY}
    \right).
\end{alignat*}
Both Mahalanobis terms are linear in $\overrightarrow{\Br}^{-1}$. Under the reference posterior, $\overrightarrow{\Br}^{-1}$ is Wishart distributed with $N + M + 1$ degrees of freedom and scale $( N \overrightarrow{\Br}_\kappa )^{-1}$ \citep{dawid1981some}, and the reference posterior therefore satisfies
\begin{alignat*}{3}
    \mathbb{E} \! \left[ \overrightarrow{\Br}^{-1} \right]
    &=
    \frac{N + M + 1}{N} \, \overrightarrow{\Br}_\kappa^{-1}
    =
    ( N + M + 1 ) \, \sigma_Y^2 \, \widetilde{\C}_{YY}^{-1} ,
\end{alignat*}
where the second equality follows from~\eqref{eq: K0}. The expected Mahalanobis term of the reference model is
\begin{alignat*}{3}
    \frac{1}{\sigma_Y^2}
    \mathbb{E} \! \left[
        \operatorname{Tr}\!\left(
        \overrightarrow{\Br}^{-1} \, \C_{YY}
        \right)
    \right]
    &=
    ( N + M + 1 ) \operatorname{Tr}\!\left(
        \widetilde{\C}_{YY}^{-1} \, \C_{YY}
    \right).
\end{alignat*}
For the paired model, using~\eqref{eq: lifted residual identity} gives
\begin{alignat*}{3}
    \mathbb{E} \! \left[
        \operatorname{Tr}\!\left(
        \overrightarrow{\Br}^{-1} \, \K_{ZY} \, \widetilde{\mathbf{R}}_{XX}^{-1} \, \K_{ZY}^\top
        \right)
    \right]
    &=
    ( N + M + 1 ) \operatorname{Tr}\!\left[
        \widetilde{\C}_{YY}^{-1}
        \left(
            \C_{YY} - \C_{YX} \, \widetilde{\C}_{XX}^{-1} \, \C_{XY}
        \right)
    \right] \\
    &=
    ( N + M + 1 ) \left[
        \operatorname{Tr}\!\left(
            \widetilde{\C}_{YY}^{-1} \, \C_{YY}
        \right)
        -
        \operatorname{Tr}\!\left(
            \widetilde{\C}_{YY}^{-1} \, \C_{YX} \, \widetilde{\C}_{XX}^{-1} \, \C_{XY}
        \right)
    \right].
\end{alignat*}
Notice here that the first trace term is common to both models and simply cancels out.

To identify the remaining trace term, observe that
\begin{alignat}{3} \label{eq: kZ VAMP2 score}
    \mathcal{S}
    \coloneqq
    \operatorname{Tr}\!\left(
        \widetilde{\C}_{YY}^{-1} \, \C_{YX} \, \widetilde{\C}_{XX}^{-1} \, \C_{XY}
    \right)
    &=
    \operatorname{Tr}\!\left[
        \overrightarrow{\Br}_\kappa^{-1} \left( \overrightarrow{\Br}_\kappa - \overrightarrow{\B}_\kappa \right)
    \right] 
    &=
    \left\|
        \widetilde{\C}_{XX}^{-1/2} \, \C_{XY} \, \widetilde{\C}_{YY}^{-1/2}
    \right\|_F^2
    =
    \sum_{i=1}^{M}\rho_i^2 .
\end{alignat}
The second equality follows from~\eqref{eq: K0} and~\eqref{eq: KYYX}. The difference $\overrightarrow{\Br}_\kappa - \overrightarrow{\B}_\kappa$ is the lifted-target covariance explained by the predecessor--successor pairing. Therefore $\mathcal{S}$ measures this explained covariance relative to the marginal covariance $\overrightarrow{\Br}_\kappa$. In the variational approach for Markov processes (VAMP), $\widetilde{\C}_{XX}^{-1/2} \, \C_{XY} \, \widetilde{\C}_{YY}^{-1/2}$ is the uncentered and normalized time-lagged operator with singular values $\rho_1,\ldots,\rho_M$ \citep{wu2020variational}, and $\mathcal{S}$ is the corresponding VAMP-$2$ score. Equivalently, $\widetilde{\C}_{YY}^{-1} \, \C_{YX} \, \widetilde{\C}_{XX}^{-1} \, \C_{XY}$ is the associated backward--forward composition with eigenvalues $\rho_i^2$. These quantities become squared \emph{canonical correlations} in the centered CCA construction reviewed in Appendix~\ref{sec: Canonical Correlation Analysis}. Under the additional RKHS range and invariance conditions discussed in Appendix~\ref{subsec: Forward-backward Operators}, the corresponding population construction is also connected to the singular functions used to identify \emph{coherent sets} \citep{klus2019kernel, 10.1063/1.4971788, FROYLAND20131}.

In~\eqref{eq: kZ score contrast}, the constant $MN\operatorname{ln}(2\pi)$ and the row log-determinant $N\operatorname{ln}|\overrightarrow{\Br}|$ cancel between the two VFEs for every $\overrightarrow{\Br}$. Moreover,
\begin{alignat*}{3}
    M \operatorname{ln}\!\left| \widetilde{\mathbf{R}}_{XX} \right|
    - MN \operatorname{ln}(\sigma_Y^2)
    &=
    M \operatorname{ln}\!\left(
        \left| \widetilde{\mathbf{R}}_{XX} \right|
        \big/
        \left| \sigma_Y^2 \I_N \right|
    \right),
\end{alignat*}
while the difference between the two expected Mahalanobis terms is $- ( N + M + 1 ) \, \mathcal{S}$. Therefore,
\begin{alignat}{3} \label{eq: kZ VAMP2 objective}
    \overrightarrow{\mathcal{L}}_{\mathbf{k}_Z}(\theta,\Z)
    =
    - ( N + M + 1 ) \sum_{i=1}^{M}\rho_i^2
    + M \operatorname{ln}\!\left(
        \left| \widetilde{\mathbf{R}}_{XX} \right|
        \big/
        \left| \sigma_Y^2 \I_N \right|
    \right)
    + \frac{M}{\sigma_Y^2}
    \operatorname{Tr}\!\left(
        \K_{XX} - \mathbf{R}_{XX}
    \right)
    .
\end{alignat}
The first term rewards a large score $\mathcal{S}$ and therefore favors representations in which the predecessor--successor pairing explains a large amount of the normalized lifted-target covariance. It consequently promotes endpoint feature spaces that capture the dominant one-step dynamical dependence. The residual interpretation of this score follows from the finite-sample identity~\eqref{eq: sparse residual geometry}, while Corollary~\ref{cor: population VAMP residual geometry} relates its population counterpart to conditional variability and Koopman closure error. The second term is the \emph{relative} model complexity, while the third is the Nystr\"om sparsity error.

At a fixed dictionary size $M$, model selection proceeds by minimizing~\eqref{eq: kZ VAMP2 objective} over the pseudo-inputs and the remaining kernel hyperparameters. On its own, the VAMP-$2$ score keeps improving as the feature space is enlarged or made more flexible, and \citet{wu2020variational} therefore guard against overfitting by cross-validation with their VAMP-E score. In~\eqref{eq: kZ VAMP2 objective}, the relative model-complexity term instead penalizes the flexibility of the feature space without held-out data. Optimizing the pseudo-inputs $\Z$ and choosing $M$ can nevertheless remain computationally demanding. As a compromise, we recommend greedy active learning methods \citep{mackay1992information, engel2004kernel, lawrence2004learning}. In the same vein, Appendix~\ref{sec: Dictionary Learning} details a multi-tiered sparsity method, which combines an active learning criterion that approximates a Bayesian D-optimal design with a final stage that scores candidates via the lifted objective~\eqref{eq: kZ VAMP2 objective}. Additionally, the analytical gradients of $\overrightarrow{\mathcal{L}}_{\mathbf{k}_Z}$ are derived in Appendix~\ref{sec: Gradient Derivations}, and Appendix~\ref{sec: GP-TCCA} describes a symmetrized objective~\eqref{eq: fb lifted objective} that adds the backward terms.

\begin{remark}
    The regularization scales of the lifted model are calibrated from the state-space regressions rather than estimated from the lifted objective. Here, $\sigma_f^2$ denotes the signal variance of the prior covariance function, while $\sigma_Y^2$ and $\sigma_X^2$ are the residual variances of the forward and backward flow-map regressions, respectively. The symmetrized lifted objective~\eqref{eq: fb lifted objective}, discussed in Section~\ref{sec: Bayesian Forward-Backward Operators}, is invariant under a common rescaling of these quantities together with the inducing-noise variance $\sigma_Z^2$ and does not determine their absolute scale. We therefore keep $\sigma_f^2$, $\sigma_Y^2$, and $\sigma_X^2$ fixed during optimization, taking $\sigma_f^2$ and $\sigma_Y^2$ from the forward flow-map VFE and $\sigma_X^2$ from the corresponding backward regression. This calibration acts as a gauge fixing of the scale redundancy.
\end{remark}

\begin{remark}
    Independently of its Bayesian derivation, the VAMP-$2$ score $\mathcal{S}$ rests on the variational principle of \citet{wu2020variational}. Among all orthonormal functions of the initial state and of the final state, the dominant singular functions of the Koopman operator between the weighted spaces of Appendix~\ref{subsec: Forward-backward Operators} maximize the sum of the squared uncentered correlations between paired functions, and the maximum is the sum of the squared singular values, as stated in~\eqref{eq: VAMP variational principle}.
\end{remark}

\begin{remark}
    For the sparse likelihood~\eqref{eq: lifted likelihood}, the Jensen step in~\eqref{eq: Bayes factor bound} can be avoided, since $\overrightarrow{\Br}^{-1}$ is Wishart distributed under the reference posterior and the average of the likelihood ratio has a closed form. The resulting bound on $\operatorname{ln} \beta_\mathcal{F}$ is tighter than $- \frac{1}{2} \overrightarrow{\mathcal{L}}_{\mathbf{k}_Z}(\theta,\Z)$ and differs from it only in the data-fit term, where the VAMP-$2$ score is replaced by
    \begin{alignat*}{3}
        2 I_{\mathcal{N}}(X;Y) 
        &=
        \operatorname{ln}\!\left( \frac{ \left| \widetilde{\C}_{YY} \right| }{ \left| \widetilde{\C}_{YY \mid X} \right| } \right)
        = - \sum_{i=1}^{M} \operatorname{ln}\!\left( 1 - \rho_i^2 \right)
        \geq
        \sum_{i=1}^{M} \rho_i^2 .
    \end{alignat*}
    Here, $I_{\mathcal{N}}(X;Y)$ denotes the mutual information of the corresponding Gaussian surrogate. Equivalently, the log-determinant measures twice the corresponding Gaussian entropy reduction \citep{cover1999elements, bach2002kernel}. This replacement diverges as any $\rho_i$ tends to one (Corollary~\ref{cor: population VAMP residual geometry}), which is one reason for using the bounded VAMP-$2$ score.
\end{remark}

\subsection{Bayesian Round-trip Operators} \label{sec: Bayesian Forward-Backward Operators}

The sparse posterior of the embedded PFO in Sections~\ref{sec: Operators as Random Variables}--\ref{sec: Finite Output Factor} supplies the forward lifted model and its Koopman representation. Gaussian process time-lagged canonical correlation analysis (GP-TCCA, Appendix~\ref{sec: GP-TCCA}) additionally uses the backward lifted model to estimate the singular structure of the source-side forward--backward operator introduced in Appendix~\ref{subsec: Forward-backward Operators}. After summarizing the backward model, we derive the posterior moments of the induced random round trip at the RKHS-operator level, using the directional posterior laws already established.

Unlike the eigendecomposition of GP-DMD, the singular value decomposition of GP-TCCA does not require a shared dictionary (Appendix~\ref{subsec: GP-TCCA Forward-Backward Symmetry}). We therefore return to the separate prior covariance functions $\overrightarrow{\kappa}_\textrm{pr}$ and $\overleftarrow{\kappa}_\textrm{pr}$, RKHSs $\mathbb{H}_X$ and $\mathbb{H}_Y$, and feature maps $\varphi_X$ and $\varphi_Y$ of Sections~\ref{sec: Kernel Transfer Operators}--\ref{sec: Sparse Transfer Operators}, now with separate dictionaries $\Phi_{Z_X}$ and $\Phi_{Z_Y}$ of $M_X$ and $M_Y$ elements. A feature vector is taken in the dictionary of the endpoint at which it is evaluated, $\mathbf{k}_Z(\bm{x}) = \Phi_{Z_X}^\ast \varphi_X(\bm{x})$ and $\mathbf{k}_Z(\bm{y}) = \Phi_{Z_Y}^\ast \varphi_Y(\bm{y})$, and $\K_{ZX}$ and $\K_{ZY}$ collect these vectors for the training pairs.

The backward lifted model mirrors Definition~\ref{def: lifted observation model} with the roles of the predecessor and the successor exchanged. For a fixed successor $Y = \bm{y}$, the backward conditional mean embedding is
\[
    \mu_{X \mid \bm{y}}
    \coloneqq
    \mathbb{E}\!\left[ \varphi_X(X) \mid Y = \bm{y} \right]
    =
    \int_{\mathbb{X}}
    \overleftarrow{p}_{\Delta t}(\bm{x}' \mid \bm{y}) \,
    \varphi_X(\bm{x}') \,
    \mathrm{d}\bm{x}',
\]
where $\overleftarrow{p}_{\Delta t}$ is the time-reversed transition density~\eqref{eq: reversed transition density}. Hence, the backward lifted likelihood mirrors Definition~\ref{def: lifted observation model},
\[
    \varphi_X(X) \mid Y = \bm{y}
    \approx
    \mu_{X \mid \bm{y}} + \overleftarrow{\nu},
    \qquad
    \overleftarrow{\nu}
    \sim
    \mathcal{N}_{\mathbb{H}_X}\!\left( 0, \, \sigma_X^2 \, \overleftarrow{\mathcal{B}} \right),
    \qquad
    \overleftarrow{\mathcal{B}} \coloneqq \sigma_X^{-2} \, \mathcal{C}_{XX \mid Y},
\]
where $\mathcal{C}_{XX \mid Y}$ follows from~\eqref{eq: Bayesian consistency covariance} by exchanging $X$ and $Y$. The corresponding backward priors are
\begin{alignat*}{3}
    \mu_{X \mid \cdot}
    &\sim
    \mathcal{GP}^\infty \big( 0 , \, \overleftarrow{\kappa}_\textrm{pr} \, \overleftarrow{\mathcal{B}} \big),
    &\qquad
    \overleftarrow{b}
    &\sim
    \mathcal{IW}^\infty \big( 2 , \, ( \sigma_Y^2 / \sigma_X^2 ) \, \overrightarrow{\kappa}_\textrm{pr} \big).
\end{alignat*}
Thus the output-covariance prior of each lifted model uses the prior covariance function of the opposite direction. The posterior formulas of Sections~\ref{sec: Operators as Random Variables}--\ref{sec: Finite Output Factor} carry over under the exchange of $X$ and $Y$, $\sigma_Y$ and $\sigma_X$, and the corresponding kernels and dictionaries.

The backward counterparts of~\eqref{eq: KYYX} and the inverse-Wishart posterior of Section~\ref{sec: Finite Output Factor} are
\begin{alignat}{3}
    \widetilde{\C}_{XX \mid Y}
    &\coloneqq
    \widetilde{\C}_{XX}
    -
    \C_{XY} \, \widetilde{\C}_{YY}^{-1} \, \C_{YX},
    \label{eq: KXXY}
    \\
    \overleftarrow{\B} \mid \mathcal{D}
    &\sim
    \mathcal{IW}^{M_X} \big( 2 + N , \, \sigma_X^{-2} \, \widetilde{\C}_{XX \mid Y} \big). \notag
\end{alignat}
Table~\ref{tab: forward backward models} summarizes the two models and their shared endpoint regularization.

\begin{table}[t]
    \centering
    \renewcommand{\arraystretch}{1.1}
    \setlength{\tabcolsep}{6pt}
    \caption{\emph{Forward and backward lifted models}.}
    \label{tab: forward backward models}
    \resizebox{0.9\linewidth}{!}{%
    \begin{tabular}{@{}l l l@{}}
        \toprule
        & Forward model & Backward model \\
        \midrule
        Conditional law & $\overrightarrow{p}_{\Delta t}(\bm{y} \mid \bm{x})$ & $\overleftarrow{p}_{\Delta t}(\bm{x} \mid \bm{y})$ \\
        Embedding & $\mu_{Y \mid \bm{x}} \in \mathbb{H}_Y$ & $\mu_{X \mid \bm{y}} \in \mathbb{H}_X$ \\
        Prior covariance function & $\overrightarrow{\kappa}_\textrm{pr}$ & $\overleftarrow{\kappa}_\textrm{pr}$ \\
        Output-covariance prior & $\overrightarrow{b} \sim \mathcal{IW}^\infty ( 2 , \, (\sigma_X^2 / \sigma_Y^2) \, \overleftarrow{\kappa}_\textrm{pr} )$ & $\overleftarrow{b} \sim \mathcal{IW}^\infty ( 2 , \, (\sigma_Y^2 / \sigma_X^2) \, \overrightarrow{\kappa}_\textrm{pr} )$ \\
        Lifted targets & $\K_{ZY}$ & $\K_{ZX}$ \\
        Residual covariance & $\sigma_Y^2 \, \overrightarrow{\B}$ & $\sigma_X^2 \, \overleftarrow{\B}$ \\
        Source-side second moment & $\widetilde{\C}_{XX} = \C_{XX} + \sigma_Y^2 \widetilde{\K}_{ZZ}$ & $\widetilde{\C}_{YY} = \C_{YY} + \sigma_X^2 \widetilde{\K}_{ZZ}$ \\
        Koopman matrix & $\overrightarrow{\U}_\kappa = \widetilde{\C}_{XX}^{-1} \C_{XY}$ & $\overleftarrow{\U}_\kappa = \widetilde{\C}_{YY}^{-1} \C_{YX}$ \\
        \addlinespace[4pt]
        Embedded PFO mean & $\overrightarrow{\mathcal{M}}_\textrm{pst}(\bm{x}) = \Phi_Y \K_{ZX}^\top \widetilde{\C}_{XX}^{-1} \mathbf{k}_Z(\bm{x})$ & $\overleftarrow{\mathcal{M}}_\textrm{pst}(\bm{y}) = \Phi_X \K_{ZY}^\top \widetilde{\C}_{YY}^{-1} \mathbf{k}_Z(\bm{y})$ \\
        \addlinespace[4pt]
        Posterior output covariance & $\overrightarrow{\B}_\kappa = ( N \sigma_Y^2 )^{-1} \, \widetilde{\C}_{YY \mid X}$ & $\overleftarrow{\B}_\kappa = ( N \sigma_X^2 )^{-1} \, \widetilde{\C}_{XX \mid Y}$ \\
        Pre-pairing covariance & $\overrightarrow{\Br}_\kappa = ( N \sigma_Y^2 )^{-1} \, \widetilde{\C}_{YY}$ & $\overleftarrow{\Br}_\kappa = ( N \sigma_X^2 )^{-1} \, \widetilde{\C}_{XX}$ \\
        \bottomrule
    \end{tabular}
    }
\end{table}

The directional sparse posteriors in Table~\ref{tab: forward backward models} induce posteriors over the corresponding random kernel Koopman operators $\overrightarrow{\mathcal U}:\mathbb H_Y\to\mathbb H_X$ and $\overleftarrow{\mathcal U}:\mathbb H_X\to\mathbb H_Y$. We use the inducing-mediated finite-rank operators whose actions on the respective dictionary spans are represented by $\overrightarrow{\U}$ and $\overleftarrow{\U}$. Their precise RKHS realization is specified in Appendix~\ref{subsec: Forward-Backward Moments}. We then define the source- and target-side round trip operators by
\begin{alignat}{3} \label{eq: Bayesian forward-backward operator}
    \overrightarrow{\mathcal T}
    &\coloneqq \overrightarrow{\mathcal U}\,\overleftarrow{\mathcal U}
    :\mathbb H_X\to\mathbb H_X,
    &\qquad
    \overleftarrow{\mathcal T}
    &\coloneqq \overleftarrow{\mathcal U}\,\overrightarrow{\mathcal U}
    :\mathbb H_Y\to\mathbb H_Y.
\end{alignat}
Each round trip is a product of two random Koopman operators and is therefore itself random, with a posterior induced by the joint posterior of the two operators. This joint posterior depends on the relation between the two directional models. The models share the regularized blocks of one joint second-moment matrix, through the opposite-direction prior kernels~\eqref{eq: output covariance prior} and the paired Schur complements~\eqref{eq: KYYX} and~\eqref{eq: KXXY}. This sharing fixes both directional posteriors but leaves the cross-covariance between the two random operators unspecified. 

As a working assumption, we treat the two operators as independent a priori and multiply their lifted likelihoods, although both are evaluated on the same transition pairs and the population operators are adjoint to each other~\eqref{eq: forward-backward pairing}. This assumption keeps both directional posteriors conjugate and in closed form, and it neglects only the cross-covariance between the two operators. Given this assumption, the joint posterior factorizes as
\begin{alignat*}{3}
    p_{\textrm{pst}}(\overrightarrow{\mathcal U}, \overleftarrow{\mathcal U})
    &=
    p_{\textrm{pst}}(\overrightarrow{\mathcal U}) \, p_{\textrm{pst}}(\overleftarrow{\mathcal U}).
\end{alignat*}

Independence then gives the posterior-mean compositions
\begin{alignat}{3} \label{eq: Bayesian forward-backward mean}
    \overrightarrow{\mathcal T}_\kappa
    &\coloneqq \mathbb E[\overrightarrow{\mathcal T}\mid\mathcal D]
    =\overrightarrow{\mathcal U}_\kappa\,\overleftarrow{\mathcal U}_\kappa,
    &\qquad
    \overleftarrow{\mathcal T}_\kappa
    &\coloneqq \mathbb E[\overleftarrow{\mathcal T}\mid\mathcal D]
    =\overleftarrow{\mathcal U}_\kappa\,\overrightarrow{\mathcal U}_\kappa.
\end{alignat}
Using the posterior-mean Koopman representations of~\eqref{eq: CME matrix} and Table~\ref{tab: forward backward models}, the source-side product is
\begin{alignat*}{3}
    \overrightarrow{\U}_\kappa\,\overleftarrow{\U}_\kappa
    &=\widetilde{\C}_{XX}^{-1}\C_{XY}\widetilde{\C}_{YY}^{-1}\C_{YX}
    =\I_{M_X}-\overleftarrow{\Br}_\kappa^{-1}\overleftarrow{\B}_\kappa.
\end{alignat*}
The target-side expression follows by symmetry, and both products share the same nonzero eigenvalues $\rho_i^2$, given by the squared GP-TCCA singular values of~\eqref{eq: kZ VAMP2 score}.\footnote{For a source eigenvector $\bm{\mathrm{w}}_i$ of $\overrightarrow{\U}_\kappa\overleftarrow{\U}_\kappa$, the residual $1-\rho_i^2 = (\bm{\mathrm{w}}_i^\top\widetilde{\C}_{XX\mid Y}\bm{\mathrm{w}}_i) / (\bm{\mathrm{w}}_i^\top\widetilde{\C}_{XX}\bm{\mathrm{w}}_i)$ is analogous to the posterior variance residual~\eqref{eq: Bayesian ResDMD residual} for Koopman eigenfunctions.} Since the trace of a finite-rank operator equals the trace of its dictionary representation, linearity and~\eqref{eq: Bayesian forward-backward mean} show that the posterior mean of the random trace recovers the VAMP-$2$ score,
\begin{alignat*}{3}
    \mathbb E\!\left[\operatorname{Tr}_{\mathbb H_X}(\overrightarrow{\mathcal T})\mid\mathcal D\right]
    &=\operatorname{Tr}(\overrightarrow{\U}_\kappa\overleftarrow{\U}_\kappa)=\mathcal S=\sum_i\rho_i^2.
\end{alignat*}
The posterior trace variance is derived in Appendix~\ref{subsec: Forward-Backward Moments} and assembled in~\eqref{eq: Bayesian forward-backward trace variance derivation}. Additionally it admits the modal decomposition
\begin{alignat}{3} \label{eq: Bayesian forward-backward trace variance}
    \operatorname{Var}\!\left[\operatorname{Tr}_{\mathbb H_X}(\overrightarrow{\mathcal T}) \mid \mathcal D\right]
    &=\frac{2}{N}\sum_i\rho_i^2(1-\rho_i^2)
      +\frac{(M_X-\mathcal S)(M_Y-\mathcal S)}{N^2}\notag\\
    &=\sum_{i=1}^{M_X}(1-\rho_i^2)
      \left[\frac{2\rho_i^2}{N}+\frac{M_Y-\mathcal S}{N^2}\right].
\end{alignat}

The posterior moments of $\operatorname{Tr}_{\mathbb H_X}(\overrightarrow{\mathcal T})$ quantify the epistemic uncertainty in the operator itself, but not the aleatoric variability of individual round trips. We therefore follow the forward--backward chain $X \rightarrow Y \rightarrow X'$, where $X'$ is a fresh draw from the backward transition law at $Y$, and take $f = \Phi_{Z_X} \bm{\mathrm{w}} \in \mathbb{H}_{X}$ in the dictionary span. Each leg of the chain is a one-step prediction of the sparse variational posterior. The backward leg predicts $f(X')$ from the successor with the conditional mean $(\overleftarrow{\mathcal U}_\kappa f)(\bm{y})$, and the forward leg predicts this conditional mean from the source state. Their variances are
\begin{alignat*}{3}
    \operatorname{Var}\!\left[ f(X') \mid Y = \bm{y}, \, \mathcal{D} \right]
    &=
    \big( \overleftarrow{\kappa}_\textrm{pst}(\bm{y}, \bm{y}) + \sigma_X^2 \big) \,
    \bm{\mathrm{w}}^\top \, \overleftarrow{\B}_\kappa \, \bm{\mathrm{w}},
    \\
    \operatorname{Var}\!\left[ (\overleftarrow{\mathcal U}_\kappa f)(Y) \mid X = \bm{x}, \, \mathcal{D} \right]
    &=
    \big( \overrightarrow{\kappa}_\textrm{pst}(\bm{x}, \bm{x}) + \sigma_Y^2 \big) \,
    ( \overleftarrow{\U}_\kappa \bm{\mathrm{w}} )^\top \, \overrightarrow{\B}_\kappa \, \overleftarrow{\U}_\kappa \bm{\mathrm{w}}.
\end{alignat*}
Here $\overrightarrow{\kappa}_\textrm{pst}$ is the sparse posterior kernel~\eqref{eq: posterior kernel VFE}, and $\overleftarrow{\kappa}_\textrm{pst}$ is its backward counterpart,
\begin{alignat*}{3}
    \overrightarrow{\kappa}_\textrm{pst}(\bm{x}, \bm{x})
    &=
    \overrightarrow{\kappa}_\textrm{pr}(\bm{x}, \bm{x})
    -
    \mathbf{k}_Z^\top(\bm{x}) \, \tilde{\Q}_{XX} \, \mathbf{k}_Z(\bm{x}),
    &\qquad
    \overleftarrow{\kappa}_\textrm{pst}(\bm{y}, \bm{y})
    &=
    \overleftarrow{\kappa}_\textrm{pr}(\bm{y}, \bm{y})
    -
    \mathbf{k}_Z^\top(\bm{y}) \, \tilde{\Q}_{YY} \, \mathbf{k}_Z(\bm{y}),
\end{alignat*}
with $\tilde{\Q}_{YY} \coloneqq ( \widetilde{\K}_{ZZ}^{-1} - \sigma_X^2 \widetilde{\C}_{YY}^{-1} )$. 

Since $X'$ depends on $X$ only through $Y$ and the directional posteriors are independent, the law of total variance splits the predictive variance into the expected conditional variance and the variance of the conditional mean,
\begin{alignat*}{3}
    \operatorname{Var}\!\left[ f(X') \mid X = \bm{x}, \, \mathcal{D} \right]
    =
    \mathbb{E}\!\left[ \operatorname{Var}\!\left( f(X') \mid Y, \, \mathcal{D} \right) \mid X = \bm{x}, \, \mathcal{D} \right]
    +
    \operatorname{Var}\!\left[ (\overleftarrow{\mathcal U}_\kappa f)(Y) \mid X = \bm{x}, \, \mathcal{D} \right].
\end{alignat*}
Inserting the one-step moments gives
\begin{alignat}{3}
    \operatorname{Var}\!\left[ f(X') \mid X = \bm{x}, \, \mathcal{D} \right]
    ={}&
    \big( \mathbb{E}\!\left[ \overleftarrow{\kappa}_\textrm{pst}(Y, \, Y) \mid X = \bm{x}, \, \mathcal{D} \right] + \sigma_X^2 \big) \,
    \bm{\mathrm{w}}^\top \, \overleftarrow{\B}_\kappa \, \bm{\mathrm{w}} \notag
    \\
    &+
    \big( \overrightarrow{\kappa}_\textrm{pst}(\bm{x}, \, \bm{x}) + \sigma_Y^2 \big) \,
    ( \overleftarrow{\U}_\kappa \bm{\mathrm{w}} )^\top \, \overrightarrow{\B}_\kappa \, \overleftarrow{\U}_\kappa \bm{\mathrm{w}}. \label{eq: Bayesian forward-backward predictive variance}
\end{alignat}
This is the round-trip counterpart of the recursion in~\eqref{eq: GP DMD cov recursive}, with both residual variances treated entirely as transition variability. The first term injects the expected one-step covariance of the backward leg, and the second propagates the one-step covariance of the forward leg through $\overleftarrow{\U}_\kappa^\top$. Similar to the multi-step forecast in Section~\ref{sec: Prediction Theory}, the expectation in the first term is in general not analytically tractable.

\section{Numerical Experiments} \label{sec: Numerical Experiments}

In all subsequent experiments, we used the $\text{Matérn}^{5/2}$ kernel \citep{matern2013spatial} with automatic relevance determination (ARD; \citet{williams2006gaussian}):
\begin{alignat*}{3}
    \kappa_\textrm{pr}(\bm{x}, \bm{x}')
    = \sigma_f^2 \left( 1 + \sqrt{5} \, r + \frac{5}{3} \, r^2 \right) \exp\!\left( - \sqrt{5} \, r \right),
\end{alignat*}
where $r^2 = (\bm{x} - \bm{x}')^\top \mathbf{T}^{-1} (\bm{x} - \bm{x}')$ and the diagonal matrix $\mathbf{T} \succ 0$ contains the squared \emph{characteristic lengthscales}. The lengthscales determine how far the input must move along each axis for the function values to become uncorrelated, and the \emph{signal variance} $\sigma_f^2$ sets the average squared distance of the regression function from its mean.

The initial conditions of the training trajectories were drawn from an optimized Latin hypercube design \citep{bates2004formulation, urquhart2020surrogate} over a box in the state space, and each trajectory was rolled out for a fixed number of steps. All data were standardized, and the hyperparameters were selected as described in Section~\ref{sec: Model Selection}. The simulated data contain no measurement noise. We therefore set $\sigma_\textrm{proc}^2 = \sigma_Y^2$ and $\sigma_\textrm{meas}^2 = 0$ throughout, which treats the residual variance entirely as transition variability. 

Figure~\ref{fig: Bayesian overview} arranges data-driven transfer operator methods by the dynamics they address, and the five benchmarks cover all three of its columns. The Duffing oscillator (Section~\ref{sec: Duffing Oscillator}), the van der Pol oscillator (Section~\ref{sec: van der Pol Oscillator}), and the Lorenz system (Section~\ref{sec: Lorenz System}) are time-homogeneous and nonreversible, and they fall into the middle column. The stochastic quadruple well (Section~\ref{sec: Stochastic Quadruple Well}) is reversible and also falls into the left column. The Bickley jet (Section~\ref{sec: Bickley Jet}) is time-inhomogeneous and falls into the right column.

\vspace{0.25\baselineskip}
\noindent\begin{minipage}{\columnwidth}
    \centering
    \includegraphics[width=0.99\columnwidth]{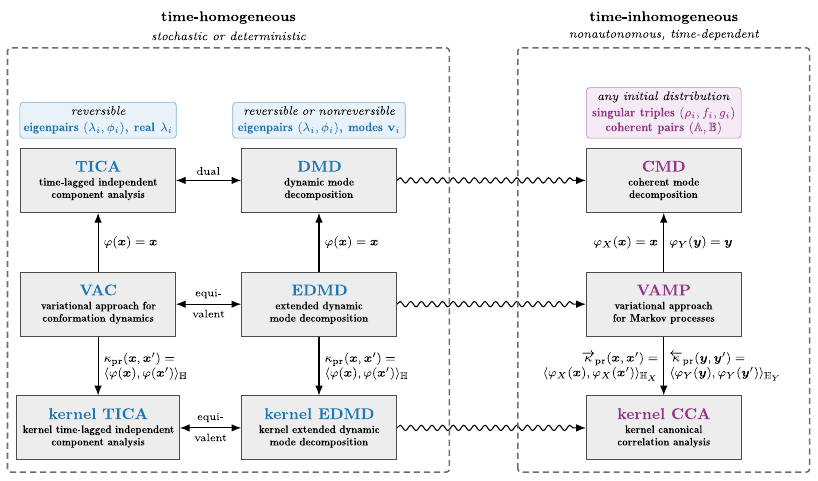}
    \captionsetup{hypcap=false}
    \phantomsection
    \captionof{figure}{\emph{Data-driven transfer operator methods}, adapted from \citet{klus2019kernel}. From top to bottom, the rows represent the state by its coordinates, by a dictionary of basis functions, and by a reproducing kernel. The posterior means of the Bayesian transfer operators recover the regularized kernel estimators.}
    \label{fig: Bayesian overview}
    \vspace{0.0\baselineskip}
\end{minipage}

\vspace{-0.5\baselineskip}
\subsection{Duffing Oscillator} \label{sec: Duffing Oscillator}

The first benchmark is the Duffing oscillator driven by additive noise,
\begin{alignat*}{3}
    \begin{split}
        \mathrm{d}x_{t,1} &= x_{t,2} \, \mathrm{d}t + \sigma_1 \, \mathrm{d}W_{t,1},\\
        \mathrm{d}x_{t,2} &= \left( - \delta \, x_{t,2} - \alpha \, x_{t,1} - \beta \, x_{t,1}^3 \right) \mathrm{d}t + \sigma_2 \, \mathrm{d}W_{t,2},
    \end{split}
\end{alignat*}
where $W_{t,1}$ and $W_{t,2}$ are two independent standard Wiener processes. The damping is $\delta = 0.5$, the linear and cubic stiffness coefficients are $\alpha = 1$ and $\beta = 5$, and the noise intensities are $\sigma_1 = 0.08$ and $\sigma_2 = 0.26$. We chose the noise intensities such that the Bayesian-consistency ratio~\eqref{eq: directional consistency ratio} of each state coordinate is close to one. The isotropic residual model of Section~\ref{sec: Sampling and the Flow Map} then approximately holds for the standardized data. The system was simulated by a numerical SDE solver and sampled in time intervals of $\Delta t = 50 \text{ [ms]}$. The training set consists of $N = 10\ts000$ one-step pairs from $500$ trajectories of $20$ steps, with initial conditions in $[-2.5, 2.5] \times [-5, 5]$, and the learned dictionary contains $M = 19$ pseudo-inputs.

Figure~\ref{fig: Duffing trajectory} shows a forecast over $100$ steps from $\bm{x}_0 = [-2, \, 4.5]^\top$, compared with the mean and spread of $500$ sample paths of the SDE. Instead of following the spectral recursion of Section~\ref{sec: Prediction Theory}, the forecast re-lifts the state at every step. Given the Gaussian approximation $X_{k-1} \sim \mathcal{N}^D(\hat{\bm{x}}_{k-1}, \K_\textrm{pst}^{k-1})$, the mean and covariance of the lifted coordinates were evaluated by quadrature with six nodes per dimension. One step of the learned dynamics, reconstructed with $\A_{\bm{F}}^0$ and using $\A_{\bm{F}}^0 \A_\kappa = \A_{\bm{F}}^1$, then gives
\begin{alignat*}{3}
    \hat{\bm{x}}_k
    &=
    \A_{\bm{F}}^1 \, \mathbb{E}\!\left[ \mathbf{k}_Z(X_{k-1}) \right],
    \\
    \K_\textrm{pst}^{k}
    &=
    \A_{\bm{F}}^1 \operatorname{Cov}\!\left[ \mathbf{k}_Z(X_{k-1}) \right] (\A_{\bm{F}}^1)^\top
    +
    \tilde{\kappa}_{k-1} \, \A_{\bm{F}}^0 \, \overrightarrow{\B}_\kappa \, (\A_{\bm{F}}^0)^\top .
\end{alignat*}
Under the Bayesian-consistency principle $\A_{\bm{F}}^0 \, \overrightarrow{\B}_\kappa \, (\A_{\bm{F}}^0)^\top = \I_D$, and consequently the second term reduces to $(\bar{\kappa}_{k-1} + \sigma_\textrm{proc}^2) \, \I_D$. Since $\bm{x} \mapsto \A_{\bm{F}}^1 \, \mathbf{k}_Z(\bm{x})$ is the posterior mean of the flow map, the forecast iterates the state-space model with uncertain inputs. Its covariance is the law of total variance that \citet{girard2002gaussian} use for multiple-step-ahead GP forecasts, i.e., the covariance of the posterior mean plus the expected one-step predictive variance, here evaluated by Gauss--Hermite quadrature.

\vspace{0.25\baselineskip}
\noindent\begin{minipage}{\columnwidth}
    \centering
    \includegraphics[trim={2 12 55 1}, clip, width=0.85\columnwidth]{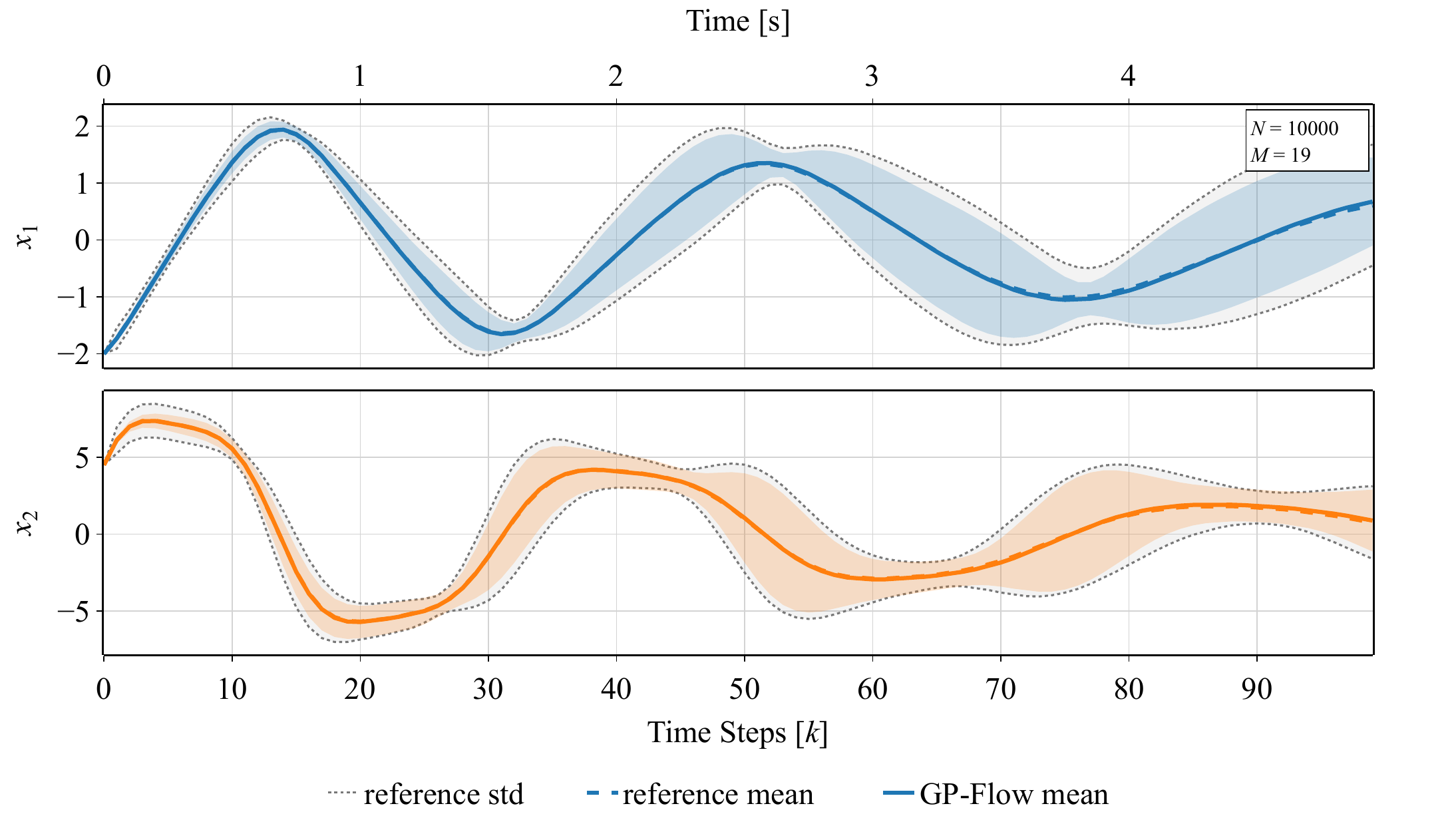}
    \captionsetup{hypcap=false}
    \phantomsection
    \captionof{figure}{\emph{Forecast of the stochastic Duffing oscillator}. The forecast re-lifts the state at every step and thereby iterates the posterior mean of the flow map. Solid lines show the GP-Flow mean and shaded regions three posterior standard deviations. Dashed lines show the reference mean of $500$ sample paths of the SDE and dotted lines three standard deviations of these paths.}\label{fig: Duffing trajectory}
    \vspace{0.25\baselineskip}
\end{minipage}

Figure~\ref{fig: Duffing spectrum} shows the eigenvalues of the learned Koopman matrix by frequency and magnitude, with complex-conjugate pairs displayed once at their positive frequency and the trivial eigenvalue $\lambda_1 = 1$ omitted. In the left panel, colored by the posterior Koopman residual~\eqref{eq: posterior Koopman residual}, the real eigenvalue with $|\lambda_i| \approx 0.97$ has the smallest residual, and the residuals tend to grow as $|\lambda_i|$ decreases. In the right panel, colored by the Bayesian-consistency ratio~\eqref{eq: directional consistency ratio} at the eigenvectors, all ratios lie at or above one, i.e., the residuals observed by the lifted model are at least as large as those predicted by the state-space model. Interestingly, the largest deviation, a ratio of about $1.4$, occurs at the eigenvalue with the smallest residual. Since the ratio compares the two models, such deviations signal a mismatch without identifying which model is at fault.

\vspace{0.75\baselineskip}
\noindent\begin{minipage}{\columnwidth}
    \centering
    \begin{minipage}{0.49\columnwidth}
        \centering
        \includegraphics[trim={2 2 15 10}, clip, width=\linewidth]{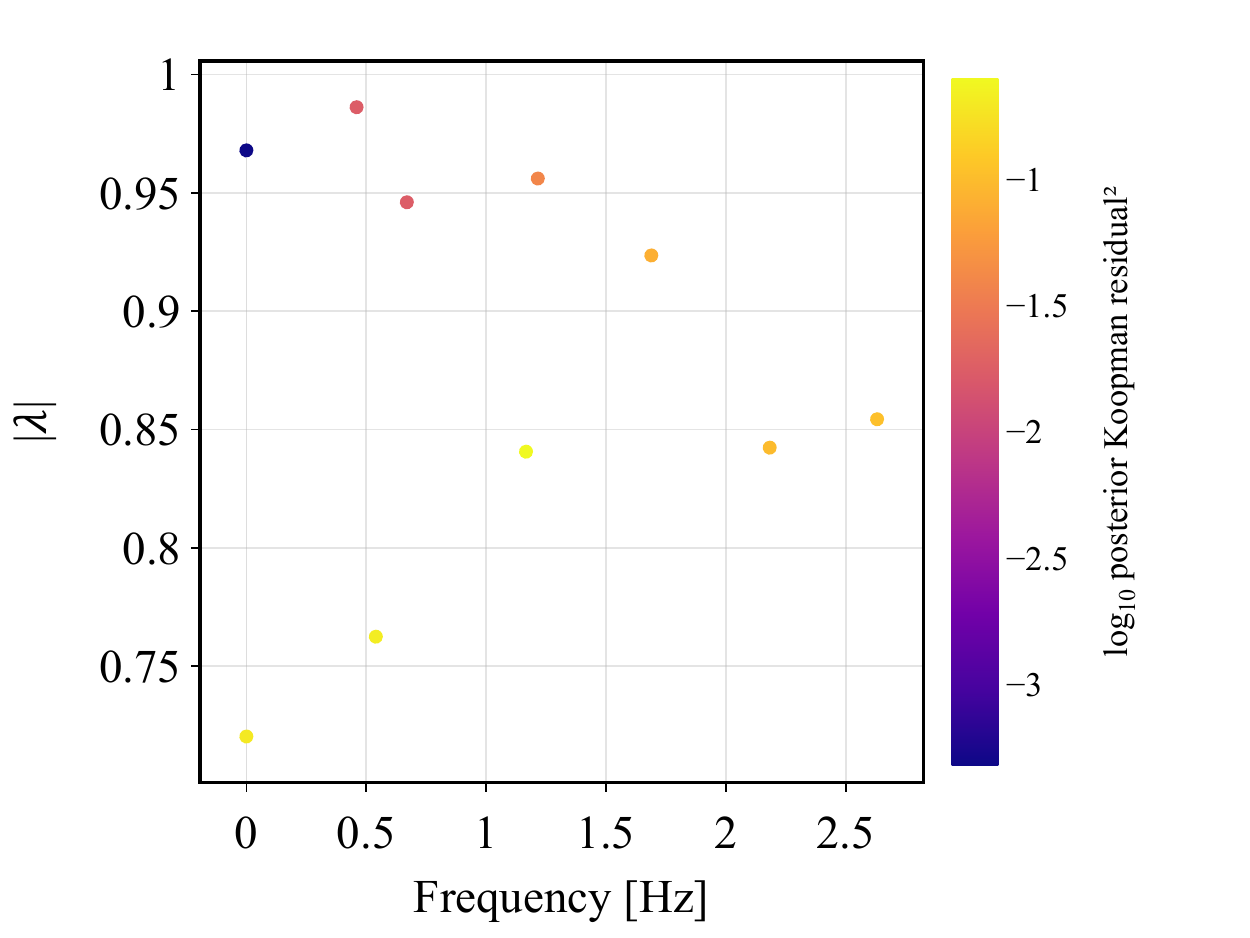}
    \end{minipage}
    \hfill
    \begin{minipage}{0.49\columnwidth}
        \centering
        \includegraphics[trim={2 2 15 10}, clip, width=\linewidth]{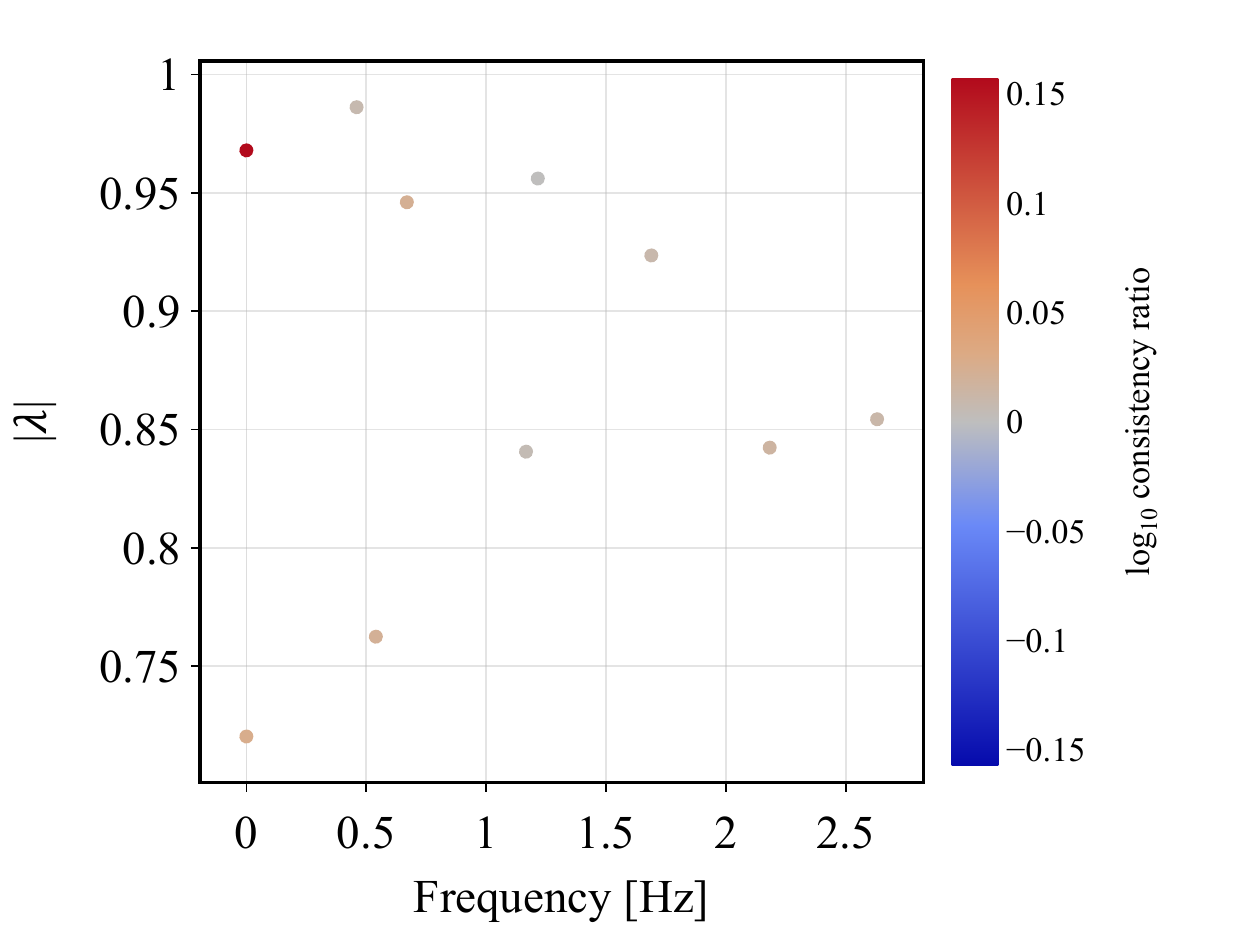}
    \end{minipage}
    \captionsetup{hypcap=false}
    \phantomsection
    \captionof{figure}{\emph{Spectral diagnostics of the Duffing oscillator}. Eigenvalues by frequency and magnitude $|\lambda_i|$, colored by the posterior Koopman residual (left, $\log_{10}$ scale) and by the Bayesian-consistency ratio at the eigenvector (right, $\log_{10}$ scale, centered at one).}\label{fig: Duffing spectrum}
\end{minipage}
\vspace{0.75\baselineskip}

\vspace{-0.5\baselineskip}
\subsection{Van der Pol Oscillator} \label{sec: van der Pol Oscillator}

The second benchmark is the deterministic van der Pol oscillator,
\begin{alignat*}{3}
    \begin{split}
        \dot{x}_{t,1} &= x_{t,2},\\
        \dot{x}_{t,2} &= \alpha \left( 1 - x_{t,1}^2 \right) x_{t,2} - x_{t,1},
    \end{split}
\end{alignat*}
where $\alpha = 2$ controls the degree of nonlinearity. Its trajectories converge to a limit cycle that alternates between slow segments and fast relaxation jumps. The system was simulated by a numerical ODE solver and sampled in time intervals of $\Delta t = 50 \text{ [ms]}$. The training set consists of $N = 15\ts000$ one-step pairs from $750$ trajectories of $20$ steps, with initial conditions in $[-3.5, 3.5] \times [-6, 6]$.

\vspace{0.25\baselineskip}
\noindent\begin{minipage}{\columnwidth}
    \centering
    \includegraphics[trim={2 12 55 1}, clip, width=0.75\columnwidth]{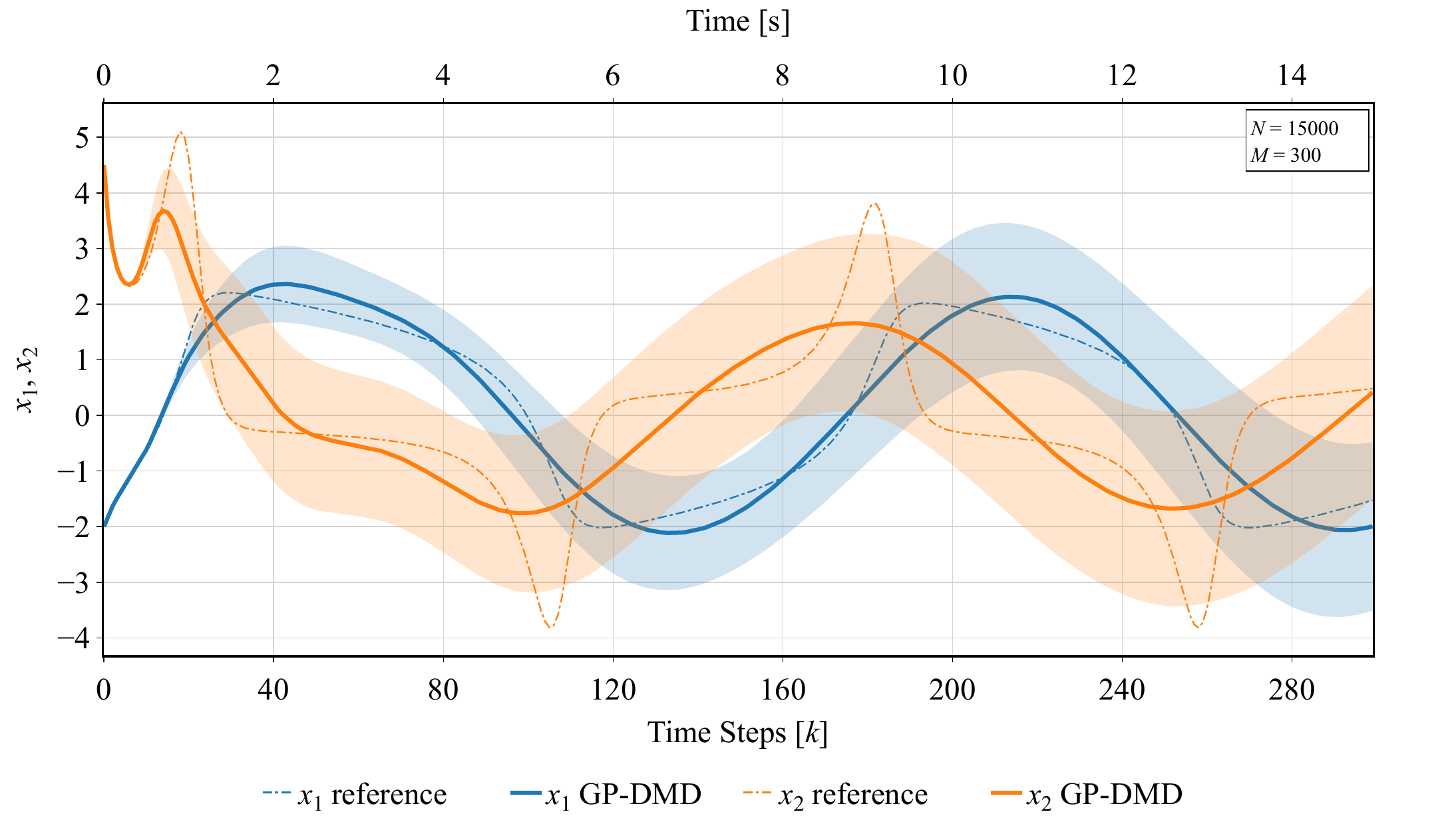}
    \captionsetup{hypcap=false}
    \phantomsection
    \captionof{figure}{\emph{Spectral forecast of the van der Pol oscillator}. The state is lifted once and propagated by the spectral recursion. Solid lines show the GP-DMD mean and shaded regions three posterior standard deviations. Dash-dotted lines show the reference trajectory.}\label{fig: vdP trajectory}
    \vspace{0.25\baselineskip}
\end{minipage}

Figure~\ref{fig: vdP trajectory} shows a forecast over $300$ steps from $\bm{x}_0 = [-2, \, 4.5]^\top$, obtained with a model whose $M = 300$ pseudo-inputs were initialized by a Latin hypercube design and then optimized with analytical gradients. Unlike the Duffing forecast, which re-lifts the state at every step, this forecast lifts the initial state only once and then propagates the mean and covariance with the spectral recursions~\eqref{eq: GP DMD} and~\eqref{eq: GP DMD cov recursive}. Since the van der Pol oscillator has a limit cycle, no finite-dimensional Koopman-invariant subspace contains the state \citep{10.1371/journal.pone.0150171}. Even an exactly invariant dictionary space would then reconstruct the state only approximately through the linear decoder $\A_{\bm{F}}^0$ of Section~\ref{sec: Bayesian Consistency}.

To examine the dictionary selection cascade of Appendix~\ref{sec: Dictionary Learning}, we trained four further models on the same data. Each model adds one stage to the previous one, and the fourth additionally optimizes all pseudo-inputs with analytical gradients. The stages are ordered from the least to the most computationally expensive. Every run used the fully adaptive version of the algorithm: the hyperparameters were optimized throughout the dictionary selection, and the algorithm selected the dictionary size $M$ from the data rather than taking it as prescribed. 

Figure~\ref{fig: vdP dictionary} shows the resulting dictionaries. With the approximate-linear-dependence (ALD) criterion~\eqref{eq: ALD residual} alone, the $M = 86$ pseudo-inputs spread almost uniformly over the sampled box, since the criterion only removes candidates that the current dictionary already explains. However, ALD alone admits the candidate with the largest residual first, which is often an outlier. Combining ALD with our information-gain (IG) criterion~\eqref{eq: joint information gain} enlarges the dictionary to $M = 100$. IG instead favors candidates that are strongly correlated with the other survivors and demotes isolated candidates, which are then admitted only after the more central pseudo-inputs. Since selection stops only once every residual falls below the ALD tolerance, the dictionary ends up larger. The third stage adds the target-aware VFE criterion of Appendix~\ref{subsec: VFE criterion} and produces $M = 102$ pseudo-inputs, with noticeably fewer of them close to the boundary. In the final and most computationally expensive stage (VFE$^\ast$), all pseudo-inputs are additionally optimized with the analytical gradients of Appendix~\ref{sec: Gradient Derivations}. The dictionary then grows to $M = 117$, and a large share of the pseudo-inputs moves into a narrow band around $x_{2} = 0$, where the trajectories reverse their horizontal direction and the slow segments of the limit cycle lie.

\vspace{0.5\baselineskip}
\noindent\begin{minipage}{\columnwidth}
    \centering
    \begin{minipage}{0.4\columnwidth}
        \centering
        \includegraphics[trim={5 8 55 18}, clip, width=\linewidth]{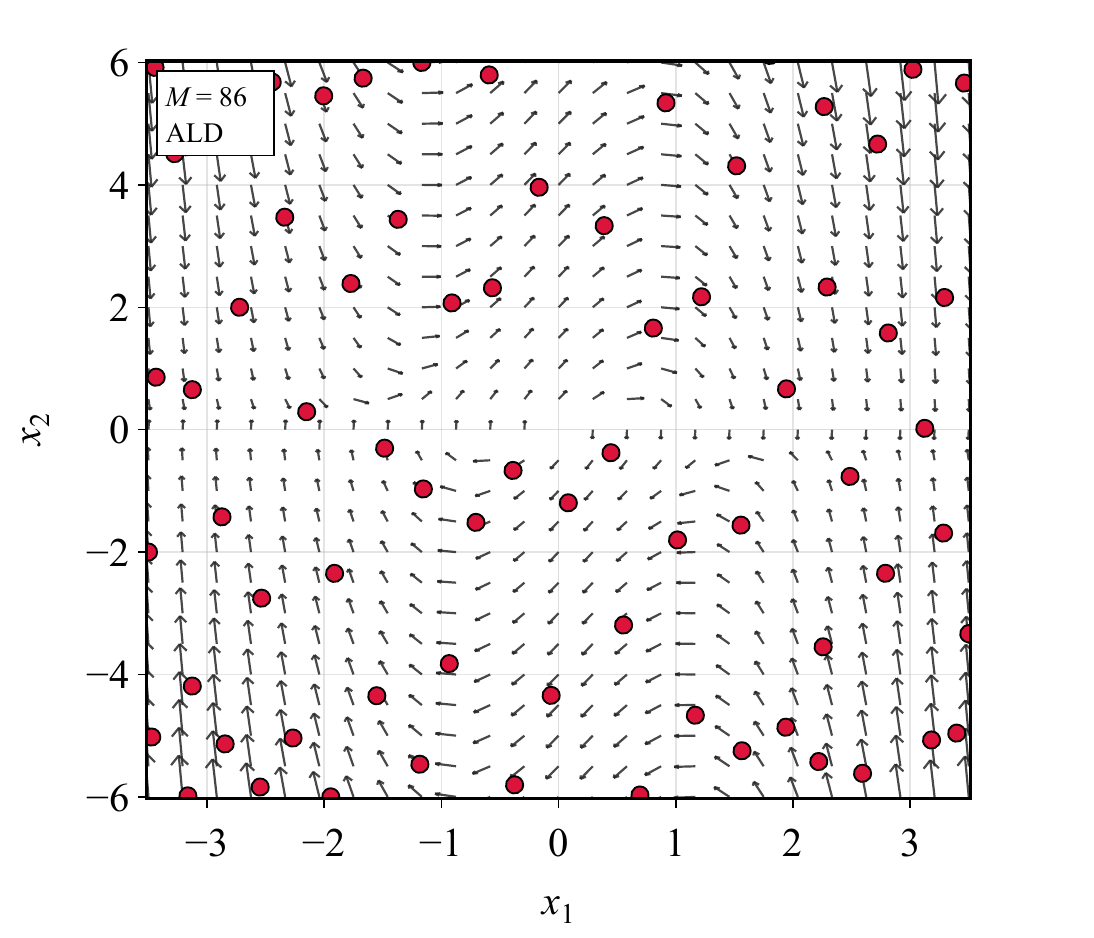}
    \end{minipage}
    \hspace{0.02\columnwidth}
    \begin{minipage}{0.4\columnwidth}
        \centering
        \includegraphics[trim={5 8 55 18}, clip, width=\linewidth]{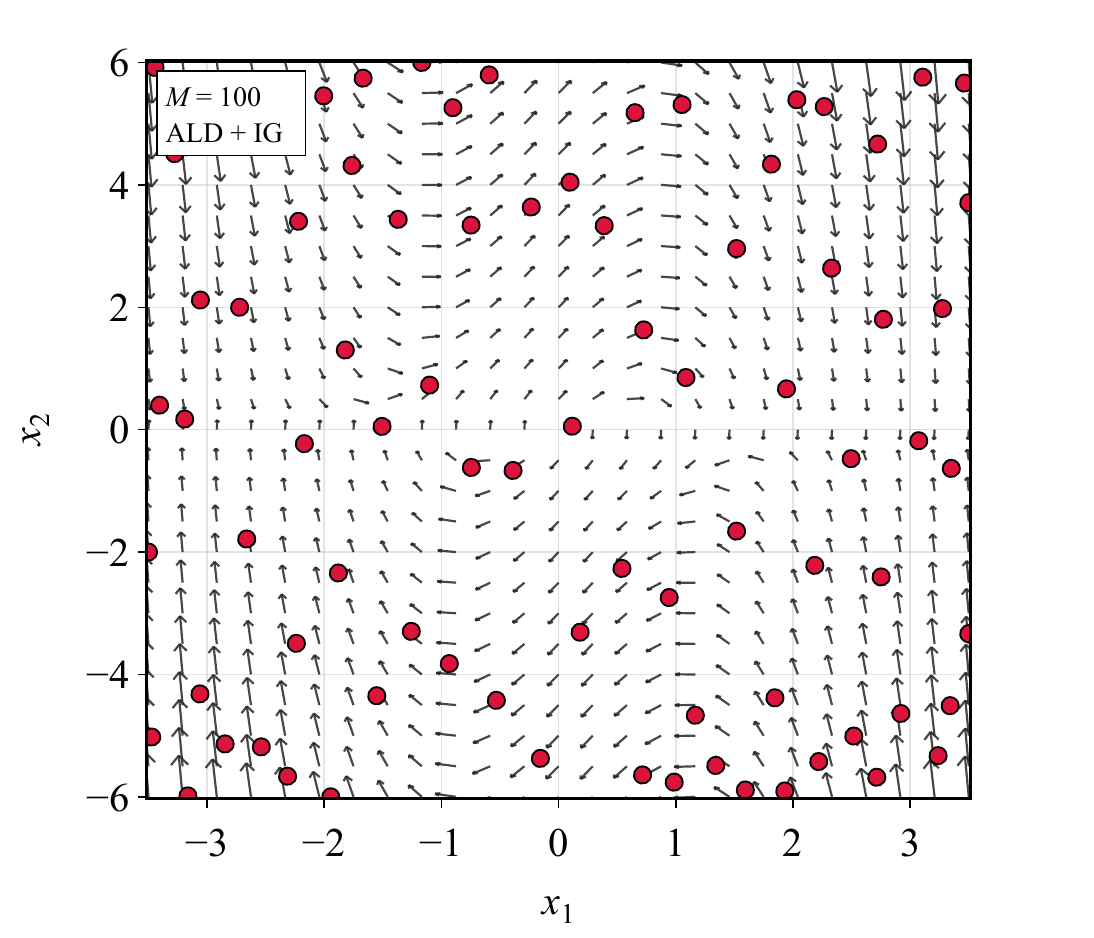}
    \end{minipage}

    \vspace{0.5\baselineskip}

    \begin{minipage}{0.4\columnwidth}
        \centering
        \includegraphics[trim={5 8 55 18}, clip, width=\linewidth]{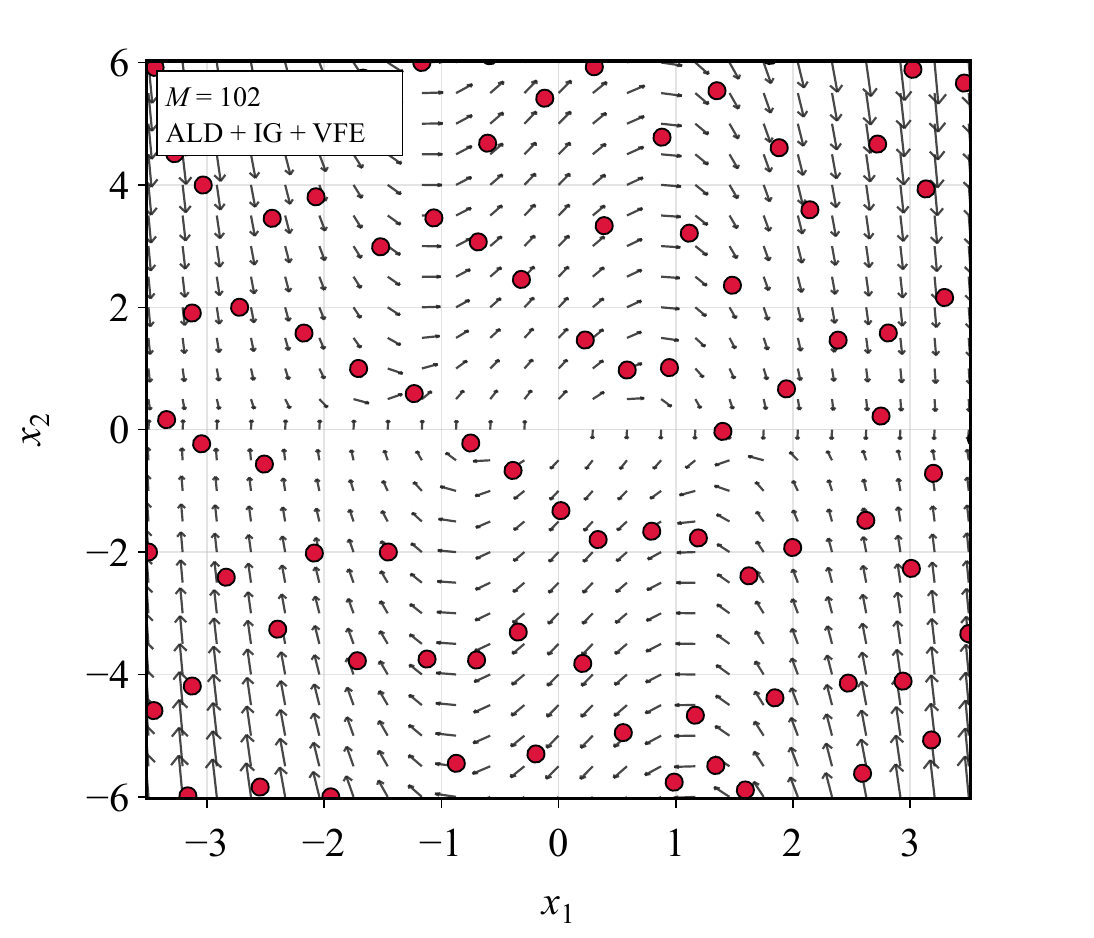}
    \end{minipage}
    \hspace{0.02\columnwidth}
    \begin{minipage}{0.4\columnwidth}
        \centering
        \includegraphics[trim={5 8 55 18}, clip, width=\linewidth]{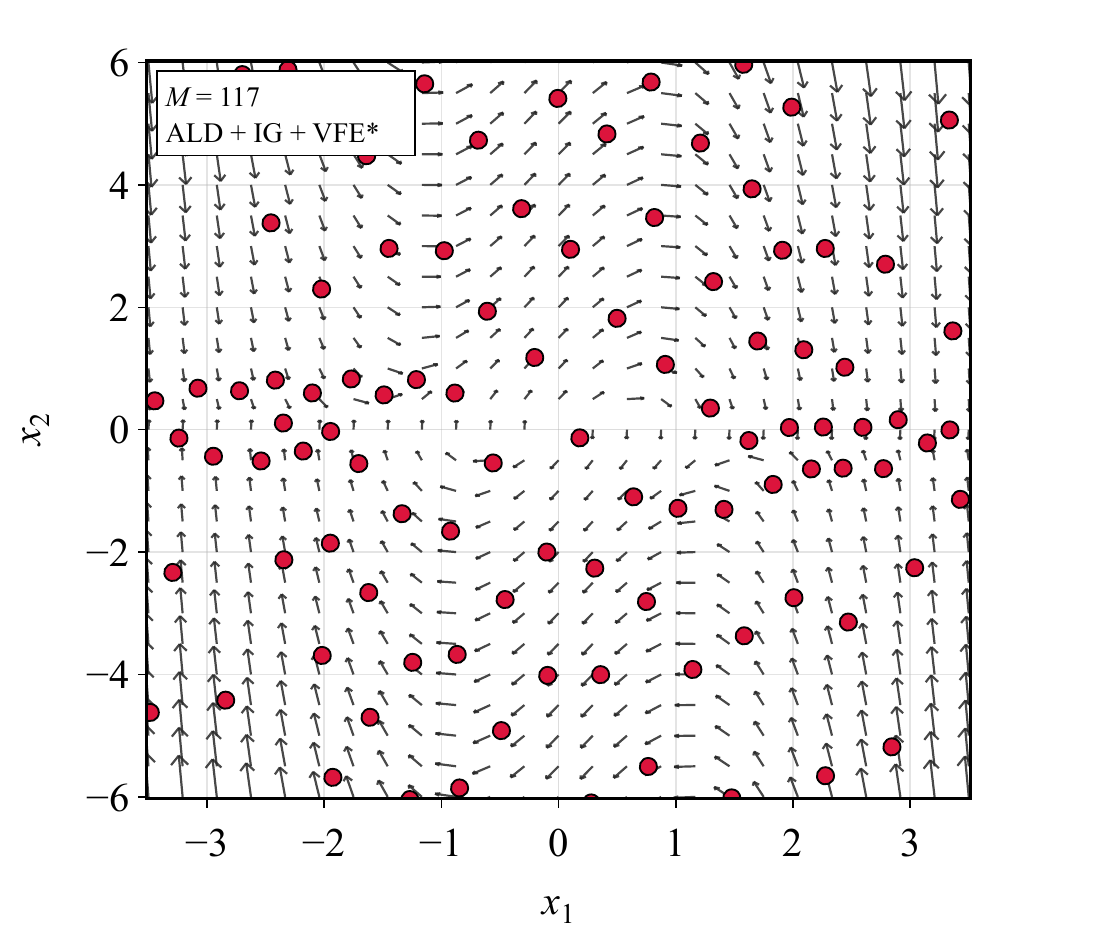}
    \end{minipage}
    \captionsetup{hypcap=false}
    \phantomsection
    \captionof{figure}{\emph{Dictionary learning for the van der Pol oscillator}. Pseudo-inputs (red markers) selected by the ALD criterion (top left), by ALD and IG (top right), by ALD, IG, and VFE (bottom left), and with all pseudo-inputs additionally optimized by gradients (VFE$^\ast$, bottom right). Arrows depict the direction of the one-step displacement, with lengths ordered by its magnitude.}\label{fig: vdP dictionary}
    \vspace{0.5\baselineskip}
\end{minipage}

\subsection{Lorenz System} \label{sec: Lorenz System}

The third benchmark is the chaotic Lorenz system,
\begin{alignat*}{3}
    \begin{split}
        \dot{x}_{t,1} &= \sigma \left( x_{t,2} - x_{t,1} \right),\\
        \dot{x}_{t,2} &= x_{t,1} \left( \rho - x_{t,3} \right) - x_{t,2},\\
        \dot{x}_{t,3} &= x_{t,1} \, x_{t,2} - \beta \, x_{t,3},
    \end{split}
\end{alignat*}
with the parameters $\sigma = 10$, $\rho = 28$, and $\beta = 8 / 3$ used by \citet{brunton2017chaos}. The system was simulated by a numerical ODE solver with a step size of $1 \text{ [ms]}$ and sampled in time intervals of $\Delta t = 10 \text{ [ms]}$. The training set consists of $N = 10\ts000$ one-step pairs from $500$ trajectories of $20$ steps, with initial conditions in $[-20, 20] \times [-27, 27] \times [0, 50]$, and the learned dictionary contains $M = 300$ pseudo-inputs.

The Duffing forecast lifts the state at every step, whereas the van der Pol forecast lifts it only once. Reprojection interpolates between these two extremes. The learned dictionary space is in general not invariant under the Koopman operator, and the propagated coordinates ${\mathbf{k}}_{Z,k}$ of~\eqref{eq: LSDE} therefore drift away from the image of the feature map $\mathbf{k}_Z$. A reprojection maps them back onto this image \citep{van2023reprojection}. We triggered a reprojection once the root-mean-square standard deviation of the standardized state forecast exceeded a predefined tolerance,
\begin{alignat*}{3}
    \sqrt{ \operatorname{Tr}\big( \K_\textrm{pst}^{k} \big) / D } > \tau_\textrm{proj}.
\end{alignat*}
In our experiments, we set $\tau_\textrm{proj} = 0.02$. 

A reprojection replaces the propagated coordinates ${\mathbf{k}}_{Z,k}$ by the expected feature vector $\mathbb{E}[\mathbf{k}_Z(X_k)]$ under the Gaussian approximation $X_k \sim \mathcal{N}^D(\hat{\bm{x}}_k, \K_\textrm{pst}^{k})$. This expectation reuses the Gauss--Hermite nodes of $\bar{\kappa}_k$, with three nodes per dimension. The propagated covariance ${\mathbf{\Xi}}_{\textrm{pst}}^{k}$ is then reset to zero, such that the forecast restarts from the one-step covariance. Without this reset, the Gaussian approximation would widen over the chaotic forecast, and each reprojection would average the features over an ever larger region. The mean would then approach the ensemble mean, which for chaotic dynamics tends to an average over the attractor. The reset keeps the reprojected mean close to the lift of a single state, at the cost that the propagated covariance then measures only the uncertainty accumulated since the last reprojection. Between reprojections, the forecast follows the spectral recursion of Section~\ref{sec: Prediction Theory}.

\vspace{0.5\baselineskip}
\noindent\begin{minipage}{\columnwidth}
    \centering
    \includegraphics[trim={113 38 51 416}, clip, width=0.7\columnwidth]{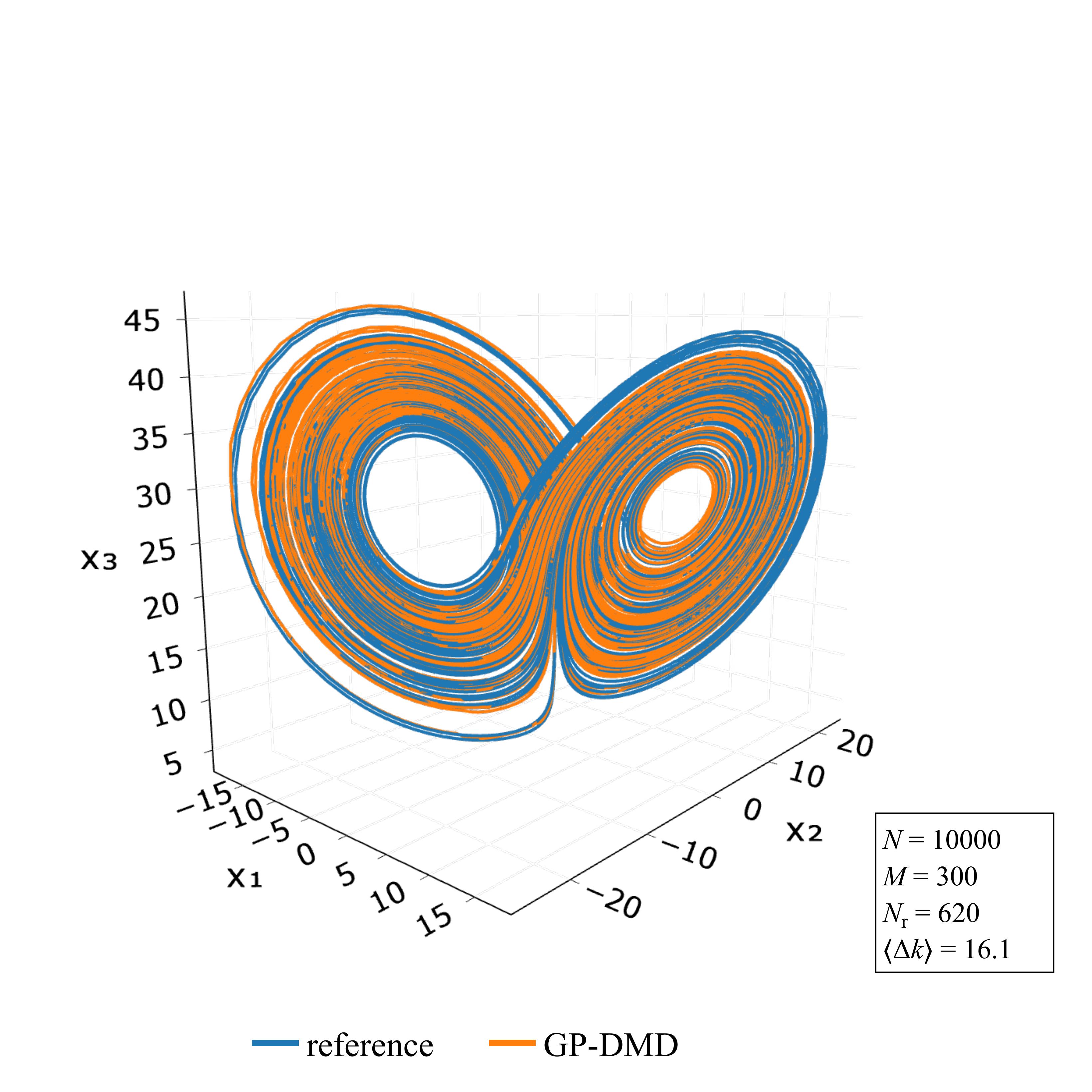}
    \captionsetup{hypcap=false}
    \phantomsection
    \captionof{figure}{\emph{Forecast of the Lorenz system}. GP-DMD forecast (orange) over $10\ts000$ steps with uncertainty-triggered reprojections, and the reference trajectory (blue). The annotation reports the number of one-step pairs $N$, the dictionary size $M$, the number $N_\textrm{r}$ of triggered reprojections, which excludes the initial lifting, and the average number $\langle \Delta k \rangle$ of steps between consecutive reprojections.}\label{fig: Lorenz}
    \vspace{0.5\baselineskip}
\end{minipage}

Figure~\ref{fig: Lorenz} shows a forecast over $10\ts000$ steps, i.e., $100 \text{ [s]}$, against the reference trajectory. The initial condition lies on the attractor and was obtained by integrating the system for $10 \text{ [s]}$ from $[-8, \, 8, \, 27]^\top$. Since the dynamics are chaotic, the forecast and the reference trajectory separate after a short time, and a pointwise comparison over this horizon carries little meaning. Instead, the forecast traces the same two-lobed geometry as the reference trajectory, i.e., the model reproduces the shape of the attractor.

\subsection{Stochastic Quadruple Well} \label{sec: Stochastic Quadruple Well}

The fourth benchmark is the overdamped Langevin equation
\begin{alignat*}{3}
    \begin{split}
        \mathrm{d}{x}_{t,1} &= -\nabla_{x_1} V(\bm{x}_t) \, \mathrm{d}t + \sigma_T \; \mathrm{d}W_{t,1},\\
        \mathrm{d}{x}_{t,2} &= -\nabla_{x_2} V(\bm{x}_t) \, \mathrm{d}t + \sigma_T \; \mathrm{d}W_{t,2},
    \end{split}
\end{alignat*}
with the quadruple-well potential
\begin{alignat*}{3}
    V(\bm{x}) = \left( x_1^2 - 1 \right)^2 + \left( x_2^2 - 1 \right)^2,
\end{alignat*}
where $W_{t,1}$ and $W_{t,2}$ are two independent standard Wiener processes. The four minima of $V$ at $(\pm 1, \pm 1)$ are the metastable states of the system. A particle lingers near one minimum and rarely crosses a barrier into a neighboring well. As in \citet{klus2018kernel}, we fixed the noise intensity to $\sigma_T = 1 / \sqrt{2}$ and sampled the system at intervals of $\Delta t = 10 \text{ [s]}$. The system was simulated with the Euler--Maruyama scheme at a step size of $1 \text{ [ms]}$. The training set consists of $N = 10\ts000$ one-step pairs from $100$ trajectories of $100$ steps, with initial conditions in $[-2, 2]^2$. The dictionary was initialized with $M = 4$ pseudo-inputs and optimized with analytical gradients. This size matches the four dominant eigenvalues of the system, which are separated from the remaining spectrum by a spectral gap \citep{klus2018kernel}.

Figure~\ref{fig: quadruple well eigenfunctions} shows the eigenfunctions $\phi_2$ and $\phi_3$ propagated over $k = 1$ step and $\phi_4$ propagated over $k = 1$ and $k = 5$ steps, all normalized to unit mean square over the training inputs. The surface height depicts the forecast mean of $\phi_i(X_k)$ given $X_0 = \bm{x}$, and the colour depicts its standard deviation, obtained from the $k$-step covariance $\W^\top \, \mathbf{\Xi}_{\textrm{pst}}^{k}(\bm{x}) \, (\W^\top)^*$ of Section~\ref{sec: Prediction Theory} with three quadrature nodes per dimension.

\vspace{0.5\baselineskip}
\noindent\begin{minipage}{\columnwidth}
    \centering
    \begin{minipage}{0.45\columnwidth}
        \centering
        \includegraphics[trim={111 124 12 109}, clip, width=\linewidth]{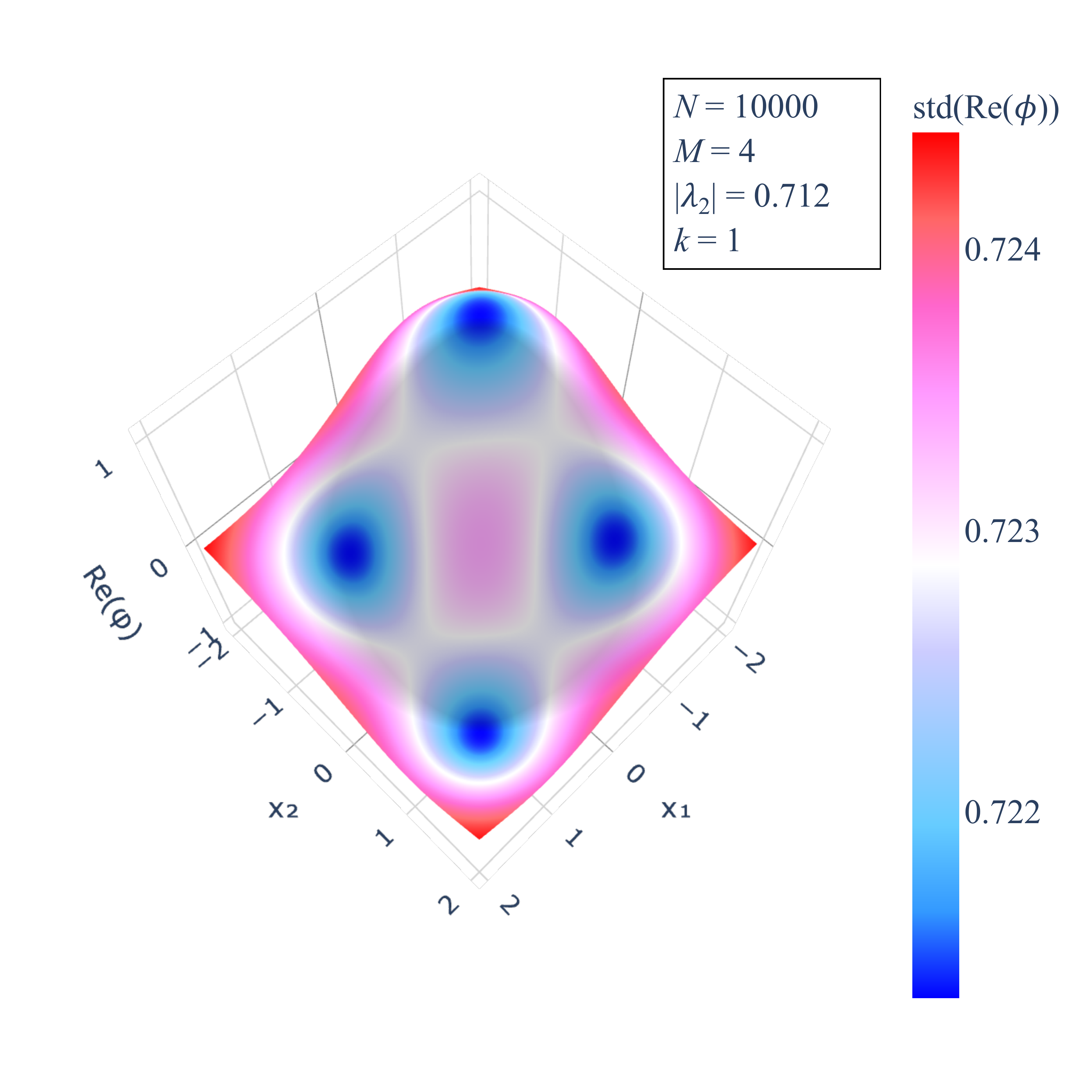}
    \end{minipage}
    \hspace{0.02\columnwidth}
    \begin{minipage}{0.45\columnwidth}
        \centering
        \includegraphics[trim={111 124 12 109}, clip, width=\linewidth]{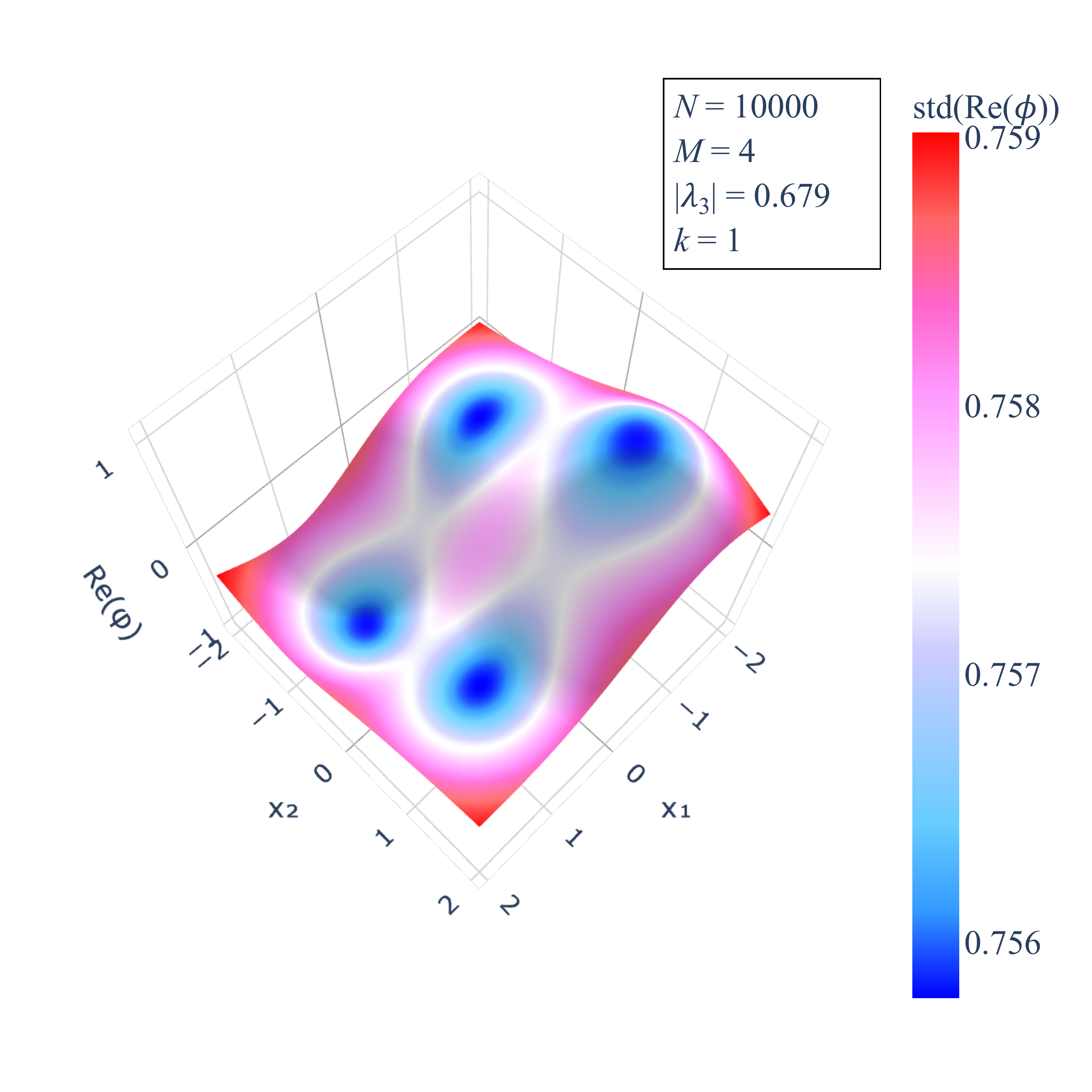}
    \end{minipage}

    \vspace{0.5\baselineskip}

    \begin{minipage}{0.45\columnwidth}
        \centering
        \includegraphics[trim={111 124 12 109}, clip, width=\linewidth]{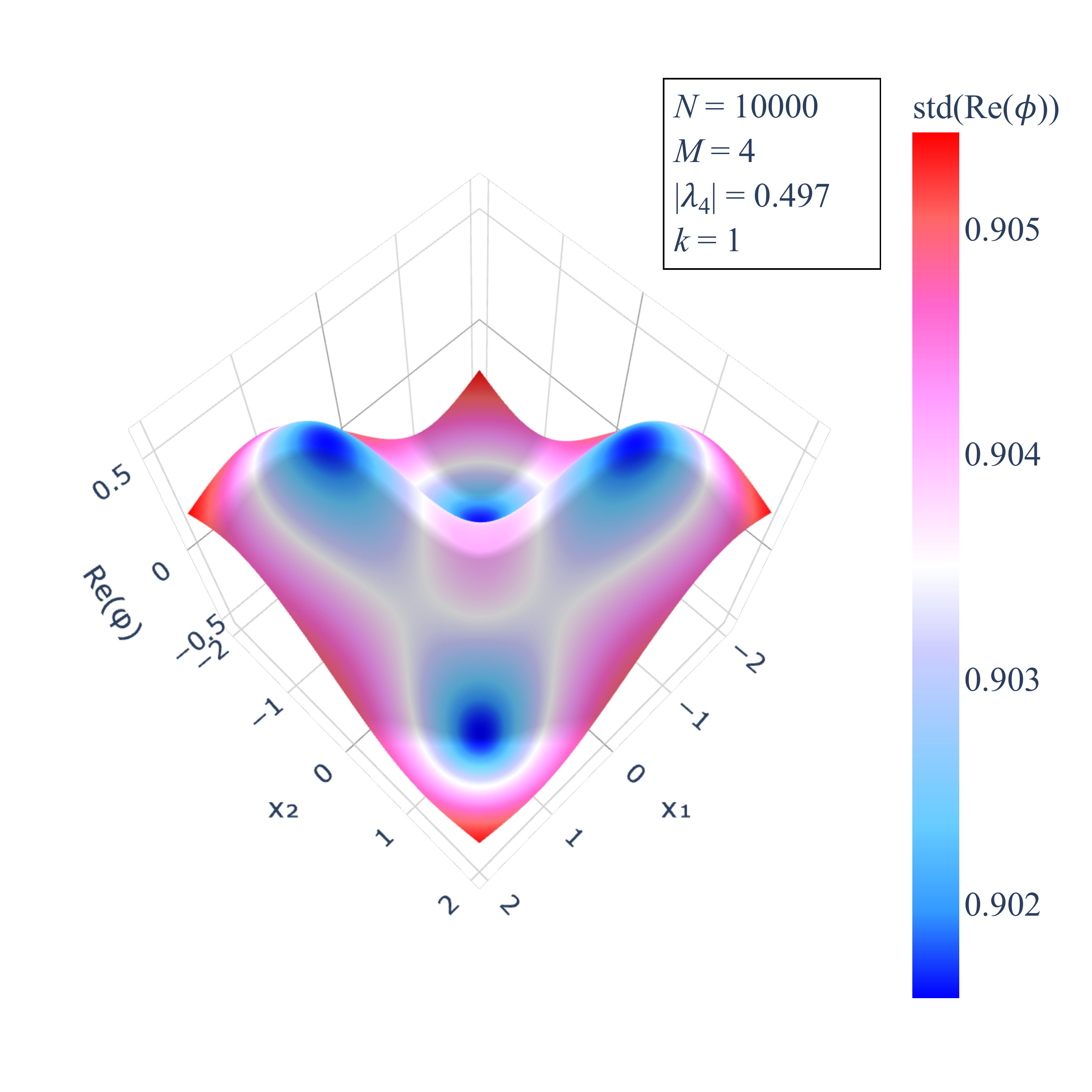}
    \end{minipage}
    \hspace{0.02\columnwidth}
    \begin{minipage}{0.45\columnwidth}
        \centering
        \includegraphics[trim={111 124 12 109}, clip, width=\linewidth]{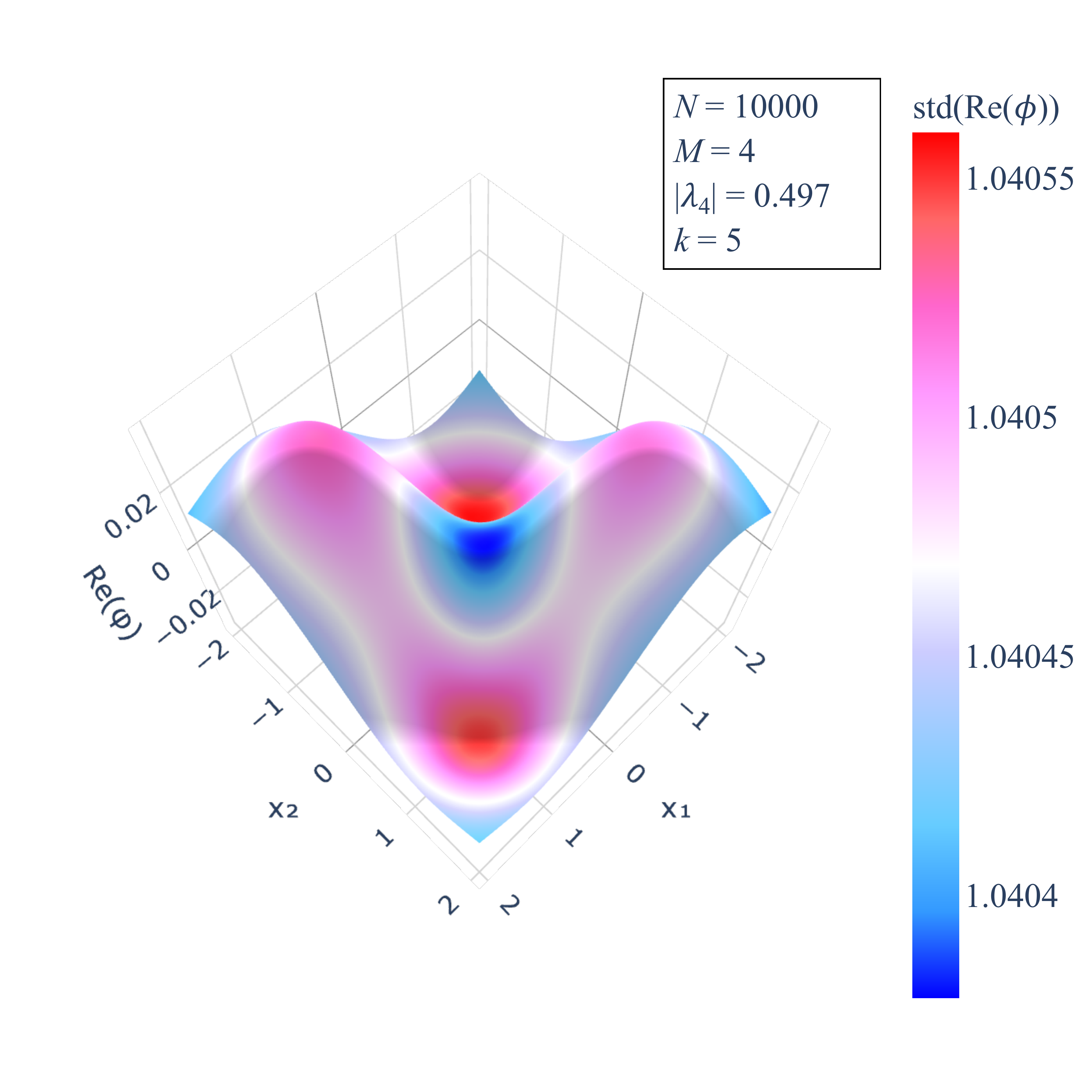}
    \end{minipage}
    \captionsetup{hypcap=false}
    \phantomsection
    \captionof{figure}{\emph{Propagated eigenfunctions of the stochastic quadruple well}. Top row: $\phi_2$ (left) and $\phi_3$ (right) at $k = 1$. Bottom row: $\phi_4$ at $k = 1$ (left) and $k = 5$ (right). The surface height depicts the forecast mean of $\phi_i(X_k)$ given $X_0 = \bm{x}$ and the colour its standard deviation, with the colour scale set per panel.}\label{fig: quadruple well eigenfunctions}
    \vspace{0.5\baselineskip}
\end{minipage}

Since the potential is separable and the noise components are independent, the coordinates $x_{t,1}$ and $x_{t,2}$ evolve as independent copies of the same one-dimensional double-well process. The eigenfunctions of the transfer operators are therefore products of one-dimensional eigenfunctions, with products of the corresponding eigenvalues. Since the dynamics are reversible, all eigenvalues are real-valued \citep{mauroy2020koopman}. Apart from the constant eigenfunction, the dominant eigenfunctions are the slowest non-trivial one-dimensional eigenfunction in either coordinate, which separates the wells along $x_1$ or along $x_2$, and the product of the two. In theory, this gives $\lambda_2 = \lambda_3$ and $\lambda_4 = \lambda_2 \, \lambda_3$. Every linear combination of the two eigenfunctions that share the eigenvalue $\lambda_2 = \lambda_3$ is again an eigenfunction, and the estimated $\phi_2$ and $\phi_3$ are determined only up to a rotation within this two-dimensional eigenspace. The estimates $|\lambda_2| = 0.7123$ and $|\lambda_3| = 0.6788$ split slightly, while the ratio $\lambda_4 / (\lambda_2 \, \lambda_3) \approx 1.028$ is close to one.

Since $\phi_i$ is an eigenfunction, its forecast mean $\lambda_i^k \, \phi_i(\bm{x})$ decays geometrically with $k$. Its variance, in contrast, receives a one-step contribution at every step, which contains the epistemic variance at the state where it is evaluated, while earlier contributions are damped by the factor $|\lambda_i|^2$ per step. At $k = 1$, the only contribution is evaluated at the initial condition $\bm{x}$, and the standard deviation is lowest at the four wells, where the long training trajectories spend most of their time and the data are densest. For larger $k$, the variance levels off, and the later contributions are evaluated at intermediate states that depend less and less on $\bm{x}$. The bottom row shows this for $\phi_4$. From $k = 1$ to $k = 5$, its mean shrinks by the factor $\lambda_4^4 \approx 0.06$, and the spatial contrast of its standard deviation fades. The forecast thereby loses the information about the initial condition as the process mixes.

Reversibility also places this benchmark where GP-DMD and GP-TCCA coincide. For pairs drawn from the invariant density, the forward and backward lifted models agree in the shared-kernel, shared-dictionary large-sample limit (Appendices~\ref{subsec: Large-sample Limit} and~\ref{subsec: Forward-backward Operators}), the singular functions of GP-TCCA coincide with the eigenfunctions, and the singular values equal the moduli of the eigenvalues. Our trajectories start from a Latin hypercube design, but with $| \lambda_2 | \approx 0.71$ per step most of their $100$ steps are taken close to the invariant density. The eigenfunctions in Figure~\ref{fig: quadruple well eigenfunctions} therefore also approximate the leading singular functions. Among the classical methods, this is the setting of the variational approach for conformation dynamics (VAC) \citep{noe2013variational}.

\subsection{Bickley Jet} \label{sec: Bickley Jet}

The last benchmark is the deterministic perturbed Bickley jet of \citet{rypina2007lagrangian}, in the form of \citet{10.1063/1.4971788}. It idealizes the zonal jet at the edge of the stratospheric polar vortex as a steady jet carrying three traveling waves,
\begin{alignat*}{3}
    \begin{split}
        \dot{x}_{t,1}
        &=
        U_0 \operatorname{sech}^2\!\left( \frac{x_{t,2}}{L_0} \right)
        \left[
            1
            + 2 \tanh\!\left( \frac{x_{t,2}}{L_0} \right)
            \sum_{n=1}^{3} A_n \cos\!\big( k_n ( x_{t,1} - c_n t ) \big)
        \right],\\
        \dot{x}_{t,2}
        &=
        - U_0 L_0 \operatorname{sech}^2\!\left( \frac{x_{t,2}}{L_0} \right)
        \sum_{n=1}^{3} A_n k_n \sin\!\big( k_n ( x_{t,1} - c_n t ) \big),
    \end{split}
\end{alignat*}
with $U_0 = 5.4138 \text{ [Mm/day]}$, $L_0 = 1.77 \text{ [Mm]}$, the wavenumbers $k_n = 2 n / r_0$ with $r_0 = 6.371 \text{ [Mm]}$, the phase speeds $( c_1, c_2, c_3 ) = ( 0.1446, 0.2053, 0.4561 ) \, U_0$, and the amplitudes $( A_1, A_2, A_3 ) = ( 0.075, 0.4, 0.3 )$. The flow is divergence free and periodic in $x_{t,1}$ with period $\pi r_0 \approx 20 \text{ [Mm]}$, and, as in \citet{klus2019kernel}, we wrap $x_{t,1}$ into $[0, 20)$. Since the flow is time dependent, its transition law depends on the initial time as well as on the lag \citep{klus2019kernel}. All pairs therefore start at the same initial time $t_0 = 0$ and end at $t_0 + \Delta t$. This gives a single transition law between two fixed times, which is the setting of coherent sets described in Appendix~\ref{subsec: Forward-backward Operators}.

The system was simulated by a numerical ODE solver and sampled once after $\Delta t = 40 \text{ [days]}$. The training set consists of $N = 10\ts000$ one-step pairs, with initial conditions drawn from a plain Latin hypercube design over $[0, 20] \times [-3.1, 3.1]$. Since the singular value decomposition of GP-TCCA does not require a shared dictionary (Section~\ref{sec: Bayesian Forward-Backward Operators}), the two endpoints have independent dictionaries $\Z_X$ and $\Z_Y$ with independent residuals, lengthscales, and signal variances.

Figure~\ref{fig: Bickley jet singular functions} shows the source singular functions $f_2, \dots, f_7$, where, as for the eigenvalues of Section~\ref{sec: Stochastic Quadruple Well}, the index $1$ belongs to the trivial pair of constant functions. With $\rho_2 = 0.978$, the function $f_2$ separates the regions above and below the meandering jet, and $f_3$ singles out the jet core. The function $f_4$ divides the jet core into segments of opposite sign. The functions $f_5$, $f_6$, and $f_7$ are nearly zero except on isolated patches above and below the jet, which mark individual vortices. Since $x_{t,1}$ is periodic, the patches of $f_6$ at the left and right edges belong to a single vortex.

The modal variance displayed in each annotation box of Figure~\ref{fig: Bickley jet singular functions} is the fraction of the posterior trace variance~\eqref{eq: Bayesian forward-backward trace variance} contributed by one mode. Dividing the $i$th summand of the modal decomposition by the total variance gives, for the source function of mode $i$, the fraction
\begin{alignat*}{3}
    \frac{
        ( 1 - \rho_i^2 ) \left[ \dfrac{2 \rho_i^2}{N} + \dfrac{M_Y - \mathcal{S}}{N^2} \right]
    }{
        \dfrac{2}{N} \sum_j \rho_j^2 ( 1 - \rho_j^2 ) + \dfrac{( M_X - \mathcal{S} ) ( M_Y - \mathcal{S} )}{N^2}
    }
    &=
    \frac{
        \rho_i^2 ( 1 - \rho_i^2 ) + \dfrac{( 1 - \rho_i^2 ) ( M_Y - \mathcal{S} )}{2 N}
    }{
        \sum_j \rho_j^2 ( 1 - \rho_j^2 ) + \dfrac{( M_X - \mathcal{S} ) ( M_Y - \mathcal{S} )}{2 N}
    } .
\end{alignat*}
These values sum to one over $i = 1, \dots, M_X$. As $N \rightarrow \infty$ at fixed dictionary sizes, the terms with $2 N$ vanish, and a source function and its paired target function share the limiting fraction
\begin{alignat*}{3}
    \frac{
        \rho_i^2 ( 1 - \rho_i^2 )
    }{
        \sum_j \rho_j^2 ( 1 - \rho_j^2 )
    } .
\end{alignat*}
A modal variance is thus largest near $\rho_i^2 = 1 / 2$ and vanishes as $\rho_i^2$ approaches one. The most coherent pairs therefore contribute little to the posterior uncertainty of $\operatorname{Tr}_{\mathbb{H}_X} ( \overrightarrow{\mathcal{T}} )$.

\vspace{0.5\baselineskip}
\noindent\begin{minipage}{\columnwidth}
    \centering
    \includegraphics[width=\columnwidth]{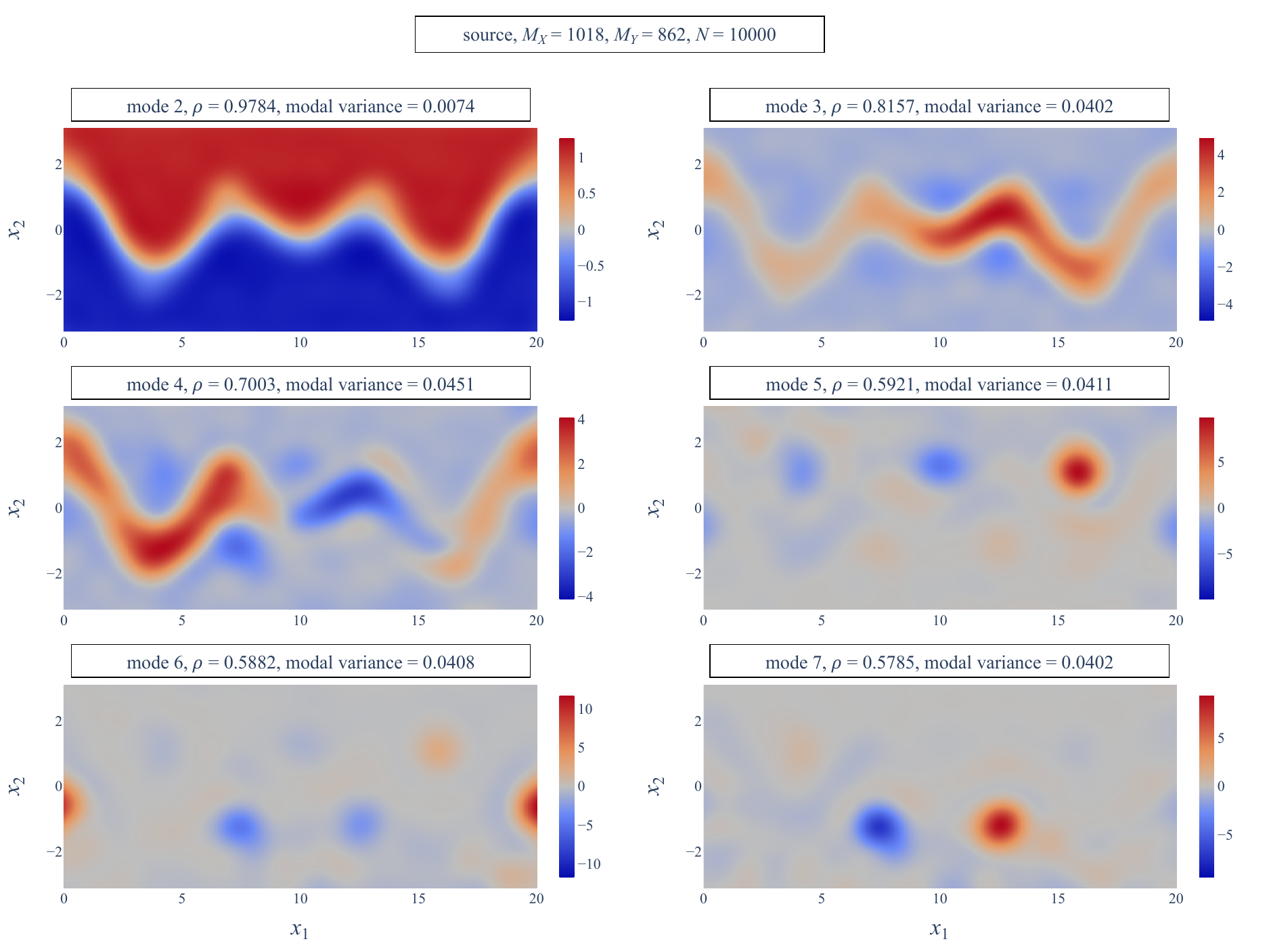}
    \captionsetup{hypcap=false}
    \phantomsection
    \vspace{-2.0\baselineskip}
    \captionof{figure}{\emph{Singular functions of the Bickley jet}. Source functions $f_i$ at $t_0$ for $i \in \{ 2, \dots, 7 \}$, scaled to unit mean square over the training states and signed to make the largest training value positive.}\label{fig: Bickley jet singular functions}
    \vspace{0.5\baselineskip}
\end{minipage}

\section{Conclusion} \label{sec: Conclusion}

We have presented and developed the mathematical theory underpinning Bayesian transfer operators, ranging from prediction to sparsification, and from analysis to model selection. We started by treating the embedded Perron--Frobenius operator \citep{klus2020eigendecompositions} as a random variable. More precisely, we placed an operator-valued GP prior on the conditional mean embedding of the one-step transition law and an inverse-Wishart process prior on its output covariance. The resulting posterior separates epistemic from aleatoric uncertainty, which we then carried into multi-step predictions, Koopman eigenfunctions, and time-lagged canonical correlation analysis. Comparing the lifted and state-space models gave the Bayesian-consistency principle, and adding the time-reversed lifted model gave Bayesian round-trip operators whose singular functions identify coherent sets, even in time-inhomogeneous systems.

The central strength of the Bayesian perspective is that results derived independently in separate lines of work emerge from a single unified model. The posterior mean recovers kernel EDMD \citep{williams2015kernel, williams2015data} and the conditional mean embedding \citep{grunewalder2012conditional, klus2020eigendecompositions}. The bound on the log Bayes factor used for model selection measures data fit through the VAMP-$2$ score \citep{wu2020variational}, and the inferred output covariance yields ResDMD residuals \citep{colbrook2023residual, colbrook2024beyond}. Methods that grew up in separate communities prove to be faces of one probabilistic model.

More broadly, the operator estimate shared by kernel EDMD and the conditional mean embedding admits both a kernel-embedding and a Bayesian GP interpretation. The former makes its conditional-expectation and operator structure explicit, while the latter identifies the estimate as a posterior mean, adds posterior uncertainty, and model selection via the Bayes factor. This dual interpretation affords a parallax on a single operator estimate, revealing connections between regularization, uncertainty, and operator approximation that are less apparent from either view alone. The same parallax extends to kernel mean embeddings themselves, whose Bayesian treatment has so far concerned marginal embeddings \citep{flaxman2016bayesian} and conditional embeddings for causal data fusion \citep{chau2021bayesimp}, and which the present work carries over to the conditional embeddings that represent transfer operators.

Since modern GP methodologies reach problems considerably more difficult than the ones treated in this work, we recommend reading this manuscript as theoretical groundwork. For example, our sparse approximation rests on the collapsed VFE of \citet{titsias2009variational}. The key insight of \citet{hensman2013gaussian}, by contrast, is to retain an explicit variational distribution over the inducing variables rather than to marginalize it. The resulting lower bound on the VFE decomposes into a sum over the observations and therefore admits stochastic variational inference with mini-batches. Carrying this idea over to the sparse lifted model is a natural route towards extremely large data sets. Beyond scalability, separating measurement noise from transition variability, which the training likelihood leaves undetermined, matters equally, since measurement noise can bias DMD spectral estimates \citep{duke2012error, dawson2016characterizing} while process noise can itself alter the Koopman spectrum \citep{bagheri2014effects}. Heteroskedastic and uncertain-input GP models \citep{lazaro2011variational, mchutchon2011gaussian} offer natural extensions of the noise model, complementing the debiasing, total-least-squares, Kalman-filter, and robust variants of DMD \citep{dawson2016characterizing, hemati2017biasing, nonomura2019extended, jiang2022correcting, scherl2020robust}. Finally, the multi-step forecasts rest on the Markov closure of Section~\ref{sec: Prediction Theory}, which explicit memory terms of the Mori--Zwanzig formalism \citep{doi:10.1137/21M1401759} would relax.

Richer output structures and learned feature maps offer a second future research direction. The model we presented here treats the flow-map channels independently and adopts a homoskedastic output covariance. Gaussian process regression networks \citep{10.5555/3042573.3042720} allow for correlations between outputs that vary with the input, and orthogonal mixing models keep exact multi-output inference tractable \citep{bruinsma2020scalable}. Generalized Wishart processes and their inverse counterparts \citep{wilson2011generalised, heaukulani2019scalable} suggest an output-covariance prior indexed by the source state. Such a prior would yield a heteroskedastic Bayesian-consistency covariance. Going one step beyond tuning the hyperparameters of a fixed kernel family, one can also learn the feature map. For example, deep kernels could endow a GP model with a learned neural feature map \citep{wilson2016deep}. Placed within the lifted objective, such kernels would offer a Bayesian counterpart to VAMPnets \citep{mardt2018vampnets}. More generally, artificial neural networks and kernel methods are closely related rather than competing approaches \citep{neal1996bayesian, Lee2018}, and each brings its own strengths to different facets of a problem.

One limitation of Bayesian transfer operator learning lies in the reconstruction of the state. A linear decoder \citep{azencot2020forecasting} requires the full-state observable to lie in the dictionary space, which is quite restrictive when that space must also be Koopman invariant \citep{10.1371/journal.pone.0150171}. Koopman autoencoders lift this restriction with a nonlinear decoder \citep{lusch2018deep, doi:10.1137/18M1177846}. In the same spirit, a GP regression from the lifted coordinates to the state, with a dedicated nonlinear prior covariance function, would act as a Bayesian decoder. For high-dimensional observations, such as the discretized fields of partial differential equations, the GP latent variable model might offer a scalable alternative \citep{pmlr-v9-titsias10a, damianou2011variational}. Either decoder would then generalize the Bayesian-consistency principle beyond linear reconstruction, and the latter would bring Bayesian transfer operators to bear on fluid flows \citep{schmid2010dynamic, rowley2009spectral}. In each of these directions, our Bayesian model provides a common and unyielding foundation for integrating modern Gaussian process methodologies.

\funding{
\begin{itemize}[leftmargin=*, nosep]
    \item SB acknowledges support by the Ministry of Culture and Science of North Rhine-Westphalia (MKW NRW) within the project SAIL under grant no. NW21-059D.
    \item SB acknowledges funding by the Federal Ministry of Research, Technology and Space (BMFTR) via the 6GEM+ transfer hub under grant no. 16KIS2412.
    \item SP acknowledges funding from the European Research Council (ERC Starting Grant ``KoOpeRaDE'') under the European Union's Horizon Europe research and innovation programme (grant agreement no. 101161457).
\end{itemize}
}

\appendix

\newpage

\bibliographystyle{unsrtnat}
\bibliography{bibliography}

\newpage
\begin{appendices}
\addtocontents{toc}{\protect\setcounter{tocdepth}{0}}
\section{Gaussian Process Regression} \label{sec: Gaussian Process Regression}

Given a \emph{training data set} $\mathcal{D} = \{(\bm{x}_{i}, y_i)\}_{i = 1}^N$ consisting of $N$ pairs of $D$-dimensional \emph{inputs} (covariates) and scalar \emph{outputs} (targets or dependent variables), the goal is to estimate the distribution of the function evaluation $f(\bm{x}_*)$ at any novel \emph{test} location $\bm{x}_* \in \mathbb{X}$, where $f \colon \mathbb{X} \rightarrow \mathbb{R}$. This appendix uses generic notation, without the arrows of the main text.

We start by grouping the inputs and outputs into the data matrices $\X \coloneqq [\bm{x}_1,\dots, \bm{x}_N] \in \mathbb{R}^{D \times N}$ and $\Y \coloneqq [y_1, \dots, y_N] \in \mathbb{R}^{1 \times N}$. Note that while GP regression supports a wide variety of input domains, we have defined $\mathbb{X} \subseteq \mathbb{R}^D$.

Our first modeling assumption is that the sensor noise on the targets is additive, independent, and Gaussian. Consequently, the relationship between the input random variable $X$ and the output random variable $Y$ is
\begin{alignat}{3} \label{eq: observation model}
    Y_i = f(X_i) + \epsilon_i, \hspace{1cm} \text{where } \epsilon_i \sim \mathcal{N}(0, \sigma_Y^2),
\end{alignat}
and $\sigma_Y^2$ is the homoskedastic variance of the sensor noise or the \emph{nugget} term for surrogate mismodeling \citep{gramacy2012cases}.

\subsection{Reproducing Kernel Hilbert Spaces}

A reproducing kernel $\kappa: \mathbb{X} \times \mathbb{X} \rightarrow \mathbb{R}$ is a bivariate positive semidefinite function. A rough sketch of the role of a kernel function is as a measure of the \emph{similarity} (or affinity) between two inputs, $\bm{x}$ and $\bm{x}'$. This notion of `similarity' is expressed through an inner product between the embeddings of the two inputs into a reproducing kernel Hilbert space (RKHS), $\mathbb{H}$ \citep{berlinet2011reproducing}. This inner product is embodied by the \emph{kernel trick}: 
\begin{alignat*}{3}
\kappa(\bm{x}, \bm{x}') = \langle \kappa(\bm{x}, \cdot), \kappa(\bm{x}', \cdot)\rangle_{\mathbb{H}}.
\end{alignat*}

The kernel trick encapsulates the properties that make RKHS methods so effective. The defining \emph{reproducing property} of an RKHS is $f(\bm{x}) = \langle f, \kappa(\bm{x}, \cdot)\rangle_{\mathbb{H}}$ for all $f \in \mathbb{H}$ \citep{berlinet2011reproducing}. What this means will become clear shortly. First define the \emph{feature map} $\varphi: \mathbb{X} \rightarrow \mathbb{H}$ according to $\varphi(\bm{x}) \coloneqq \kappa(\bm{x}, \cdot) \in \mathbb{H}$. Now consider that the reproducing property allows us to evaluate any function in $\mathbb{H}$. Since both $\varphi(\bm{x})$ and $\varphi(\bm{x}')$ are elements of $\mathbb{H}$, the reproducing property allows each to be evaluated at the point associated with the other.

Alternatively, one may think of any two points $\bm{x},\bm{x}' \in \mathbb{X}$ as vertices on a symmetric graph where the weight of the edge connecting them, $\kappa(\bm{x}, \bm{x}')$, is a quantitative comparison between two more complex objects. In this sense, a reproducing kernel defines a local geometry on $\mathbb{X}$. The practical benefit of the feature map is that while the input data may have very few discernible structures, we can represent more nonlinear structures with $\varphi(\cdot)$. 

\subsection{Bayes' Law}

Informally, \emph{Gaussian processes} (GPs) are mathematical objects that are generalizations of multivariate normal distributions. Recall that a multivariate normal distribution describes random variables that are vectors of \emph{finite} dimensionality. If the input domain were real numbers, then roughly speaking the random variable described by a GP is a ``vector'' of \emph{infinite} dimensionality indexed not by a natural number but by a real number. More formally, according to the \emph{function-space view}, a GP defines a probability distribution over functions such that every finite collection of function evaluations is jointly Gaussian; its covariance kernel $\kappa_\textrm{pr}$ induces the RKHS $\mathbb{H}$ used below for the kernel geometry.

GP regression proceeds following a \emph{Bayesian} philosophy, which identifies the \emph{likelihood} with the probability density $p_\textrm{om}(y_i \mid f, \bm{x}_i) = \mathcal{N}(y_i; f(\bm{x}_i), \sigma_Y^2)$. As indicated by the subscript, the likelihood captures the \emph{observation model}, i.e., the probabilistic mapping between a noise-free latent function evaluation and the noisy target. Through Bayes' law, the likelihood determines how the observed data turn prior beliefs about the regression function into posterior beliefs.

It follows that the next step is placing a GP \emph{prior} over the function $f(\cdot)$ such that $p_\textrm{pr}(f) = \mathcal{GP}\left(f; m_\textrm{pr}(\cdot), \kappa_\textrm{pr}(\cdot, \cdot) \right)$.  Here $m_\textrm{pr} : \mathbb{X} \rightarrow \mathbb{R}$ is the prior mean function $\mathbb{E}[f(\cdot)]$. The second component of the GP prior, the reproducing kernel $\kappa_\textrm{pr}(\bm{x}, \bm{x}')$, encodes $\mathbb{E}[(f(\bm{x}) - m_\textrm{pr}(\bm{x}))(f(\bm{x}') - m_\textrm{pr}(\bm{x}'))]$, i.e., the prior covariance between function values.

The combination of the likelihood and the prior defines the \emph{generative model}. Accordingly, we refer to $\theta$ as the set of \emph{generative} hyperparameters, which parameterize both the prior kernel function $\kappa_\textrm{pr}(\cdot,\cdot)$ and the characteristics of the sensor noise. To lighten the notation, we may omit the dependence of the probability densities on $\theta$.

An important feature is that a GP evaluated on the finite subset $\{\bm{x}_i\}_{i = 1}^N \subset \mathbb{X}$ is an $N$-dimensional multivariate Gaussian random variable. For simplicity, we choose the GP prior mean function to be identically zero, $m_\textrm{pr} \equiv 0$, so that the prior distribution over the function values is
\begin{align*}
   p_\textrm{pr}(\bm{f}_X) &= \mathcal{N}^N(\bm{f}_X ;\bm{0}, \K_{XX}),
\end{align*}
where we have introduced two new variables that are best understood at hand of the \emph{dictionary}:
\begin{align*}
    \Phi_X \coloneqq \begin{bmatrix} \varphi(\bm{x}_1) & \ldots & \varphi(\bm{x}_N) \end{bmatrix} \in \mathbb{H}^{1 \times N}.
\end{align*}
In essence, the dictionary collects the lifting functions and is a generalization of the \emph{design matrix} in linear regression. 

The adjoint of the dictionary operator is called the \emph{observation operator}, since $\Phi_X^\ast:\mathbb{H} \rightarrow \mathbb{R}^N$ maps an RKHS element to its evaluations at the sample locations. By the reproducing property, $\Phi_X^\ast h = [h(\bm{x}_1), \ldots, h(\bm{x}_N)]^\top$ for every $h \in \mathbb{H}$. For the GP latent function, we denote the corresponding finite vector of evaluations directly by $\bm{f}_X \coloneqq [f(\bm{x}_1), \ldots, f(\bm{x}_N)]^\top$, and define the \emph{variance-covariance matrix} by $\K_{XX} \coloneqq \big[\kappa_\textrm{pr}(\bm{x}_i, \bm{x}_j)\big]_{i,j = 1}^N = \kappa_\textrm{pr}(\X, \X) = \Phi_X^\ast \Phi_X$.
Simply put, GPs treat every latent function value as a random variable.

The final product of a GP model is the conditional probability density of novel function evaluations \citep{williams2006gaussian}. This posterior predictive distribution can be inferred via Bayes' rule, the prior GP (over both training and test latent values), and the likelihood.  Under the conditionally independent observation model, the likelihood factorizes over all $N$ samples, and the exact posterior process over $f(\cdot)$ can be expressed as
\begin{alignat*}{3}
    p_\textrm{pst}(f \mid \mathcal{D}) = \frac{\prod_{i = 1}^N  p_\textrm{om}\left(y_i \mid f(\bm{x}_i \right))p_\textrm{pr}(f) }{\int \prod_{i = 1}^N p_\textrm{om}\left(y_i \mid f(\bm{x}_i)\right) p_\textrm{pr}(f) \, \mathrm{d}f}. 
\end{alignat*}

Because the likelihood was Gaussian to begin with, the posterior process is also a GP admitting a closed-form expression. To derive this expression, we marginalize out the latent function values at the training points, resulting in a posterior distribution that is fully specified by
\begin{alignat}{3}
    m_\textrm{pst}(\bm{x}) &\coloneqq {\A}_f \, \mathbf{k}_{X}(\bm{x}) \text{ }  , \label{eq: GP mean}\\
    {\kappa}_\textrm{pst}(\bm{x}, \bm{x}') &\coloneqq \kappa_\textrm{pr}(\bm{x}, \bm{x}') - \mathbf{k}_{X}^\top(\bm{x}) \, \tilde{\K}_{XX}^{-1} \, \mathbf{k}_{X}(\bm{x}'), \label{eq: GP cov}
\end{alignat}
where ${\A}_f \coloneqq \Y \:\tilde{\K}_{XX}^{-1} \in \mathbb{R}^{1 \times N}$ is the \emph{weight matrix}, $\tilde{\K}_{XX} \coloneqq \K_{XX} + \sigma_Y^2 \I_N$ is the covariance matrix for the noisy outputs\footnote{The \emph{noise kernel} models a constant covariance over the state space $\mathbb{X}$ with $\kappa_\sigma(\bm{x},\bm{x}') \coloneqq \sigma^2\delta(\bm{x},\bm{x}')$, which is a scaled Kronecker delta function. Couple this with the fact that the sum of kernel functions is a valid kernel.}, and $\mathbf{k}_{X}(\bm{x}) \coloneqq \Phi_X^\ast \varphi(\bm{x}) 
 = [\kappa_\textrm{pr}(\bm{x}_1, \bm{x}), \ldots, \kappa_\textrm{pr}(\bm{x}_N, \bm{x})]^\top$ is a \emph{feature vector}. Now, all our questions about the \emph{noisy} predictive posterior distribution at some unseen input $\bm{x}_*$  can be answered by querying the expression:
 \begin{alignat*}{3}
    p_\textrm{pst}(y_* \mid \mathcal{D}) = \mathcal{N}(y_*; m_\textrm{pst}(\bm{x}_*), {\kappa}_\textrm{pst}(\bm{x}_*, \bm{x}_*) + \sigma_Y^2).
\end{alignat*}

Since the quality of the posterior solution is sensitive to $\theta$, it is typically prudent to first attempt to infer the optimal hyperparameters, $\theta^*$. The denominator in Bayes' rule, $p_\textrm{ml}(\Y)$, known as the \emph{marginal likelihood} (or \emph{evidence}), paves a path to craft an optimization algorithm for model selection, since it takes into account the entire generative model. In other words, it is natural to choose $\theta^*$ such that it maximizes the marginal probability of observing the targets under the model, after integrating over the latent function values according to the prior. To this end, the closed-form expression for the evidence is
\begin{alignat}{3} 
    \operatorname{ln}\left(p_\textrm{ml}\left(\Y\right)\right) &= \operatorname{ln} \left(\int p_\textrm{om}(\Y \mid \bm{f}_X) p_\textrm{pr}\left(\bm{f}_X\right) \, \mathrm{d}\bm{f}_X \right) \label{eq: exact evidence}\\
    &= -\frac{1}{2}
    \left[
    \underbrace{N \operatorname{ln}(2\pi)}_{\text{constant}} + 
    \underbrace{\operatorname{ln}\left(\left|\tilde{\K}_{XX}\right|\right)}_{\text{model complexity}} +  
    \underbrace{\Y \tilde{\K}_{XX}^{-1} \Y^\top}_{\text{data mismatch}}
    \right]. \nonumber
\end{alignat}
Maximizing \eqref{eq: exact evidence} naturally embodies Occam's razor, and can be interpreted as engineering the prior GP such that it maximizes the volume of data-matching functions.\!\footnote{One can often find an analytical expression for the gradient of \eqref{eq: exact evidence} and use optimization algorithms such as conjugate gradient ascent or BFGS \citep{zhu1997algorithm}.}

\subsection{Sparse Regression} \label{sec: VFE}

The most salient downside of GP regression is the computational complexity of identifying the posterior. Computing the exact posterior GP in conventional GP regression requires the inversion of an $N \!\times\! N$ matrix, which takes $\mathcal{O}(N^3)$ runtime, and $\mathcal{O}(N^2)$ memory space to store the weights and the training inputs. The cubic time-complexity and quadratic memory consumption create a bottleneck that hinders GPs from being deployed on data sets larger than a few thousand points.

To mitigate these computational drawbacks, practitioners resort to approximation techniques that employ an additional smaller \emph{pseudo-data set}. This data set is comprised of \emph{pseudo-outputs} (i.e., inducing variables) $\bm{f}_Z = \Phi_Z^\ast f$ that are unknown evaluations of the \emph{pseudo-inputs} (i.e., inducing inputs or points) $\Z \coloneqq \{\bm{z}_i \in \mathbb{X}\}_{i = 1}^M$. Such pseudo-data set schemes are widely used to reduce a kernel-based algorithm's training time-complexity to $\mathcal{O}(N M^2)$ and the memory-complexity to $\mathcal{O}(NM)$ \citep{snelson2005sparse, quinonero2005unifying, bui2017unifying}. The variable $M$, representing the cardinality of the pseudo-data set, governs the expressiveness of the sparse GP. The pseudo-inputs, $\Z$, play a similar role in support or relevance vector machines \citep{scholkopf2002learning,tipping2001sparse}.\!\footnote{Speed-ups can also be realized through low-rank methods, such as partial Gram--Schmidt orthogonalization \citep{hardoon2004canonical, shawe2004kernel}, random Fourier features \citep{rahimi2007random, nuske2023efficient}, or the LASSO-like algorithm of \citet{grunewalder2012conditional}.}

The challenge now becomes to construct an expression for the sparse posterior and jointly infer the hyperparameters and the pseudo-inputs. One approach is to design the approximated posterior GP so that it is as close as possible to the true posterior GP. The phrase ``as close as possible'' can be, for example, interpreted as minimizing the \emph{Kullback--Leibler} (KL) divergence between the sparse posterior GP and the exact posterior GP \citep{cover1999elements}. This particular optimization objective is explored in the \emph{variational free energy} (VFE) method of \citet{titsias2009variational}.

In the process-level interpretation of sparse variational GP inference \citep{matthews2016sparse}, a distribution $q(f)$ is introduced over the entire function $f$ as an approximation to $p_\textrm{pst}(f \mid \mathcal{D})$. The primary advantage of the VFE framework is that the pseudo-inputs are treated as \emph{variational} parameters of an approximated posterior rather than as generative parameters. Because the pseudo-inputs are treated as variational rather than generative parameters, their locations can be optimized directly through the variational objective without introducing them as additional parameters of the generative model \citep{titsias2009variational}. Another advantage of the VFE method is that increasing $M$ by adding inducing points cannot decrease the variational lower bound, and therefore monotonically improves, or leaves unchanged, the variational approximation to the exact GP posterior \citep{titsias2009variational}.

\begin{definition}\label{def: variational free energy} 
    The \emph{variational free energy} (VFE; \citet{titsias2009variational}):
    \begin{alignat*}{3}
       \mathcal{F}_f(\theta, \Z)  \coloneqq \int \operatorname{ln}\left(\frac{p(\Y, f \mid \theta)}{q(f)}\right) q(f) \, \mathrm{d}f \leq \operatorname{ln}(p_\textrm{ml}(\Y)),
    \end{alignat*}
    is often referred to as the \emph{evidence lower bound} (ELBO).
\end{definition}

The name variational free energy comes from statistical physics, and the connection is exact. Treat $- \operatorname{ln}(p(\Y, f \mid \theta))$ as the energy of a function $f$. The posterior $p_\textrm{pst}(f \mid \mathcal{D})$ is then a Boltzmann distribution at unit temperature, whose normalizing constant, the partition function, is the evidence $p_\textrm{ml}(\Y)$. The negative log-evidence is therefore the Helmholtz free energy. For any other distribution $q(f)$, the negative VFE is the expected energy under $q(f)$ minus the entropy of $q(f)$, which is the variational free energy of statistical physics. The variational free energy exceeds the Helmholtz free energy by $KL\left(q(f) \,||\, p_\textrm{pst}(f \mid \mathcal{D})\right)$ and equals it only when $q(f)$ is the posterior \citep{neal1998view}. In machine learning, \citet{hinton1994autoencoders} recognized this structure in the minimum description length principle, where the expected length of a randomly chosen code, net of the entropy of the choice, takes the form of the Helmholtz free energy. \citet{neal1998view} then showed that both steps of the EM algorithm maximize a single function of this form. The bound of \citet{titsias2009variational} carries the same name in, e.g., \citet{bauer2016understanding}. Two readings carry over from physics. The free energy is an expected description length \citep{hinton1994autoencoders}, and the log-ratio of two evidences is minus a free-energy difference, the quantity compared in the model selection of Section~\ref{sec: Model Selection}.

Under the conditionally independent observation model, and with the noisy inducing variables $\tilde{\bm{f}}_Z = \bm{f}_Z + \bm{\epsilon}_Z$ of Section~\ref{sec: Sparse Transfer Operators}, the VFE can be written as
\begin{alignat*}{3}
    \mathcal{F}_f(\theta, \Z) &= \sum_{i = 1}^N \int \text{ ln}(p_\textrm{om}(y_i \mid f_{i}, \theta)) q(f_{i} \mid \theta) \,
 \mathrm{d}f_i - KL\left(q(\tilde{\bm{f}}_Z) \,||\, p_\textrm{pr}(\tilde{\bm{f}}_Z \mid \theta)\right),
\end{alignat*}
where $q(\tilde{\bm{f}}_Z)$ is a free-form multivariate Gaussian distribution, and $p_\textrm{pr}(\tilde{\bm{f}}_Z \mid \theta) = \mathcal{N}^M(\tilde{\bm{f}}_Z; \bm{0}, \widetilde{\K}_{ZZ})$ refers to the prior Gaussian distribution over the inducing variables, with $\widetilde{\K}_{ZZ}$ of~\eqref{eq: regularized pseudo cov}. Setting $\sigma_Z = 0$ recovers the noise-free pseudo-outputs $\bm{f}_Z$ of \citet{titsias2009variational}, and for $\sigma_Z > 0$ the bound keeps its form with $\K_{ZZ}$ replaced by $\widetilde{\K}_{ZZ}$ \citep{titsias2008modelselection}. Since $\tilde{\bm{f}}_Z$ is then not a deterministic function of $f$, the gap between the bound and the log-evidence is instead a Kullback--Leibler divergence over the pair $(f, \tilde{\bm{f}}_Z)$. By the chain rule, this divergence is at least the process-level divergence of \citet{matthews2016sparse}. 

Due to the Gaussian assumptions, the calculus of variations can be used to analytically solve for the optimal approximate posterior Gaussian process. This results in the following posterior mean and covariance functions:
\begin{alignat}{3}
    m_\textrm{pst}(\bm{x})  &\coloneqq {\A}_f \: \mathbf{k}_{Z}(\bm{x}), \label{eq: VFE mean}\\
    {\kappa}_\textrm{pst}(\bm{x}, \bm{x}') &\coloneqq \kappa_\textrm{pr}(\bm{x}, \bm{x}') - \mathbf{k}_{Z}^\top(\bm{x}) \text{ }\tilde{\Q}\text{ }\mathbf{k}_{Z}(\bm{x}'). \label{eq: VFE cov}
\end{alignat}
Here, we are deviating from the notation in \citet{titsias2009variational} to highlight the similarities with (kernel) ridge-regression. The \emph{weight matrix}, ${\A}_f \!  \coloneqq \! \Y \, \K_{ZX}^\top \, \widetilde{\C}_{XX}^{-1}$, contains the regression coefficients obtained from linear operations on matrices. The noisy \emph{Gramian} is defined by $\widetilde{\C}_{XX} \!  \coloneqq \!  \K_{ZX}\K_{ZX}^{\top} \!  + \!  \sigma_Y^2\widetilde{\K}_{ZZ}$, and the \emph{cross-covariance} matrix by $\K_{ZX} \!  \coloneqq \!  \kappa_\textrm{pr}(\Z, \X) \!  \in \!  \mathbb{R}^{M\times N}$. Lastly, to compute the posterior kernel the weights for the \emph{information gain} term are $\tilde{\Q} \! \coloneqq \,  \left(\widetilde{\K}_{ZZ}^{-1} -\sigma_Y^2 \, \widetilde{\C}_{XX}^{-1}\right)$.

The VFE, i.e., the quantity that we maximize, is
\begin{alignat}{3} \label{eq: VFE}
        &\mathcal{F}_f(\theta, \Z) = \operatorname{ln}\left( \mathcal{N}^N(\Y ; \bm{0}, \widetilde{\mathbf{R}}_{XX})\right) - \frac{1}{2\sigma_Y^2} \operatorname{Tr}\left(\K_{XX} - \mathbf{R}_{XX} \right),
\end{alignat}
where $\mathbf{R}_{XX} \coloneqq \K_{ZX}^\top\widetilde{\K}_{ZZ}^{-1}\K_{ZX}$ is the \emph{Nyström approximation} to the exact prior covariance matrix, and $\widetilde{\mathbf{R}}_{XX} \coloneqq \mathbf{R}_{XX} + \sigma_Y^2\I_N$ is its noisy variant \citep{williams2000using}. Notice that the closed-form expression in \eqref{eq: VFE} is very similar to that of the exact log-marginal likelihood \eqref{eq: exact evidence}, with the addition of the trace term. The trace term encourages pseudo-inputs to better capture the training data, and is proportional to the sum of the variances of the training function values given the pseudo-targets, $p\left(\bm{f}_X \mid \bm{f}_Z\right)$, i.e., the total squared error of predicting $\bm{f}_X$ given $\bm{f}_Z$.

\section{Transfer Operator Theory} \label{sec: Transfer Operator Theory}

GP regression as presented in Appendix~\ref{sec: Gaussian Process Regression} can directly be applied to time-series data. However, if we merge GP regression with transfer operator theory we can efficiently perform multi-step predictions and gain a deeper understanding of the dynamical system.

To that end, we recall the definition of the Koopman operator for random dynamical systems (RDSs), starting with time-homogeneous Markov processes \citep{vcrnjaric2020Koopman, Klus2016Numerical}. Examples include stochastic differential equations driven by Gaussian white noise and discrete-time systems generated by i.i.d. random maps. The finite-time construction introduced below subsequently allows the source and image laws to differ, and for time-inhomogeneous dynamics the transition law additionally depends on the starting time \citep{FROYLAND20131, klus2019kernel}.

Consider the continuous-time stochastic dynamical systems represented by the time-homogeneous Itô stochastic differential equation (SDE),
\begin{alignat}{3} \label{eq: Ito}
    \mathrm{d}\bm{x}_t = \mathbf{r}(\bm{x}_t) \, \mathrm{d}t + \mathbf{h}(\bm{x}_t) \, \mathrm{d}W_t, \quad t \geq 0,
\end{alignat}
where $\{ \bm{x}_t \}_{t \geq 0}$ is a stochastic process defined on the state space $\mathbb{X}$, and $W_t$ is the $D$-dimensional standard Wiener process (i.e., Brownian motion). Such stochastic dynamics arise when there is process noise present in the system, which enters in the form of a diffusion term $\mathbf{h}(\bm{x}_t)$. The first term $\mathbf{r}(\bm{x}_t)$ is commonly referred to as the drift term. Here the drift coefficient $\mathbf{r} \colon \mathbb{X} \to \mathbb{R}^{D}$ and diffusion coefficient $\mathbf{h} \colon \mathbb{X} \to \mathbb{R}^{D \times D}$ are assumed to be smooth and time-invariant.

While Koopman theory also covers continuous-time dynamics, we convert to discrete-time dynamics by sampling trajectories at uniform time intervals, $\Delta t$, such that $ t_k \coloneqq 
\Delta t \, k$, which implies $t_k + \Delta t = t_{k + 1}, \text{ } \forall \text{ } k\in \mathbb{N}$.  The resulting discrete-time Markov process can be represented by the \emph{flow map} $\bm{F}_\alpha \colon \mathbb{X} \to \mathbb{X}$, which is indexed by the random variable $\alpha \in \Omega$ such that $Y = \bm{F}_\alpha(X)$, where $\Omega$ is the probability space associated with the stochastic dynamics. Assuming that its transition kernel is absolutely continuous with respect to the Lebesgue measure, the resulting discrete-time RDS admits a \emph{transition density function}, $\overrightarrow{p}_{\Delta t}: \mathbb{X} \times \mathbb{X} \rightarrow \mathbb{R}_{\geq 0}$, where $\overrightarrow{p}_{\Delta t} ( \bm{x}_{k+1} \mid \bm{x}_k )$ denotes the conditional density of the next state $\bm{x}_{k+1}$ given the current state $\bm{x}_k$.

\subsection{The Koopman and Perron--Frobenius Operators}

Koopman theory reformulates the evolution of a dynamical system in terms of functions of the state, referred to as \emph{observables}. Although the underlying state dynamics may be nonlinear, their evolution is described by a linear operator acting on a generally infinite-dimensional function space. In short, operator theory can be viewed as a trade-off between nonlinear dynamics in the state space and linear dynamics in a richer space of observables. We denote an observable by $g \in \mathbb{G}$ and take $\mathbb{G} = \mathcal{L}^\infty(\mathbb{X})$.\!\footnote{Other Banach spaces for $\mathbb{G}$ are also valid, e.g., the Hilbert space of Lebesgue square-integrable functions, i.e., $\mathbb{G} \coloneqq L^2(\mathbb{X})$ \citep{klus2020eigendecompositions, ikeda2022Koopman}.}

\begin{definition}\label{def: stochastic Koopman operator} 
    The \emph{stochastic Koopman operator} (SKO) $\mathcal{U} \colon \mathbb{G} \rightarrow \mathbb{G}$ is defined as the conditional expectation of any observable composed with the flow \citep{williams2015data, mezic2005spectral},
    \begin{alignat*}{3}
        (\mathcal{U}g)(\bm{x}) &= \mathbb{E}\left[g(Y) \mid X = \bm{x}\right]\\
        &=\int  \overrightarrow{p}_{\Delta t}(\bm{y} \mid \bm{x}) \, g(\bm{y}) \, \mathrm{d}\bm{y}.
    \end{alignat*} 
\end{definition}

In general, an observable may be a vector-valued function, $\mathbf{g} \in \mathbb{G}^P$. Of particular interest are the \emph{full-state observable}, $\mathbf{g}_{\bm{F}}(\bm{x}) = \bm{x}$, and the \emph{extended-state observable}, $\mathbf{g}_{\kappa}(\bm{x}) = \mathbf{k}_{X}(\bm{x})$, i.e., the kernel feature vector of~\eqref{eq: GP mean} in Appendix~\ref{sec: Gaussian Process Regression}, whose components are the kernel functions $\kappa_\textrm{pr}(\bm{x}_i, \cdot)$ centered at the training inputs. From a theoretical perspective, we often only consider scalar-valued observables, $g \colon \mathbb{X} \to \mathbb{R}$, and have $\mathcal{U}$ act component-wise on the set of observables $\{ g_i\}_{i=1}^P$.

\begin{definition}\label{def: Perron-Frobenius operator} 
    The \emph{Perron--Frobenius operator} (PFO) is concerned with an ensemble of trajectories and acts on the conjugate space of probability densities $\mathcal{P}: \mathcal{L}^1(\mathbb{X}) \rightarrow \mathcal{L}^1(\mathbb{X})$:
    \begin{alignat*}{3}
        (\mathcal{P}p) (\bm{y})
        &=\int \overrightarrow{p}_{\Delta t}(\bm{y} \mid \bm{x}) \, p(\bm{x}) \, \mathrm{d}\bm{x}.
    \end{alignat*} 
\end{definition}

In particular, the SKO is the adjoint of the PFO under the duality pairing between $L^1 ( \mathbb{X} )$ and $L^\infty ( \mathbb{X} )$ \citep{lasota1994chaos}. In comparison, the PFO evolves (probability) densities over the state space, presenting another lens by which the SKO can be analyzed. In particular, it is well suited for studying chaotic dynamical systems. 

Hence we call any operator, transporting some object (a distribution, or an observable) by the dynamics, a \emph{transfer operator}, i.e., the action of the process on functions of the state. A kernel-based theory of transfer operators can be formulated by representing observables in reproducing kernel Hilbert spaces, which enables transfer operators and their spectral properties to be expressed in terms of kernel covariance and cross-covariance operators. Following \citet{klus2020eigendecompositions}, the eigenvalues and eigenfunctions of these operators can then be studied within the RKHS setting. For a bounded measurable kernel, the Tikhonov-regularized kernel transfer operators used in this work (Appendix~\ref{subsec: Conditional Distributions}) are trace class, and hence compact with a discrete nonzero spectrum \citep{hsing2015theoretical}.

Strictly speaking, the definitions of kernel transfer operators are formulated in the language of kernel mean embeddings (KMEs) \citep{Muandet20171}. Kernel mean embeddings provide a nonparametric representation of probability distributions and conditional distributions in an RKHS. KME is closely associated with kernel principal component analysis (KPCA) and kernel canonical correlation analysis (KCCA), both of which are considered to be unsupervised learning techniques \citep{scholkopf1998nonlinear, scholkopf1997kernel, mika1998kernel, hardoon2004canonical, bach2002kernel, fukumizu2007statistical}. For the sake of completeness, the theory describing KMEs is recapitulated in Appendix \ref{sec: Embedding Probability Distributions}. 

\subsection{Koopman Mode Decomposition}

The crux of transfer operator analysis rests upon the fact that the nonlinear dynamics are linear in the eigenfunction coordinates \citep{mauroy2020koopman}. These \emph{eigenfunctions} $\{ \phi_i \in \mathbb{G}\}_{i = 1}^\infty$, and the corresponding \emph{eigenvalues} $\{ \lambda_i \in \mathbb{C}\}_{i = 1}^\infty$, of $\mathcal{U}$, obey the following equation in discrete time:
\begin{alignat}{3}\label{eq:koopman eigen}
    \left(\mathcal{U} \phi_i\right)(\bm{x}_k) = \mathbb{E}\left[\phi_i(X_{k+1}) \mid X_k = \bm{x}_k\right] = \lambda_i \, \phi_i(\bm{x}_k).
\end{alignat}
Assuming that the dynamical systems we are dealing with have discrete point spectra and that the corresponding eigenfunctions span the observable subspace of interest, the evolution of observables in this subspace can be expanded in terms of the eigenfunctions and eigenvalues. Specifically we can write the infinite series:
\begin{alignat*}{3}
    (\mathcal{U}\mathbf{g})(\bm{x}) = \sum_{i = 1}^\infty \mathbf{v}_i \lambda_i \phi_i(\bm{x}).
\end{alignat*}

Every \emph{Koopman mode}, $\mathbf{v}_i \in \mathbb{C}^P$, is determined by the dictionary of observables and is associated with an eigenpair $(\lambda_i, \phi_i)$. The modes are in essence the coefficients used to construct $\mathbf{g}(\cdot)$ using the Koopman eigenfunction basis. In other words, the observables can be expanded into a weighted sum of eigenfunctions:
\begin{equation*}
    \mathbf{g} = \sum_{i = 1}^\infty \mathbf{v}_i \phi_i \in \mathbb{G}^P.
\end{equation*}

Altogether, within a Koopman spectral representation, the modes and eigenpairs allow us to reconstruct and propagate the system's state over multiple time steps \citep{williams2015data}. Where the eigenvalues describe the temporal behavior of the dynamical system, the modes capture the spatial behavior. Once we have found the sequence of triplets $\{(\lambda_i, \phi_i, \mathbf{v}_i)\}_{i = 1}^\infty$ for a dynamical system, we have completed its \emph{Koopman mode decomposition} (KMD; \citet{mezic2005spectral, mezic2004comparison}). As a last point, note that the constant observable is always a Koopman eigenfunction with unit eigenvalue \citep{mauroy2020koopman}.

\subsection{Finite-Dimensional Approximation} \label{sec: koopman matrix}

The challenge is that the Koopman operator is infinite-dimensional. As far as numerical methods are concerned, we want to find a reduced but finite set $\{(\lambda_i, \phi_i, \mathbf{v}_i)\}_{i = 1}^M$, such that the evolution of the observables of interest can be reasonably well approximated. That is, the goal is to derive a matrix representation of the SKO by projecting it onto a finite-dimensional subspace, such that a finite-dimensional linear system is induced \citep{colbrook2024rigorous}.

A finite-dimensional approximation of the Koopman operator can be obtained by compressing its action to a chosen finite-dimensional subspace \citep{mauroy2020koopman, colbrook2024rigorous}. Here, the chosen subspace is spanned by the dictionary $\Phi_Z$ of Section~\ref{sec: Sparse Transfer Operators}, i.e., the kernel functions centered at the pseudo-inputs $\bm{z}_1, \ldots, \bm{z}_M$ of the sparse GP (Appendix~\ref{sec: VFE}),
\begin{alignat*}{3}
    \mathbb{G}_M
    \coloneqq
    \operatorname{span}\left\{
        \varphi(\bm{z}_1), \ldots, \varphi(\bm{z}_M)
    \right\},
\end{alignat*}
so that every $g \in \mathbb{G}_M$ can be written as
\begin{alignat*}{3}
    g = \sum_{i = 1}^M q_i \, \varphi(\bm{z}_i)
    = \Phi_Z \, \bm{q},
    \quad
    \bm{q} \in \mathbb{R}^M.
\end{alignat*}
In general, $\mathcal{U}g$ need not belong to $\mathbb{G}_M$, so the action of the Koopman operator must be projected back onto the chosen subspace. The special case in which
\begin{alignat*}{3}
    \mathcal{U}g \in \mathbb{G}_M
    \qquad
    \forall \, g \in \mathbb{G}_M
\end{alignat*}
defines a Koopman-invariant subspace. In that case the compression coincides with the restriction of $\mathcal{U}$ to $\mathbb{G}_M$ and introduces no closure error on this subspace \citep{mauroy2020koopman, colbrook2024rigorous}. When $\mathbb{G}_M$ is not Koopman invariant, the finite-dimensional model represents only the projected action of $\mathcal{U}$ on the chosen subspace. Moreover, if the Koopman operator has continuous or mixed spectral components, these cannot in general be represented exactly by a finite set of eigenfunctions \citep{mezic2005spectral, colbrook2024rigorous}.

En route to a finite-dimensional matrix representation, we define a bounded linear map $\Gamma_Z \colon \mathbb{G} \to \mathbb{R}^M$ that yields the coordinates of its projected approximation in the basis $\{ \varphi(\bm{z}_i)\}_{i=1}^M$. Applying $\Gamma_Z$ to an observable $g \in \mathbb{G}$ returns the coefficients, $\bm{q} = \Gamma_Z \; g \in \mathbb{R}^M$. In turn, having the dictionary operator act on $\bm{q}$ will recover its projected approximation in $\mathbb{G}_M$, $\Phi_Z \bm{q} = \Pi_Z g \approx g$, where $\Pi_Z \coloneqq \Phi_Z \,  \Gamma_Z$.  Provided the dictionary functions $\{\varphi(\bm{z}_i)\}_{i=1}^M$ are linearly independent, $\Gamma_Z$ is their coordinate map and $\Gamma_Z \, \Phi_Z = \I_M$, so $\Pi_Z$ is the projector onto $\mathbb{G}_M$. If in addition $\mathbb{G}_M$ is invariant, $\mathcal{U} \, \Pi_Z \,g \in \mathbb{G}_M$ for every $g \in \mathbb{G}$, and hence $\Pi_Z \, \mathcal{U} \, \Pi_Z = \mathcal{U} \, \Pi_Z$; the compression below is then exact on $\mathbb{G}_M$ \citep{mauroy2020koopman}.

Altogether, the compression of the Koopman operator to $\mathbb{G}_M$ defines the finite-dimensional operator $\mathcal{U}_M : \mathbb{G}_M \rightarrow \mathbb{G}_M$ by
\begin{alignat*}{3}
    \mathcal{U}_M g
    \coloneqq
    \Pi_Z \, \mathcal{U} g,
    \qquad
    g \in \mathbb{G}_M.
\end{alignat*}
From here we can define the finite-dimensional matrix representation $\U: \mathbb{R}^M \rightarrow \mathbb{R}^M$ of $\mathcal{U}$ s.t.
\begin{alignat}{3} \label{eq: projected Koopman representation}
    \mathcal{U}_M \, g = \Pi_Z \, \mathcal{U} \, \Pi_Z \, g = \Phi_Z \, \U \, \Gamma_Z \, g, \quad g \in \mathbb{G}_M.
\end{alignat}
which defines the \emph{Koopman matrix} as
\begin{alignat}{3} \label{eq: koopman matrix}
    \U \coloneqq \Gamma_Z \, \mathcal{U} \, \Phi_Z \in \mathbb{R}^{M \times M}.
\end{alignat}

The spectral properties of the finite-dimensional compression $\mathcal{U}_M$ can be obtained from its matrix representation $\U$. Let $\{ \phi_i\}_{i = 1}^M$ denote the eigenfunctions of $\mathcal{U}_M$, which can be written as
\begin{alignat*}{3}
    \phi_i = \sum_{j =1}^M \mathrm{w}_{ij} \, \varphi(\bm{z}_j) = \Phi_Z \,\bm{\mathrm{w}}_i, \quad \bm{\mathrm{w}}_i \in \mathbb{C}^M,
\end{alignat*}
and then apply the relationship between eigenvalues and eigenfunctions \eqref{eq:koopman eigen}:
\begin{alignat*}{3}
    \mathcal{U}_M \phi_i &= \lambda_i \phi_i, \\
    \Phi_Z \U \Gamma_Z \phi_i &= \lambda_i \Phi_Z \Gamma_Z \phi_i, \\
    \U \Gamma_Z \phi_i &= \lambda_i \Gamma_Z \phi_i, \\
    \U \bm{\mathrm{w}}_i &= \lambda_i \bm{\mathrm{w}}_i.
\end{alignat*}
Hence, the right eigenvectors of $\U$ give the coordinates of the eigenfunctions of $\mathcal{U}_M$ in the dictionary basis, while its eigenvalues are the eigenvalues of the compressed operator $\mathcal{U}_M$. When $\mathbb{G}_M$ is Koopman invariant, $\mathcal{U}_M$ coincides with the restriction of $\mathcal{U}$ to $\mathbb{G}_M$, and these eigenvalues and eigenfunctions are therefore also eigenvalues and eigenfunctions of the underlying Koopman operator. The left eigenvectors of $\U$ define the modes of the extended-state observable and, in the invariant case, are related to the eigenfunctionals of the dual Koopman operator \citep{mauroy2020koopman}.

\subsection{Forward--backward Operators} \label{subsec: Forward-backward Operators}

The preceding spectral discussion concerns eigenfunctions of a single transfer operator. Such eigenfunctions reveal metastable, or almost-invariant, sets \citep{FROYLAND20131}. An invariant set, such as a fixed point or an attractor, is never left by a trajectory that starts in it. An almost-invariant set $\mathbb{A}$ may be left, but a trajectory that starts in it is likely to still be in it after one step. For an initial state that follows the invariant density, this means $\mathbb{P}[Y \in \mathbb{A} \mid X \in \mathbb{A}] \approx 1$ \citep{froyland2009almost}. An attractor thus describes where trajectories end up, whereas almost-invariant sets describe how trajectories move in the long run. They divide the state space into regions between which trajectories pass only rarely.

For reversible dynamics, the eigenvalues of the Koopman operator $\mathcal{U}$ are real and its eigenfunctions are orthonormal with respect to the invariant density \citep{noe2013variational, wu2020variational}. Let the initial state $X$ follow the invariant density, and order the eigenvalues as $1 = \lambda_1 \geq \lambda_2 \geq \cdots$, where $\lambda_1$ belongs to the constant eigenfunction $\id$. If a function $f$ is normalized and orthogonal to the eigenfunctions of $\lambda_1, \ldots, \lambda_{j-1}$, all with respect to the invariant density, then
\begin{alignat*}{3}
    \mathbb{E}\left[f(X) \, f(Y)\right]
    &\leq
    \lambda_j ,
\end{alignat*}
with equality when $f$ is the eigenfunction of $\lambda_j$. This variational principle underlies the variational approach for conformation dynamics, which finds the slow processes by maximizing the autocorrelation of trial functions \citep{noe2013variational}.

This variational principle rests on reversibility \citep{noe2013variational, wu2020variational}. For any dynamics and any initial distribution, one may instead ask which functions of the initial state can still be recovered after one step of length $\Delta t$ from the final state. The forward--backward operator addresses this question by propagating a function one step forward and pulling it back with the time-reversed dynamics. It maps functions of the initial state to functions of the initial state and is itself reversible, so its eigenfunctions are orthogonal and satisfy the same variational principle. These eigenfunctions are the source-side singular functions of the normalized transfer operator defined below, for which \citet{wu2020variational} give the variational principle~\eqref{eq: VAMP variational principle}.

Almost-invariant sets describe structures that persist over all times. A \emph{coherent set} instead describes transport between a fixed initial and a fixed final time, and therefore extends to time-dependent dynamics, whose transition law also depends on the initial time \citep{FROYLAND20131}. For time-homogeneous dynamics, the transition law depends only on the elapsed time, here a single step of length $\Delta t$, and the initial time enters only through the distribution of the initial state. For time-dependent dynamics, fixing the initial and the final time again yields a single transition law, and the construction below carries over with $\Delta t$ read as the lag between these two times. Whereas an almost-invariant region is mapped approximately onto itself, a coherent pair consists of a region $\mathbb{A} \subset \mathbb{X}$ at the initial time and a region $\mathbb{B} \subset \mathbb{X}$ at the final time, which may differ in position and shape. In a standard formulation, the two regions carry equal probability, $\mathbb{P}_X(\mathbb{A}) = \mathbb{P}_Y(\mathbb{B})$, and the pair is chosen so that as much as possible of the probability mass in $\mathbb{A}$ is transported into $\mathbb{B}$, or equivalently so that as little of this mass as possible ends up outside of $\mathbb{B}$ \citep{FROYLAND20131}. Coherent sets are therefore finite-time, minimally dispersive analogues of almost-invariant sets, of which moving atmospheric and oceanic vortices are typical examples \citep{FROYLAND20131, 10.1063/1.4971788}.

Let the initial state $X$ have density $p_X$, so that its successor has density $p_Y=\mathcal P p_X$ under the density-evolving PFO of Definition~\ref{def: Perron-Frobenius operator}. The two endpoint spaces are $L^2_{p_X}(\mathbb X)$ and $L^2_{p_Y}(\mathbb X)$. On these weighted spaces we denote the ordinary forward and Bayesian-reversed conditional-expectation operators by the same directional Koopman symbols used in the main text, with their functional setting made explicit:
\begin{alignat}{3} \label{eq: normalized forward operator}
    (\overrightarrow{\mathcal U}g)(\bm x)
    &\coloneqq\int_{\mathbb X}\overrightarrow p_{\Delta t}(\bm y\mid\bm x)g(\bm y)\,\mathrm d\bm y,
    &\qquad \overrightarrow{\mathcal U}&:L^2_{p_Y}\to L^2_{p_X},\\
    (\overleftarrow{\mathcal U}f)(\bm y)
    &\coloneqq\frac{1}{p_Y(\bm y)}\int_{\mathbb X}\overrightarrow p_{\Delta t}(\bm y\mid\bm x)f(\bm x)p_X(\bm x)\,\mathrm d\bm x,
    &\qquad \overleftarrow{\mathcal U}&:L^2_{p_X}\to L^2_{p_Y}.
\end{alignat}
The second expression is also $\mathbb E[f(X)\mid Y=\bm y]$ by Bayes' rule, and equals the normalized forward PFO acting on the density contrast $f$: $\overleftarrow{\mathcal U}f=\mathcal P(p_Xf)/p_Y$ wherever $p_Y>0$. It is not the same operator as the original PFO $\mathcal P$ on densities, nor the embedded PFO on the RKHS. The first expression agrees with the action of the forward Koopman operator of Appendix~\ref{sec: Transfer Operator Theory} on these weighted spaces. In contrast, its RKHS counterpart in Section~\ref{sec: Kernel Transfer Operators} acts between feature spaces and its adjoint uses the RKHS inner products.

Exchanging the order of integration gives the exact endpoint-weighted adjoint pairing
\begin{alignat}{3} \label{eq: forward-backward pairing}
    \langle\overleftarrow{\mathcal U}f,g\rangle_{p_Y}
    &=\langle f,\overrightarrow{\mathcal U}g\rangle_{p_X}
    =\mathbb E[f(X)g(Y)],
    &\qquad \overleftarrow{\mathcal U}&=\overrightarrow{\mathcal U}^{*}.
\end{alignat}
Both conditional-expectation operators also preserve the constant function. For measurable endpoint sets $\mathbb A,\mathbb B\subset\mathbb X$ with $\mathbb P_X(\mathbb A)>0$, the pairing implies
\begin{alignat}{3} \label{eq: coherent pair probability}
    \frac{\langle\overleftarrow{\mathcal U}\id_{\mathbb A},\id_{\mathbb B}\rangle_{p_Y}}{\mathbb P_X(\mathbb A)}
    &=\mathbb P[Y\in\mathbb B\mid X\in\mathbb A].
\end{alignat}
The normalized conditional-expectation formulation therefore connects the coherent-pair probability to the two endpoint distributions, without assuming reversibility of the original dynamics \citep{FROYLAND20131, 10.1063/1.4971788, klus2019kernel}.

To assess whether an initial region survives this finite-time transport, one may first propagate it to the final endpoint and then pull it back using the Bayesian time-reversed transition density
\begin{alignat}{3} \label{eq: reversed transition density}
    \overleftarrow{p}_{\Delta t}(\bm{x}' \mid \bm{y})
    &\coloneqq
    \frac{
        \overrightarrow{p}_{\Delta t}(\bm{y} \mid \bm{x}') \, p_X(\bm{x}')
    }{
        p_Y(\bm{y})
    }.
\end{alignat}
The corresponding \emph{forward--backward operator} acts on the initial weighted space and is
\begin{alignat}{3} \label{eq: forward-backward operator}
    \overrightarrow{\mathcal T}
    &\coloneqq\overrightarrow{\mathcal U}\,\overleftarrow{\mathcal U}
    =\overleftarrow{\mathcal U}^{*}\overleftarrow{\mathcal U},
    &\qquad
    (\overrightarrow{\mathcal T}f)(\bm x)
    &=\int_{\mathbb X}p_{\textrm{fb}}(\bm x'\mid\bm x)f(\bm x')\,\mathrm d\bm x',
\end{alignat}
where
\begin{alignat*}{3}
    p_{\textrm{fb}}(\bm{x}' \mid \bm{x})
    &\coloneqq
    \int_{\mathbb{X}}
        \overrightarrow{p}_{\Delta t}(\bm{y} \mid \bm{x}) \,
    \overleftarrow{p}_{\Delta t}(\bm{x}' \mid \bm{y}) \,
    \mathrm{d}\bm{y}.
\end{alignat*}
Thus one step of the original dynamics is followed by one step of the Bayesian time-reversed dynamics \citep{10.1063/1.4971788}. The auxiliary transition satisfies $p_X(\bm{x}) \, p_{\textrm{fb}}(\bm{x}' \mid \bm{x}) = p_X(\bm{x}') \, p_{\textrm{fb}}(\bm{x} \mid \bm{x}')$, and is reversible with respect to $p_X$ even when the original dynamics is not \citep{10.1063/1.4971788}. Moreover, $\langle \overrightarrow{\mathcal T} \id_{\mathbb{A}}, \id_{\mathbb{A}} \rangle_{p_X} / \mathbb{P}_X(\mathbb{A})$ is the probability that the auxiliary chain $X \rightarrow Y \rightarrow X'$ returns to $\mathbb{A}$ after starting there. A highly coherent source region therefore becomes an almost-invariant, or metastable, region of this reversible forward--backward process \citep{10.1063/1.4971788, klus2019kernel}.

Coherent pairs are awkward to optimize over directly, because indicator functions are restricted to two values and the admissible pairs are constrained by $\mathbb{P}_X(\mathbb{A}) = \mathbb{P}_Y(\mathbb{B})$. The standard remedy is to relax the set-based problem to a function-based one: instead of indicators, one maximizes the pairing~\eqref{eq: forward-backward pairing} over normalized functions of the initial and the final state, and recovers candidate regions afterwards from the optimizers \citep{FROYLAND20131}. This relaxation turns the search for coherent pairs into a singular value problem for $\overleftarrow{\mathcal U}$, and it plays the role that the eigenfunction relaxation plays for metastable sets.

Since $\overrightarrow{\mathcal T} = \overleftarrow{\mathcal U}^* \overleftarrow{\mathcal U}$, it is positive and self-adjoint, with $\langle f, \overrightarrow{\mathcal T} f \rangle_{p_X} = \| \overleftarrow{\mathcal U} f \|_{p_Y}^2$. If the normalized transition kernel $\overrightarrow{p}_{\Delta t}(\bm{y} \mid \bm{x}) / p_Y(\bm{y})$ is square integrable with respect to $p_X \times p_Y$, then $\overleftarrow{\mathcal U}$ is Hilbert--Schmidt and therefore compact, admitting singular triples $(\rho_i, f_i, g_i)$ such that $\overleftarrow{\mathcal U} f_i = \rho_i \, g_i$ and $\overrightarrow{\mathcal U} g_i = \rho_i \, f_i$ \citep{FROYLAND20131}. The source-side functions $f_i$ are eigenfunctions of $\overrightarrow{\mathcal T}$ with eigenvalues $\rho_i^2$. The final-endpoint companion $\overleftarrow{\mathcal{T}} \coloneqq \overleftarrow{\mathcal U} \, \overrightarrow{\mathcal U} = \overleftarrow{\mathcal U} \overleftarrow{\mathcal U}^*$ has the same nonzero spectrum and the corresponding target-side eigenfunctions $g_i$. Since $\overleftarrow{\mathcal U} \id = \id$ and $\overrightarrow{\mathcal U} \id = \id$, the constant function forms the trivial singular triple $(\rho_1, f_1, g_1) = (1, \id, \id)$, and no singular value exceeds one \citep{FROYLAND20131, wu2020variational}. This triple only expresses the conservation of probability mass and carries no information about the dynamics. The informative part of the spectrum therefore begins at $\rho_2$. The dominant nontrivial singular functions describe the least-dispersive finite-time directions, and the leading nontrivial singular value makes the relaxation quantitative: for any partitions $\{\mathbb{A}_1, \mathbb{A}_2\}$ and $\{\mathbb{B}_1, \mathbb{B}_2\}$ of the two endpoints with $\mathbb{P}_X(\mathbb{A}_k) = \mathbb{P}_Y(\mathbb{B}_k)$, the coherence probabilities \eqref{eq: coherent pair probability} satisfy $\sum_k \mathbb{P}[Y \in \mathbb{B}_k \mid X \in \mathbb{A}_k] \leq 1 + \rho_2$ \citep{FROYLAND20131}. Concrete regions are obtained afterwards by thresholding or clustering these functions \citep{FROYLAND20131, klus2019kernel}.

The singular triples satisfy the variational principle of \citet{wu2020variational}, which extends the variational principle for reversible dynamics to arbitrary endpoint laws. Order the singular values as $1 = \rho_1 \geq \rho_2 \geq \cdots$, and let $u_1, \ldots, u_j$ be orthonormal in $L^2_{p_X}(\mathbb{X})$ and $v_1, \ldots, v_j$ be orthonormal in $L^2_{p_Y}(\mathbb{X})$. For any positive integer $r$, the VAMP-$r$ score of these functions satisfies
\begin{alignat}{3} \label{eq: VAMP variational principle}
    \sum_{i=1}^{j}
    \mathbb{E}_{YX}\left[u_i(X) \, v_i(Y)\right]^r
    &\leq
    \sum_{i=1}^{j}
    \rho_i^r ,
\end{alignat}
with equality for $u_i = f_i$ and $v_i = g_i$ \citep{wu2020variational}. The dominant singular functions are thus the pairs of functions of the initial and the final state with the largest uncentered correlations, and the VAMP-$2$ score~\eqref{eq: kZ VAMP2 score} corresponds to $r = 2$. For reversible dynamics at equilibrium, the singular value decomposition is equivalent to the eigendecomposition \citep{wu2020variational}, and~\eqref{eq: VAMP variational principle} generalizes the variational principle for eigenfunctions.

For an invertible deterministic flow, with the final law the push-forward of the initial law, the transport is lossless in the sense that $\overleftarrow{\mathcal U}^* \overleftarrow{\mathcal U} = \mathcal{I}$ \citep{FROYLAND20131}. Every singular value then equals one, and no partition stands out as more coherent than another. To single out coherent sets, \citet{FROYLAND20131} adds a small diffusion before and after the flow, and \citet{10.1063/1.4971788} motivate the same step by requiring coherent sets to be robust under a small random perturbation of the flow. A stochastic model already contains such spreading in its transition density, whereas a deterministic model needs it added explicitly. In both cases, the amount of diffusion shapes the coherent sets that are found \citep{FROYLAND20131, 10.1063/1.4971788}.

The squared Hilbert--Schmidt norm of $\overleftarrow{\mathcal U}$, i.e., the VAMP-$2$ score of the full spectrum \citep{wu2020variational}, admits two further readings. In terms of the normalized transition kernel,
\begin{alignat*}{3}
    \sum_{i \geq 1} \rho_i^2
    =
    \int_{\mathbb{X}} \int_{\mathbb{X}}
    \left(
        \frac{ \overrightarrow{p}_{\Delta t}(\bm{y} \mid \bm{x}) }{ p_Y(\bm{y}) }
    \right)^{2}
    p_X(\bm{x}) \, p_Y(\bm{y}) \,
    \mathrm{d}\bm{x} \, \mathrm{d}\bm{y} ,
\end{alignat*}
which compares the joint law of $(X, Y)$ with the product of its marginals, i.e., with the law the endpoints would have if they were independent. The sum equals one plus the $\chi^2$-divergence between these two laws. The trivial triple contributes the one. The remaining singular values measure how strongly the final state depends on the initial state. They vanish when $X$ and $Y$ are independent, and the sum of their squares is infinite for the lossless transport above, where the joint law concentrates on the graph of the flow. The logarithm of the sum is the R\'enyi divergence of order two \citep{vanerven2014renyi}, which is never smaller than the mutual information of $X$ and $Y$ \citep{cover1999elements, vanerven2014renyi}. The full-spectrum VAMP-$2$ score is thus a measure of the statistical dependence between the two endpoints.

The second reading concerns reversible dynamics at equilibrium, i.e., with both endpoint laws equal to the invariant density. Then $\overleftarrow{\mathcal U} = \overrightarrow{\mathcal U}$ is self-adjoint on $L^2_{p_X}(\mathbb{X})$, and the $\rho_i$ are the moduli of its eigenvalues. For overdamped Langevin dynamics, a reversible instance of the SDE~\eqref{eq: Ito}, the negative Koopman generator is unitarily equivalent to a Schr\"odinger operator shifted by its ground-state energy \citep{klus2020kernel}. This correspondence maps the relaxation rates of the diffusion to the energy levels of a quantum system, and the square root of the invariant density to its ground state with zero energy, which corresponds to the trivial triple. Each squared singular value then becomes the Boltzmann weight of an energy level at inverse temperature $2 \Delta t$, and the sum, whenever it is finite, is the partition function of the quantum system. Since fast modes correspond to high energies, a longer lag time suppresses them in the same way that a lower temperature suppresses the excited states.

Both weighted operators are conditional expectations, $\overleftarrow{\mathcal U} f = \mathbb{E}[f(X) \mid Y = \cdot\,]$ and $\overrightarrow{\mathcal U} g = \mathbb{E}[g(Y) \mid X = \cdot\,]$, and in an RKHS a conditional expectation becomes an inner product with a conditional mean embedding (Definition~\ref{def: cme}). For $g \in \mathbb{H}_Y$, for instance, $(\overrightarrow{\mathcal U} g)(\bm{x}) = \langle g, \mu_{Y \mid \bm{x}} \rangle_{\mathbb{H}_Y}$, where $\mu_{Y \mid \bm{x}}$ embeds the distribution of the successor of $\bm{x}$. The weighted operators can therefore be expressed through covariance operators of the features. Accordingly, Proposition~\ref{prop: CME} and its counterpart with $X$ and $Y$ interchanged give $\mathcal{C}_{YY} \overleftarrow{\mathcal U} f = \mathcal{C}_{YX} f$ and $\mathcal{C}_{XX} \overrightarrow{\mathcal U} g = \mathcal{C}_{XY} g$, provided that the corresponding conditional expectations belong to $\mathbb{H}_Y$ and $\mathbb{H}_X$, respectively \citep{klus2019kernel}. In particular, the embedded PFO $\overrightarrow{\mathcal{P}}_\varepsilon$ of Section~\ref{sec: Kernel Transfer Operators} satisfies $\overrightarrow{\mathcal{P}}_\varepsilon ( \mathcal{C}_{XX} + \varepsilon \mathcal{I} ) f = \mathcal{C}_{YY} \overleftarrow{\mathcal U} f$, making explicit the relationship between the two representations. In an infinite-dimensional RKHS the covariance operators are compact and generally do not possess globally defined bounded inverses, so the relevant kernel approximations use operators regularized by Tikhonov parameters $\varepsilon_X, \varepsilon_Y > 0$, such as $(\mathcal{C}_{YY} + \varepsilon_Y \mathcal{I})^{-1} \mathcal{C}_{YX}$ and $(\mathcal{C}_{XX} + \varepsilon_X \mathcal{I})^{-1} \mathcal{C}_{XY}$ \citep{klus2019kernel}. Their two-sided composition provides a regularized RKHS approximation of the forward--backward problem. From here, exact identification with the coherent-set eigenproblem in the unregularized limit additionally requires the corresponding RKHS range and invariance conditions \citep{klus2019kernel}. This connection leads naturally to canonical correlation analysis, which is reviewed in Appendix~\ref{sec: Canonical Correlation Analysis}.

\section{Embedding Probability Distributions} \label{sec: Embedding Probability Distributions}

Kernel mean embeddings represent probability distributions and conditional expectations as elements or operators in reproducing kernel Hilbert spaces, with empirical estimates obtained from samples. 

From the outset, KME allows for featurization of the output variable, while GP regression is typically presented with pure scalar-valued outputs\footnote{In this work, we bridge the gap by providing a Bayesian interpretation of $\varphi(Y)$.}. In other words, the output space $\mathbb{Y}$ is not limited to $\mathbb{Y} \subseteq \mathbb{R}$ \citep{10.1145/1553374.1553497}. In contrast with \eqref{eq: GP mean} we directly model $\mathbb{E}_{Y \mid \bm{x}}[g(Y) \mid X = \bm{x}], \; g \in \mathbb{H}_Y$. Apart from this, the mathematical formalisms for the predictive mean \eqref{eq: GP mean} and the kernel CME \eqref{eq: CME eval} are exactly the same, while the predictive variance \eqref{eq: GP cov} is the regularization parameter times a regularized squared Mahalanobis norm of the input feature in the RKHS \citep{chowdhury2017kernelized, pmlr-v189-chowdhury23a}. It is a case where the GP regression and KME communities have developed overlapping tools while having different goals in mind.

A limitation of the KME approach is that it recovers only part of the structure available in GP regression: in particular, other concepts such as the marginal likelihood do not translate directly into the embedding framework.  

\subsection{Marginal Distributions}

The idea of KMEs is to identify probability distributions with elements of an RKHS that fully capture their statistical features. Operationally, we map the data to an RKHS and then we compute the mean. The embeddings are not direct estimates of probability distributions, but rather representative points or elements in an RKHS. For readers interested in embedding probability distributions in RKHSs, excellent resources are \citet{Muandet20171, 6530747, 10.1145/1553374.1553497}.

\begin{definition}
\emph{Marginal Mean Embedding}. \label{def: mean_embedding} 
Let $\mathbb{H}_X$ be an RKHS with a bounded kernel\footnote{Stationary kernels (functions only depending on the distance between the inputs), such as the Matérn or squared exponential kernels, are bounded $\text{sup}_{\bm{x}, \bm{y} \in \mathbb{X}} \kappa(\bm{x},\bm{y}) < \infty$ \citep{williams2006gaussian, ghojogh2021reproducing}.}. Then the kernel mean embedding $\mu_X \in \mathbb{H}_X$ is defined as 
    \begin{alignat*}{3}
        \mu_X \coloneqq \mathbb{E}_X\left[\varphi(X)\right] = \int_{\mathbb{X}}^{} \varphi(\bm{x}) \, \mathrm{d}\mathbb{P}_{X}(\bm{x}),
    \end{alignat*}
where the integral is a Bochner integral \citep{hsing2015theoretical, diestel1977vector}.
\end{definition}

For a large class of kernel functions known as \emph{characteristic kernels} \citep{fukumizu2004dimensionality, sriperumbudur2008injective}, the mean embedding captures all the necessary information about the distribution $\mathbb{P}_{X}$. In other words, the mean map $\mathbb{P}_{X} \rightarrow \mu_X$ is injective and there is no information lost when mapping the distribution into $\mathbb{H}_X$.

By virtue of the reproducing property and linearity of expectation, the marginal embedding $\mu_X$ has the property that the expectation of any Hilbert space function $f\in \mathbb{H}_X$ can be evaluated as an inner product, i.e., $\mathbb{E}_X \left[f(X)\right]= \langle f,\mu_X \rangle_{\mathbb{H}_X}$.

For samples drawn independently and identically distributed (i.i.d.) from $\mathbb{P}_{X}$, the standard unbiased and consistent empirical estimate $\hat{\mu}_X$ of $\mu_X$ is
\begin{alignat*}{3}
        \hat{\mu}_X \coloneqq \frac{1}{N} \sum\limits_{i = 1}^N \kappa(\bm{x}_i, \cdot).
\end{alignat*} 

\subsection{Joint Distributions} \label{subsec: Joint Distributions}

The next step is to generalize to two (or more) random variables. For joint probability distributions we compute covariance operators over tensor product spaces of RKHSs \citep{baker1973joint}. 

\begin{definition}\emph{Joint Mean Embedding}. \label{def: joint_embedding} 
    Let $(Y, X)$ be a joint random variable on $\mathbb{Y}\times \mathbb{X}$ with the joint distribution $\mathbb{P}_{YX}$. Let $\overrightarrow{\kappa}_\textrm{pr}$ and $\overleftarrow{\kappa}_\textrm{pr}$ be characteristic kernels, the prior covariance functions of the forward and the backward lifted model of Section~\ref{sec: Bayesian Forward-Backward Operators}, with respective RKHSs $\mathbb{H}_X$ and $\mathbb{H}_Y$, parameters $\theta_X$ and $\theta_Y$, and associated feature maps $\varphi_X$ and $\varphi_Y$. Following the uncentered covariance-operator convention used for conditional mean embeddings and Kernel Bayes' rule \citep{fukumizu2013kernel}, for which \citet{klebanov2020rigorous} give a rigorous treatment --- while noting that the centered convention requires weaker hypotheses --- the trace-class \emph{(uncentered) covariance operator} $\mathcal{C}_{XX}:\mathbb{H}_X \rightarrow \mathbb{H}_X$ and Hilbert--Schmidt \emph{(uncentered) cross-covariance operator} $\mathcal{C}_{YX} : \mathbb{H}_X \rightarrow \mathbb{H}_Y$ are defined as
    \begin{alignat*}{3}
    \begin{split}
        \mathcal{C}_{YX} &\coloneqq \mathbb{E}_{YX}\left[\varphi_{YX}(Y,X)\right] = \int_{\mathbb{X} \times \mathbb{Y}}^{} \varphi_{YX}(\bm{y},\bm{x}) \, \mathrm{d}\mathbb{P}_{YX}(\bm{y},\bm{x}), \\
        \mathcal{C}_{XX} &\coloneqq \mathbb{E}_X\left[\varphi_{XX}(X,X)\right] = \int_{\mathbb{X}}^{} \varphi_{XX}(\bm{x},\bm{x}) \, \mathrm{d}\mathbb{P}_{X}(\bm{x}),
    \end{split}
    \end{alignat*} 
where the \emph{joint feature map} is defined according to $\varphi_{YX}(\bm{y}, \bm{x}) \coloneqq \varphi_Y(\bm{y}) \otimes \varphi_X(\bm{x})$. Through the joint feature map, a pair of realizations $(\bm{y},\bm{x})$ can be lifted into the tensor-product feature space $\mathbb{H}_{YX} \coloneqq \mathbb{H}_Y \otimes \mathbb{H}_X$, which is identified with the space of Hilbert--Schmidt operators from $\mathbb{H}_X$ to $\mathbb{H}_Y$.

\end{definition}

The cross-covariance operator encodes the joint second moments between functions in $\mathbb{H}_X$ and $\mathbb{H}_Y$, and therefore captures the dependence between $X$ and $Y$ that is represented by the chosen feature spaces. If both $\overrightarrow{\kappa}_\textrm{pr}$ and $\overleftarrow{\kappa}_\textrm{pr}$ are linear kernels such that the feature maps $\varphi_X$ and $\varphi_Y$ are identity maps, we recover the corresponding second-moment matrices. Hence, one may think of $\mathcal{C}_{XX}$ and $\mathcal{C}_{YX}$ as nonlinear generalizations of these matrices. Using these ideas, the uncentered cross-covariance between two functions $f \in \mathbb{H}_X$ and $g \in \mathbb{H}_Y$ is
\begin{alignat*}{3}
    \mathbb{E}_{YX} \left[f(X) \, g(Y) \right] &= \langle f, \mathcal{C}_{XY} \, g \rangle_{\mathbb{H}_X} = \langle \mathcal{C}_{YX} \, f, \, g \rangle_{\mathbb{H}_Y}\\
    &=\langle g \otimes f, \mathcal{C}_{YX} \rangle_{\mathbb{H}_{YX}}.
\end{alignat*}

Therefore, $\mathcal{C}_{XX}$ is a self-adjoint operator and $\mathcal{C}_{YX}$ is the adjoint of $\mathcal{C}_{XY}$. Note that the joint feature map is a rank one operator. Consequently, there is an interesting duality of perspectives at play, because just like the mean embedding was an element in the RKHS $\mu_X \in \mathbb{H}_X$, so the cross-covariance operator is an element in the tensor product feature space $\mathcal{C}_{YX} \in \mathbb{H}_{YX}$.

The empirical estimators for the covariance operators given i.i.d. samples $\{ (\bm{x}_i,\bm{y}_i) \}_{i = 1}^N$ are:
\begin{alignat}{3} \label{eq: estimate covariance operators}
    \begin{split}
            \hat{\mathcal{C}}_{YX} &\coloneqq \frac{1}{N}\sum\limits_{i = 1}^N \varphi_{YX}(\bm{y}_i, \bm{x}_i) = \frac{1}{N}\Phi_Y\Phi_X^\ast,\\
            \hat{\mathcal{C}}_{XX}  &\coloneqq \frac{1}{N}\Phi_X \Phi_X^\ast, \; \text{and} \; \hat{\mathcal{C}}_{YY}  \coloneqq \frac{1}{N}\Phi_Y \Phi_Y^\ast.
    \end{split}
\end{alignat} 

Two conventions are used in what follows. The conditional mean embeddings of the next subsection retain these uncentered second moments, whereas canonical correlation analysis, reviewed in Appendix~\ref{sec: Canonical Correlation Analysis}, uses their centered counterparts.

\subsection{Conditional Distributions} \label{subsec: Conditional Distributions}

We are now in a position to introduce RKHS embeddings of conditional probability distributions, known as \emph{conditional mean embeddings} (CMEs). To do so, we require the following result, which relates the two operators $\mathcal{C}_{XX}$ and $\mathcal{C}_{XY}$.
\begin{proposition} \label{prop: CME}
  If $\, \mathbb{E}_{Y \mid X}\left[g(Y) \! \mid X = \!  \cdot \right] \,  \in \!  \mathbb{H}_X \; \forall \; g \in \mathbb{H}_Y$, then $\mathcal{C}_{XX} \,  \mathbb{E}_{Y \mid X} \left[g(Y) \mid X \!  = \!  \cdot \right] = \mathcal{C}_{XY} \, g$ \citep{fukumizu2004dimensionality, fukumizu2004erratum, fukumizu2013kernel}.
\end{proposition}

\begin{definition} \emph{Conditional Mean Embedding} \citep{10.1145/1553374.1553497}.\label{def: cme}
Assuming that Proposition \ref{prop: CME} holds, the CME of $\mathbb{P}_{(Y \mid X=\cdot)}$ is the operator $\mathcal{C}_{Y \mid X} : \mathbb{H}_X \rightarrow \mathbb{H}_Y$, whereas the evaluation of the conditional distribution $\mathbb{P}_{Y \mid \bm{x}}$ corresponds to the element $\mu_{Y \mid \bm{x}} \in \mathbb{H}_Y$:
    \begin{alignat*}{3}
    \begin{split}
        \mathcal{C}_{Y \mid X} &\coloneqq  \mathcal{C}_{YX} \mathcal{C}_{XX}^{-1}, \\
        \mu_{Y \mid \bm{x}} &\coloneqq \mathcal{C}_{Y \mid X} \text{ }\varphi_X(\bm{x}).
    \end{split}
    \end{alignat*}   
\end{definition}

From the reproducing property it follows that 
\begin{alignat}{3}
    \begin{split}
        \mathbb{E}_{Y \mid \bm{x}}\left[g(Y) \mid X = \bm{x}\right] &= \langle \mathcal{C}_{XX}^{-1} \mathcal{C}_{XY} g, \varphi_X(\bm{x}) \rangle_{\mathbb{H}_X}\\
        &= \langle g, \mu_{Y \mid \bm{x}} \rangle_{\mathbb{H}_Y} \; \forall \; g \in \mathbb{H}_Y \label{eq: CME eval}.
    \end{split}
\end{alignat}

The condition $\mathbb{E}_{Y \mid X}\left[g(Y) \mid X = \cdot \right] \in \mathbb{H}_X \; \forall \; g \in \mathbb{H}_Y$ is restrictive and need not hold in general \citep{fukumizu2004dimensionality, fukumizu2004erratum, fukumizu2013kernel, Muandet20171}. When it does not hold, a standard approach is to replace $\mathcal{C}_{XX}^{-1}$ by the regularized inverse $(\mathcal{C}_{XX} + \varepsilon \mathcal{I})^{-1}$, where $\mathcal{I}$ is the identity operator on $\mathbb{H}_X$ and $\varepsilon > 0$ is a Tikhonov regularization parameter \citep{fukumizu2013kernel, tikhonov1977solutions}. The resulting expression is then a regularized approximation of the conditional expectation rather than the exact identity of Proposition~\ref{prop: CME}. Moreover, $\mathcal{C}_{XX}$ is a compact trace-class operator, with eigenvalues that accumulate at zero when $\mathbb{H}_X$ is infinite dimensional \citep{hsing2015theoretical}, so regularization also renders the inverse problem well posed.

All that is missing from the framework are the estimators for the CME. The empirical estimator of  $\mathcal{C}_{Y \mid X}$ is given by,
\begin{alignat*}{3}
    \hat{\mathcal{C}}_{Y \mid X} &= \hat{\mathcal{C}}_{YX}(\hat{\mathcal{C}}_{XX} + \varepsilon \mathcal{I})^{-1} \\
    &= \frac{1}{N} \Phi_Y \Phi_X^\ast \left(\frac{1}{N}\Phi_X \Phi_X^\ast + \varepsilon \mathcal{I} \right)^{-1}\\
    &= \Phi_Y \left(\K_{XX} + N\varepsilon \I_N \right)^{-1} \Phi_X^\ast\\
    &= \Phi_Y \tilde{\K}_{XX}^{-1} \Phi_X^\ast.
\end{alignat*}
Under the Gaussian observation model of Appendix~\ref{sec: Gaussian Process Regression}, identifying $N\varepsilon$ with the observation variance $\sigma_Y^2$ makes the regularized CME estimator coincide algebraically with the corresponding GP posterior-mean estimator, so that $\varepsilon = \sigma_Y^2/N$.

The empirical estimator of ${\mu}_{Y \mid \bm{x}}$ permits the intuitive description as a weighted sum of dictionary elements $\Phi_Y$, where the weights are dependent on the data point in $\mathbb{X}$ on which we are conditioning. This line of reasoning leads us to
\begin{alignat*}{3}
    \hat{\mu}_{Y \mid \bm{x}} \coloneqq \sum\limits_{i = 1}^{N} \varphi_Y(\bm{y}_i) \alpha_i(\bm{x})  = \Phi_Y \; \bm{\alpha}(\bm{x}),
\end{alignat*}
where the weight vector is
\begin{alignat*}{3}
    \bm{\alpha}(\bm{x}) \coloneqq \tilde{\K}_{XX}^{-1} \; \mathbf{k}_{X}(\bm{x}) \in \mathbb{R}^N.
\end{alignat*}
Now the resemblance between \eqref{eq: CME eval} and \eqref{eq: GP mean} should be apparent. Moreover, it is easy to see that the CME coincides with the expression for the embedded PFO found in \eqref{eq: embedded PFO} when the input and output feature spaces are equal, i.e., $\mathbb{H}_X = \mathbb{H}_Y$. In this case the feature space is $\mathbb{H}_X = \mathbb{H}_Y =: \mathbb{H}$, which together with $\mathbb{G} = \mathbb{H}$ gives the common finite-dimensional dictionary subspace
\[
    \mathbb{G}_M
    =
    \operatorname{span}(\Phi_Z)
    =
    \mathbb{H}_M.
\]

\section{Canonical Correlation Analysis} \label{sec: Canonical Correlation Analysis}

Canonical correlation analysis (CCA) seeks functions of two random variables whose centered evaluations are maximally correlated. In the dynamical setting, CCA provides the statistical counterpart of the forward--backward singular-function problem of Appendix~\ref{subsec: Forward-backward Operators} \citep{klus2019kernel}. Kernel CCA sets the admissible functions to RKHSs, $f \in \mathbb{H}_X$ and $g \in \mathbb{H}_Y$ \citep{bach2002kernel, fukumizu2007statistical}. For the dynamical application considered later, $X$ and its time-lagged successor $Y$ both take values in the same physical state space $\mathbb{X}$, while their endpoint representations may nevertheless use different feature maps $\varphi_X \colon \mathbb{X} \to \mathbb{H}_X$ and $\varphi_Y \colon \mathbb{X} \to \mathbb{H}_Y$. With the mean embeddings $\mu_X$ and $\mu_Y$, let $\varphi_X^\circ(\bm{x}) \coloneqq \varphi_X(\bm{x}) - \mu_X$ and $\varphi_Y^\circ(\bm{y}) \coloneqq \varphi_Y(\bm{y}) - \mu_Y$, and denote the corresponding centered covariance operators by,
\begin{alignat}{3} \label{eq: centered covariance operators}
    \begin{split}
        \mathcal{C}_{XX}^\circ
        &\coloneqq
        \mathbb{E}_X\left[\varphi_X^\circ(X) \otimes \varphi_X^\circ(X)\right]
        =
        \mathcal{C}_{XX} - \mu_X \otimes \mu_X, \\
        \mathcal{C}_{YX}^\circ
        &\coloneqq
        \mathbb{E}_{YX}\left[\varphi_Y^\circ(Y) \otimes \varphi_X^\circ(X)\right]
        =
        \mathcal{C}_{YX} - \mu_Y \otimes \mu_X,
    \end{split}
\end{alignat}
with $\mathcal{C}_{YY}^\circ$ defined analogously and $\mathcal{C}_{XY}^\circ = (\mathcal{C}_{YX}^\circ)^*$, such that $\langle g, \, \mathcal{C}_{YX}^\circ f \rangle_{\mathbb{H}_Y} = \operatorname{Cov}[f(X), \ g(Y)]$.

Then the first canonical correlation is defined as the value of
\begin{alignat}{3} \label{eq: population CCA}
    \begin{split}
        \rho_1
        &\coloneqq
        \sup_{f \in \mathbb{H}_X, \, g \in \mathbb{H}_Y}
        \langle g, \mathcal{C}_{YX}^\circ f \rangle_{\mathbb{H}_Y} \\
        &\phantom{\coloneqq}
        \text{ subject to }
        \langle f, \mathcal{C}_{XX}^\circ f \rangle_{\mathbb{H}_X}
        =
        \langle g, \mathcal{C}_{YY}^\circ g \rangle_{\mathbb{H}_Y}
        =
        1,
    \end{split}
\end{alignat}
whose stationary points satisfy $\mathcal{C}_{YX}^\circ f = \rho \, \mathcal{C}_{YY}^\circ g$ and $\mathcal{C}_{XY}^\circ g = \rho \, \mathcal{C}_{XX}^\circ f$. The problem is then generalized by obtaining further triples $(\rho_i, f_i, g_i)$ that are orthogonal in the corresponding covariance-induced inner products \citep{fukumizu2007statistical, klus2019kernel}.

Under the moment assumptions required for the covariance operators above, these operators are compact in an infinite-dimensional RKHS and generally do not possess globally defined bounded inverses \citep{fukumizu2007statistical, klus2019kernel}. Kernel CCA therefore regularizes the endpoint variances. With the Tikhonov parameters $\varepsilon_X, \varepsilon_Y > 0$ introduced in Appendix~\ref{subsec: Forward-backward Operators}, the constraints are replaced by
\begin{alignat*}{3}
    \left\langle
        f,
        \left(
            \mathcal{C}_{XX}^\circ + \varepsilon_X \mathcal{I}
        \right) f
    \right\rangle_{\mathbb{H}_X}
    &=
    1, \qquad
    &
    \left\langle
        g,
        \left(
            \mathcal{C}_{YY}^\circ + \varepsilon_Y \mathcal{I}
        \right) g
    \right\rangle_{\mathbb{H}_Y}
    &=
    1,
\end{alignat*}
and eliminating $g$ yields the regularized operator,
\begin{alignat}{3} \label{eq: regularized kernel CCA}
    \left(\mathcal{C}_{XX}^\circ + \varepsilon_X \mathcal{I}\right)^{-1}
    \mathcal{C}_{XY}^\circ
    \left(\mathcal{C}_{YY}^\circ + \varepsilon_Y \mathcal{I}\right)^{-1}
    \mathcal{C}_{YX}^\circ \, f
    &=
    \rho^2 f.
\end{alignat}
Equivalently, the regularized canonical correlations are the singular values of the normalized and centered cross-covariance operator
\begin{alignat}{3} \label{eq: whitened CCA operator}
    \mathcal{S}_{YX}^{\circ}
    &\coloneqq
    \left(\mathcal{C}_{YY}^\circ + \varepsilon_Y \mathcal{I}\right)^{-1/2}
    \mathcal{C}_{YX}^\circ
    \left(\mathcal{C}_{XX}^\circ + \varepsilon_X \mathcal{I}\right)^{-1/2}.
\end{alignat}
Thus kernel CCA first whitens the two endpoint feature spaces with their respective regularized covariance operators and then performs a singular value decomposition of the resulting cross-covariance operator \citep{fukumizu2007statistical, klus2019kernel}. Written in the adjoint orientation, $\mathcal{S}_{YX}^{\circ}$ has the same singular values as the finite matrix appearing in \eqref{eq: kZ VAMP2 score}, so this two-sided whitening and singular-value structure is the part of CCA that reappears in the GP-TCCA construction of Appendix~\ref{sec: GP-TCCA}.

When $Y$ is the time-lagged successor of $X$, the canonical functions acquire a dynamical interpretation. By the pairing identity~\eqref{eq: forward-backward pairing}, the quantity maximized by the singular functions of the normalized forward operator is a correlation between a function of the initial state and a function of the final state. Because we are restricted to functions with zero mean, the largest attainable value is the leading nontrivial singular value $\rho_2$ \citep{FROYLAND20131}, and in the unregularized limit below, the first canonical correlation $\rho_1$ of~\eqref{eq: population CCA} equals this value. Centering removes the constant direction, so that the associated unit eigenvalue disappears from the Gram-matrix formulation \citep{klus2019kernel}, and the canonical functions describe only the nontrivial correlated variation between the two endpoints. 

Now suppose that the conditional expectations of Appendix~\ref{subsec: Forward-backward Operators} remain in the respective RKHSs, so that $\overleftarrow{\mathcal U}$ and $\overrightarrow{\mathcal U}$ admit exact kernel representations. Then, as $\varepsilon_X, \varepsilon_Y \to 0$, the kernel-CCA eigenproblem becomes $\overrightarrow{\mathcal{T}} f = \rho^2 f$ restricted to functions with zero mean under $p_X$, the subspace on which the trivial triple $(\rho_1, f_1, g_1) = (1, \id, \id)$ has been removed \citep{klus2019kernel}.

This is also the setting of the variational approach for Markov processes, where the dominant singular components provide optimal low-rank coordinates under the VAMP-$r$ scores \citep{wu2020variational, klus2019kernel}. When coherent sets themselves are required, these coordinates may subsequently be thresholded or clustered.

\newpage

\section{Residual Geometry of the Output Covariance} \label{sec: Output Covariance Geometry}

The preceding appendices review background material, whereas this appendix and the following ones contain the derivations specific to this work. Here we characterize the output covariance $\overrightarrow{\mathcal{B}}$ of Definition~\ref{def: lifted observation model}. The population results are stated for a predecessor RKHS $\mathbb{H}_X$ and a successor RKHS $\mathbb{H}_Y$ with feature maps $\varphi_X$ and $\varphi_Y$, as in Section~\ref{sec: Kernel Transfer Operators} and Appendix~\ref{sec: Embedding Probability Distributions}. From Subsection~\ref{subsec: Sparse Residual Geometry} on, we use the shared prior covariance function $\kappa_\textrm{pr}$, RKHS $\mathbb{H}$, and feature map $\varphi$ of Section~\ref{sec: Finite Output Factor}. We first recall the pointwise conditional covariance, then define the global residual operator and relate it to averaged conditional variance and RKHS closure error. We then derive the posterior of the output covariance function, its sparse counterpart on the dictionary space, the large-sample limit of the finite Bayesian output covariance, the consistency of its posterior, and the posterior variance of the residuals of Section~\ref{sec: Bayesian Consistency}. Finally, we use the same residual geometry to interpret the population VAMP-$2$ score and its relation to Koopman closure. As before, $X$ and $Y$ denote random variables of the underlying joint law, whereas $\X$, $\Y$, and the corresponding matrix quantities contain the finite samples.

\subsection{Local Conditional Covariance} \label{subsec: Local Conditional Covariance}

The conditional mean embedding of Appendix~\ref{subsec: Conditional Distributions} describes the conditional first moment at a fixed source state. For the second-order quantity needed below, we retain only the covariance specialization of the local conditional cross-covariance embedding of \citet{10.1145/1553374.1553497}. For a fixed conditioning value $\bm{x}$, define the conditional second-moment embedding
\begin{alignat*}{3}
    \mu_{YY \, \mid \, \bm{x}}
    &\coloneqq
    \mathbb{E}_{Y \, \mid \, \bm{x}}
    \left[
        \varphi_Y(Y) \otimes \varphi_Y(Y)
        \mid X = \bm{x}
    \right].
\end{alignat*}
The pointwise conditional covariance embedding is
\begin{alignat}{3}
    \mathcal{C}_{YY \, \mid \, \bm{x}}
    &\coloneqq
    \mu_{YY \, \mid \, \bm{x}}
    -
    \mu_{Y \, \mid \, \bm{x}} \otimes \mu_{Y \, \mid \, \bm{x}},
    \label{eq:conditional-covariance-embedding}
\end{alignat}
or, equivalently,
\begin{alignat*}{3}
    \mathcal{C}_{YY \, \mid \, \bm{x}}
    =
    \mathbb{E}_{Y \, \mid \, \bm{x}}
    \left[
        \left(
            \varphi_Y(Y)-\mu_{Y \, \mid \, \bm{x}}
        \right)
        \otimes
        \left(
            \varphi_Y(Y)-\mu_{Y \, \mid \, \bm{x}}
        \right)
        \mid X = \bm{x}
    \right].
\end{alignat*}
Therefore, for any $g, g' \in \mathbb{H}_Y$,
\begin{alignat}{3}
    \left\langle
        g, \,
        \mathcal{C}_{YY \, \mid \, \bm{x}} \, g'
    \right\rangle_{\mathbb{H}_Y}
    &=
    \operatorname{Cov}_{Y \, \mid \, \bm{x}}
    \left[
        g(Y), \, g'(Y)
        \mid X = \bm{x}
    \right].
    \label{eq:conditional-covariance-evaluation}
\end{alignat}
In particular, $\mathcal{C}_{YY \, \mid \, \bm{x}}$ is the RKHS analogue of the conditional covariance matrix of $Y$ at the fixed conditioning value $\bm{x}$.

\subsection{Global Conditional Covariance}

Averaging the local operator~\eqref{eq:conditional-covariance-embedding} over the source states gives the global summary $\mathbb{E}_{X}[\mathcal{C}_{YY \, \mid \, \bm{x}}]$. The global (uncentered) conditional covariance operator $\mathcal{C}_{YY \mid X}$ is instead built from the uncentered second moments of the lifted pair $(\varphi_Y(Y), \varphi_X(X))$, through the residual of the best prediction of $g(Y)$ that is linear, rather than affine, in the lifted source feature $\varphi_X(X)$:
\begin{alignat}{3} \label{eq: global conditional covariance}
    \big\langle g, \, \mathcal{C}_{YY \mid X} \, g \big\rangle_{\mathbb{H}_Y}
    \coloneqq
    \inf_{h \in \mathbb{H}_X}
    \mathbb{E}_{YX}\big[ \big( g(Y) - h(X) \big)^2 \big], \quad \forall g \in \mathbb{H}_Y.
\end{alignat}

In general, $\mathcal{C}_{YY \mid X}$ is only the residual second-moment operator of this prediction problem, and the infimum need not be attained in $\mathbb{H}_X$. Since the predictions carry no free constant term, the optimal residual need not have mean zero. Admitting such a constant turns the second moment into a variance and yields the centered counterpart
\begin{alignat}{3} \label{eq: centered global conditional covariance}
    \big\langle g, \, \mathcal{C}_{YY \mid X}^\circ \, g \big\rangle_{\mathbb{H}_Y}
    \coloneqq
    \inf_{h \in \mathbb{H}_X}
    \mathbb{E}_{YX}\big[ \big( g(Y) - h(X) - c \big)^2 \big],
    \quad \textrm{where } c = \langle g, \, \mu_Y \rangle_{\mathbb{H}_Y} - \langle h, \, \mu_X \rangle_{\mathbb{H}_X},
\end{alignat}
which is the variational form \citep{fukumizu2009kernel7} of the centered Schur complement of \citet{fukumizu2004dimensionality}. Both operators equal $\mathbb{E}_{X}[\mathcal{C}_{YY \, \mid \, \bm{x}}]$ exactly when $\mathcal{U} g$ (Definition~\ref{def: stochastic Koopman operator}) lies in the $L^2(\mathbb{P}_X)$-closure of $\mathbb{H}_X$, or of $\mathbb{H}_X$ together with the constant functions, respectively, for every $g \in \mathbb{H}_Y$. The centered condition holds for every source distribution if the kernel of $\mathbb{H}_X$ is characteristic \citep{fukumizu2009kernel7}. Moreover, the two operators coincide whenever the constant function lies in the closure of $\mathbb{H}_X$. We use $\mathcal{C}_{YY \mid X}$, since a centered operator is linear only on the deviations from the generally different mean embeddings $\mu_X$ and $\mu_Y$. This suffices for singular value analyses such as the kernel CCA of Appendix~\ref{sec: Canonical Correlation Analysis}, whereas the repeated application of the operator in Sections~\ref{sec: Koopman Mode Decomposition} and~\ref{sec: Prediction Theory} requires $\mu_X = \mu_Y$.

\begin{proposition} \label{prop: residual geometry}
    Let the kernels be bounded and let $\mathbb{H}_X$ and $\mathbb{H}_Y$ be separable. Then the following hold.
    \begin{enumerate}[label=(\roman*)]
        \item Equation~\eqref{eq: global conditional covariance} defines a unique bounded self-adjoint operator $\mathcal{C}_{YY \mid X}$ on $\mathbb{H}_Y$. In particular, $\mathcal{C}_{YY \mid X}$ is positive and trace class, and $0 \preceq \mathcal{C}_{YY \mid X} \preceq \mathcal{C}_{YY}$. 
        \item \label{item: residual geometry ii} For every $g \in \mathbb{H}_Y$, with $\mathcal{U} g = \mathbb{E}_{Y \mid X}[g(Y) \mid X = \cdot \,]$ as in Definition~\ref{def: stochastic Koopman operator},
        \[
            \big\langle g, \, \mathcal{C}_{YY \mid X} \, g \big\rangle_{\mathbb{H}_Y}
            =
            \mathbb{E}_{X}\big[ \operatorname{Var}\big( g(Y) \mid X \big) \big]
            +
            \inf_{h \in \mathbb{H}_X}
            \mathbb{E}_{X}\big[ \big( (\mathcal{U} g)(X) - h(X) \big)^2 \big].
        \]
        Hence $\mathcal{C}_{YY \mid X} \succeq \mathbb{E}_{X}[\mathcal{C}_{YY \, \mid \, \bm{x}}]$, with equality if and only if $\mathcal{U} g$ lies in the $L^2(\mathbb{P}_X)$-closure of $\mathbb{H}_X$ for every $g \in \mathbb{H}_Y$.
        \item \label{item: residual geometry iii} If the range of $\mathcal{C}_{XY}$ is contained in the range of $\mathcal{C}_{XX}$, the infimum in~\eqref{eq: global conditional covariance} is attained at $h = \mathcal{C}_{XX}^{-1} \mathcal{C}_{XY} \, g$, and
        \[
            \mathcal{C}_{YY \mid X}
            =
            \mathcal{C}_{YY}
            -
            \mathcal{C}_{YX} \, \mathcal{C}_{XX}^{-1} \, \mathcal{C}_{XY}.
        \]
        Under the condition of Proposition~\ref{prop: CME}, the minimizer is $\mathcal{U} g = \langle g, \mu_{Y \mid \cdot} \rangle_{\mathbb{H}_Y}$, $\mathbb{P}_X$-almost everywhere, i.e., the best linear prediction is given by the CME.
    \end{enumerate}
\end{proposition}

\begin{proof} \leavevmode

    \begin{enumerate}[label=(\roman*)]
        \item By the reproducing property and the boundedness of the kernels, $g \mapsto g(Y)$ and $h \mapsto h(X)$ are bounded linear maps into $L^2(\mathbb{P}_{YX})$. The right-hand side of~\eqref{eq: global conditional covariance} is therefore the squared $L^2(\mathbb{P}_{YX})$-distance of $g(Y)$ from the closure of $\{ h(X) : h \in \mathbb{H}_X \}$, i.e., the squared norm of a residual of orthogonal projection that depends linearly and boundedly on $g$. It is thus the quadratic form of a bounded, symmetric, positive bilinear form on $\mathbb{H}_Y$, which is represented by a unique bounded self-adjoint operator. The choice $h \equiv 0$ gives
        \[
            \big\langle g, \, \mathcal{C}_{YY \mid X} \, g \big\rangle_{\mathbb{H}_Y}
            \leq
            \mathbb{E}_{Y}\big[ g(Y)^2 \big]
            =
            \big\langle g, \, \mathcal{C}_{YY} \, g \big\rangle_{\mathbb{H}_Y},
        \]
        and since $\mathcal{C}_{YY}$ is positive and trace class (Definition~\ref{def: joint_embedding}), so is $\mathcal{C}_{YY \mid X}$.
    
        \item The function $\mathcal{U} g$ is bounded because $g$ is, and $g(Y) - (\mathcal{U} g)(X)$ is orthogonal in $L^2(\mathbb{P}_{YX})$ to every square-integrable function of $X$. Hence, for every $h \in \mathbb{H}_X$,
        \[
            \mathbb{E}_{YX}\big[ \big( g(Y) - h(X) \big)^2 \big]
            =
            \mathbb{E}_{YX}\big[ \big( g(Y) - (\mathcal{U} g)(X) \big)^2 \big]
            +
            \mathbb{E}_{X}\big[ \big( (\mathcal{U} g)(X) - h(X) \big)^2 \big],
        \]
        and by~\eqref{eq:conditional-covariance-evaluation} the first term on the right equals
        \[
            \mathbb{E}_{X}\big[ \operatorname{Var}\big( g(Y) \mid X \big) \big]
            =
            \big\langle g, \, \mathbb{E}_{X}[\mathcal{C}_{YY \, \mid \, \bm{x}}] \, g \big\rangle_{\mathbb{H}_Y}.
        \]
        Taking the infimum over $h$ gives the identity, and the operator statements follow because self-adjoint operators with equal quadratic forms are equal.
    
        \item By the range inclusion, $\mathcal{C}_{XX} \, \mathcal{C}_{XX}^{-1} \mathcal{C}_{XY} \, g = \mathcal{C}_{XY} \, g$. Expanding the square with $\mathbb{E}_{YX}[h(X) \, g(Y)] = \langle h, \mathcal{C}_{XY} \, g \rangle_{\mathbb{H}_X}$ gives
        \begin{alignat*}{3}
            \mathbb{E}_{YX}\big[ \big( g(Y) - h(X) \big)^2 \big]
            &=
            \big\langle g, \, \big( \mathcal{C}_{YY} - \mathcal{C}_{YX} \, \mathcal{C}_{XX}^{-1} \, \mathcal{C}_{XY} \big) \, g \big\rangle_{\mathbb{H}_Y}
            \\
            &\phantom{=}
            +
            \big\langle h - \mathcal{C}_{XX}^{-1} \mathcal{C}_{XY} \, g, \, \mathcal{C}_{XX} \, \big( h - \mathcal{C}_{XX}^{-1} \mathcal{C}_{XY} \, g \big) \big\rangle_{\mathbb{H}_X},
        \end{alignat*}
        whose last term is nonnegative and vanishes at $h = \mathcal{C}_{XX}^{-1} \mathcal{C}_{XY} \, g$. Under the condition of Proposition~\ref{prop: CME}, $\mathcal{C}_{XX} \, \mathcal{U} g = \mathcal{C}_{XY} \, g$, and hence $\mathcal{U} g - \mathcal{C}_{XX}^{-1} \mathcal{C}_{XY} \, g$ lies in the null space of $\mathcal{C}_{XX}$ and vanishes $\mathbb{P}_X$-almost everywhere. Hence $\mathcal{U} g$ is also a minimizer, and it equals $\langle g, \mu_{Y \mid \cdot} \rangle_{\mathbb{H}_Y}$ by~\eqref{eq: CME eval}.
    \end{enumerate}
\end{proof}

Consequently, for every $g \in \mathbb{H}_Y$ whose conditional expectation $\mathcal{U} g$ lies in the $L^2(\mathbb{P}_X)$-closure of $\mathbb{H}_X$, in particular under the condition of Proposition~\ref{prop: CME},
\begin{alignat}{3} \label{eq: global covariance characterization}
    \left\langle g, \, \mathcal{C}_{YY \mid X} \, g\right\rangle_{\mathbb{H}_Y}
    =
    \left\langle g, \, \mathcal{C}_{YY \mid X}^\circ \, g\right\rangle_{\mathbb{H}_Y}
    =
    \mathbb{E}_{X}\left[\operatorname{Var}\left(g(Y)\mid X\right)\right],
\end{alignat}
where the first equality follows from $\mathbb{E}_{X}[\mathcal{C}_{YY \, \mid \, \bm{x}}] \preceq \mathcal{C}_{YY \mid X}^\circ \preceq \mathcal{C}_{YY \mid X}$, and if this holds for every $g \in \mathbb{H}_Y$, the population relation
\begin{alignat}{3} \label{eq: global local covariance relation}
    \mathcal{C}_{YY \mid X}
    =
    \mathcal{C}_{YY \mid X}^\circ
    =
    \mathbb{E}_{X}\left[\mathcal{C}_{YY \, \mid \, \bm{x}}\right]
\end{alignat}
holds in the weak operator sense. Thus, whereas $\mathcal{C}_{YY \, \mid \, \bm{x}}$ describes the conditional second-order structure at a particular conditioning value, $\mathcal{C}_{YY \mid X}$ summarizes that structure over the source distribution. In general, the additional term in Proposition~\ref{prop: residual geometry}\ref{item: residual geometry ii} is the second moment of the error of the best $L^2(\mathbb{P}_X)$ approximation of the conditional expectations $\mathcal{U} g$ by elements of $\mathbb{H}_X$, i.e., a closure error of the Koopman operator on the RKHS. Together with the averaged conditional variance, this term mirrors the split of the one-step risk into an irreducible part and an excess part by \citet{kostic2023sharp}. 

Proposition~\ref{prop: residual geometry} and its proof remain valid when the predictions $h$ are restricted to a closed subspace of $\mathbb{H}_X$, such as the dictionary space $\mathbb{H}_M$, with $\mathcal{C}_{XX}$ and $\mathcal{C}_{XY}$ compressed to that subspace. In finite dimensions the range inclusion of Proposition~\ref{prop: residual geometry}\ref{item: residual geometry iii} always holds, because $\mathcal{C}_{XY} \, g$ is orthogonal to the null space of $\mathcal{C}_{XX}$. The infimum in $\mathbb{H}_M$ is therefore attained, and the additional term is the second moment of the dictionary closure error.

The operator $\mathcal{C}_{YY \mid X}$ also justifies the point-independent Gaussian working likelihood of Definition~\ref{def: lifted observation model}. Since $\mathbb{E}_{YX}[(g(Y) - h(X))^2]$ is the quadratic form of the block operator of uncentered second moments of $(\varphi_Y(Y), \varphi_X(X))$ on $\mathbb{H}_Y \oplus \mathbb{H}_X$,
\[
    \begin{bmatrix}
        \mathcal{C}_{YY} & \mathcal{C}_{YX} \\
        \mathcal{C}_{XY} & \mathcal{C}_{XX}
    \end{bmatrix},
\]
evaluated at $(g, -h)$, definition~\eqref{eq: global conditional covariance} is the variational characterization of the shorted operator of this block operator to $\mathbb{H}_Y$ \citep{owhadi2018conditioning}. For bounded kernels the block operator is positive and trace class and therefore the covariance operator of a zero-mean Gaussian random element of $\mathbb{H}_Y \oplus \mathbb{H}_X$. The conditional covariance of its first component given the second is this shorted operator for every conditioning value \citep{owhadi2018conditioning}. Hence $\mathcal{C}_{YY \mid X}$ is the conditional covariance of a Gaussian pair with the same uncentered second moments as the lifted pair. The working likelihood thus matches the lifted pair in its second moments, even though the lifted pair is generally not Gaussian.

\subsection{Posterior of the Output Covariance Function}

Besides being unknown, the population geometry $\overrightarrow{\mathcal{B}}$ can be singular, which is a further reason for treating the output covariance as random in Section~\ref{sec: Operators as Random Variables}. For deterministic dynamics, an observable $g \in \mathbb{H}_Y$ with $\mathcal{U} g \in \mathbb{H}_X$ satisfies $\langle g, \overrightarrow{\mathcal{B}} \, g \rangle_{\mathbb{H}_Y} = 0$ by Proposition~\ref{prop: residual geometry}\ref{item: residual geometry ii}, since both the conditional variance and the closure error vanish. Consider a model in which the output covariance is fixed at $\overrightarrow{\mathcal{B}}$. By Definition~\ref{def: lifted observation model}, the evaluation of a lifted target in the direction $g$ splits into a prediction and a residual,
\[
    g(\bm{y}_i)
    =
    \langle g, \mu_{Y \mid \bm{x}_i} \rangle_{\mathbb{H}_Y}
    +
    \langle g, \overrightarrow{\nu} \rangle_{\mathbb{H}_Y} ,
\]
whose prior and residual variances are
\[
    \overrightarrow{\kappa}_\textrm{pr}(\bm{x}_i, \bm{x}_i) \, \langle g, \overrightarrow{\mathcal{B}} \, g \rangle_{\mathbb{H}_Y}
    =
    0 ,
    \qquad
    \sigma_Y^2 \, \langle g, \overrightarrow{\mathcal{B}} \, g \rangle_{\mathbb{H}_Y}
    =
    0 .
\]
Both terms have mean zero, and the model therefore fixes $g(\bm{y}_i)$ at zero. The data instead satisfy $g(\bm{y}_i) = (\mathcal{U} g)(\bm{x}_i)$, which is generally nonzero, and such a model assigns zero likelihood to the observed transitions. Under the inverse-Wishart process prior, by contrast, $\overrightarrow{\B}$ is almost surely nonsingular for each finite target-side set $\Z_Y$ of distinct evaluation points \citep{dawid1981some}, and the finite-evaluation likelihood therefore remains nondegenerate.

The finite-sample identities needed below are summarized in the following proposition.

\begin{proposition} \label{prop: posterior residual geometry}
    Let $\hat{\mathcal{C}}_{XY} \coloneqq \hat{\mathcal{C}}_{YX}^{*}$. For every $g \in \mathbb{H}_Y$, the minimum in~\eqref{eq: posterior conditional covariance} is uniquely attained at
    \[
        h_g
        =
        \left(
            \hat{\mathcal{C}}_{XX}
            +
            \frac{\sigma_Y^2}{N}\mathcal{I}
        \right)^{-1}
        \hat{\mathcal{C}}_{XY} \, g,
    \]
    and
    \begin{alignat*}{3}
        \hat{\mathcal{C}}_{YY \mid X}
        &=
        \hat{\mathcal{C}}_{YY}
        +
        \frac{\sigma_X^2}{N}\mathcal{I}
        -
        \hat{\mathcal{C}}_{YX}
        \left(
            \hat{\mathcal{C}}_{XX}
            +
            \frac{\sigma_Y^2}{N}\mathcal{I}
        \right)^{-1}
        \hat{\mathcal{C}}_{XY}
        \\
        &=
        \frac{1}{N}
        \left(
            \sigma_Y^2
            \Phi_Y
            \tilde{\K}_{XX}^{-1}
            \Phi_Y^\ast
            +
            \sigma_X^2\mathcal{I}
        \right).
    \end{alignat*}

    Therefore for every finite target-side set $\Z_Y$ of distinct points, the posterior marginal of the output covariance is
    \[
        \overrightarrow{\B} \mid \mathcal{D}
        \sim
        \mathcal{IW}^{|\Z_Y|}
        \left(
            2+N,\,
            N\,\overrightarrow{b}_\textrm{pst}(\Z_Y,\Z_Y)
        \right).
    \]
    Hence, for every $\bm{a} \neq \bm{0}$ with $g = \Phi_{Z_Y} \bm{a}$, the one-step residual variance of $g(Y)$ averaged over the source states, in units of $\sigma_Y^2$, i.e., $\bm{a}^\top \overrightarrow{\B} \bm{a}$, satisfies
        \[
        \bm{a}^\top \overrightarrow{\B} \bm{a} \mid \mathcal{D}
        \sim
        \mathcal{IW}^1 \big( 2 + N , \, N \langle g, \, \overrightarrow{\mathcal{B}}_\kappa \, g \rangle_{\mathbb{H}_Y} \big) ,
    \]
    In particular,
    \[
        \mathbb{E}
        \left[
            \bm{a}^\top \overrightarrow{\B}\bm{a}
            \mid
            \mathcal{D}
        \right]
        =
        \langle g, \, \overrightarrow{\mathcal{B}}_\kappa \, g \rangle_{\mathbb{H}_Y},
    \]
    while for $N>2$,
    \[
        \operatorname{Var}
        \left[
            \bm{a}^\top \overrightarrow{\B}\bm{a}
            \mid
            \mathcal{D}
        \right]
        =
        \frac{
            2\langle g, \, \overrightarrow{\mathcal{B}}_\kappa \, g \rangle_{\mathbb{H}_Y}^2
        }{
            N-2
        }.
    \]
\end{proposition}

\begin{proof}
For fixed $g$, the definition~\eqref{eq: posterior conditional covariance} reads, after expanding the square,
\begin{alignat*}{3}
    \big\langle g, \, \hat{\mathcal{C}}_{YY \mid X} \, g \big\rangle_{\mathbb{H}_Y}
    &=
    \min_{h \in \mathbb{H}_X}
    \Big\{
        \frac{1}{N} \sum_{i=1}^{N} \big( g(\bm{y}_i) - h(\bm{x}_i) \big)^2
        +
        \frac{\sigma_Y^2}{N} \, \| h \|_{\mathbb{H}_X}^2
    \Big\}
    +
    \frac{\sigma_X^2}{N} \, \| g \|_{\mathbb{H}_Y}^2
    \\
    &=
    \min_{h \in \mathbb{H}_X}
    \Big\{
        \frac{1}{N} \sum_{i=1}^{N} g(\bm{y}_i)^2
        - \frac{2}{N} \sum_{i=1}^{N} h(\bm{x}_i) \, g(\bm{y}_i)
        + \frac{1}{N} \sum_{i=1}^{N} h(\bm{x}_i)^2
        + \frac{\sigma_Y^2}{N} \, \| h \|_{\mathbb{H}_X}^2
    \Big\}
    +
    \frac{\sigma_X^2}{N} \, \| g \|_{\mathbb{H}_Y}^2 .
\end{alignat*}
By the reproducing property, $g(\bm{y}_i) = \langle g, \, \varphi_Y(\bm{y}_i) \rangle_{\mathbb{H}_Y}$ and $h(\bm{x}_i) = \langle h, \, \varphi_X(\bm{x}_i) \rangle_{\mathbb{H}_X}$, and the empirical operators~\eqref{eq: estimate covariance operators} turn the three sums into quadratic forms,
\begin{alignat*}{3}
    \frac{1}{N} \sum_{i=1}^{N} g(\bm{y}_i)^2
    &=
    \langle g, \, \hat{\mathcal{C}}_{YY} \, g \rangle_{\mathbb{H}_Y} ,
    \quad
    \frac{1}{N} \sum_{i=1}^{N} h(\bm{x}_i) \, g(\bm{y}_i)
    &=
    \langle h, \, \hat{\mathcal{C}}_{XY} \, g \rangle_{\mathbb{H}_X} ,
    \quad
    \frac{1}{N} \sum_{i=1}^{N} h(\bm{x}_i)^2
    &=
    \langle h, \, \hat{\mathcal{C}}_{XX} \, h \rangle_{\mathbb{H}_X} .
\end{alignat*}
Together with $\| h \|_{\mathbb{H}_X}^2 = \langle h, \, \mathcal{I} \, h \rangle_{\mathbb{H}_X}$, the expression in braces becomes
\begin{alignat*}{3}
    \langle g, \, \hat{\mathcal{C}}_{YY} \, g \rangle_{\mathbb{H}_Y}
    - 2 \, \langle h, \, \hat{\mathcal{C}}_{XY} \, g \rangle_{\mathbb{H}_X}
    + \Big\langle h, \, \Big( \hat{\mathcal{C}}_{XX} + \frac{\sigma_Y^2}{N} \mathcal{I} \Big) \, h \Big\rangle_{\mathbb{H}_X} .
\end{alignat*}
Since $\sigma_Y > 0$, the regularized operator in the third term is strictly positive and therefore invertible. With the minimizer $h_g$ stated in the proposition, the terms that depend on $h$ satisfy
\begin{alignat*}{3}
    &- 2 \, \langle h, \, \hat{\mathcal{C}}_{XY} \, g \rangle_{\mathbb{H}_X}
    +
    \Big\langle h, \, \Big( \hat{\mathcal{C}}_{XX} + \frac{\sigma_Y^2}{N} \mathcal{I} \Big) \, h \Big\rangle_{\mathbb{H}_X}
    \\
    &\quad =
    \Big\langle h - h_g, \, \Big( \hat{\mathcal{C}}_{XX} + \frac{\sigma_Y^2}{N} \mathcal{I} \Big) \, ( h - h_g ) \Big\rangle_{\mathbb{H}_X}
    -
    \Big\langle g, \, \hat{\mathcal{C}}_{YX} \Big( \hat{\mathcal{C}}_{XX} + \frac{\sigma_Y^2}{N} \mathcal{I} \Big)^{-1} \hat{\mathcal{C}}_{XY} \, g \Big\rangle_{\mathbb{H}_Y},
\end{alignat*}
which follows by expanding the first term on the right and using $\big( \hat{\mathcal{C}}_{XX} + \frac{\sigma_Y^2}{N} \mathcal{I} \big) \, h_g = \hat{\mathcal{C}}_{XY} \, g$ and $\hat{\mathcal{C}}_{YX} = \hat{\mathcal{C}}_{XY}^\ast$. The first term on the right is nonnegative and vanishes only at $h = h_g$, which is therefore the unique minimizer. Adding the target-side term to the minimum value gives the quadratic form of the first expression for $\hat{\mathcal{C}}_{YY \mid X}$, and self-adjoint operators with equal quadratic forms are equal.

For the second expression, we insert the empirical operators~\eqref{eq: estimate covariance operators}, whose factors $1 / N$ combine to
\begin{alignat*}{3}
    \hat{\mathcal{C}}_{YX} \Big( \hat{\mathcal{C}}_{XX} + \frac{\sigma_Y^2}{N} \mathcal{I} \Big)^{-1} \hat{\mathcal{C}}_{XY}
    &=
    \frac{1}{N} \Phi_Y \Phi_X^\ast
    \Big( \frac{1}{N} \big( \Phi_X \Phi_X^\ast + \sigma_Y^2 \mathcal{I} \big) \Big)^{-1}
    \frac{1}{N} \Phi_X \Phi_Y^\ast
    \\
    &=
    \frac{1}{N} \Phi_Y \Phi_X^\ast
    \big( \Phi_X \Phi_X^\ast + \sigma_Y^2 \mathcal{I} \big)^{-1}
    \Phi_X \Phi_Y^\ast .
\end{alignat*}
By the push-through identity
\begin{alignat*}{3}
    \Phi_X^\ast \big( \Phi_X \Phi_X^\ast + \sigma_Y^2 \mathcal{I} \big)^{-1}
    =
    \tilde{\K}_{XX}^{-1} \Phi_X^\ast
\end{alignat*}
and $\Phi_X^\ast \Phi_X = \K_{XX} = \tilde{\K}_{XX} - \sigma_Y^2 \I_N$, the product above becomes
\begin{alignat*}{3}
    \frac{1}{N} \Phi_Y \tilde{\K}_{XX}^{-1} \K_{XX} \Phi_Y^\ast
    =
    \frac{1}{N} \Phi_Y \big( \I_N - \sigma_Y^2 \tilde{\K}_{XX}^{-1} \big) \Phi_Y^\ast .
\end{alignat*}
Subtracting it from $\hat{\mathcal{C}}_{YY} = \frac{1}{N} \Phi_Y \Phi_Y^\ast$ and adding the target-side term gives
\begin{alignat*}{3}
    \hat{\mathcal{C}}_{YY \mid X}
    =
    \frac{1}{N}
    \big(
        \sigma_Y^2 \,
        \Phi_Y \tilde{\K}_{XX}^{-1} \Phi_Y^\ast
        +
        \sigma_X^2 \mathcal{I}
    \big).
\end{alignat*}

Now fix a finite target-side set $\Z_Y$, at which the output covariance function enters only through $\overrightarrow{\B} = \overrightarrow{b}(\Z_Y, \Z_Y)$. By Section~\ref{sec: Operators as Random Variables}, the lifted targets and the embedded PFO satisfy
\begin{alignat*}{3}
    \operatorname{vec} \K_{ZY} \mid \mu_{Y \mid \X} , \overrightarrow{\B}
    &\sim
    \mathcal{N} \big( \operatorname{vec} ( \Phi_{Z_Y}^\ast \mu_{Y \mid \X} ) , \, \I_N \otimes \sigma_Y^2 \overrightarrow{\B} \big) ,
    \\
    \operatorname{vec} ( \Phi_{Z_Y}^\ast \mu_{Y \mid \X} ) \mid \overrightarrow{\B}
    &\sim
    \mathcal{N} \big( \bm{0} , \, \K_{XX} \otimes \overrightarrow{\B} \big) .
\end{alignat*}
Integrating out the embedded PFO adds the two covariances, and their common factor $\overrightarrow{\B}$ gives
\begin{alignat*}{3}
    \operatorname{vec} \K_{ZY} \mid \overrightarrow{\B}
    \sim
    \mathcal{N} \big( \bm{0} , \, ( \K_{XX} + \sigma_Y^2 \I_N ) \otimes \overrightarrow{\B} \big)
    =
    \mathcal{N} \big( \bm{0} , \, \tilde{\K}_{XX} \otimes \overrightarrow{\B} \big) .
\end{alignat*}
Whitening the transitions with the symmetric inverse square root of $\tilde{\K}_{XX}$ gives
\begin{alignat*}{3}
    \operatorname{vec} \big( \K_{ZY} \tilde{\K}_{XX}^{-1/2} \big) \mid \overrightarrow{\B}
    \sim
    \mathcal{N} \big( \bm{0} , \, \I_N \otimes \overrightarrow{\B} \big) ,
\end{alignat*}
i.e., the $N$ columns of $\K_{ZY} \tilde{\K}_{XX}^{-1/2}$ are independent with distribution $\mathcal{N} ( \bm{0} , \overrightarrow{\B} )$. The scatter matrix is the sum of their outer products,
\begin{alignat*}{3}
    \K_{ZY} \tilde{\K}_{XX}^{-1} \K_{ZY}^\top
    =
    \big( \K_{ZY} \tilde{\K}_{XX}^{-1/2} \big) \big( \K_{ZY} \tilde{\K}_{XX}^{-1/2} \big)^\top ,
\end{alignat*}
and by the definition of the Wishart distribution,
\begin{alignat*}{3}
    \K_{ZY} \tilde{\K}_{XX}^{-1} \K_{ZY}^\top \mid \overrightarrow{\B}
    \sim
    \mathcal{W} \big( N , \, \overrightarrow{\B} \big) .
\end{alignat*}
The inverse-Wishart conjugate update therefore raises the index from $2$ to $2+N$ and adds this scatter matrix to the prior scale \citep{dawid1981some}. Using the expression for $\hat{\mathcal{C}}_{YY \mid X}$ above,
\[
    \overrightarrow{b}_\textrm{pr}(\Z_Y,\Z_Y)
    +
    \K_{ZY}\tilde{\K}_{XX}^{-1}\K_{ZY}^\top
    =
    \Phi_{Z_Y}^\ast
    \left(
        \frac{\sigma_X^2}{\sigma_Y^2}\mathcal{I}
        +
        \Phi_Y\tilde{\K}_{XX}^{-1}\Phi_Y^\ast
    \right)
    \Phi_{Z_Y}
    =
    N\,\overrightarrow{b}_\textrm{pst}(\Z_Y,\Z_Y),
\]
which gives the stated finite-dimensional posterior. These marginals are consistent under marginalization \citep{dawid1981some,shah2014student}, and give the inverse-Wishart-process posterior stated in Section~\ref{sec: Operators as Random Variables}.

Finally, the affine scale property of the inverse-Wishart distribution gives the one-dimensional law stated above, which is a scaled inverse chi-squared distribution \citep{dawid1981some},
\[
    \frac{
        N \langle g, \, \overrightarrow{\mathcal{B}}_\kappa \, g \rangle_{\mathbb{H}_Y}
    }{
        \bm{a}^\top \overrightarrow{\B} \bm{a}
    }
    \mid
    \mathcal{D}
    \overset{d}{=}
    \chi_{N+2}^{2},
\]
where $\chi_{N+2}^{2}$ denotes a chi-squared random variable with $N + 2$ degrees of freedom. For $N > 2$, its reciprocal has the moments
\[
    \mathbb{E}\left[ ( \chi_{N+2}^{2} )^{-1} \right]
    =
    \frac{1}{N} ,
    \qquad
    \operatorname{Var}\left[ ( \chi_{N+2}^{2} )^{-1} \right]
    =
    \frac{2}{N^2 (N - 2)} .
\]
Given the training data set $\mathcal{D}$, the quadratic form $\bm{a}^\top \overrightarrow{\B} \bm{a}$ is distributed as $N \langle g, \, \overrightarrow{\mathcal{B}}_\kappa \, g \rangle_{\mathbb{H}_Y} \, ( \chi_{N+2}^{2} )^{-1}$, and its posterior moments are
\begin{alignat*}{3}
    \mathbb{E}\left[ \bm{a}^\top \overrightarrow{\B} \bm{a} \mid \mathcal{D} \right]
    &=
    N \langle g, \, \overrightarrow{\mathcal{B}}_\kappa \, g \rangle_{\mathbb{H}_Y} \, \frac{1}{N}
    =
    \langle g, \, \overrightarrow{\mathcal{B}}_\kappa \, g \rangle_{\mathbb{H}_Y} ,
    \\
    \operatorname{Var}\left[ \bm{a}^\top \overrightarrow{\B} \bm{a} \mid \mathcal{D} \right]
    &=
    N^2 \langle g, \, \overrightarrow{\mathcal{B}}_\kappa \, g \rangle_{\mathbb{H}_Y}^2 \, \frac{2}{N^2 (N - 2)}
    =
    \frac{2 \langle g, \, \overrightarrow{\mathcal{B}}_\kappa \, g \rangle_{\mathbb{H}_Y}^2}{N - 2} ,
\end{alignat*}
as stated in the proposition.
\end{proof}

Its posterior law, whose spread shrinks with $N$, quantifies the epistemic uncertainty about the residual variance of $g(Y)$, which sets the error bars of the residual diagnostics in Appendix~\ref{subsec: Posterior Variance of the Residuals}. The epistemic uncertainty of the transfer operator is instead carried by the posterior kernel $\overrightarrow{\kappa}_\textrm{pst}$.

\subsection{Sparse Residual Geometry} \label{subsec: Sparse Residual Geometry}

For the sparse model of Section~\ref{sec: Finite Output Factor}, the matrix $\widetilde{\C}_{YY \mid X}$ defined in~\eqref{eq: KYYX} plays the corresponding role in the RKHS $\widetilde{\mathbb{H}}$ of the augmented kernel $\widetilde{\kappa}_{\textrm{pr}} = \kappa_{\textrm{pr}} + \sigma_Z^2 \delta$, where $\sigma_Z^2$ is the variance of the noisy inducing variables of Section~\ref{sec: Sparse Transfer Operators} and the Kronecker delta acts on the dictionary elements only. The corresponding feature operator satisfies $\widetilde{\Phi}_Z^{\ast} \widetilde{\Phi}_Z = \widetilde{\K}_{ZZ}$, and since the Kronecker delta never couples a dictionary element with a state, $\widetilde{\Phi}_Z \bm{c}$ agrees with $\Phi_Z \bm{c}$ on the state space for every $\bm{c} \in \mathbb{R}^M$. For every $\bm{a} \in \mathbb{R}^M$ and $g = \widetilde{\Phi}_Z \bm{a}$,
\begin{alignat}{3} \label{eq: sparse residual geometry}
    \bm{a}^\top \widetilde{\C}_{YY \mid X} \, \bm{a}
    &=
    \min_{h \in \operatorname{span}(\widetilde{\Phi}_Z)}
    \Big\{
        \sum_{i = 1}^{N}
        \big( g(\bm{y}_i) - h(\bm{x}_i) \big)^2
        +
        \sigma_Y^2 \, \| h \|_{\widetilde{\mathbb{H}}}^2
    \Big\}
    +
    \sigma_X^2 \, \| g \|_{\widetilde{\mathbb{H}}}^2 .
\end{alignat}
The minimum is attained at $h = \widetilde{\Phi}_Z \overrightarrow{\U}_\kappa \bm{a}$, with the Koopman matrix of~\eqref{eq: CME matrix}. On the state space, this minimizer agrees with the one-step prediction $\overrightarrow{\mathcal{U}}_\kappa \Phi_Z \bm{a}$ of the estimated kernel Koopman operator.

To verify this identity, write each candidate function as $h = \widetilde{\Phi}_Z \bm{c}$. The reproducing property gives $g(\bm{y}_i) = \bm{a}^\top \mathbf{k}_Z(\bm{y}_i)$ and $h(\bm{x}_i) = \bm{c}^\top \mathbf{k}_Z(\bm{x}_i)$, whereas $\| h \|_{\widetilde{\mathbb{H}}}^2 = \bm{c}^\top \widetilde{\K}_{ZZ} \bm{c}$ and $\| g \|_{\widetilde{\mathbb{H}}}^2 = \bm{a}^\top \widetilde{\K}_{ZZ} \bm{a}$. The expression in braces therefore equals
\[
    \left\|
        \K_{ZY}^\top \bm{a}
        -
        \K_{ZX}^\top \bm{c}
    \right\|^2
    +
    \sigma_Y^2 \,
    \bm{c}^\top \widetilde{\K}_{ZZ} \bm{c}.
\]
The normal equations are
\[
    \widetilde{\C}_{XX} \bm{c}
    =
    \C_{XY} \bm{a},
\]
whose solution $\bm{c} = \widetilde{\C}_{XX}^{-1} \C_{XY} \bm{a} = \overrightarrow{\U}_\kappa \bm{a}$ attains the minimum value
\[
    \bm{a}^\top
    \left(
        \C_{YY}
        -
        \C_{YX}
        \widetilde{\C}_{XX}^{-1}
        \C_{XY}
    \right)
    \bm{a}.
\]
Adding $\sigma_X^2 \, \bm{a}^\top \widetilde{\K}_{ZZ} \bm{a}$ gives
\[
    \bm{a}^\top
    \left(
        \widetilde{\C}_{YY}
        -
        \C_{YX}
        \widetilde{\C}_{XX}^{-1}
        \C_{XY}
    \right)
    \bm{a}
    =
    \bm{a}^\top
    \widetilde{\C}_{YY \mid X}
    \bm{a},
\]
which proves~\eqref{eq: sparse residual geometry}.

Hence~\eqref{eq: sparse residual geometry} is $N$ times the restriction of~\eqref{eq: posterior conditional covariance}, posed on $\widetilde{\mathbb{H}}$, to the dictionary span, and $N^{-1} \widetilde{\C}_{YY \mid X}$ is the sparse counterpart of $\hat{\mathcal{C}}_{YY \mid X}$.

\subsection{Large-sample Behavior of the Output Covariance} \label{subsec: Large-sample Limit}

What does the output covariance actually learn from infinitely many sampled transitions? In correctly specified multivariate regression, the posterior of the error covariance concentrates at the true error covariance \citep{sarkar2025posterior}. However, the lifted regression is generally misspecified, since the conditional expectation $\mathcal{U} g$ need not lie in the dictionary space $\mathbb{H}_M$, and the posterior instead concentrates at the residual geometry of Proposition~\ref{prop: residual geometry} restricted to $\mathbb{H}_M$. More data therefore sharpens the posterior, but only a better dictionary lowers its limit. This limit underlies the large-sample readings of Section~\ref{sec: Bayesian Consistency}, where a lasting deviation of the decoded covariance from the state-space residual variance points to the dictionary, the linear decoder, or the state-space model.

\begin{theorem} \label{thm: output covariance consistency}
    Let the transitions be independent draws from the joint law of $(X, Y)$, or a stationary ergodic sequence with this law, let the kernel be bounded with fixed hyperparameters, and let the pseudo-inputs, $\sigma_Z$, $\sigma_X$, and $\sigma_Y$ be fixed with $\Phi_Z^\ast \mathcal{C}_{XX} \Phi_Z$ positive definite. Then, almost surely, for every $\bm{a} \in \mathbb{R}^M$ and $g = \Phi_Z \bm{a}$, the posterior law of $\bm{a}^\top \overrightarrow{\B} \bm{a}$ in the sparse model converges weakly to the point mass at
    \[
        \frac{1}{\sigma_Y^2}
        \Big(
            \mathbb{E}_X \big[ \operatorname{Var} \big( g(Y) \mid X \big) \big]
            +
            \inf_{h \in \mathbb{H}_M}
            \mathbb{E}_X \big[ \big( (\mathcal{U} g)(X) - h(X) \big)^2 \big]
        \Big) .
    \]
\end{theorem}

\begin{proof}
    Fix $\bm{a} \neq \bm{0}$ with $g = \Phi_Z \bm{a}$, the case $\bm{a} = \bm{0}$ being trivial. The proof combines the convergence of the posterior mean with the vanishing of the posterior variance.

    First, by~\eqref{eq: KYYX}, the posterior mean satisfies
    \begin{alignat*}{3}
        \sigma_Y^2 \, \overrightarrow{\B}_\kappa
        &=
        \frac{1}{N} \C_{YY}
        + \frac{\sigma_X^2}{N} \widetilde{\K}_{ZZ}
        - \frac{1}{N} \C_{YX}
        \Big(
            \frac{1}{N} \C_{XX}
            + \frac{\sigma_Y^2}{N} \widetilde{\K}_{ZZ}
        \Big)^{-1}
        \frac{1}{N} \C_{XY} .
    \end{alignat*}
    The empirical second moments $\C_{YY} / N$, $\C_{XX} / N$, and $\C_{XY} / N$ converge almost surely, by the strong law of large numbers for independent transitions and by the ergodic theorem for stationary ergodic sequences, whereas the two regularization terms are fixed matrices divided by $N$ and vanish. Since the limit $\Phi_Z^\ast \mathcal{C}_{XX} \Phi_Z$ of $\C_{XX} / N$ is positive definite, $\bm{a}^\top \overrightarrow{\B}_\kappa \bm{a}$ tends to the residual of least-squares prediction from the dictionary, which the orthogonal decomposition of Proposition~\ref{prop: residual geometry}\ref{item: residual geometry ii}, with the predictions restricted to $\mathbb{H}_M$, splits into
    \begin{alignat*}{3}
        \bm{a}^\top \overrightarrow{\B}_\kappa \bm{a}
        &\rightarrow
        \frac{1}{\sigma_Y^2}
        \inf_{h \in \mathbb{H}_M}
        \mathbb{E}_{YX} \big[ \big( g(Y) - h(X) \big)^2 \big]
        \\
        &=
        \frac{1}{\sigma_Y^2}
        \Big(
            \mathbb{E}_X \big[ \operatorname{Var} \big( g(Y) \mid X \big) \big]
            +
            \inf_{h \in \mathbb{H}_M}
            \mathbb{E}_X \big[ \big( (\mathcal{U} g)(X) - h(X) \big)^2 \big]
        \Big) ,
    \end{alignat*}
    the point stated in the theorem.

    Second, the sparse posterior $\mathcal{IW}^M ( 2 + N , \, N \overrightarrow{\B}_\kappa )$ of Section~\ref{sec: Finite Output Factor} has the form of the posterior in Proposition~\ref{prop: posterior residual geometry}, whose one-dimensional law gives, for $N > 2$,
    \begin{alignat*}{3}
        \operatorname{Var} \big[ \bm{a}^\top \overrightarrow{\B} \bm{a} \mid \mathcal{D} \big]
        =
        \frac{ 2 \, ( \bm{a}^\top \overrightarrow{\B}_\kappa \bm{a} )^2 }{ N - 2 } .
    \end{alignat*}
    The numerator converges by the first step, and the variance therefore tends to zero almost surely.

    Third, fix $\tau > 0$. Chebyshev's inequality bounds the posterior probability of a deviation from the posterior mean,
    \begin{alignat*}{3}
        \mathbb{P} \Big[ \big| \bm{a}^\top \overrightarrow{\B} \bm{a} - \bm{a}^\top \overrightarrow{\B}_\kappa \bm{a} \big| \geq \frac{\tau}{2} \;\Big|\; \mathcal{D} \Big]
        \leq
        \frac{4}{\tau^2} \operatorname{Var} \big[ \bm{a}^\top \overrightarrow{\B} \bm{a} \mid \mathcal{D} \big] .
    \end{alignat*}
    Once the posterior mean lies within $\tau / 2$ of the limit point, every deviation from the limit point by at least $\tau$ requires a deviation from the posterior mean by at least $\tau / 2$. In other words, the posterior probability of the former therefore tends to zero for every $\tau > 0$, and the posterior law converges weakly to the point mass at the limit point.
\end{proof}

The closure error in the theorem depends on the choice of dictionary but not on the number of transitions $N$. Since $\mathbb{H}_M \subset \mathbb{H}$, the limit is bounded below by $\langle g, \overrightarrow{\mathcal{B}} \, g \rangle_{\mathbb{H}}$, with equality when the best prediction of $g(Y)$ from $\mathbb{H}$ is already attained in $\mathbb{H}_M$. For the full model of Section~\ref{sec: Operators as Random Variables}, the reverse bound $\limsup_{N \rightarrow \infty} \langle g, \overrightarrow{\mathcal{B}}_\kappa \, g \rangle_{\mathbb{H}} \leq \langle g, \overrightarrow{\mathcal{B}} \, g \rangle_{\mathbb{H}}$ holds almost surely for every $g \in \mathbb{H}$. Indeed, the right-hand side of~\eqref{eq: posterior conditional covariance} is at most its value at any fixed $h \in \mathbb{H}$, which converges almost surely to $\mathbb{E}_{YX} [ ( g(Y) - h(X) )^2 ]$, and the infimum of this expectation over a countable dense subset of $\mathbb{H}$ equals $\langle g, \mathcal{C}_{YY \mid X} \, g \rangle_{\mathbb{H}}$ for a bounded kernel. Whether $\overrightarrow{\mathcal{B}}_\kappa$ also attains $\overrightarrow{\mathcal{B}}$ in an infinite-dimensional $\mathbb{H}$ depends on how quickly the ridge regression with $\varepsilon = \sigma_Y^2 / N$ fits the residuals in-sample.

The point of concentration is the population residual geometry whenever the empirical second moments converge, whether or not the lifted residuals are Gaussian. The spread, by contrast, is that of the Gaussian working likelihood of Definition~\ref{def: lifted observation model} and is not a calibrated frequentist interval when the lifted residuals are not Gaussian. The split of the limit into averaged conditional variance and closure error parallels the variance and residual decomposition of \citet{colbrook2024beyond} and the risk decomposition of \citet{kostic2023sharp}.
\subsection{Posterior Variance of the Residuals} \label{subsec: Posterior Variance of the Residuals}

The residuals of Section~\ref{sec: Bayesian Consistency} evaluate the output covariance at its posterior mean $\overrightarrow{\B}_\kappa = ( N \sigma_Y^2 )^{-1} \, \widetilde{\C}_{YY \mid X}$. Evaluating~\eqref{eq: Bayesian ResDMD residual} at the output covariance $\overrightarrow{\B}$ instead of $\overrightarrow{\B}_\kappa$ defines the random variance residual
\[
    [ \operatorname{res}^{\textrm{var}}_{b}(\lambda_i, \phi_i) ]^2
    \coloneqq
    \frac{
        N \, \sigma_Y^2 \, \bm{\mathrm{w}}_i^* \, \overrightarrow{\B} \, \bm{\mathrm{w}}_i
    }{
        \bm{\mathrm{w}}_i^* \, \widetilde{\C}_{XX} \, \bm{\mathrm{w}}_i
    } .
\]
The residual evaluations of the lifted targets at the pseudo-inputs have the covariance $\sigma_Y^2 \overrightarrow{\B}$ (Section~\ref{sec: Operators as Random Variables}), and the numerator is therefore $N$ times the one-step residual variance of $\phi_i(Y)$ under the output covariance $\overrightarrow{\B}$. Since $\mathbb{E}[ \overrightarrow{\B} \mid \mathcal{D} ] = \overrightarrow{\B}_\kappa$, the posterior mean of $[ \operatorname{res}^{\textrm{var}}_{b}(\lambda_i, \phi_i) ]^2$ is $[ \operatorname{res}^{\textrm{var}}_{\textrm{pst}}(\lambda_i, \phi_i) ]^2$, and its posterior variance follows from the inverse-Wishart posterior of $\overrightarrow{\B}$.

\begin{proposition} \label{prop: residual variance}
    Let $N > 2$. For every eigenpair $(\lambda_i, \bm{\mathrm{w}}_i)$ of $\overrightarrow{\U}_\kappa$,
    \begin{alignat*}{3}
        \operatorname{Var}\big[ [ \operatorname{res}^{\textrm{var}}_{b}(\lambda_i, \phi_i) ]^2 \mid \mathcal{D} \big]
        =
        \frac{N + 2}{(N + 1)(N - 2)} \, [ \operatorname{res}^{\textrm{var}}_{\textrm{pst}}(\lambda_i, \phi_i) ]^4
        +
        \frac{N}{(N + 1)(N - 2)}
        \left|
            \frac{
                \bm{\mathrm{w}}_i^\top \, \widetilde{\C}_{YY \mid X} \, \bm{\mathrm{w}}_i
            }{
                \bm{\mathrm{w}}_i^* \, \widetilde{\C}_{XX} \, \bm{\mathrm{w}}_i
            }
        \right|^2 ,
    \end{alignat*}
    and
    \[
        \operatorname{Var}\big[ [ \operatorname{res}^{\textrm{var}}_{b}(\lambda_i, \phi_i) ]^2 \mid \mathcal{D} \big]
        \leq
        \frac{2}{N - 2} \, [ \operatorname{res}^{\textrm{var}}_{\textrm{pst}}(\lambda_i, \phi_i) ]^4 .
    \]
    Equality holds if and only if $\bm{\mathrm{w}}_i$ is a complex multiple of a real vector, which can always be arranged for a real eigenvalue $\lambda_i$. In this case,
    \[
        [ \operatorname{res}^{\textrm{var}}_{b}(\lambda_i, \phi_i) ]^2 \mid \mathcal{D}
        \overset{d}{=}
        \frac{
            N \, [ \operatorname{res}^{\textrm{var}}_{\textrm{pst}}(\lambda_i, \phi_i) ]^2
        }{
            \chi_{N+2}^{2}
        } .
    \]
\end{proposition}

\begin{proof}
    As for the reference posterior of Section~\ref{sec: Model Selection}, under the posterior of Section~\ref{sec: Finite Output Factor}, which is inverse-Wishart with index $2 + N$ and scale $N \overrightarrow{\B}_\kappa$, the precision $\overrightarrow{\B}^{-1}$ is Wishart distributed with $N + M + 1$ degrees of freedom and scale $(N \overrightarrow{\B}_\kappa)^{-1}$ \citep{dawid1981some}. The second moments of the inverse of a Wishart matrix are given by \citet{matsumoto2012general}, and for these parameters they read
    \begin{alignat*}{3}
        \mathbb{E}\big[ [\overrightarrow{\B}]_{jk} \, [\overrightarrow{\B}]_{lm} \mid \mathcal{D} \big]
        &=
        \frac{N}{(N + 1)(N - 2)}
        \Big(
            (N - 1) \, [\overrightarrow{\B}_\kappa]_{jk} \, [\overrightarrow{\B}_\kappa]_{lm}
            +
            [\overrightarrow{\B}_\kappa]_{jl} \, [\overrightarrow{\B}_\kappa]_{km}
            +
            [\overrightarrow{\B}_\kappa]_{jm} \, [\overrightarrow{\B}_\kappa]_{kl}
        \Big) .
    \end{alignat*}
    Contracting with $\bar{\mathrm{w}}_{ij} \, \mathrm{w}_{ik} \, \bar{\mathrm{w}}_{il} \, \mathrm{w}_{im}$, where $\mathrm{w}_{ij}$ denotes the $j$-th entry of $\bm{\mathrm{w}}_i$, turns the first and the third product into $( \bm{\mathrm{w}}_i^* \, \overrightarrow{\B}_\kappa \, \bm{\mathrm{w}}_i )^2$ and the second into
    \[
        ( \bar{\bm{\mathrm{w}}}_i^\top \, \overrightarrow{\B}_\kappa \, \bar{\bm{\mathrm{w}}}_i )
        ( \bm{\mathrm{w}}_i^\top \, \overrightarrow{\B}_\kappa \, \bm{\mathrm{w}}_i )
        =
        | \bm{\mathrm{w}}_i^\top \, \overrightarrow{\B}_\kappa \, \bm{\mathrm{w}}_i |^2 .
    \]
    Subtracting the squared posterior mean $( \bm{\mathrm{w}}_i^* \, \overrightarrow{\B}_\kappa \, \bm{\mathrm{w}}_i )^2$ and using $N^2 - (N + 1)(N - 2) = N + 2$ gives
    \[
        \operatorname{Var}\big[ \bm{\mathrm{w}}_i^* \, \overrightarrow{\B} \, \bm{\mathrm{w}}_i \mid \mathcal{D} \big]
        =
        \frac{
            (N + 2) \, ( \bm{\mathrm{w}}_i^* \, \overrightarrow{\B}_\kappa \, \bm{\mathrm{w}}_i )^2
            +
            N \, | \bm{\mathrm{w}}_i^\top \, \overrightarrow{\B}_\kappa \, \bm{\mathrm{w}}_i |^2
        }{
            (N + 1)(N - 2)
        } .
    \]
    Multiplication by $\big( N \sigma_Y^2 / ( \bm{\mathrm{w}}_i^* \, \widetilde{\C}_{XX} \, \bm{\mathrm{w}}_i ) \big)^2$ together with $N \sigma_Y^2 \, \overrightarrow{\B}_\kappa = \widetilde{\C}_{YY \mid X}$ yields the stated variance. For the bound, $\widetilde{\C}_{YY \mid X}$ is positive definite by~\eqref{eq: KYYX}, and the Cauchy--Schwarz inequality for the inner product $(\bm{u}, \bm{v}) \mapsto \bm{u}^* \, \widetilde{\C}_{YY \mid X} \, \bm{v}$, applied to $\bar{\bm{\mathrm{w}}}_i$ and $\bm{\mathrm{w}}_i$, gives $| \bm{\mathrm{w}}_i^\top \, \widetilde{\C}_{YY \mid X} \, \bm{\mathrm{w}}_i | \leq \bm{\mathrm{w}}_i^* \, \widetilde{\C}_{YY \mid X} \, \bm{\mathrm{w}}_i$. The second term of the variance is therefore at most $N / ((N + 1)(N - 2))$ times $[ \operatorname{res}^{\textrm{var}}_{\textrm{pst}}(\lambda_i, \phi_i) ]^4$, and $(N + 2) + N = 2 (N + 1)$ gives the bound. Equality holds if and only if $\bar{\bm{\mathrm{w}}}_i$ is a multiple of $\bm{\mathrm{w}}_i$, i.e., if and only if $\bm{\mathrm{w}}_i$ is a complex multiple of a real vector $\bm{a}$. In that case, $\bm{\mathrm{w}}_i^* \, \overrightarrow{\B} \, \bm{\mathrm{w}}_i$ is a positive multiple of $\bm{a}^\top \, \overrightarrow{\B} \, \bm{a}$, and the inverse chi-squared law follows from the affine scale property used in Proposition~\ref{prop: posterior residual geometry}.
\end{proof}

The bound holds for every $N > 2$ and every eigenpair, and it limits the coefficient of variation of the variance residual by a constant that depends on $N$ alone,
\[
    \frac{
        \sqrt{ \operatorname{Var}\big[ [ \operatorname{res}^{\textrm{var}}_{b}(\lambda_i, \phi_i) ]^2 \mid \mathcal{D} \big] }
    }{
        [ \operatorname{res}^{\textrm{var}}_{\textrm{pst}}(\lambda_i, \phi_i) ]^2
    }
    \leq
    \sqrt{ \frac{2}{N - 2} } ,
\]
with equality for real eigenpairs. The posterior uncertainty of the variance residual thus decays at the rate $N^{-1/2}$, which is the rate implicit in the proof of Theorem~\ref{thm: output covariance consistency} for a fixed real direction.

The same bound holds for the Bayesian-consistency ratio~\eqref{eq: directional consistency ratio}, which compares the observed and the predicted residual of every observable $g = \Phi_Z \bm{\mathrm{w}}$ in $\mathbb{H}_M$. Evaluating its numerator at $\overrightarrow{\B}$ gives the random ratio
\[
    \frac{
        N \sigma_Y^2 \, \bm{\mathrm{w}}^* \, \overrightarrow{\B} \, \bm{\mathrm{w}}
    }{
        \sum_{j=1}^{N} \mathbb{E}_{\bm{F}} \big[ | g(Y) - (\overrightarrow{\mathcal{U}}_\kappa g)(\bm{x}_j) |^2 \mid X = \bm{x}_j \big]
        +
        \gamma(\bm{\mathrm{w}})
    } ,
\]
whose denominator is fixed given the data. Since $\mathbb{E}[ \overrightarrow{\B} \mid \mathcal{D} ] = \overrightarrow{\B}_\kappa$ and $N \sigma_Y^2 \, \overrightarrow{\B}_\kappa = \widetilde{\C}_{YY \mid X}$, its posterior mean is $c(\bm{\mathrm{w}})$. Its posterior standard deviation is the same fixed multiple of the standard deviation of $\bm{\mathrm{w}}^* \, \overrightarrow{\B} \, \bm{\mathrm{w}}$, and the fixed multiple cancels from the coefficient of variation, which is therefore
\[
    \frac{
        \sqrt{ \operatorname{Var}\big[ \bm{\mathrm{w}}^* \, \overrightarrow{\B} \, \bm{\mathrm{w}} \mid \mathcal{D} \big] }
    }{
        \bm{\mathrm{w}}^* \, \overrightarrow{\B}_\kappa \, \bm{\mathrm{w}}
    } .
\]
The variance of $\bm{\mathrm{w}}^* \, \overrightarrow{\B} \, \bm{\mathrm{w}}$ in the proof of Proposition~\ref{prop: residual variance} and the Cauchy--Schwarz inequality used there, written for $\overrightarrow{\B}_\kappa$ as $| \bm{\mathrm{w}}^\top \, \overrightarrow{\B}_\kappa \, \bm{\mathrm{w}} | \leq \bm{\mathrm{w}}^* \, \overrightarrow{\B}_\kappa \, \bm{\mathrm{w}}$, hold for every nonzero $\bm{\mathrm{w}} \in \mathbb{C}^M$ and give
\begin{alignat*}{3}
    \operatorname{Var}\big[ \bm{\mathrm{w}}^* \, \overrightarrow{\B} \, \bm{\mathrm{w}} \mid \mathcal{D} \big]
    &=
    \frac{
        (N + 2) \, ( \bm{\mathrm{w}}^* \, \overrightarrow{\B}_\kappa \, \bm{\mathrm{w}} )^2
        +
        N \, | \bm{\mathrm{w}}^\top \, \overrightarrow{\B}_\kappa \, \bm{\mathrm{w}} |^2
    }{
        (N + 1)(N - 2)
    }
    \\
    &\leq
    \frac{
        (N + 2) + N
    }{
        (N + 1)(N - 2)
    } \,
    ( \bm{\mathrm{w}}^* \, \overrightarrow{\B}_\kappa \, \bm{\mathrm{w}} )^2
    =
    \frac{2}{N - 2} \,
    ( \bm{\mathrm{w}}^* \, \overrightarrow{\B}_\kappa \, \bm{\mathrm{w}} )^2 .
\end{alignat*}
Taking the square root and dividing by $\bm{\mathrm{w}}^* \, \overrightarrow{\B}_\kappa \, \bm{\mathrm{w}}$ bounds the coefficient of variation of the ratio by $\sqrt{2 / (N - 2)}$. Consequently, a deviation of $c(\bm{\mathrm{w}})$ from one by much more than $c(\bm{\mathrm{w}}) \sqrt{2 / (N - 2)}$ therefore cannot be attributed to the posterior uncertainty of the output covariance and points to a mismatch between the lifted model and the state-space model.

Two limitations apply. First, the eigenpair is held at the posterior-mean operator~\eqref{eq: CME matrix}, which does not depend on $\overrightarrow{\B}$, and the epistemic uncertainty of the operator, quantified by $\overrightarrow{\kappa}_\textrm{pst}$, is not propagated into the residuals. Second, as after Theorem~\ref{thm: output covariance consistency}, the spread is that of the Gaussian working likelihood and is not a calibrated frequentist interval when the lifted residuals are not Gaussian.

\subsection{Population VAMP-2 Geometry} \label{subsec: Population VAMP residual geometry}

For a fixed dictionary space $\mathbb{H}_M$, the population counterpart of the VAMP-$2$ trace in~\eqref{eq: kZ VAMP2 score} is obtained from the projected population second moments. Its residual interpretation follows directly from Proposition~\ref{prop: residual geometry}.

\begin{corollary} \label{cor: population VAMP residual geometry}
    Suppose that $\Phi_Z^\ast \mathcal{C}_{XX} \Phi_Z$ and $\Phi_Z^\ast \mathcal{C}_{YY} \Phi_Z$ are positive definite, and let $g_j = \Phi_Z \bm{a}_j$, $j=1,\ldots,M$, be orthonormal in $L^2(\mathbb{P}_Y)$. Then the population VAMP-$2$ trace
    \begin{alignat}{3}
        \mathcal{S}_\infty
        &\coloneqq
        \operatorname{Tr}\!\left[
            \big( \Phi_Z^\ast \mathcal{C}_{YY} \Phi_Z \big)^{-1}
            \Phi_Z^\ast \mathcal{C}_{YX} \Phi_Z
            \big( \Phi_Z^\ast \mathcal{C}_{XX} \Phi_Z \big)^{-1}
            \Phi_Z^\ast \mathcal{C}_{XY} \Phi_Z
        \right]
        \nonumber
        \\
        &=
        M
        -
        \sum_{j=1}^{M}
        \left(
            \mathbb{E}_X \big[ \operatorname{Var}\big( g_j(Y) \mid X \big) \big]
            +
            \inf_{h \in \mathbb{H}_M}
            \mathbb{E}_X \big[
                \big( (\mathcal{U} g_j)(X) - h(X) \big)^2
            \big]
        \right).
        \label{eq: population VAMP residual geometry}
    \end{alignat}
    For deterministic dynamics, $\mathcal{S}_\infty=M$ if and only if $\mathbb{H}_M$ is invariant under $\mathcal{U}$ in $L^2(\mathbb{P}_X)$.
\end{corollary}

\begin{proof}
    Orthonormality gives
    \[
        \bm{a}_j^\top
        \Phi_Z^\ast \mathcal{C}_{YY} \Phi_Z
        \bm{a}_k
        =
        \mathbb{E}_Y[g_j(Y)g_k(Y)]
        =
        \delta_{jk}.
    \]
    Hence
    \[
        \big( \Phi_Z^\ast \mathcal{C}_{YY} \Phi_Z \big)^{-1}
        =
        \sum_{j=1}^{M}
        \bm{a}_j \bm{a}_j^\top .
    \]
    Therefore,
    \[
        M-\mathcal{S}_\infty
        =
        \sum_{j=1}^{M}
        \bm{a}_j^\top
        \left(
            \Phi_Z^\ast \mathcal{C}_{YY} \Phi_Z
            -
            \Phi_Z^\ast \mathcal{C}_{YX} \Phi_Z
            \big( \Phi_Z^\ast \mathcal{C}_{XX} \Phi_Z \big)^{-1}
            \Phi_Z^\ast \mathcal{C}_{XY} \Phi_Z
        \right)
        \bm{a}_j .
    \]
    By Proposition~\ref{prop: residual geometry}\ref{item: residual geometry iii} with the predictions restricted to $\mathbb{H}_M$, each summand equals
    \[
        \inf_{h \in \mathbb{H}_M}
        \mathbb{E}_{YX}\big[
            \big( g_j(Y)-h(X) \big)^2
        \big].
    \]
    Proposition~\ref{prop: residual geometry}\ref{item: residual geometry ii} gives the stated decomposition into averaged conditional variance and dictionary closure error. For deterministic dynamics the conditional-variance term vanishes. Since $\{g_j\}_{j=1}^M$ spans $\mathbb{H}_M$, the trace equals $M$ exactly when $\mathcal{U}g_j$ lies in $\mathbb{H}_M$ in $L^2(\mathbb{P}_X)$ for every $j$, which is equivalent to invariance of $\mathbb{H}_M$ under $\mathcal{U}$.
\end{proof}

Thus the population VAMP-$2$ score rewards a dictionary with small aggregate one-step residuals. For stochastic dynamics these residuals combine conditional variability with dictionary closure error, whereas for deterministic dynamics only the closure error remains. This parallels the residual--variance distinction for stochastic Koopman approximations discussed by \citet{colbrook2024beyond}.

\section{Gaussian Process Time-lagged Canonical Correlation Analysis} \label{sec: GP-TCCA}

Here we detail the time-lagged canonical correlation analysis (TCCA) algorithm used to obtain a GP-DMD model. It follows the feature TCCA algorithm of \citet{wu2020variational}, with the endpoint regularization of regularized kernel CCA \citep{bach2002kernel, klus2019kernel}.

TCCA is a multivariate statistical method applicable to two sets of `featurized' variables that are related by the flow map of a dynamical process. The algorithm is especially attractive, because it performs a decomposition that is similar to DMD, but with a valid interpretation for both reversible and \emph{irreversible} processes, and for stationary and \emph{non-stationary} processes \citep{wu2020variational}. It constructs source- and target-side whitening transformations and applies an SVD to the resulting normalized time-lagged second-moment matrix, from which the finite-dimensional Koopman matrix is reconstructed. As in Section~\ref{sec: Model Selection}, the second moments below are uncentered, so the singular values $\rho_i$ collected in $\mathbf{P}$ are those of the normalized TCCA matrix rather than centered canonical correlations.

\subsection{Estimation Algorithm} \label{sec: Estimation Algorithm} 
\begin{multicols}{2}
\begin{enumerate}
    \item Using the transformation $\bm{\Psi}_Z(\cdot) \coloneqq \mathbf{L}_{ZZ}^{-1} \, \mathbf{k}_Z(\cdot)$ start by preconditioning the cross-covariance matrices:
    \begin{alignat*}{3}
        \mathbf{\Psi}_{ZX} &\coloneqq \mathbf{L}^{-1}_{ZZ} \K_{ZX}, \\
        \mathbf{\Psi}_{ZY} &\coloneqq \mathbf{L}^{-1}_{ZZ} \K_{ZY}.
    \end{alignat*}
    The lower triangular matrix $\mathbf{L}_{ZZ}$ is obtained from the \emph{Cholesky decomposition} $\widetilde{\K}_{ZZ} = \mathbf{L}_{ZZ} \mathbf{L}_{ZZ}^\top$ of the regularized pseudo-covariance matrix~\eqref{eq: regularized pseudo cov}. This step simplifies upcoming computations, because the regularized Gramian matrix factorizes according to
    \begin{alignat*}{3}
        \widetilde{\C}_{XX} 
        &= \K_{ZX} \K_{ZX}^\top + \sigma_{Y}^2 \widetilde{\K}_{ZZ}\\
        &= \mathbf{L}_{ZZ} \, \left(\mathbf{\Psi}_{ZX} \mathbf{\Psi}_{ZX}^\top + \sigma_{Y}^2\I \right) \, \mathbf{L}_{ZZ}^\top.
    \end{alignat*}
  
  \item In this basis construct the $M \times M$ Gramian and stiffness matrices:
  \begin{alignat*}{3}
        \mathbf{G}_{XX} &\coloneqq \mathbf{\Psi}_{ZX} \, \mathbf{\Psi}_{ZX}^\top,\\
        \mathbf{G}_{YY} &\coloneqq \mathbf{\Psi}_{ZY} \, \mathbf{\Psi}_{ZY}^\top,\\
        \mathbf{G}_{XY} &\coloneqq \mathbf{\Psi}_{ZX} \, \mathbf{\Psi}_{ZY}^\top.
    \end{alignat*}

    \item Proceed by performing a thin singular value decomposition (SVD) on the preconditioned covariance matrices:
    \begin{alignat*}{3}
        \mathbf{\Psi}_{ZX} &= \mathbf{M}_X \mathbf{\Sigma}_X \mathbf{N}_{X}^\top,\\
        \mathbf{\Psi}_{ZY} &= \mathbf{M}_Y \mathbf{\Sigma}_Y \mathbf{N}_{Y}^\top.
    \end{alignat*}

    \item We regularize both source- and target-side covariance matrices with ridges that correspond to the observation-noise scales:
    \begin{alignat*}{3}
        \mathbf{\tilde{\Sigma}}_{X,i} &= \sqrt{\mathbf{\Sigma}_{X,i}^2 + \sigma_{Y}^2}, \; \mathbf{\tilde{\Sigma}}_{Y,i} = \sqrt{\mathbf{\Sigma}_{Y,i}^2 + \sigma_X^2}.
    \end{alignat*}
    This implies that the inverse and inverse square root of the (noisy) Gramian matrices can be expressed as
    \begin{alignat*}{3}
        \tilde{\mathbf{G}}_{XX}^{-1} &= \mathbf{M}_X \mathbf{\tilde{\Sigma}}_{X}^{-2} \mathbf{M}_X^\top, \\
        \tilde{{\mathbf{G}}}_{YY}^{-1/2} &= \mathbf{M}_Y \tilde{\mathbf{{\Sigma}}}_{Y}^{-1} \mathbf{M}_Y^\top.
    \end{alignat*}

    \item After normalizing the preconditioned cross-covariance matrices with $\tilde{\mathbf{G}}_{XX}^{-1/2}$ and $\tilde{\mathbf{G}}_{YY}^{-1/2}$, the whitened Koopman matrix can be computed with
    \begin{alignat*}{3}
        \U' &=  \tilde{\mathbf{G}}_{XX}^{-1/2} \, \mathbf{G}_{XY} \, \tilde{\mathbf{G}}_{YY}^{-1/2},
    \end{alignat*}
    which has the SVD:
    \begin{alignat*}{3}
        \U' &= \mathbf{W}_X' \,\mathbf{P} \, \mathbf{W}_Y'^\top.
    \end{alignat*}
    
    \item Compute the transformation matrices:
    \begin{alignat*}{3}
        \mathbf{W}_X = {\tilde{\mathbf{G}}}_{XX}^{-1/2} \mathbf{W}_X', \: \mathbf{W}_Y = \tilde{\mathbf{G}}_{YY}^{-1/2} \mathbf{W}_Y'.
    \end{alignat*}

    \item In the finite-dimensional TCCA model, the mean dynamics are represented by
    \begin{alignat*}{3}
        \mathbb{E} \left[ \mathbf{W}_Y^\top \bm{\Psi}_Z(Y) \right] 
        &\approx \mathbf{P} \, \mathbb{E} \left[ \mathbf{W}_X^\top \bm{\Psi}_Z(X) \right], \\
        \mathbb{E} \left[ \bm{\Psi}_Z(Y) \right] 
        &\approx \left( \mathbf{W}_X \mathbf{P} \mathbf{W}_Y^{\dagger} \right)^\top \mathbb{E} \left[ \bm{\Psi}_Z(X) \right] \\ 
        &= \overrightarrow{\U}_\kappa^\top \mathbb{E} \left[ \bm{\Psi}_Z(X) \right],
    \end{alignat*}
    the final Koopman matrix in the preconditioned basis is
    \begin{alignat*}{3}
        \overrightarrow{\U}_\kappa &= \mathbf{W}_X \,\mathbf{P} \, \mathbf{W}_Y^\top \, \tilde{\mathbf{G}}_{YY}.
    \end{alignat*}

    \item \label{item: GP-TCCA eigendecomposition} The eigenvalue decomposition of $\overrightarrow{\U}_\kappa$ returns ${\V}_{\kappa}$. For the retained nonzero eigenvalues, the corresponding modes with respect to the full-state observable are
    \begin{alignat*}{3}
        {\V}_{\bm{F}}  &= \mathbf{\Lambda}^{-1} \, {\V}_{\kappa}  \, \tilde{\mathbf{G}}_{XX}^{-1} \, \mathbf{\Psi}_{ZX} \Y^\top.
    \end{alignat*}  
\end{enumerate}
\end{multicols}

\newpage

Besides for the Koopman matrix, the algorithm provides the singular structure of the normalized time-lagged second-moment matrix. The source- and target-side singular functions of Appendix~\ref{subsec: Forward-backward Operators} are approximated by $\mathbf{W}_X^\top \bm{\Psi}_Z(\cdot)$ and $\mathbf{W}_Y^\top \bm{\Psi}_Z(\cdot)$, respectively, and the diagonal matrix $\mathbf{P}$ contains the associated singular values $\rho_i$ \citep{wu2020variational}. The normalizations $\mathbf{W}_X^\top \tilde{\mathbf{G}}_{XX} \mathbf{W}_X = \I$ and $\mathbf{W}_Y^\top \tilde{\mathbf{G}}_{YY} \mathbf{W}_Y = \I$ are the finite, regularized, and uncentered analogues of the constraints in~\eqref{eq: population CCA}. These singular functions can serve as relaxed coordinates from which \emph{coherent sets} are extracted by thresholding or clustering \citep{FROYLAND20131, klus2019kernel}.

\subsection{Forward--Backward Symmetry} \label{subsec: GP-TCCA Forward-Backward Symmetry}

The algorithm in Appendix~\ref{sec: Estimation Algorithm} whitens the two endpoints separately, which extends to distinct endpoint dictionaries $\Z_X$ and $\Z_Y$, whereas the eigendecomposition in step~\ref{item: GP-TCCA eigendecomposition} requires the shared dictionary $\Z$. The ridges in $\widetilde{\C}_{XX} = \C_{XX} + \sigma_Y^2 \, \widetilde{\K}_{ZZ}$ and $\widetilde{\C}_{YY} = \C_{YY} + \sigma_X^2 \, \widetilde{\K}_{ZZ}$ stem from the paired likelihood of Section~\ref{sec: Finite Output Factor} and from the inverse-Wishart prior~\eqref{eq: output covariance prior}, respectively. The prior leaves $\sigma_X$ free, and we calibrate it by symmetry. Just as $\sigma_Y^2$ is the residual variance of the forward full-state GP regression $X \mapsto Y$, we set $\sigma_X^2$ to the residual variance of the backward regression $Y \mapsto X$,
\[
    X_i
    =
    \overleftarrow{\bm{F}}(Y_i)
    +
    \overleftarrow{\bm{\epsilon}}_i,
    \qquad
    \overleftarrow{\bm{\epsilon}}_i
    \sim
    \mathcal{N}^D\!\left( \bm{0}, \, \sigma_X^2 \I_D \right).
\]
The predecessors are then regularized by $\sigma_Y^2 \widetilde{\K}_{ZZ}$ and the successors by $\sigma_X^2 \widetilde{\K}_{ZZ}$, both measured relative to $\sigma_Y^2$ through the normalization in Definition~\ref{def: lifted observation model}. Section~\ref{sec: Bayesian Forward-Backward Operators} lifts the backward regression to a backward lifted transfer operator, whose likelihood regularizes the successors with this same ridge. Neither model assumes reversibility, and the backward regression also supplies the backward model-complexity and sparsity terms of~\eqref{eq: fb lifted objective} below. The VAMP-$2$ contribution~\eqref{eq: kZ VAMP2 score} used for the lifted representation reads
\begin{alignat}{3} \label{eq: fb VAMP2 score}
    \mathcal{S}
    &\coloneqq
    \operatorname{Tr}\!\left(
        \widetilde{\C}_{YY}^{-1}
        \, \C_{YX}
        \, \widetilde{\C}_{XX}^{-1}
        \, \C_{XY}
    \right)
    =
    \left\|
        \widetilde{\C}_{XX}^{-1/2}
        \, \C_{XY}
        \, \widetilde{\C}_{YY}^{-1/2}
    \right\|_F^2 .
\end{alignat}

By cyclic invariance of the trace, reversing the direction gives the same scalar score, and the data-mismatch contribution is $- ( N + M + 1 ) \, \mathcal{S}$ in either direction, because the backward reference posterior has the same index $2 + N$ and dimension $M$. The forward--backward structure thus arises from the contrast of the paired and the reference VFE through the two endpoint covariances. The shared kernel and dictionary provide a common lifted space for the left and right singular structures, which need not coincide for nonreversible dynamics but align under the appropriate equilibrium-weighted inner product in the reversible case, where the singular value decomposition of the Koopman operator reduces to its eigendecomposition \citep{wu2020variational}. In the terminology of Appendix~\ref{subsec: Forward-backward Operators}, the product inside the trace in \eqref{eq: fb VAMP2 score} represents the backward--forward operator $\overleftarrow{\mathcal T}$, whose nonzero spectrum coincides with that of $\overrightarrow{\mathcal{T}}$, which is the operator-theoretic reason for the orientation independence of the score.

For the backward contribution to the transfer operator objective, define the Nystr\"om approximation
\begin{alignat}{3}
    \mathbf{R}_{YY}
    &\coloneqq
    \K_{ZY}^{\top}
    \, \widetilde{\K}_{ZZ}^{-1}
    \, \K_{ZY}.
\end{alignat}
Then the forward and backward model-complexity terms are
\begin{alignat}{3}
    \overrightarrow{\mathcal{L}}_{\mathrm{c}}
    &\coloneqq
    M \operatorname{ln}\!\left|
        \I_N
        +
        \sigma_Y^{-2}
        \, \mathbf{R}_{XX}
    \right|,
    \qquad
    \overleftarrow{\mathcal{L}}_{\mathrm{c}}
    \coloneqq
    M \operatorname{ln}\!\left|
        \I_N
        +
        \sigma_X^{-2}
        \, \mathbf{R}_{YY}
    \right|,
\end{alignat}
while the corresponding Nystr\"om sparsity terms are
\begin{alignat}{3}
    \overrightarrow{\mathcal{L}}_{\mathrm{s}}
    &\coloneqq
    \frac{M}{\sigma_Y^2}
    \operatorname{Tr}\!\left(
        \K_{XX}
        -
        \mathbf{R}_{XX}
    \right),
    \qquad
    \overleftarrow{\mathcal{L}}_{\mathrm{s}}
    \coloneqq
    \frac{M}{\sigma_X^2}
    \operatorname{Tr}\!\left(
        \K_{YY}
        -
        \mathbf{R}_{YY}
    \right).
\end{alignat}

To preserve the relative balance between the data-mismatch, model-complexity, and sparsity contributions of the directional objective derived in Section~\ref{sec: Model Selection}, we use the average of the forward and backward directional contrasts:
\begin{alignat}{3} \label{eq: fb lifted objective}
    \mathcal{L}_{\mathbf{k}_Z}^{\textrm{fb}}
    &\coloneqq
    - ( N + M + 1 ) \, \mathcal{S}
    +
    \frac{1}{2}
    \left(
        \overrightarrow{\mathcal{L}}_{\mathrm{c}}
        +
        \overleftarrow{\mathcal{L}}_{\mathrm{c}}
    \right)
    +
    \frac{1}{2}
    \left(
        \overrightarrow{\mathcal{L}}_{\mathrm{s}}
        +
        \overleftarrow{\mathcal{L}}_{\mathrm{s}}
    \right).
\end{alignat}
The factors of $1 / 2$ arise because $\mathcal{S}$ is already common to the two directional contrasts. Equivalently, Equation~\eqref{eq: fb lifted objective} is one half of their sum and therefore preserves the relative weighting of the three contributions in the directional construction. The kernel lengthscales and the variational parameters are then optimized with respect to $\mathcal{L}_{\mathbf{k}_Z}^{\textrm{fb}}$. The signal variance $\sigma_f^2$ and the endpoint ridges $\sigma_Y^2$ and $\sigma_X^2$ remain fixed at their calibrated values. Releasing that calibration would make the two ridges free parameters of the lifted model. The score alone is monotone in the inverse regularization, and the ridges would then be determined only by the countervailing complexity and sparsity terms of~\eqref{eq: fb lifted objective}.

\subsection{Moments of the Random Forward--Backward Operator} \label{subsec: Forward-Backward Moments}

Following the projected Koopman representation~\eqref{eq: projected Koopman representation}, we apply the same construction to the random directional Koopman matrices, allowing distinct input and output dictionaries. For the augmented kernels $\widetilde{\kappa}_{\textrm{pr}} = \kappa_{\textrm{pr}} + \sigma_Z^2 \delta$, the corresponding feature operators satisfy $\widetilde{\Phi}_Z^{\ast} \widetilde{\Phi}_Z = \widetilde{\K}_{ZZ}$, and the coefficient map is $\widetilde{\Gamma}_Z = \widetilde{\K}_{ZZ}^{-1} \widetilde{\Phi}_Z^{\ast}$. Given that these pseudo-covariance matrices are invertible, the finite-rank representations of the kernel Koopman operators are
\begin{alignat}{3} \label{eq: lifted Bayesian directional operators}
    \overrightarrow{\mathcal U}
    &\coloneqq \widetilde{\Phi}_{Z_X} \overrightarrow{\U} \widetilde{\K}_{ZZ}^{-1} \widetilde{\Phi}_{Z_Y}^{\ast},
    &\qquad
    \overleftarrow{\mathcal U}
    &\coloneqq \widetilde{\Phi}_{Z_Y} \overleftarrow{\U} \widetilde{\K}_{ZZ}^{-1} \widetilde{\Phi}_{Z_X}^{\ast}.
\end{alignat}
In each expression, $\widetilde{\K}_{ZZ}$ is formed from the dictionary of the domain of the operator, i.e., from $\Z_Y$ for $\overrightarrow{\mathcal U}$ and from $\Z_X$ for $\overleftarrow{\mathcal U}$. This construction realizes the random coefficient matrices as operators on the augmented RKHSs without changing their finite-dimensional posterior laws.

By construction, $\overrightarrow{\mathcal U} \widetilde{\Phi}_{Z_Y} = \widetilde{\Phi}_{Z_X} \overrightarrow{\U}$ and $\overleftarrow{\mathcal U} \widetilde{\Phi}_{Z_X} = \widetilde{\Phi}_{Z_Y} \overleftarrow{\U}$. Both operators vanish on the orthogonal complements of their domain dictionary spans. Their composition is therefore
\begin{alignat*}{3}
    \overrightarrow{\mathcal T}
    &= \widetilde{\Phi}_{Z_X} \overrightarrow{\U} \overleftarrow{\U}
       \widetilde{\K}_{ZZ}^{-1} \widetilde{\Phi}_{Z_X}^{\ast} = \widetilde{\Phi}_{Z_X} \overrightarrow{\U} \overleftarrow{\U} \, \widetilde{\Gamma}_{Z_X}.
\end{alignat*}
Cyclicity of the finite-rank trace yields
\begin{alignat}{3} \label{eq: Bayesian forward-backward trace identity}
    \operatorname{Tr}_{\widetilde{\mathbb H}_X}(\overrightarrow{\mathcal T})
    &= \operatorname{Tr}(\overrightarrow{\U} \overleftarrow{\U}).
\end{alignat}
The target-side identity follows by symmetry. Section~\ref{sec: Bayesian Forward-Backward Operators} writes this trace as $\operatorname{Tr}_{\mathbb H_X}$, which is exact for $\sigma_Z = 0$, where the augmented spaces reduce to $\mathbb H_X$ and $\mathbb H_Y$. Thus the trace of the forward--backward operator agrees with the trace of the coefficient-matrix product for every realization, allowing its posterior moments to be calculated entirely through finite-dimensional matrices.

It follows that the variance of $\operatorname{Tr}_{{\mathbb H}_X}(\overrightarrow{\mathcal T})$ requires the posterior second moments of the directional Koopman matrices. We begin with the forward matrix-valued random variable $\overrightarrow{\U} \in \mathbb R^{M_X \times M_Y}$. Expressing the noisy inducing values in coefficient coordinates gives the matrix-normal prior, and the lifted observations provide the Gaussian likelihood
\begin{alignat*}{3}
    \operatorname{vec} \overrightarrow{\U}^{\top} \mid \overrightarrow{\B}
    &\sim \mathcal{N}^{M_X M_Y}\!\left(
        \bm{0}, \,
        \widetilde{\K}_{ZZ}^{-1} \otimes \overrightarrow{\B}
    \right),\\
    \operatorname{vec} \K_{ZY} \mid \overrightarrow{\U}, \overrightarrow{\B}
    &\sim \mathcal{N}^{M_Y N}\!\left(
        \operatorname{vec} ( \overrightarrow{\U}^{\top} \K_{ZX} ), \,
        \sigma_Y^2 \I_N \otimes \overrightarrow{\B}
    \right).
\end{alignat*}
In both Kronecker products, the first factor correlates the source coefficients or the transitions, and the coregionalization matrix $\overrightarrow{\B}$ correlates the target channels. Here the prior source covariance $\widetilde{\K}_{ZZ}^{-1}$ follows from applying the dictionary coefficient map $\widetilde{\K}_{ZZ}^{-1}$ to the noisy inducing values with covariance $\widetilde{\K}_{ZZ}$. Conditioning on $\K_{ZY}$ then gives the inducing-mediated posterior and its entrywise covariance
\begin{alignat*}{3}
    \operatorname{vec} \overrightarrow{\U}^{\top} \mid \mathcal D, \overrightarrow{\B}
    &\sim \mathcal{N}^{M_X M_Y}\!\left(
        \operatorname{vec} \overrightarrow{\U}_\kappa^{\top}, \,
        \sigma_Y^2 \widetilde{\C}_{XX}^{-1} \otimes \overrightarrow{\B}
    \right),\\
    \operatorname{Cov}\!\left(
        [\overrightarrow{\U}]_{ik}, [\overrightarrow{\U}]_{jl}
        \mid \mathcal D, \overrightarrow{\B}
    \right)
    &= \sigma_Y^2 [\widetilde{\C}_{XX}^{-1}]_{ij} [\overrightarrow{\B}]_{kl}.
\end{alignat*}
The backward posterior follows by exchanging the endpoints and residual scales.

We now calculate the posterior trace variance directly from the coefficient matrices using~\eqref{eq: Bayesian forward-backward trace identity}. Define $\Delta_{Y \mid X} \coloneqq \overrightarrow{\U} - \overrightarrow{\U}_\kappa$ and $\Delta_{X \mid Y} \coloneqq \overleftarrow{\U} - \overleftarrow{\U}_\kappa$. Independence and centering give
\begin{alignat*}{3}
    \operatorname{Tr}(\overrightarrow{\U} \overleftarrow{\U}) - \mathcal S
    ={}& \operatorname{Tr}(\Delta_{Y \mid X} \overleftarrow{\U}_\kappa)
       + \operatorname{Tr}(\overrightarrow{\U}_\kappa \Delta_{X \mid Y})
       + \operatorname{Tr}(\Delta_{Y \mid X} \Delta_{X \mid Y}).
\end{alignat*}
The three centered trace terms have zero posterior means and are pairwise uncorrelated, since $\Delta_{Y \mid X}$ and $\Delta_{X \mid Y}$ are independent and centered. Consequently, the variance of their sum equals the sum of the individual variances. To calculate these variances, we contract the entrywise directional covariances. For the forward contribution,
\begin{alignat*}{3}
    \overrightarrow{\sigma}^{\,2} \mid \overrightarrow{\B} &\coloneqq \operatorname{Var}\!\left[
        \operatorname{Tr}(\Delta_{Y \mid X} \overleftarrow{\U}_\kappa)
        \mid \mathcal D, \overrightarrow{\B}
    \right]\\
    & = \sum_{i,j=1}^{M_X} \sum_{k,l=1}^{M_Y}
        [\overleftarrow{\U}_\kappa]_{ki}
        [\overleftarrow{\U}_\kappa]_{lj}
        \operatorname{Cov}\!\left(
            [\Delta_{Y \mid X}]_{ik},
            [\Delta_{Y \mid X}]_{jl}
            \mid \mathcal D, \overrightarrow{\B}
        \right)\\
    & = \sigma_Y^2 \operatorname{Tr}\!\left(
        \overleftarrow{\U}_\kappa^{\top}
        \overrightarrow{\B}
        \overleftarrow{\U}_\kappa
        \widetilde{\C}_{XX}^{-1}
    \right).
\end{alignat*}
The backward contribution $\overleftarrow{\sigma}^{\,2} \mid \overleftarrow{\B}$ follows by exchanging the endpoints. For the interaction term, independence of the directional coefficient matrices yields
\begin{alignat*}{3}
    \overleftrightarrow{\sigma}^{\,2} \mid \overrightarrow{\B}, \, \overleftarrow{\B}
    &\coloneqq \operatorname{Var}\!\left[
        \operatorname{Tr}(\Delta_{Y \mid X} \Delta_{X \mid Y})
        \mid \mathcal D, \overrightarrow{\B}, \overleftarrow{\B}
    \right]\\
    &\quad = \sigma_Y^2 \sigma_X^2
        \operatorname{Tr}\!\left(
            \overrightarrow{\B} \widetilde{\C}_{YY}^{-1}
        \right)
        \operatorname{Tr}\!\left(
            \overleftarrow{\B} \widetilde{\C}_{XX}^{-1}
        \right).
\end{alignat*}
Finally, since the conditional means vanish, averaging over the inverse-Wishart output-covariance posteriors replaces these covariances by their posterior means $(\overrightarrow{\B}, \, \overleftarrow{\B}) \rightarrow (\overrightarrow{\B}_\kappa, \, \overleftarrow{\B}_\kappa)$.

The posterior output-covariance means satisfy $\overrightarrow{\B}_\kappa = (N \sigma_Y^2)^{-1} \widetilde{\C}_{YY \mid X}$ and $\overleftarrow{\B}_\kappa = (N \sigma_X^2)^{-1} \widetilde{\C}_{XX \mid Y}$. Using these Schur complements and $\overrightarrow{\U}_\kappa = \widetilde{\C}_{XX}^{-1} \C_{XY}$, $\overleftarrow{\U}_\kappa = \widetilde{\C}_{YY}^{-1} \C_{YX}$, the forward term reduces to
\begin{alignat*}{3}
    \overrightarrow{\sigma}^{\,2}
    &= \frac{1}{N} \operatorname{Tr}\!\left(
        \overleftarrow{\U}_\kappa^{\top}
        \widetilde{\C}_{YY \mid X}
        \overleftarrow{\U}_\kappa
        \widetilde{\C}_{XX}^{-1}
    \right)\\
    &= \frac{1}{N} \operatorname{Tr}\!\left(
        \overleftarrow{\U}_\kappa^{\top}
        \big( \widetilde{\C}_{YY} - \C_{YX} \, \widetilde{\C}_{XX}^{-1} \, \C_{XY} \big)
        \overleftarrow{\U}_\kappa
        \widetilde{\C}_{XX}^{-1}
    \right)\\
    &= \frac{1}{N} \left[
        \operatorname{Tr}(\overrightarrow{\U}_\kappa \overleftarrow{\U}_\kappa)
        - \operatorname{Tr}\!\left(
            (\overrightarrow{\U}_\kappa \overleftarrow{\U}_\kappa)^2
        \right)
    \right]\\
    &= \frac{1}{N} \sum_i \rho_i^2 (1 - \rho_i^2).
\end{alignat*}
The backward contribution is identical because the products $\overrightarrow{\U}_\kappa \overleftarrow{\U}_\kappa$ and $\overleftarrow{\U}_\kappa \overrightarrow{\U}_\kappa$ have the same nonzero eigenvalues. For the interaction term, the Schur complements give
\begin{alignat*}{3}
    \sigma_Y^2 \operatorname{Tr}\!\left(
        \overrightarrow{\B}_\kappa \widetilde{\C}_{YY}^{-1}
    \right)
    = \frac{M_Y - \mathcal S}{N},\qquad
    \sigma_X^2 \operatorname{Tr}\!\left(
        \overleftarrow{\B}_\kappa \widetilde{\C}_{XX}^{-1}
    \right)
    = \frac{M_X - \mathcal S}{N}.
\end{alignat*}
Hence $\overleftrightarrow{\sigma}^{\,2} = (M_X - \mathcal S)(M_Y - \mathcal S) / N^2$. Combining the three uncorrelated contributions gives the posterior trace variance and its modal decomposition:
\begin{alignat}{3}
    \operatorname{Var}\!\left[
        \operatorname{Tr}(\overrightarrow{\U} \overleftarrow{\U})
        \mid \mathcal D
    \right]
    &= \overrightarrow{\sigma}^{\,2}
       + \overleftarrow{\sigma}^{\,2}
       + \overleftrightarrow{\sigma}^{\,2} \notag\\
    &= \frac{2}{N} \sum_i \rho_i^2 (1 - \rho_i^2)
       + \frac{(M_X - \mathcal S)(M_Y - \mathcal S)}{N^2} \notag\\
    &= \sum_{i=1}^{M_X} (1 - \rho_i^2)
       \left[
           \frac{2 \rho_i^2}{N}
           + \frac{M_Y - \mathcal S}{N^2}
       \right]. \label{eq: Bayesian forward-backward trace variance derivation}
\end{alignat}
The final equality follows from $\mathcal S = \sum_{i=1}^{M_X} \rho_i^2$, with the singular values extended by zeros when necessary. Moreover, each summand has a direct interpretation. For a source eigenvector $\bm{\mathrm{w}}_i$ of $\overrightarrow{\U}_\kappa \overleftarrow{\U}_\kappa$, define the random round trip evaluated along this fixed direction,
\begin{alignat*}{3}
    \varrho_i
    &\coloneqq
    \frac{
        \bm{\mathrm{w}}_i^\top \, \widetilde{\C}_{XX} \, \overrightarrow{\U} \overleftarrow{\U} \, \bm{\mathrm{w}}_i
    }{
        \bm{\mathrm{w}}_i^\top \, \widetilde{\C}_{XX} \, \bm{\mathrm{w}}_i
    } .
\end{alignat*}
Since the random trace is the sum of these pairwise uncorrelated evaluations, $\operatorname{Tr}(\overrightarrow{\U} \overleftarrow{\U}) = \sum_{i=1}^{M_X} \varrho_i$, its posterior mean and variance decompose into the contributions
\begin{alignat*}{3}
    \mathbb{E} [ \varrho_i \mid \mathcal{D} ]
    &=
    \rho_i^2,
    &\qquad
    \operatorname{Var} [ \varrho_i \mid \mathcal{D} ]
    &=
    ( 1 - \rho_i^2 )
    \left[
        \frac{2 \rho_i^2}{N} + \frac{M_Y - \mathcal{S}}{N^2}
    \right].
\end{alignat*}

\section{Dictionary Learning} \label{sec: Dictionary Learning}

Treating the pseudo-inputs $\Z$ as free variational parameters and repeatedly optimizing the model's hyperparameter objective can be quite computationally demanding. The variational scheme of \citet{titsias2009variational} also admits a greedy discrete alternative in which candidates are selected from a working set and ranked by the increase they produce in the variational lower bound. Titsias shows that adding an inducing point cannot decrease the bound and that the VFE can be evaluated in $\mathcal{O}(NM)$ time for each candidate in the working set. In Appendix~\ref{subsec: VFE criterion} we derive the corresponding rank-one candidate score for the regularized inducing-variable construction used here, and Appendix~\ref{subsec: transfer dictionary criterion} carries it over to the lifted targets of the transfer operator objective~\eqref{eq: kZ VAMP2 objective}. A scalarization of the two scores forms the final stage of the multi-tiered dictionary-selection cascade.

We further propose to construct the dictionary with a three-stage selection cascade with increasing computational complexity. The first two stages are inexpensive active-learning criteria that require only rank-one linear algebra and are used to whittle down the candidate set in the forward-selection spirit of \citet[Alg.~8.1]{williams2006gaussian}. The first criterion removes inputs that are already well explained by the current dictionary, and the second criterion ranks the remaining inputs by the uncertainty reduction they induce over the batch. Only a retained subset is then passed to the third criterion, which evaluates the exact rank-one increase of a scalarization of the flow-map VFE and the paired lifted VFE of Section~\ref{sec: Model Selection}. Hence, rather than initializing $\Z$ at random, the resulting dictionary can provide an informed warm start for the comparatively expensive gradient-based refinement of the variational hyperparameters.

Throughout, we work with the whitened feature vector $\bm{\Psi}_Z(\bm{x}) \coloneqq \Lb_{ZZ}^{-1}\, \mathbf{k}_Z(\bm{x})$ of Appendix~\ref{sec: Estimation Algorithm}, where $\Lb_{ZZ}$ is the lower-triangular Cholesky factor of the regularized pseudo-covariance matrix~\eqref{eq: regularized pseudo cov}, $\widetilde{\K}_{ZZ} = \Lb_{ZZ} \Lb_{ZZ}^\top$. The inducing-noise variance $\sigma_Z^2$ of Section~\ref{sec: Sparse Transfer Operators} is adapted alongside $\Z$ as the model is optimized. Conditioning the prior on the noisy inducing variables associated with the pseudo-inputs $\Z$ yields the latent conditional variance and covariance
\begin{alignat}{3}
    \sigma_M^2(\bm{x}) &= \kappa_\textrm{pr}(\bm{x}, \bm{x}) - \mathbf{k}_Z^\top(\bm{x})\, \widetilde{\K}_{ZZ}^{-1}\, \mathbf{k}_Z(\bm{x}) = \kappa_\textrm{pr}(\bm{x}, \bm{x}) - \lVert \bm{\Psi}_Z(\bm{x}) \rVert^2, \label{eq: post var M} \\
    \sigma_M(\bm{x}, \bm{x}') &= \kappa_\textrm{pr}(\bm{x}, \bm{x}') - \langle \bm{\Psi}_Z(\bm{x}), \, \bm{\Psi}_Z(\bm{x}') \rangle. \label{eq: post cov M}
\end{alignat}

\subsection{Fast GP updates}

The dictionary criteria below repeatedly test the effect of appending a single candidate pseudo-input. It is therefore useful to express the size-$(M+1)$ posterior covariance in terms of quantities already available at size $M$. Let $\Z_+ \coloneqq \Z \cup \{\bm{z}\}$ denote the dictionary obtained after admitting a candidate $\bm{z}$. The enlarged regularized pseudo-covariance matrix and feature vector are
\begin{alignat*}{3}
    \widetilde{\K}_{Z_+Z_+}
    =
    \begin{bmatrix}
        \widetilde{\K}_{ZZ} & \mathbf{k}_Z(\bm{z})\\
        \mathbf{k}_Z^\top(\bm{z}) & \kappa_\textrm{pr}(\bm{z},\bm{z})+\sigma_Z^2
    \end{bmatrix},
    \qquad
    \mathbf{k}_{Z_+}(\bm{x})
    =
    \begin{bmatrix}
        \mathbf{k}_Z(\bm{x})\\
        \kappa_\textrm{pr}(\bm{z},\bm{x})
    \end{bmatrix}.
\end{alignat*}
Thus admitting $\bm{z}$ increases the pseudo-covariance matrix from size $M$ to size $M+1$, and the updated covariance is
\begin{alignat*}{3}
    \sigma_{M+1}(\bm{x}, \, \bm{x}' \mid \bm{z})
    =
    \kappa_\textrm{pr}(\bm{x},\bm{x}')
    -
    \mathbf{k}_{Z_+}^\top(\bm{x})\,\widetilde{\K}_{Z_+Z_+}^{-1} \, \mathbf{k}_{Z_+}(\bm{x}').
\end{alignat*}
Applying the block-inverse formula to $\widetilde{\K}_{Z_+Z_+}$ gives
\begin{alignat}{3} \label{eq: block inverse}
    &\mathbf{k}_{Z_+}^\top(\bm{x})\,\widetilde{\K}_{Z_+Z_+}^{-1}\,\mathbf{k}_{Z_+}(\bm{x}') \nonumber\\
    &\quad =
    \mathbf{k}_Z^\top(\bm{x})\,\widetilde{\K}_{ZZ}^{-1}\,\mathbf{k}_Z(\bm{x}')
    +
    \frac{
    \left[
        \kappa_\textrm{pr}(\bm{x},\bm{z})
        -
        \mathbf{k}_Z^\top(\bm{x})\,\widetilde{\K}_{ZZ}^{-1}\,\mathbf{k}_Z(\bm{z})
    \right]
    \left[
        \kappa_\textrm{pr}(\bm{z},\bm{x}')
        -
        \mathbf{k}_Z^\top(\bm{z})\,\widetilde{\K}_{ZZ}^{-1}\,\mathbf{k}_Z(\bm{x}')
    \right]
    }{
        \widetilde{\sigma}^2_M({\bm{z}})
    },
\end{alignat}
where the Schur complement associated with the newly added coordinate is
\begin{alignat*}{3}
    \widetilde{\sigma}^2_M({\bm{z}})
    &\coloneqq
    \kappa_\textrm{pr}(\bm{z},\bm{z})
    -
    \mathbf{k}_Z^\top(\bm{z})\,\widetilde{\K}_{ZZ}^{-1}\,\mathbf{k}_Z(\bm{z})
    + 
    \sigma_Z^2
    =
    \sigma_M^2(\bm{z})+\sigma_Z^2.
\end{alignat*}
The bracketed terms in~\eqref{eq: block inverse} are the size-$M$ posterior covariances $\sigma_M(\bm{x},\bm{z})$ and $\sigma_M(\bm{z},\bm{x}')$ from~\eqref{eq: post cov M}. Substitution into the definition of $\sigma_{M+1}$ therefore gives the rank-one variance and covariance updates
\begin{alignat}{3}
    \sigma_{M+1}^2(\bm{x} \mid \bm{z})
    &=
    \sigma_M^2(\bm{x})
    -
    \frac{
        \sigma_M(\bm{x},\bm{z})^2
    }{
        \widetilde{\sigma}^2_M({\bm{z}})
    },
    \label{eq: post var M+1}\\
    \sigma_{M+1}(\bm{x},\bm{x}' \mid \bm{z})
    &=
    \sigma_M(\bm{x},\bm{x}')
    -
    \frac{
        \sigma_M(\bm{x},\bm{z})\,\sigma_M(\bm{z},\bm{x}')
    }{
        \widetilde{\sigma}^2_M({\bm{z}})
    }.
    \label{eq: post cov M+1}
\end{alignat}
Scoring a candidate therefore does not require an inverse or Cholesky factor of the enlarged matrix. The additional pseudo-input enters only through the Schur complement $\widetilde{\sigma}^2_M({\bm{z}})$ in the denominator and a rank-one correction.

\subsection{Approximate Linear Dependence}
The first criterion \citep{engel2004kernel} measures how well a candidate input $\bm{x}$ is already represented by the current dictionary $\Phi_Z = \{\varphi(\bm{z}_i)\}_{i=1}^M$. We use the corresponding regularized ALD criterion, given by the latent conditional variance~\eqref{eq: post var M} under the noisy inducing-variable model,
\begin{alignat}{3} \label{eq: ALD residual}
    \sigma_M^2(\bm{x}) \;=\; \kappa_\textrm{pr}(\bm{x}, \bm{x}) - \lVert \bm{\Psi}_Z(\bm{x}) \rVert^2.
\end{alignat}

Let $\mathcal{J}$ denote the current batch of distinct candidate inputs. Then a candidate $\bm{x} \in \mathcal{J}$ whose residual falls below a fixed tolerance, $\sigma_M^2(\bm{x}) < \sigma_\textrm{tol}^2$, is well approximated by the existing pseudo-inputs and is removed from $\mathcal{J}$. The survivors --- those carrying appreciable residual, capped at the $b$ highest --- form the reduced set $\mathcal{J}_b \subseteq \mathcal{J}$ passed to the second criterion. The tolerance governs the sparsity of the model: a smaller $\sigma_\textrm{tol}^2$ retains more candidates and yields a larger, more expressive dictionary.

The approximate-linear-dependence (ALD) residual~\eqref{eq: ALD residual} scores each candidate against the current dictionary alone, and so cannot detect redundancy \emph{among} the candidates themselves: two nearby inputs may both clear the tolerance, yet once either is admitted the other becomes superfluous. Admitting candidates one at a time resolves this without any explicit pairwise comparison. After a single pseudo-input is appended to $\Z$, the residuals of the remaining candidates are recomputed. A near-duplicate of the newly admitted pseudo-input then has its residual fall below $\sigma_\textrm{tol}^2$ and is removed from $\mathcal{J}_b$ on the next pass. The ALD residual therefore acts as a cheap pre-filter: it ranks candidates by their own residual variance, but it does not determine which admissible candidate is most valuable relative to the remaining candidates. That comparison is the role of the second criterion.

\subsection{Active Learning Cohn}
Since the ALD approach is a local `whack-a-mole' active learning scheme, we may consider refining the subset by implementing \emph{active learning Cohn} \citep{861310, cohn1996neural}, which is a globally oriented methodology. The second criterion selects, from $\mathcal{J}_b$, the pseudo-input whose inclusion most reduces the predictive variance \emph{across a reference set} --- the active-data-selection principle of \citet{861310}, and an instance of \emph{active learning Cohn} (ALC) \citep{gramacy2020surrogates}. Unlike a pure maximum-variance rule, which returns the single most uncertain candidate and tends to favor isolated outliers, ALC rewards a candidate that lowers the uncertainty at \emph{all} reference sites simultaneously. In its original formulation the reference set is a space-filling sample over $\mathbb{X}$ \citep{861310, gramacy2020surrogates}. We instead reuse the ALD-reduced set $\mathcal{J}_b$, whose members already mark the locations of highest residual variance. When scoring a candidate $\bm{z} \in \mathcal{J}_b$, we take the reference set to be the remaining survivors $\mathcal{J}_b^{-\bm{z}} \coloneqq \mathcal{J}_b\setminus\{\bm{z}\}$, so that the score measures the uncertainty reduction the candidate induces at the \emph{remaining} sites.

After admitting a candidate $\bm{z}$, the reduction in variance at a reference site $\bm{x} \in \mathcal{J}_b^{-\bm{z}}$ becomes
\begin{alignat*}{3}
    \Delta \sigma^2(\bm{x} \mid \bm{z})
    \;\coloneqq\;
    \sigma_M^2(\bm{x}) - \sigma_{M+1}^2(\bm{x} \mid \bm{z})
    \;=\;
    \frac{\sigma_M(\bm{x}, \bm{z})^2}{\widetilde{\sigma}^2_M({\bm{z}})},
\end{alignat*}
where $\sigma_{M+1}^2(\bm{x} \mid \bm{z})$ is the rank-one update in~\eqref{eq: post var M+1}. This expression only requires the cached size-$M$ posterior covariance $\sigma_M(\bm{x},\bm{z})$ and the scalar $\sigma_M^2(\bm{z})+\sigma_Z^2$, both obtained from~\eqref{eq: post var M} and~\eqref{eq: post cov M}. Hence the reduction can be evaluated in $\mathcal{O}(M)$ per reference site without re-forming the enlarged factor.

Consequently, the ALC criterion is the aggregate reduction over the reference set,
\begin{alignat}{3} \label{eq: ALC}
    \textrm{ALC}(\bm{z}) \;=\; \sum_{\bm{x}_j \in \mathcal{J}_b^{-\bm{z}}} \Delta \sigma^2(\bm{x}_j \mid \bm{z}),
\end{alignat}
which is the variance-reduction form of ALC. It approximates, by a sum over the finite reference set, the integrated variance reduction
\[
    \int_{\mathbb{X}} \big[ \sigma_M^2(\bm{x}) - \sigma_{M+1}^2(\bm{x} \mid \bm{z}) \big] \, \mathrm{d}\bm{x} ,
\]
which an integrated-mean-squared-error ($I$-optimal) design maximizes and which is intractable in general \citep{861310, gramacy2020surrogates}. Over the reference set itself, \eqref{eq: ALC} is the reduction in the trace of the posterior covariance and is therefore the $A$-optimal counterpart of the log-determinant criterion derived next. Our implementation ranks candidates by this closely related information-gain criterion rather than by the ALC score.

\subsection{The Information Gain Criterion}

The ALC score aggregates marginal variance reductions over the reference set, and therefore only uses the diagonal entries of the updated posterior covariance. To account for the off-diagonal correlations between reference sites, we instead measure the reduction in the differential entropy of the joint Gaussian posterior. Fix a candidate $\bm{z}\in\mathcal{J}_b$ and enumerate the remaining survivors $\mathcal{J}_b^{-\bm{z}}=\{\bm{x}_1,\ldots,\bm{x}_{b-1}\}$. Define the size-$M$ posterior covariance matrix over this reference set by
\begin{alignat*}{3}
    [\K_M]_{ij}
    \; \coloneqq \;
    \sigma_M(\bm{x}_i,\bm{x}_j),
    \qquad
    i,j = 1,\ldots,b-1.
\end{alignat*}
Using the rank-one covariance update in~\eqref{eq: post cov M+1}, the posterior covariance over $\mathcal{J}_b^{-\bm{z}}$ after including $\bm{z}$ is
\begin{alignat*}{3}
    \K_{M+1}
    =
    \K_M
    -
    \frac{
        \mathbf{k}_{\textrm{pst}}(\bm{z}) \, \mathbf{k}_{\textrm{pst}}^\top(\bm{z})
    }{
        \widetilde{\sigma}^2_M({\bm{z}})
    } \in \mathbb{R}^{(b-1) \times (b-1)}, \;\; \text{where} \;\; \mathbf{k}_{\textrm{pst}}(\bm{z})
    \;\coloneqq\;
    \begin{bmatrix}
        \sigma_M(\bm{x}_1,\bm{z}) \, 
        \ldots \,
        \sigma_M(\bm{x}_{b-1},\bm{z})
    \end{bmatrix}^\top.
\end{alignat*}

Assuming the posterior covariance matrices are positive definite, the entropy of the Gaussian posterior before and after the inclusion of $\bm{z}$ over $\mathcal{J}_b^{-\bm{z}}$ is
\begin{alignat*}{3}
    H_M(\mathcal{J}_b^{-\bm{z}})
    =
    \frac{1}{2}
    \ln\!\left((2\pi e)^{b-1} \left| \K_M \right| \right), \qquad
    H_{M+1}(\mathcal{J}_b^{-\bm{z}} \mid \bm{z})
    =
    \frac{1}{2}
    \ln\!\left((2\pi e)^{b-1} \left|\K_{M+1} \right| \right).
\end{alignat*}
We can therefore rank candidates by the joint information gain
\begin{equation} \label{eq: joint information gain}
\begin{aligned}
    \operatorname{IG}(\bm{z})
    &\coloneqq
    H_M(\mathcal{J}_b^{-\bm{z}})
    -
    H_{M+1}(\mathcal{J}_b^{-\bm{z}} \mid \bm{z}) \\
    &=
    \frac{1}{2}
    \ln
    \frac{
        \left\lvert \K_M \right\rvert
    }{
        \left\lvert \K_{M+1} \right\rvert
    } \\
    &=
    \frac{1}{2}
    \ln \left\lvert \K_M \right\rvert
    -
    \frac{1}{2}
    \ln
    \left\lvert
        \K_M
        -
        \frac{
            \mathbf{k}_{\textrm{pst}}(\bm{z}) \, \mathbf{k}_{\textrm{pst}}^\top(\bm{z})
        }{
            \widetilde{\sigma}^2_M(\bm{z})
        }
    \right\rvert.
\end{aligned}
\end{equation}

By the matrix determinant lemma, \eqref{eq: joint information gain} can be written as
\begin{alignat*}{3}
    \operatorname{IG}(\bm{z})
    =
    -\frac{1}{2}
    \ln
    \left(
        1
        -
        \frac{
            \mathbf{k}_{\textrm{pst}}^\top(\bm{z}) \, \K_M^{-1} \, \mathbf{k}_{\textrm{pst}}(\bm{z})
        }{
            \widetilde{\sigma}^2_M({\bm{z}})
        }
    \right)
    =
    \frac{1}{2}
    \ln
    \left(
        \frac{
            \widetilde{\sigma}^2_M({\bm{z}})
        }{
            \widetilde{\sigma}^2_M({\bm{z}})
            -
            \mathbf{k}_{\textrm{pst}}^\top(\bm{z}) \, \K_M^{-1} \, \mathbf{k}_{\textrm{pst}}(\bm{z})
        }
    \right).
\end{alignat*}
In practice, we add a small numerical jitter to $\K_M$. The ranking of $\mathcal{J}_b$ by $\operatorname{IG}(\bm{z})$ can be used in two ways. The single top-ranked candidate gives the greedy choice
\begin{alignat*}{3}
    \bm{z}^\star
    =
    \arg\max_{\bm{z}\in\mathcal{J}_b}
    \operatorname{IG}(\bm{z}),
\end{alignat*}
whereas the top-ranked subset forms the set $\mathcal{J}_{\mathrm{vfe}}$ that the third criterion of Appendix~\ref{subsec: VFE criterion} scores.

However, evaluating $\operatorname{IG}(\bm{z})$ directly would require, for each of the $b$ candidates, assembling and factorizing the $(b-1)\times(b-1)$ reference covariance $\K_M$ of its own reduced set $\mathcal{J}_b^{-\bm{z}}$, at a cost of $\mathcal{O}(b^4)$. All $b$ scores can instead be obtained from a single factorization. Collect the latent posterior covariance over the entire survivor set,
\begin{alignat*}{3}
    \big[\K_b\big]_{ij}
    \;\coloneqq\;
    \sigma_M(\bm{x}_i,\bm{x}_j),
    \qquad
    i,j = 1,\ldots,b.
\end{alignat*}
and take the candidate $\bm{z}=\bm{x}_i$. The matrix $\K_b$ splits into three blocks. Removing its $i$-th row and column leaves the latent reduced-set covariance $\K_M$, the off-diagonal entries of its $i$-th column form $\mathbf{k}_{\textrm{pst}}(\bm{x}_i)$, and its $i$-th diagonal entry is
\[
    [\K_b]_{ii}
    =
    \sigma_M^2(\bm{x}_i).
\]
The block-inverse (Schur-complement) identity therefore gives
\begin{alignat*}{3}
    \big[\K_b^{-1}\big]_{ii}
    \;=\;
    \frac{1}{
        \sigma_M^2(\bm{x}_i)
        -
        \mathbf{k}_{\textrm{pst}}^\top(\bm{x}_i)
        \, \K_M^{-1}
        \, \mathbf{k}_{\textrm{pst}}(\bm{x}_i)
    }.
\end{alignat*}
Since
\[
    \widetilde{\sigma}_M^2(\bm{x}_i)
    =
    \sigma_M^2(\bm{x}_i)
    +
    \sigma_Z^2,
\]
substitution into~\eqref{eq: joint information gain} yields
\begin{alignat}{3} \label{eq: IG fast}
    \operatorname{IG}(\bm{x}_i)
    \;=\;
    \frac{1}{2}
    \ln\!\left(
        \frac{
            \big(
                [\K_b]_{ii}
                +
                \sigma_Z^2
            \big)
            [\K_b^{-1}]_{ii}
        }{
            1
            +
            \sigma_Z^2
            [\K_b^{-1}]_{ii}
        }
    \right).
\end{alignat}
A single Cholesky factorization
\[
    \K_b
    =
    {\Lb} \, {\Lb}^\top
\]
therefore yields every candidate's score: the diagonal $[\K_b]_{ii}$ is read directly, while
\[
    [\K_b^{-1}]_{ii}
    =
    \lVert {\Lb}^{-1} \bm{e}_i \rVert^2
\]
is the squared norm of the $i$-th column of ${\Lb}^{-1}$.

Note that \eqref{eq: joint information gain} defines a greedy leave-one-out D-optimal-style criterion. Bayesian D-optimality minimizes the determinant of the posterior covariance matrix, which geometrically minimizes the volume of the confidence ellipsoid \citep{chaloner1995bayesian}. Because it is the reduction in the log-determinant rather than the trace, $\operatorname{IG}(\bm{z})$ is sensitive to the full reference-set covariance, and includes the off-diagonal posterior correlations that $\operatorname{ALC}(\bm{z})$ in~\eqref{eq: ALC} discards.

The single-site version of this differential-entropy score --- ranking each candidate by its own marginal entropy reduction --- is the greedy selection rule of the informative vector machine \citep{lawrence2002fast, lawrence2004learning}. Our criterion~\eqref{eq: joint information gain} instead ranks a candidate by the reduction in the joint entropy of the \emph{remaining} reference sites $\mathcal{J}_b^{-\bm{z}}$ that its inclusion induces --- scoring the candidate by its effect on the rest of the set rather than on itself. An isolated candidate --- of high residual variance but weakly correlated with the other survivors --- is accordingly demoted rather than preferred, so the rule resists the tendency of a pure maximum-variance score to place inducing points at isolated outliers.

\subsection{The Variational Free Energy Criterion} \label{subsec: VFE criterion}

The information-gain criterion is deliberately target agnostic: it ranks candidates by how much including them reduces posterior uncertainty over the surviving inputs, but it does not use the evolved targets $\Y$. The VFE criterion adds this target dependence. To keep the computational cost low, the dictionary update is implemented as a cascade of three criteria. The ALD criterion first removes the candidates that the current dictionary already explains. The information-gain criterion then ranks the remaining candidates, and its top-ranked subset $\mathcal{J}_{\mathrm{vfe}} \subseteq \mathcal{J}_b$ is passed to the final, target-aware VFE criterion.

Following Appendix~\ref{sec: Gradient Derivations}, we write $\theta$ for the generative parameters and $\vartheta_M$ for the variational parameter collection with dictionary size $M$. Thus $\vartheta_M$ contains the current pseudo-inputs $\Z$ and the inducing-noise parameter $\sigma_Z$. For a candidate $\bm{z}\in\mathcal{J}_{\mathrm{vfe}}$, let $\vartheta_{M+1}(\bm{z})$ denote the variational parameter collection obtained by appending $\bm{z}$ to the current dictionary while keeping $\theta$ and $\sigma_Z^2$ fixed.

Let $\Y_d \in \mathbb{R}^{1\times N}$ denote the $d$-th row of the one-step target matrix $\Y$. The flow-map free energy is the sum of the independent scalar VFEs of the components of $\overrightarrow{\bm{F}}$,
\begin{alignat}{3} \label{eq: dict-vfe-flow-objective}
    \mathcal{F}_{\bm{F}}(\theta,\vartheta_M)
    &\coloneqq
    \sum_{d=1}^{D}
    \left[
        \operatorname{ln}\left(\mathcal{N}^{N}(\Y_d;\bm{0},\widetilde{\mathbf{R}}_{XX})\right)
        -
        \frac{1}{2\sigma_Y^2}\operatorname{Tr}\left(\K_{XX}-\mathbf{R}_{XX}\right)
    \right].
\end{alignat}
One may equally work with the reciprocal-dimension scaling $\tfrac{1}{D}\mathcal{F}_{\bm{F}}(\theta,\vartheta_M)$, which places the flow-map objective on a per-dimension footing. This scaling is constant over candidates and therefore does not affect the greedy maximizer, so the derivation below works directly with $\mathcal{F}_{\bm{F}}$.

With the inducing-noise variance $\sigma_Z^2$ contained in $\vartheta_M$, we recall the quantities of~\eqref{eq: regularized pseudo cov} and Appendix~\ref{sec: Estimation Algorithm},
\begin{alignat*}{3}
    \widetilde{\K}_{ZZ}
    &\coloneqq
    \K_{ZZ}+\sigma_Z^2\I_M
    =
    \Lb_{ZZ}\Lb_{ZZ}^{\top},
    \qquad
    \bm{\Psi}_Z(\bm{x})
    \coloneqq
    \Lb_{ZZ}^{-1}\mathbf{k}_Z(\bm{x}),
    \qquad
    \mathbf{\Psi}_{ZX}
    \coloneqq
    \Lb_{ZZ}^{-1}\K_{ZX}.
\end{alignat*}
With this notational convention,
\begin{alignat*}{3}
    \mathbf{R}_{XX}
    &=
    \K_{ZX}^{\top}\widetilde{\K}_{ZZ}^{-1}\K_{ZX}
    =
    \mathbf{\Psi}_{ZX}^{\top}\mathbf{\Psi}_{ZX},
    \qquad
    \widetilde{\mathbf{R}}_{XX}
    =
    \mathbf{R}_{XX} + \sigma_Y^2\I_N .
\end{alignat*}
Because $\mathbf{R}_{XX}$ here carries the inducing-noise variance $\sigma_Z^2$, the flow-map objective~\eqref{eq: dict-vfe-flow-objective} is the collapsed variational bound evaluated with this regularized Nystr\"om approximation.

For any candidate $\bm{z}\in\mathcal{J}_{\mathrm{vfe}}$, recall that
\begin{alignat*}{3}
    \sigma_M(\bm{x},\bm{z})
    &\coloneqq
    \kappa_{\textrm{pr}}(\bm{x},\bm{z})
    -
    \langle \bm{\Psi}_Z(\bm{x}),\bm{\Psi}_Z(\bm{z})\rangle,
    \\
    \sigma_M^2(\bm{z})
    &\coloneqq
    \kappa_{\textrm{pr}}(\bm{z},\bm{z})
    -
    \lVert\bm{\Psi}_Z(\bm{z})\rVert^2,
    \qquad
    \widetilde{\sigma}_M^2(\bm{z})
    \coloneqq
    \sigma_M^2(\bm{z})+\sigma_Z^2 .
\end{alignat*}
Equivalently, collecting the posterior covariances between all training inputs and the candidate gives
\begin{alignat*}{3}
    \bm{\sigma}_M(\X,\bm{z})
    &\coloneqq
    \begin{bmatrix}
        \sigma_M(\bm{x}_1,\bm{z}) & \ldots & \sigma_M(\bm{x}_N,\bm{z})
    \end{bmatrix}^{\top}
    =
    \mathbf{k}_X(\bm{z})
    -
    \mathbf{\Psi}_{ZX}^{\top}\bm{\Psi}_Z(\bm{z}),
    \\
    \mathbf{k}_X(\bm{z})
    &\coloneqq
    \begin{bmatrix}
        \kappa_{\textrm{pr}}(\bm{x}_1,\bm{z}) & \ldots & \kappa_{\textrm{pr}}(\bm{x}_N,\bm{z})
    \end{bmatrix}^{\top}.
\end{alignat*}

For fixed $\bm{z}$, let $\Z_+\coloneqq \Z\cup\{\bm{z}\}$, then the enlarged pseudo-covariance and its Cholesky factor are
\begin{alignat*}{3}
    \widetilde{\K}_{Z_+Z_+}
    &\coloneqq
    \begin{bmatrix}
        \widetilde{\K}_{ZZ} & \mathbf{k}_Z(\bm{z}) \\
        \mathbf{k}_Z(\bm{z})^{\top} & \kappa_{\textrm{pr}}(\bm{z},\bm{z})+\sigma_Z^2
    \end{bmatrix}
    =
    \Lb_{Z_+Z_+}\Lb_{Z_+Z_+}^{\top},
    \qquad
    \Lb_{Z_+Z_+}
    &=
    \begin{bmatrix}
        \Lb_{ZZ} & \bm{0} \\
        \bm{\Psi}_Z(\bm{z})^{\top} & \widetilde{\sigma}_M(\bm{z})
    \end{bmatrix}.
\end{alignat*}

The enlarged cross-covariance and whitened feature matrices are
\begin{alignat*}{3}
    \K_{Z_+X}
    &\coloneqq
    \begin{bmatrix}
        \K_{ZX} \\
        \mathbf{k}_X(\bm{z})^{\top}
    \end{bmatrix},
    \qquad
    \mathbf{\Psi}_{Z_+X}
    \coloneqq
    \Lb_{Z_+Z_+}^{-1}\K_{Z_+X}.
\end{alignat*}
Solving the enlarged triangular system gives
\begin{alignat*}{3}
    \mathbf{\Psi}_{Z_+X}
    &=
    \begin{bmatrix}
        \Lb_{ZZ}^{-1}\K_{ZX}
        \\
        \dfrac{
            \mathbf{k}_X(\bm{z})^{\top}
            -
            \bm{\Psi}_Z(\bm{z})^{\top}\mathbf{\Psi}_{ZX}
        }{
            \widetilde{\sigma}_M(\bm{z})
        }
    \end{bmatrix}
    =
    \begin{bmatrix}
        \mathbf{\Psi}_{ZX}
        \\
        \bm{u}_{\bm{z}}^{\top}
    \end{bmatrix},
    \qquad
    \bm{u}_{\bm{z}}
    \coloneqq
    \frac{\bm{\sigma}_M(\X,\bm{z})}{\widetilde{\sigma}_M(\bm{z})}.
\end{alignat*}
Consequently, the Nystr\"om covariance induced by the enlarged dictionary satisfies
\begin{alignat}{3} \label{eq: dict-vfe-rank-one-update}
    \mathbf{R}_{{+}XX}
    &\coloneqq
    \K_{Z_+X}^{\top}
    \widetilde{\K}_{Z_+Z_+}^{-1}
    \K_{Z_+X}
    =
    \mathbf{\Psi}_{Z_+X}^{\top}\mathbf{\Psi}_{Z_+X}
    \\
    &=
    \mathbf{\Psi}_{ZX}^{\top}\mathbf{\Psi}_{ZX}
    +
    \bm{u}_{\bm{z}}\bm{u}_{\bm{z}}^{\top}
    =
    \mathbf{R}_{XX}
    +
    \bm{u}_{\bm{z}}\bm{u}_{\bm{z}}^{\top}.
\end{alignat}
This is the central identity: the enlarged dictionary changes the $N\times N$ Nystr\"om covariance only through one outer product. Hence
\begin{alignat*}{3}
    \widetilde{\mathbf{R}}_{{+}XX}
    &\coloneqq
    \mathbf{R}_{{+}XX}+\sigma_Y^2\I_N
    =
    \widetilde{\mathbf{R}}_{XX}
    +
    \bm{u}_{\bm{z}}\bm{u}_{\bm{z}}^{\top}.
\end{alignat*}

Define the scalar contractions with respect to the current dictionary,
\begin{alignat*}{3}
    s_{\bm{z}}
    &\coloneqq
    \bm{u}_{\bm{z}}^{\top} \, \widetilde{\mathbf{R}}_{XX}^{-1} \, \bm{u}_{\bm{z}},
    \qquad
    g_{d,\bm{z}}
    \coloneqq
    \Y_d \, \widetilde{\mathbf{R}}_{XX}^{-1} \, \bm{u}_{\bm{z}} .
\end{alignat*}
The matrix determinant lemma and the Sherman--Morrison identity give
\begin{alignat*}{3}
    \operatorname{ln}\left|\widetilde{\mathbf{R}}_{{+}XX}\right|
    &=
    \operatorname{ln}\left|\widetilde{\mathbf{R}}_{XX}\right|
    +
    \operatorname{ln}(1+s_{\bm{z}}),
    \\
    \widetilde{\mathbf{R}}_{{+}XX}^{-1}
    &=
    \widetilde{\mathbf{R}}_{XX}^{-1}
    -
    \frac{
        \widetilde{\mathbf{R}}_{XX}^{-1} \,
        \bm{u}_{\bm{z}} \,
        \bm{u}_{\bm{z}}^{\top} \,
        \widetilde{\mathbf{R}}_{XX}^{-1}
    }{
        1+s_{\bm{z}}
    },
    \\
    \Y_d \, \widetilde{\mathbf{R}}_{{+}XX}^{-1} \, \Y_d^{\top}
    &=
    \Y_d \, \widetilde{\mathbf{R}}_{XX}^{-1} \, \Y_d^{\top}
    -
    \frac{g_{d,\bm{z}}^2}{1+s_{\bm{z}}},
    \\
    \operatorname{Tr}\left(\K_{XX}-\mathbf{R}_{{+}XX}\right)
    &=
    \operatorname{Tr}\left(\K_{XX}-\mathbf{R}_{XX}\right)
    -
    \lVert\bm{u}_{\bm{z}}\rVert^2 .
\end{alignat*}
Substituting these identities into~\eqref{eq: dict-vfe-flow-objective} cancels all terms that do not depend on the candidate. The VFE gain from appending $\bm{z}$ is therefore
\begin{alignat}{3} \label{eq: dict-vfe-fast-score}
    \Delta\mathcal{F}_{\bm{F}}(\bm{z})
    &\coloneqq
    \mathcal{F}_{\bm{F}}(\theta,\vartheta_{M+1}(\bm{z}))
    -
    \mathcal{F}_{\bm{F}}(\theta,\vartheta_M) \nonumber
    \\
    &=
    \sum_{d=1}^{D}
    \left[
        -\frac{1}{2}\operatorname{ln}(1+s_{\bm{z}})
        +
        \frac{1}{2} \left(\frac{g_{d,\bm{z}}^2}{1+s_{\bm{z}}} \right)
        +
        \frac{1}{2\sigma_Y^2}\lVert\bm{u}_{\bm{z}}\rVert^2
    \right].
\end{alignat}
Thus, with the flow-map targets alone, the third dictionary criterion is
\begin{alignat*}{3}
    \bm{z}^{\star}
    &=
    \arg\max_{\bm{z}\in\mathcal{J}_{\mathrm{vfe}}}
    \Delta\mathcal{F}_{\bm{F}}(\bm{z}).
\end{alignat*}
The three terms in~\eqref{eq: dict-vfe-fast-score} have the same interpretation as in the ordinary VFE: the first penalizes the increase in model complexity, the second rewards improvement in the target fit, and the third rewards the reduction of the trace error. Since $\Delta\mathcal{F}_{\bm{F}}(\bm{z})$ is the exact change in the flow-map objective~\eqref{eq: dict-vfe-flow-objective} for fixed hyperparameters $\theta$, it recovers the greedy model-selection criterion \citep{titsias2009variational, titsias2008modelselection}.

The score can be evaluated without forming or factorizing the enlarged covariance. Using the singular value decomposition of $\mathbf{\Psi}_{ZX}$ from Appendix~\ref{sec: GP-TCCA},
\begin{alignat*}{3}
    \mathbf{\Psi}_{ZX}
    &=
    \mathbf{M}_X \, \mathbf{\Sigma}_X \, \mathbf{N}_{X}^{\top},
\end{alignat*}
the inverse action needed above is
\begin{alignat*}{3}
    \widetilde{\mathbf{R}}_{XX}^{-1}\bm{v}
    &=
    \sigma_Y^{-2}
    \left[
        \bm{v}
        -
        \mathbf{N}_{X}
        \operatorname{diag}\left(
            \frac{\mathbf{\Sigma}_{X,j}^2}{\mathbf{\Sigma}_{X,j}^2+\sigma_Y^2}
        \right)_{j=1}^{M}
        \mathbf{N}_{X}^{\top}\bm{v}
    \right].
\end{alignat*}
For each candidate in $\mathcal{J}_{\mathrm{vfe}}$, computing $\bm{\sigma}_M(\X,\bm{z})$ costs $\mathcal{O}(NM)$ using the cached whitened matrix $\mathbf{\Psi}_{ZX}$, and the inverse action also costs $\mathcal{O}(NM)$. The target contractions $g_{d,\bm{z}}$ can be obtained either directly or by caching $\Y\,\widetilde{\mathbf{R}}_{XX}^{-1}$. Consequently, if $b_{\mathrm{vfe}}=\lvert\mathcal{J}_{\mathrm{vfe}}\rvert$, all VFE scores can be obtained in $\mathcal{O}(b_{\mathrm{vfe}} \, NM)$, rather than recomputing the full VFE objective for every candidate in the original batch.

\subsection{The Transfer Operator Criterion} \label{subsec: transfer dictionary criterion}

The flow-map gain~\eqref{eq: dict-vfe-fast-score} rewards candidates that improve the one-step regression of the states $\Y$, whereas Section~\ref{sec: Model Selection} selects the representation with the transfer operator objective~\eqref{eq: kZ VAMP2 objective}. The lifted counterpart of~\eqref{eq: dict-vfe-fast-score} follows from the same rank-one update~\eqref{eq: dict-vfe-rank-one-update}, provided that the lifted targets $\K_{ZY}$ and the reference posterior of $\overrightarrow{\Br}$, with mean $\overrightarrow{\Br}_\kappa$ of~\eqref{eq: K0}, are held at their size-$M$ values throughout a candidate sweep. A candidate $\bm{z} \in \mathcal{J}_{\mathrm{vfe}}$ then enlarges only the source dictionary, and the auxiliary model scored in the sweep has $M$ lifted output channels and $M + 1$ source inducing points. The reference-posterior average of~\eqref{eq: kZ null VFE} depends only on $\K_{ZY}$, $\overrightarrow{\Br}_\kappa$, and $\sigma_Y$, so it is common to all candidates and need not be evaluated. The change of the contrast~\eqref{eq: kZ score contrast} caused by $\bm{z}$ is therefore $- 2 \, \Delta\mathcal{F}_{\mathbf{k}_Z}^{\textrm{pair}}(\bm{z})$, where
\begin{alignat*}{3}
    \Delta\mathcal{F}_{\mathbf{k}_Z}^{\textrm{pair}}(\bm{z})
    &\coloneqq
    \mathbb{E} \! \left[ \mathcal{F}_{\mathbf{k}_Z}^{\textrm{pair}}(\theta, \vartheta_{M+1}(\bm{z})) \right]
    -
    \mathbb{E} \! \left[ \mathcal{F}_{\mathbf{k}_Z}^{\textrm{pair}}(\theta, \vartheta_M) \right]
\end{alignat*}
denotes the increase of the reference-posterior average of the paired variational free energy~\eqref{eq: kZ paired score} at fixed $\K_{ZY}$ and fixed reference posterior.

The paired likelihood~\eqref{eq: lifted likelihood} shares the column covariance $\widetilde{\mathbf{R}}_{XX}$ with the flow-map objective~\eqref{eq: dict-vfe-flow-objective} and differs from it only in the targets and the row covariance. By~\eqref{eq: K0} and~\eqref{eq: kZ whitened endpoint moments}, the reference posterior satisfies $\mathbb{E} [ \overrightarrow{\Br}^{-1} ] = ( N + M + 1 ) \, \sigma_Y^2 \, \Lb_{ZZ}^{-\top} \, \tilde{\mathbf{G}}_{YY}^{-1} \, \Lb_{ZZ}^{-1}$. Up to the candidate-independent term $- \tfrac{N}{2} \mathbb{E} [ \operatorname{ln} \lvert \overrightarrow{\Br}\rvert ]$, the reference-posterior average of the matrix-normal log-density in~\eqref{eq: kZ paired score} is therefore the sum of $M$ log-densities $\operatorname{ln} \mathcal{N}^{N}( \cdot \, ; \bm{0}, \widetilde{\mathbf{R}}_{XX} )$, one for each row of the whitened lifted targets
\begin{alignat*}{3}
    \sqrt{N + M + 1} \, \sigma_Y \, \tilde{\mathbf{G}}_{YY}^{-1/2} \, \mathbf{\Psi}_{ZY}
    &\in
    \mathbb{R}^{M \times N},
\end{alignat*}
where $\mathbf{\Psi}_{ZY} = \Lb_{ZZ}^{-1} \K_{ZY}$ as in Appendix~\ref{sec: Estimation Algorithm}. Consequently, $\Delta\mathcal{F}_{\mathbf{k}_Z}^{\textrm{pair}}(\bm{z})$ is obtained from~\eqref{eq: dict-vfe-fast-score} by replacing $D$ with $M$ and the rows $\Y_d$ with the rows of the whitened lifted targets, which gives
\begin{alignat}{3} \label{eq: dict-transfer-fast-score}
    \Delta\mathcal{F}_{\mathbf{k}_Z}^{\textrm{pair}}(\bm{z})
    &=
    - \frac{M}{2} \operatorname{ln}(1 + s_{\bm{z}})
    + \frac{( N + M + 1 ) \, \sigma_Y^2}{2 \, (1 + s_{\bm{z}})}
    \left\lVert
        \tilde{\mathbf{G}}_{YY}^{-1/2} \, \mathbf{\Psi}_{ZY} \, \widetilde{\mathbf{R}}_{XX}^{-1} \, \bm{u}_{\bm{z}}
    \right\rVert^2
    + \frac{M}{2 \sigma_Y^2} \lVert \bm{u}_{\bm{z}} \rVert^2 .
\end{alignat}
Writing $\mathcal{S} = \operatorname{Tr} [ \widetilde{\C}_{YY}^{-1} \, \K_{ZY} ( \I_N - \sigma_Y^2 \, \widetilde{\mathbf{R}}_{XX}^{-1} ) \K_{ZY}^{\top} ]$ with~\eqref{eq: kZ Woodbury identity} and applying the Sherman--Morrison identity above shows that the middle term of~\eqref{eq: dict-transfer-fast-score} equals $\tfrac{N + M + 1}{2} \, \Delta\mathcal{S}(\bm{z})$, where
\begin{alignat*}{3}
    \Delta\mathcal{S}(\bm{z})
    &\coloneqq
    \frac{\sigma_Y^2}{1 + s_{\bm{z}}}
    \left\lVert
        \tilde{\mathbf{G}}_{YY}^{-1/2} \, \mathbf{\Psi}_{ZY} \, \widetilde{\mathbf{R}}_{XX}^{-1} \, \bm{u}_{\bm{z}}
    \right\rVert^2
\end{alignat*}
is the increase of the VAMP-$2$ score~\eqref{eq: kZ VAMP2 score} induced by $\bm{z}$. Hence $- 2 \, \Delta\mathcal{F}_{\mathbf{k}_Z}^{\textrm{pair}}(\bm{z})$ consists of the changes of the VAMP-$2$, model-complexity, and sparsity terms of~\eqref{eq: kZ VAMP2 objective}. The coefficient $M$ in these terms stays fixed, because the number of lifted output channels does not change during the sweep. Under frozen targets, the backward terms of~\eqref{eq: fb lifted objective} are the same for every candidate and do not affect the ranking.

The singular value decomposition $\mathbf{\Psi}_{ZY} = \mathbf{M}_Y \, \mathbf{\Sigma}_Y \, \mathbf{N}_{Y}^{\top}$ and the inverse square root $\tilde{\mathbf{G}}_{YY}^{-1/2} = \mathbf{M}_Y \, \mathbf{\tilde{\Sigma}}_{Y}^{-1} \, \mathbf{M}_Y^{\top}$ of Appendix~\ref{sec: GP-TCCA} share the factor $\mathbf{M}_Y$, whose columns are orthonormal. For every $\bm{v} \in \mathbb{R}^{N}$, this gives
\begin{alignat*}{3}
    \lVert \tilde{\mathbf{G}}_{YY}^{-1/2} \, \mathbf{\Psi}_{ZY} \, \bm{v} \rVert
    &=
    \lVert \mathbf{M}_Y \, \mathbf{\tilde{\Sigma}}_{Y}^{-1} \, \mathbf{M}_Y^{\top} \, \mathbf{M}_Y \, \mathbf{\Sigma}_Y \, \mathbf{N}_{Y}^{\top} \, \bm{v} \rVert
    \\
    &=
    \lVert \mathbf{M}_Y \, \mathbf{\tilde{\Sigma}}_{Y}^{-1} \, \mathbf{\Sigma}_Y \, \mathbf{N}_{Y}^{\top} \, \bm{v} \rVert
    \\
    &=
    \lVert \mathbf{\tilde{\Sigma}}_{Y}^{-1} \, \mathbf{\Sigma}_Y \, \mathbf{N}_{Y}^{\top} \, \bm{v} \rVert ,
\end{alignat*}
where the second step uses $\mathbf{M}_Y^{\top} \, \mathbf{M}_Y = \I_M$ and the third uses that $\mathbf{M}_Y$ preserves norms. In $\Delta\mathcal{S}(\bm{z})$, the matrix $\tilde{\mathbf{G}}_{YY}^{-1/2} \, \mathbf{\Psi}_{ZY}$ can thus be replaced by the diagonally scaled factor $\mathbf{\tilde{\Sigma}}_{Y}^{-1} \, \mathbf{\Sigma}_Y \, \mathbf{N}_{Y}^{\top}$, which is available from the singular value decomposition without further factorization. Caching the product of this factor with $\widetilde{\mathbf{R}}_{XX}^{-1}$ through the inverse action above costs $\mathcal{O}(NM^2)$ once per sweep. The lifted term of each candidate then costs $\mathcal{O}(NM)$. All lifted scores are thus obtained in $\mathcal{O}(NM^2 + b_{\mathrm{vfe}} \, NM)$.

\begin{remark} \label{rem: dictionary scalarization}
    Since~\eqref{eq: dict-vfe-fast-score} and~\eqref{eq: dict-transfer-fast-score} are sums over $D$ and $M$ output channels, respectively, we combine them per output channel. With a weight $\beta \in [0, 1]$, the third criterion becomes
    \begin{alignat}{3} \label{eq: dict-scalarized-score}
        \bm{z}^{\star}
        &=
        \arg\max_{\bm{z} \in \mathcal{J}_{\mathrm{vfe}}}
        \left[
            \frac{1 - \beta}{D} \, \Delta\mathcal{F}_{\bm{F}}(\bm{z})
            + \frac{\beta}{M} \, \Delta\mathcal{F}_{\mathbf{k}_Z}^{\textrm{pair}}(\bm{z})
        \right]
        \nonumber\\
        &=
        \arg\max_{\bm{z} \in \mathcal{J}_{\mathrm{vfe}}}
        \left[
            - \frac{1}{2} \operatorname{ln}(1 + s_{\bm{z}})
            + \frac{\lVert \bm{u}_{\bm{z}} \rVert^2}{2 \sigma_Y^2}
            + \frac{1 - \beta}{2 D} \sum_{d=1}^{D} \frac{g_{d,\bm{z}}^2}{1 + s_{\bm{z}}}
            + \frac{\beta \, ( N + M + 1 )}{2 M} \, \Delta\mathcal{S}(\bm{z})
        \right].
    \end{alignat}
    The per-channel normalization leaves the model-complexity and sparsity terms independent of $\beta$, so the weight only apportions the target-fit reward between the states $\Y$ and the lifted targets $\K_{ZY}$. At $\beta = 0$ the criterion recovers the flow-map ranking of Appendix~\ref{subsec: VFE criterion}, and at $\beta = 1$ it ranks candidates by the reduction of the contrast~\eqref{eq: kZ score contrast} per lifted output channel. We use an intermediate weight because the lifted reward alone has two limitations. First, the lifted targets are features of the current dictionary, so the lifted reward measures how much a candidate improves the prediction of features that are already present, whereas the states $\Y$ are exogenous targets that do not move with the dictionary. Second, as noted after~\eqref{eq: kZ VAMP2 objective}, the lifted objective does not by itself guarantee the reconstruction property, whereas the flow-map reward is the reduction of the data mismatch of the regularized regression of $\Y$ on $\mathbf{k}_Z(\X)$ that also defines the projected modes~\eqref{eq: GP DMD modes}.

    The backward direction has an analogous lifted gain, obtained from~\eqref{eq: dict-transfer-fast-score} by exchanging $\X \leftrightarrow \Y$ and $\sigma_Y \leftrightarrow \sigma_X$. In that gain, the candidate enlarges the target-side dictionary, while $\K_{ZX}$ and $\widetilde{\C}_{XX}$ are held at their size-$M$ values. Averaging the forward and backward lifted gains would mirror~\eqref{eq: fb lifted objective}. The model-complexity and sparsity terms of the backward gain, however, are built from $\Y$ and the ridge $\sigma_X$, and differ from those of the flow-map gain. Keeping these terms independent of $\beta$ would then also require the backward full-state GP regression of Appendix~\ref{sec: GP-TCCA} in the flow-map term. For simplicity, a shared dictionary uses only the forward lifted gain.

    With separate dictionaries $\Z_X$ and $\Z_Y$, each dictionary has its own selection sweep and its own full-state regression, and the other dictionary is held fixed during the sweep. A candidate for $\Z_X$ raises the score $\mathcal{S}$, which enters both contrasts, changes only the forward penalties, and raises $M_X$ by one. Up to a term that does not depend on the candidate,~\eqref{eq: fb lifted objective} then decreases by
    \begin{alignat*}{3}
        \left( N + \frac{M_X + M_Y + 1}{2} + 1 \right) \Delta\mathcal{S}(\bm{z})
        - \frac{M_Y}{2} \operatorname{ln}(1 + s_{\bm{z}})
        + \frac{M_Y \, \lVert \bm{u}_{\bm{z}} \rVert^2}{2 \sigma_Y^2} .
    \end{alignat*}
    Dividing by $M_Y$ gives the lifted reward per output channel, which replaces the score term of~\eqref{eq: dict-scalarized-score} and keeps the model-complexity and sparsity terms independent of $\beta$. The sweep for $\Z_Y$ follows under the exchange $\X \leftrightarrow \Y$, $\sigma_Y \leftrightarrow \sigma_X$ and $M_X \leftrightarrow M_Y$.
\end{remark}

\begin{remark} \label{rem: dictionary transfer bounds}
    Since $\widetilde{\mathbf{R}}_{XX} \succeq \sigma_Y^2 \, \I_N$, we have $s_{\bm{z}} \leq \lVert \bm{u}_{\bm{z}} \rVert^2 / \sigma_Y^2$, so the first two terms of~\eqref{eq: dict-scalarized-score} are bounded below by $\tfrac{1}{2} ( s_{\bm{z}} - \operatorname{ln}(1 + s_{\bm{z}}) ) \geq 0$. The target-fit terms are nonnegative as well. The criterion is therefore nonnegative for every $\beta$, which is in line with \citet{titsias2009variational}, and it ranks candidates without deciding whether the dictionary should grow. Growth is controlled by the tolerance $\sigma_\textrm{tol}^2$, the size of the batch, and the cap $M_\textrm{max}$. Moreover, the Cauchy--Schwarz inequality in the inner product induced by $\widetilde{\mathbf{R}}_{XX}^{-1}$ bounds each target-fit reward by the fraction $s_{\bm{z}} / (1 + s_{\bm{z}})$ of the current data mismatch of its targets. For the lifted targets,~\eqref{eq: lifted residual identity} identifies this data mismatch and gives
    \begin{alignat*}{3}
        0
        \leq
        \Delta\mathcal{S}(\bm{z})
        &\leq
        \frac{s_{\bm{z}}}{1 + s_{\bm{z}}}
        \left(
            \operatorname{Tr}\!\left( \widetilde{\C}_{YY}^{-1} \, \C_{YY} \right)
            - \mathcal{S}
        \right),
        \qquad
        \mathcal{S}
        \leq
        \operatorname{Tr}\!\left( \widetilde{\C}_{YY}^{-1} \, \C_{YY} \right)
        <
        M .
    \end{alignat*}
    Each admission therefore closes at most the fraction $s_{\bm{z}} / (1 + s_{\bm{z}})$ of the gap between $\mathcal{S}$ and its ceiling $\operatorname{Tr}( \widetilde{\C}_{YY}^{-1} \, \C_{YY} )$. As this gap closes, the lifted reward vanishes and, at $\beta = 1$, the ranking falls back on the target-agnostic first two terms of~\eqref{eq: dict-scalarized-score}. A weight $\beta < 1$ keeps an exogenous target-fit reward in the ranking.
\end{remark}

\subsection{Scheduling and Warm-starts}
After $\bm{z}^\star$ is appended, the cross-covariance rows, the factor $\Lb_{ZZ}$, and the residuals $\sigma_M^2(\cdot)$ of the surviving candidates are refreshed by the same rank-one formulas, and the lifted targets $\K_{ZY}$ and the output covariance $\overrightarrow{\Br}_\kappa$ of Appendix~\ref{subsec: transfer dictionary criterion} are rebuilt with the enlarged dictionary. The loop --- filter, rank, retain, score, append --- repeats until no candidate clears the tolerance $\sigma_\textrm{tol}^2$, the batch is exhausted, or the dictionary reaches its cap $M_\textrm{max}$. To manage the computational cost, only once a preset number of new pseudo-inputs have been accumulated is the objective function invoked with analytical gradients to retune the generative and variational hyperparameters, warm-started by the greedy dictionary. The residuals and the lifted targets are then recomputed and selection resumes, interleaving inexpensive rank-one dictionary growth, retained-subset scoring by~\eqref{eq: dict-scalarized-score}, and occasional gradient-based refinement.

\section{Analytical Gradients of the Transfer Operator Objective} \label{sec: Gradient Derivations}

Here we detail the derivations of the analytical gradients of the transfer operator objective function. We write $\theta$ for the generative parameters and $\vartheta$ for the variational parameters. The optimized generative parameters $\theta$ contain the kernel lengthscales, while the signal variance $\sigma_f^2$ and the forward and backward observation-noise parameters $\sigma_Y$ and $\sigma_X$ are fixed at the values estimated from the corresponding full-state flow-map GPs. The variational parameters $\vartheta$ contain the pseudo-inputs $\Z$ and the inducing-noise parameter $\sigma_Z$ of Section~\ref{sec: Sparse Transfer Operators}, which is adapted alongside $\Z$. For reference, the regularized quantities of Sections~\ref{sec: Sparse Transfer Operators}--\ref{sec: Model Selection} are
\begin{alignat}{3}
    \widetilde{\K}_{ZZ}
    &\coloneqq
    \K_{ZZ}
    +
    \sigma_Z^2 \, \I_M
    =
    \Lb_{ZZ} \, \Lb_{ZZ}^{\top},
    \qquad
    \W_{ZX}
    \coloneqq
    \widetilde{\K}_{ZZ}^{-1} \, \K_{ZX},
    \label{eq: kZ regularized inducing definitions}
    \\
    \mathbf R_{XX}
    &\coloneqq
    \K_{ZX}^{\top}
    \, \widetilde{\K}_{ZZ}^{-1}
    \, \K_{ZX}
    =
    \K_{ZX}^{\top} \, \W_{ZX},
    \qquad
    \widetilde{\mathbf R}_{XX}
    \coloneqq
    \mathbf R_{XX}
    +
    \sigma_Y^2 \, \I_N,
    \label{eq: kZ regularized Nystrom definitions}
    \\
    \widetilde{\C}_{XX}
    &\coloneqq
    \K_{ZX} \, \K_{ZX}^{\top}
    +
    \sigma_Y^2 \, \widetilde{\K}_{ZZ},
    \qquad
    \widetilde{\C}_{YY}
    \coloneqq
    \K_{ZY} \, \K_{ZY}^{\top}
    +
    \sigma_X^2 \, \widetilde{\K}_{ZZ}.
    \label{eq: kZ regularized mass matrix}
\end{alignat}
Setting $\sigma_Z = 0$ removes the inducing noise, while the endpoint ridges $\sigma_Y^2$ and $\sigma_X^2$ remain in the regularized VAMP-$2$ normalization. With the same $\widetilde{\K}_{ZZ}$ throughout, the forward Woodbury identity gives
\begin{alignat}{3}
    \I_N
    -
    \K_{ZX}^{\top}
    \widetilde{\C}_{XX}^{-1}
    \K_{ZX}
    &=
    \sigma_Y^2
    \, \widetilde{\mathbf R}_{XX}^{-1}.
    \label{eq: kZ Woodbury identity}
\end{alignat}

We assume throughout that $\widetilde{\K}_{ZZ} \succ 0$, $\sigma_Y > 0$, $\sigma_X > 0$, and $M \leq N$. The positive regularization implies $\widetilde{\mathbf R}_{XX} \succ 0$, $\widetilde{\C}_{XX} \succ 0$, and $\widetilde{\C}_{YY} \succ 0$ without requiring $\K_{ZX}$ or $\K_{ZY}$ to have full row rank.

For the detailed gradient derivation, we consider one directional contrast explicitly. Its data-mismatch contribution uses the endpoint-regularized VAMP-$2$ score, while its model-complexity and sparsity contributions retain the forward expressions derived in Section~\ref{sec: Model Selection}. The corresponding backward penalty derivatives follow by exchanging $\X \leftrightarrow \Y$ and $\sigma_Y \leftrightarrow \sigma_X$ and therefore need not be derived separately. The reference-posterior average in~\eqref{eq: kZ score contrast} has been evaluated in closed form, so the gradients below differentiate this closed form. The dependence of the reference posterior on the parameters enters through $\widetilde{\C}_{YY}$ and requires no further term.
\begin{alignat}{3}
    \overrightarrow{\mathcal{L}}_{\mathbf{k}_Z}(\theta,\vartheta)
    &=
    \underbrace{
        - ( N + M + 1 ) \, \mathcal{S}
    }_{\mathcal{L}_{\mathrm{d}} \; \text{ -- data mismatch}}
    +
    \underbrace{
        M\operatorname{ln}
        \left|
            \I_N+\sigma_Y^{-2}\mathbf R_{XX}
        \right|
    }_{\overrightarrow{\mathcal{L}}_{\mathrm{c}} \; \text{ -- relative model complexity}}
    +
    \underbrace{
        \frac{M}{\sigma_Y^2}
        \operatorname{Tr}
        \left(
            \K_{XX}
            -
            \mathbf R_{XX}
        \right)
    }_{\overrightarrow{\mathcal{L}}_{\mathrm{s}} \; \text{ -- sparsity error}},
\end{alignat}
where
\begin{alignat*}{3}
    \mathcal{S}
    &=
    \operatorname{Tr}
    \left(
        \widetilde{\C}_{YY}^{-1}
        \C_{YX}
        \widetilde{\C}_{XX}^{-1}
        \C_{XY}
    \right)
    =
    \sum_{i = 1}^{M} \rho_i^2
\end{alignat*}
is the endpoint-regularized VAMP-$2$ contribution of Equation~\eqref{eq: fb VAMP2 score}. In the present method, the target-side ridge is fixed at the calibrated value $\sigma_X^2$. The gradient derivation itself requires only the resulting regularized covariance $\widetilde{\C}_{YY}$. The regularized quantities in the present appendix depend additionally on the inducing-noise parameter $\sigma_Z$ through $\widetilde{\K}_{ZZ}$. The detailed expressions below concern the forward model-complexity and sparsity terms, and the full forward--backward objective is assembled from the two directional penalty contributions as defined in Equation~\eqref{eq: fb lifted objective}. The derivatives of the data-mismatch, model-complexity, and sparsity-error contributions are treated in turn below.

\subsection{Model complexity} \label{sec: model complexity}

The model-complexity term is
\begin{alignat*}{3}
    \overrightarrow{\mathcal{L}}_{\mathrm{c}}
    &=
    M\operatorname{ln}
    \left|
        \I_N+\sigma_Y^{-2}\mathbf R_{XX}
    \right|
    =
    M\operatorname{ln}
    \left|
        \widetilde{\mathbf R}_{XX}
    \right|
    -
    MN\operatorname{ln}(\sigma_Y^2).
\end{alignat*}

\subsubsection*{Efficient computation}

Recall that
\begin{alignat*}{3}
    \widetilde{\mathbf R}_{XX}
    &=
    \mathbf\Psi_{ZX}^{\top}\mathbf\Psi_{ZX}
    +
    \sigma_Y^2\I_N,
    \qquad
    \mathbf\Psi_{ZX}
    =
    \Lb_{ZZ}^{-1}\K_{ZX}.
\end{alignat*}
Using the singular value decomposition already defined in Appendix~\ref{sec: GP-TCCA},
\begin{alignat*}{3}
    \mathbf\Psi_{ZX}
    &=
    \mathbf{M}_X
    \mathbf{\Sigma}_X
    \mathbf{N}_{X}^{\top},
\end{alignat*}
the eigenvalues of $\mathbf\Psi_{ZX}^{\top}\mathbf\Psi_{ZX}$ consist of $\mathbf{\Sigma}_{X,1}^2,\ldots,\mathbf{\Sigma}_{X,M}^2$ together with $N-M$ additional zeros, so that
\begin{alignat}{3}
    \operatorname{ln}
    \left|
        \widetilde{\mathbf R}_{XX}
    \right|
    &=
    N\operatorname{ln}(\sigma_Y^2)
    +
    \sum_{j=1}^{M}
    \operatorname{ln}
    \left(
        1+
        \frac{\mathbf{\Sigma}_{X,j}^2}{\sigma_Y^2}
    \right),
    \label{eq: kZ efficient logdet R}
\end{alignat}
and the $N\operatorname{ln}(\sigma_Y^2)$ cancels against the reference model:
\begin{alignat}{3}
    \boxed{
    \overrightarrow{\mathcal{L}}_{\mathrm{c}}
    =
    M
    \sum_{j=1}^{M}
    \operatorname{ln}
    \left(
        1+
        \frac{\mathbf{\Sigma}_{X,j}^2}{\sigma_Y^2}
    \right).
    }
    \label{eq: kZ efficient complexity}
\end{alignat}
Hence the model complexity requires only the $M$ singular values of $\mathbf\Psi_{ZX}$ and never the formation or factorization of a dense $N\times N$ matrix.

\subsubsection*{Differential}

By Jacobi's formula,
\begin{alignat}{3}
    \mathrm d
    \operatorname{ln}
    \left|
        \widetilde{\mathbf R}_{XX}
    \right|
    &=
    \operatorname{Tr}
    \left(
        \widetilde{\mathbf R}_{XX}^{-1}
        \mathrm d\widetilde{\mathbf R}_{XX}
    \right).
    \label{eq: kZ dlogdet R}
\end{alignat}
Since the source-side ridge is held fixed when differentiating with respect to a kernel parameter or a pseudo-input coordinate, the differential of the model-complexity contribution is
\begin{alignat}{3}
    \mathrm d\mathcal L_{\mathrm c}
    &=
    M
    \operatorname{Tr}
    \left(
        \widetilde{\mathbf R}_{XX}^{-1}
        \mathrm d\widetilde{\mathbf R}_{XX}
    \right).
    \label{eq: kZ differential model complexity}
\end{alignat}

\subsection{Sparsity-error} \label{sec: sparsity error}

The sparsity-error term is
\begin{alignat*}{3}
    \overrightarrow{\mathcal{L}}_{\mathrm{s}}
    =
    \frac{M}{\sigma_Y^2}
    \operatorname{Tr}
    \left(
        \K_{XX}
        -
        \mathbf{R}_{XX}
    \right),
    \qquad
    \mathbf{R}_{XX}
    =
    \K_{ZX}^{\top}
    \widetilde{\K}_{ZZ}^{-1}
    \K_{ZX}.
\end{alignat*}

\subsubsection*{Efficient Computation}

The full $N \times N$ matrices $\K_{XX}$ and $\mathbf{R}_{XX}$ do not need to be formed. The first trace only requires the diagonal of the exact prior covariance. For the second Nystr\"om term we use the singular values in the decomposition $\mathbf{\Psi}_{ZX}=\mathbf{M}_X\mathbf{\Sigma}_X\mathbf{N}_{X}^{\top}$ defined in Appendix~\ref{sec: GP-TCCA}. Thus the sparsity-error term can be evaluated as
\begin{alignat}{3} \label{eq: efficient sparsity error}
    \boxed{
        \overrightarrow{\mathcal{L}}_{\mathrm{s}}
        =
        \frac{M}{\sigma_Y^2}
        \left(
            \sum_{n=1}^{N}
            \kappa_{\textrm{pr}}(\bm{x}_n,\bm{x}_n)
            -
            \sum_{j=1}^{M}
            \mathbf{\Sigma}_{X,j}^2
        \right).
    }
\end{alignat}
Hence the sparsity error only requires the diagonal of $\K_{XX}$ and singular values of $\mathbf{\Psi}_{ZX}$.

\subsubsection*{Differential}

Taking the differential of the sparsity-error term gives
\begin{alignat*}{3}
    \boxed{
        \mathrm{d}\overrightarrow{\mathcal{L}}_{\mathrm{s}}
        =
        \frac{M}{\sigma_Y^2}
        \operatorname{Tr}
        \left(
            \mathrm{d}\K_{XX}
            -
            \mathrm{d}\mathbf{R}_{XX}
        \right).
    }
\end{alignat*}

\subsection{Derivatives of Pseudo-inputs and Kernel Parameters}

We consider a scalar parameter $\eta_i$ that is either a kernel lengthscale in $\theta$ or a pseudo-input coordinate $Z_{m,r}$ contained in $\vartheta$. The endpoint ridges $\sigma_Y$ and $\sigma_X$, together with $\sigma_Z$, are held fixed throughout this subsection. The parameter $\eta_i$ may affect the covariance matrices $\K_{ZX}$, $\K_{ZY}$, $\K_{ZZ}$, and $\K_{XX}$. Accordingly, we define
\begin{alignat*}{3}
    \partial_i\K_{ZX}
    &\coloneqq
    \frac{\partial\K_{ZX}}{\partial\eta_i},
    \qquad
    \partial_i\K_{ZY}
    \coloneqq
    \frac{\partial\K_{ZY}}{\partial\eta_i},
    \qquad
    \partial_i\widetilde{\K}_{ZZ}
    &\coloneqq
    \frac{\partial\widetilde{\K}_{ZZ}}{\partial\eta_i}
    =
    \frac{\partial\K_{ZZ}}{\partial\eta_i},
    \qquad
    \partial_i\K_{XX}
    \coloneqq
    \frac{\partial\K_{XX}}{\partial\eta_i}.
\end{alignat*}
The equality $\partial_i\widetilde{\K}_{ZZ}=\partial_i\K_{ZZ}$ follows because $\sigma_Z$ has been fixed. For a pseudo-input coordinate, $\partial_i\K_{XX}=\bm{0}$, whereas a kernel hyperparameter may contribute to all four derivatives.

\subsubsection*{Noisy Nystr\"om Covariance}

The Nystr\"om covariance is
\begin{alignat*}{3}
    \mathbf R_{XX}
    &=
    \K_{ZX}^{\top}
    \widetilde{\K}_{ZZ}^{-1}
    \K_{ZX},
    \qquad
    \widetilde{\mathbf R}_{XX}
    =
    \mathbf R_{XX}
    +
    \sigma_Y^2\I_N,
    \qquad
    \W_{ZX}
    =
    \widetilde{\K}_{ZZ}^{-1}\K_{ZX}.
\end{alignat*}
Since $\sigma_Y$ is fixed when differentiating with respect to $\eta_i$,
\begin{alignat*}{3}
    \partial_i\widetilde{\mathbf R}_{XX}
    &=
    \partial_i\mathbf R_{XX}.
\end{alignat*}
Using
\begin{alignat*}{3}
    \partial_i
    \left(
        \widetilde{\K}_{ZZ}^{-1}
    \right)
    &=
    -
    \widetilde{\K}_{ZZ}^{-1}
    \left(
        \partial_i\widetilde{\K}_{ZZ}
    \right)
    \widetilde{\K}_{ZZ}^{-1},
\end{alignat*}
we obtain
\begin{alignat}{3}
    \boxed{
    \partial_i\widetilde{\mathbf R}_{XX}
    =
    \left(
        \partial_i\K_{ZX}
    \right)^{\top}
    \W_{ZX}
    +
    \W_{ZX}^{\top}
    \partial_i\K_{ZX}
    -
    \W_{ZX}^{\top}
    \left(
        \partial_i\widetilde{\K}_{ZZ}
    \right)
    \W_{ZX}.
    }
    \label{eq: kZ master Nystrom differential}
\end{alignat}
This identity is the differential required by the model-complexity and sparsity-error contributions.

\subsection{Data mismatch} \label{sec: data mismatch}

The data-mismatch term is $\mathcal L_{\mathrm d} = - ( N + M + 1 ) \, \mathcal{S}$ with $\mathcal{S}$ given by~\eqref{eq: fb VAMP2 score}. For numerical evaluation we use the $\Lb_{ZZ}$-preconditioned basis of Appendix~\ref{sec: GP-TCCA}, with the whitened matrices $\mathbf\Psi_{ZX} = \Lb_{ZZ}^{-1} \K_{ZX}$ and $\mathbf\Psi_{ZY} = \Lb_{ZZ}^{-1} \K_{ZY}$ and the Gramian and stiffness matrices $\mathbf{G}_{XX}$, $\mathbf{G}_{YY}$, and $\mathbf{G}_{XY}$ defined there. Since $\K_{ZX} = \Lb_{ZZ} \mathbf\Psi_{ZX}$, $\K_{ZY} = \Lb_{ZZ} \mathbf\Psi_{ZY}$, and $\widetilde{\K}_{ZZ} = \Lb_{ZZ} \Lb_{ZZ}^{\top}$,
\begin{alignat}{3}
    \widetilde{\C}_{XX}
    &=
    \Lb_{ZZ}
    \, \tilde{\mathbf{G}}_{XX}
    \, \Lb_{ZZ}^{\top},
    \qquad
    \widetilde{\C}_{YY}
    =
    \Lb_{ZZ}
    \, \tilde{\mathbf{G}}_{YY}
    \, \Lb_{ZZ}^{\top},
    \nonumber\\
    \C_{XY}
    &=
    \Lb_{ZZ}
    \, \mathbf{G}_{XY}
    \, \Lb_{ZZ}^{\top},
    \label{eq: kZ whitened endpoint moments}
\end{alignat}
where $\tilde{\mathbf{G}}_{XX} = \mathbf{G}_{XX} + \sigma_Y^2 \I_M$ and $\tilde{\mathbf{G}}_{YY} = \mathbf{G}_{YY} + \sigma_X^2 \I_M$. Since $M \leq N$, the left singular factors $\mathbf{M}_X$ and $\mathbf{M}_Y$ of the thin singular value decompositions $\mathbf\Psi_{ZX} = \mathbf{M}_X \mathbf{\Sigma}_X \mathbf{N}_{X}^{\top}$ and $\mathbf\Psi_{ZY} = \mathbf{M}_Y \mathbf{\Sigma}_Y \mathbf{N}_{Y}^{\top}$ of Appendix~\ref{sec: GP-TCCA} are square and orthogonal, whether or not $\mathbf\Psi_{ZX}$ or $\mathbf\Psi_{ZY}$ has full row rank. Hence $\tilde{\mathbf{G}}_{XX} = \mathbf{M}_X \mathbf{\tilde{\Sigma}}_{X}^{2} \mathbf{M}_X^{\top}$ and $\tilde{\mathbf{G}}_{YY} = \mathbf{M}_Y \mathbf{\tilde{\Sigma}}_{Y}^{2} \mathbf{M}_Y^{\top}$, with the regularized singular values of Appendix~\ref{sec: GP-TCCA}, and the endpoint inverse-square-root actions are
\begin{alignat}{3}
    \tilde{\mathbf{G}}_{XX}^{-1/2}
    &=
    \mathbf{M}_X
    \, \mathbf{\tilde{\Sigma}}_{X}^{-1}
    \, \mathbf{M}_X^{\top},
    \qquad
    \tilde{\mathbf{G}}_{YY}^{-1/2}
    =
    \mathbf{M}_Y
    \, \mathbf{\tilde{\Sigma}}_{Y}^{-1}
    \, \mathbf{M}_Y^{\top}.
    \label{eq: kZ SVD endpoint actions}
\end{alignat}
Every diagonal entry of $\mathbf{\tilde{\Sigma}}_{X}$ and $\mathbf{\tilde{\Sigma}}_{Y}$ is bounded below by $\sigma_Y$ and $\sigma_X$, respectively, so the endpoint-regularized score is evaluated without factorizing either endpoint matrix:
\begin{alignat}{3}
    \mathcal S
    &=
    \operatorname{Tr}
    \left(
        \tilde{\mathbf{G}}_{XX}^{-1}
        \, \mathbf{G}_{XY}
        \, \tilde{\mathbf{G}}_{YY}^{-1}
        \, \mathbf{G}_{XY}^{\top}
    \right)
    =
    \left\|
        \U'
    \right\|_F^2,
    \label{eq: kZ whitened score}
\end{alignat}
with $\U' = \tilde{\mathbf{G}}_{XX}^{-1/2} \mathbf{G}_{XY} \tilde{\mathbf{G}}_{YY}^{-1/2}$ as in Appendix~\ref{sec: GP-TCCA}.

For the algebraic differential, define
\begin{alignat}{3}
    \mathbf T_{XY}
    \coloneqq
    \widetilde{\C}_{XX}^{-1}
    \C_{XY}
    \widetilde{\C}_{YY}^{-1}
    &=
    \Lb_{ZZ}^{-\top}
    \, \tilde{\mathbf{G}}_{XX}^{-1/2}
    \, \U'
    \, \tilde{\mathbf{G}}_{YY}^{-1/2}
    \, \Lb_{ZZ}^{-1},
    \nonumber\\
    \mathcal S
    &=
    \operatorname{Tr}
    \left(
        \mathbf T_{XY}
        \C_{YX}
    \right),
    \label{eq: kZ TXY}
\end{alignat}
although $\mathbf T_{XY}$ itself need not be formed numerically. Define further the symmetric positive-semidefinite matrices
\begin{alignat}{3}
    \bm\Gamma_X
    &\coloneqq
    \mathbf T_{XY}
    \widetilde{\C}_{YY}
    \mathbf T_{XY}^{\top}
    =
    \Lb_{ZZ}^{-\top}
    \, \tilde{\mathbf{G}}_{XX}^{-1/2}
    \, \U' \U'^{\top}
    \, \tilde{\mathbf{G}}_{XX}^{-1/2}
    \, \Lb_{ZZ}^{-1},
    \nonumber\\
    \bm\Gamma_Y
    &\coloneqq
    \mathbf T_{XY}^{\top}
    \widetilde{\C}_{XX}
    \mathbf T_{XY}
    =
    \Lb_{ZZ}^{-\top}
    \, \tilde{\mathbf{G}}_{YY}^{-1/2}
    \, \U'^{\top} \U'
    \, \tilde{\mathbf{G}}_{YY}^{-1/2}
    \, \Lb_{ZZ}^{-1},
    \label{eq: kZ Gamma XY}
\end{alignat}
which satisfy $\operatorname{Tr}(\bm\Gamma_X \, \widetilde{\C}_{XX}) = \operatorname{Tr}(\bm\Gamma_Y \, \widetilde{\C}_{YY}) = \mathcal{S}$. In the preconditioned form, $\bm\Gamma_X$ and $\bm\Gamma_Y$ are the Gram matrices of $\Lb_{ZZ}^{-\top} \tilde{\mathbf{G}}_{XX}^{-1/2} \U'$ and $\Lb_{ZZ}^{-\top} \tilde{\mathbf{G}}_{YY}^{-1/2} \U'^{\top}$, so they are positive semidefinite by construction.

\subsubsection*{Differential}

Using $\mathrm d(\mathbf X^{-1})=-\mathbf X^{-1}(\mathrm d\mathbf X)\mathbf X^{-1}$ and the cyclic property of the trace, the differential of~\eqref{eq: fb VAMP2 score} is
\begin{alignat}{3}
    \mathrm d\mathcal{S}
    &=
    2\operatorname{Tr}
    \left(
        \mathbf T_{XY}^{\top}
        \,\mathrm d\C_{XY}
    \right)
    -
    \operatorname{Tr}
    \left(
        \bm\Gamma_Y
        \,\mathrm d\widetilde{\C}_{YY}
    \right)
    -
    \operatorname{Tr}
    \left(
        \bm\Gamma_X
        \,\mathrm d\widetilde{\C}_{XX}
    \right).
    \label{eq: kZ dS master}
\end{alignat}
The differential is taken in the original covariance variables and is therefore independent of the numerical factorization used to evaluate the matrices above. Neither $\Lb_{ZZ}$ nor the singular vectors in the decompositions of $\mathbf\Psi_{ZX}$ and $\mathbf\Psi_{ZY}$ are differentiated. The singular value decompositions are used only to evaluate the inverse-square-root actions~\eqref{eq: kZ SVD endpoint actions}, and through them $\mathcal{S}$, $\bm\Gamma_X$, $\bm\Gamma_Y$, $\mathbf A_X$, and $\mathbf A_Y$, at the current parameters; the individual $\rho_i$ are never differentiated or required. Consequently $\mathcal{S}$ remains differentiable where two of the singular values $\rho_i$ coincide, at which points the singular vectors themselves are not.

\subsubsection*{Derivatives with Respect to Kernel Parameters and Pseudo-inputs}

Let $\eta_i$ be a kernel hyperparameter or a pseudo-input coordinate, with the endpoint ridges $\sigma_Y$, $\sigma_X$, and the inducing-noise parameter $\sigma_Z$ held fixed. The second moments differentiate as
\begin{alignat*}{3}
    \partial_i\C_{XY}
    &=
    \left(
        \partial_i\K_{ZX}
    \right)
    \K_{ZY}^{\top}
    +
    \K_{ZX}
    \left(
        \partial_i\K_{ZY}
    \right)^{\top},
    \\
    \partial_i\widetilde{\C}_{YY}
    &=
    \left(
        \partial_i\K_{ZY}
    \right)
    \K_{ZY}^{\top}
    +
    \K_{ZY}
    \left(
        \partial_i\K_{ZY}
    \right)^{\top}
    +
    \sigma_X^2
    \,
    \partial_i\widetilde{\K}_{ZZ},
    \\
    \partial_i\widetilde{\C}_{XX}
    &=
    \left(
        \partial_i\K_{ZX}
    \right)
    \K_{ZX}^{\top}
    +
    \K_{ZX}
    \left(
        \partial_i\K_{ZX}
    \right)^{\top}
    +
    \sigma_Y^2
    \,
    \partial_i\widetilde{\K}_{ZZ}.
\end{alignat*}
Substituting these into~\eqref{eq: kZ dS master} and collecting the coefficients of $\partial_i\K_{ZY}$ and $\partial_i\K_{ZX}$, we introduce the $M\times N$ matrices
\begin{alignat}{3}
    \mathbf A_Y
    &\coloneqq
    \mathbf T_{XY}^{\top}\K_{ZX}
    -
    \bm\Gamma_Y\K_{ZY},
    \qquad
    \mathbf A_X
    \coloneqq
    \mathbf T_{XY}\K_{ZY}
    -
    \bm\Gamma_X\K_{ZX},
    \label{eq: kZ AX AY}
\end{alignat}
and obtain
\begin{alignat}{3}
    \boxed{
    \partial_i\mathcal{S}
    =
    2\operatorname{Tr}
    \left(
        \mathbf A_Y
        \left(
            \partial_i\K_{ZY}
        \right)^{\top}
    \right)
    +
    2\operatorname{Tr}
    \left(
        \mathbf A_X
        \left(
            \partial_i\K_{ZX}
        \right)^{\top}
    \right)
    -
    \sigma_Y^2
    \operatorname{Tr}
    \left(
        \bm\Gamma_X
        \,
        \partial_i\widetilde{\K}_{ZZ}
    \right)
    -
    \sigma_X^2
    \operatorname{Tr}
    \left(
        \bm\Gamma_Y
        \,
        \partial_i\widetilde{\K}_{ZZ}
    \right).
    }
    \label{eq: kZ generic data mismatch derivative}
\end{alignat}
The first two traces are Frobenius inner products of $M\times N$ matrices, so once $\mathbf A_X$, $\mathbf A_Y$, $\bm\Gamma_X$, and $\bm\Gamma_Y$ have been formed they are reused across every scalar parameter, and each additional gradient component costs $\mathcal O(MN)$.

For numerical evaluation, substituting $\K_{ZX} = \Lb_{ZZ} \mathbf\Psi_{ZX}$ and $\K_{ZY} = \Lb_{ZZ} \mathbf\Psi_{ZY}$ together with~\eqref{eq: kZ TXY} and~\eqref{eq: kZ Gamma XY} into~\eqref{eq: kZ AX AY} gives $\mathbf A_X$ and $\mathbf A_Y$ directly in the preconditioned basis, without forming $\mathbf T_{XY}$:
\begin{alignat}{3}
    \mathbf A_X
    &=
    \Lb_{ZZ}^{-\top}
    \, \tilde{\mathbf{G}}_{XX}^{-1/2}
    \, \U'
    \left(
        \tilde{\mathbf{G}}_{YY}^{-1/2}
        \, \mathbf\Psi_{ZY}
        -
        \U'^{\top}
        \, \tilde{\mathbf{G}}_{XX}^{-1/2}
        \, \mathbf\Psi_{ZX}
    \right),
    \nonumber\\
    \mathbf A_Y
    &=
    \Lb_{ZZ}^{-\top}
    \, \tilde{\mathbf{G}}_{YY}^{-1/2}
    \, \U'^{\top}
    \left(
        \tilde{\mathbf{G}}_{XX}^{-1/2}
        \, \mathbf\Psi_{ZX}
        -
        \U'
        \, \tilde{\mathbf{G}}_{YY}^{-1/2}
        \, \mathbf\Psi_{ZY}
    \right).
    \label{eq: kZ SVD AX AY}
\end{alignat}
The only operation involving $\Lb_{ZZ}$ is the triangular solve with $\Lb_{ZZ}^{\top}$, as for $\W_{ZX} = \Lb_{ZZ}^{-\top} \mathbf\Psi_{ZX}$. Since $\C_{XX} = \widetilde{\C}_{XX} - \sigma_Y^2 \widetilde{\K}_{ZZ}$ and $\C_{YY} = \widetilde{\C}_{YY} - \sigma_X^2 \widetilde{\K}_{ZZ}$, the definitions~\eqref{eq: kZ AX AY} and~\eqref{eq: kZ Gamma XY} imply $\operatorname{Tr}(\mathbf A_X \K_{ZX}^{\top}) = \sigma_Y^2 \operatorname{Tr}(\bm\Gamma_X \widetilde{\K}_{ZZ})$ and $\operatorname{Tr}(\mathbf A_Y \K_{ZY}^{\top}) = \sigma_X^2 \operatorname{Tr}(\bm\Gamma_Y \widetilde{\K}_{ZZ})$, a useful implementation check that involves only $\K_{ZX}$, $\K_{ZY}$, and $\widetilde{\K}_{ZZ}$.
Combining~\eqref{eq: kZ generic data mismatch derivative} with~\eqref{eq: kZ differential model complexity}, the master Nyström differential~\eqref{eq: kZ master Nystrom differential}, and the sparsity-error differential of Appendix~\ref{sec: sparsity error}, the generic gradient component is
\begin{alignat}{3}
    \boxed{
    \frac{\partial\mathcal L_{\mathbf k_Z}}{\partial\eta_i}
    =
    - ( N + M + 1 ) \, \partial_i\mathcal{S}
    +
    M\operatorname{Tr}
    \left(
        \widetilde{\mathbf R}_{XX}^{-1}
        \partial_i\widetilde{\mathbf R}_{XX}
    \right)
    +
    \frac{M}{\sigma_Y^2}
    \operatorname{Tr}
    \left(
        \partial_i\K_{XX}
        -
        \partial_i\widetilde{\mathbf R}_{XX}
    \right).
    }
    \label{eq: kZ generic gradient}
\end{alignat}
For a pseudo-input coordinate, $\partial_i\K_{XX}=\bm 0$. The backward model-complexity and sparsity gradients follow from the corresponding forward expressions under the exchange $\X \leftrightarrow \Y$ and $\sigma_Y \leftrightarrow \sigma_X$. In the full objective~\eqref{eq: fb lifted objective}, the two directional penalty gradients are averaged, whereas the derivative of the common endpoint-regularized VAMP-$2$ score is included only once.

With separate dictionaries $\Z_X$ and $\Z_Y$, each endpoint is whitened with the Cholesky factor of its own dictionary, the forward and backward contrasts carry $M_Y$ and $M_X$ in place of $M$, and $\U'$ has size $M_X \times M_Y$. The definitions~\eqref{eq: kZ TXY},~\eqref{eq: kZ Gamma XY} and~\eqref{eq: kZ AX AY} hold unchanged in the covariance variables, with $\mathbf T_{XY}$ of size $M_X \times M_Y$ and $\mathbf A_X$ of size $M_X \times N$. Each dictionary and its lengthscales are optimized with those of the other dictionary held fixed, and the dictionary sizes are constant during this step. For the parameters of $\Z_X$, the derivatives $\partial_i\K_{ZY}$ and $\partial_i\K_{YY}$ vanish, as does the derivative of the pseudo-covariance matrix of $\Z_Y$. Writing $\widetilde{\K}_{ZZ}$ for the pseudo-covariance matrix of $\Z_X$, equations~\eqref{eq: kZ generic data mismatch derivative} and~\eqref{eq: kZ generic gradient} then give
\begin{alignat*}{3}
    \frac{\partial\mathcal L_{\mathbf k_Z}^{\textrm{fb}}}{\partial\eta_i}
    &=
    - \left( N + \frac{M_X + M_Y}{2} + 1 \right)
    \left[
        2\operatorname{Tr}
        \left(
            \mathbf A_X
            \left(
                \partial_i\K_{ZX}
            \right)^{\top}
        \right)
        -
        \sigma_Y^2
        \operatorname{Tr}
        \left(
            \bm\Gamma_X
            \,
            \partial_i\widetilde{\K}_{ZZ}
        \right)
    \right]
    \\
    &\quad
    +
    \frac{M_Y}{2}
    \operatorname{Tr}
    \left(
        \widetilde{\mathbf R}_{XX}^{-1}
        \partial_i\widetilde{\mathbf R}_{XX}
    \right)
    +
    \frac{M_Y}{2 \sigma_Y^2}
    \operatorname{Tr}
    \left(
        \partial_i\K_{XX}
        -
        \partial_i\widetilde{\mathbf R}_{XX}
    \right).
\end{alignat*}
The gradient for the parameters of $\Z_Y$ follows under the exchange $\X \leftrightarrow \Y$ and $\sigma_Y \leftrightarrow \sigma_X$.

\subsection{Derivatives of Noise Parameters}

\subsubsection*{Variational Inducing-Noise Parameter}

Only $\widetilde{\K}_{ZZ}$ depends on $\sigma_Z$, with $\partial\widetilde{\K}_{ZZ}/\partial\sigma_Z=2\sigma_Z\I_M$. It enters $\widetilde{\C}_{XX}$ through the term $\sigma_Y^2 \, \widetilde{\K}_{ZZ}$, $\widetilde{\C}_{YY}$ through $\sigma_X^2 \, \widetilde{\K}_{ZZ}$, and $\mathbf R_{XX}$ through the inverse, so
\begin{alignat}{3}
    \frac{\partial\mathcal{S}}{\partial\sigma_Z}
    =
    -2 \, \sigma_Z
    \left[
        \sigma_Y^2
        \operatorname{Tr}
        \left(
            \bm\Gamma_X
        \right)
        +
        \sigma_X^2
        \operatorname{Tr}
        \left(
            \bm\Gamma_Y
        \right)
    \right],
    \qquad
    \frac{\partial\mathbf R_{XX}}{\partial\sigma_Z}
    =
    -2\sigma_Z
    \W_{ZX}^{\top}
    \W_{ZX}.
    \label{eq: kZ sigmaZ covariance derivative}
\end{alignat}
Consequently
\begin{alignat}{3}
    \frac{\partial\mathcal L_{\mathrm c}}{\partial\sigma_Z}
    &=
    -2M\sigma_Z
    \operatorname{Tr}
    \left(
        \widetilde{\mathbf R}_{XX}^{-1}
        \W_{ZX}^{\top}
        \W_{ZX}
    \right),
    \qquad
    \frac{\partial\mathcal L_{\mathrm s}}{\partial\sigma_Z}
    =
    \frac{2M\sigma_Z}{\sigma_Y^2}
    \left\|
        \W_{ZX}
    \right\|_F^2,
    \label{eq: kZ sigmaZ term derivatives}
\end{alignat}
and
\begin{alignat}{3}
    \boxed{
            \frac{\partial\mathcal L_{\mathbf k_Z}}{\partial\sigma_Z}
        =
        - ( N + M + 1 )
        \frac{\partial\mathcal{S}}{\partial\sigma_Z}
        +
        \frac{\partial\mathcal L_{\mathrm c}}{\partial\sigma_Z}
        +
        \frac{\partial\mathcal L_{\mathrm s}}{\partial\sigma_Z}
    }
    \label{eq: kZ sigmaZ gradient}
\end{alignat}

\subsection{Efficient Inverse Actions and Traces}

The objective and its gradients require inverse actions of $\widetilde{\mathbf R}_{XX}$, but do not require the explicit formation of its dense inverse. Recall from Appendix~\ref{sec: GP-TCCA} that
\begin{alignat*}{3}
    \mathbf\Psi_{ZX}
    &=
    \mathbf{M}_X
    \mathbf{\Sigma}_X
    \mathbf{N}_{X}^{\top},
\end{alignat*}
where $\mathbf{N}_{X}\in\mathbb R^{N\times M}$ has orthonormal columns. Since
\begin{alignat*}{3}
    \widetilde{\mathbf R}_{XX}
    &=
    \mathbf{N}_{X}
    \mathbf{\Sigma}_X^{2}
    \mathbf{N}_{X}^{\top}
    +
    \sigma_Y^2\I_N,
\end{alignat*}
its inverse is
\begin{alignat}{3}
    \widetilde{\mathbf R}_{XX}^{-1}
    &=
    \frac1{\sigma_Y^2}
    \left[
        \I_N
        -
        \mathbf{N}_{X}
        \operatorname{diag}
        \left(
            \frac{\mathbf{\Sigma}_{X,j}^2}
            {\mathbf{\Sigma}_{X,j}^2+\sigma_Y^2}
        \right)_{j=1}^{M}
        \mathbf{N}_{X}^{\top}
    \right].
    \label{eq: kZ efficient inverse R}
\end{alignat}
Consequently, for any matrix $\mathbf B$ with $N$ columns,
\begin{alignat}{3}
    \mathbf B \, \widetilde{\mathbf R}_{XX}^{-1}
    &=
    \frac1{\sigma_Y^2}
    \left[
        \mathbf B
        -
        \left(
            \mathbf B\mathbf{N}_{X}
        \right)
        \operatorname{diag}
        \left(
            \frac{\mathbf{\Sigma}_{X,j}^2}
            {\mathbf{\Sigma}_{X,j}^2+\sigma_Y^2}
        \right)_{j=1}^{M}
        \mathbf{N}_{X}^{\top}
    \right].
    \label{eq: kZ efficient inverse action}
\end{alignat}
In particular, setting $\mathbf B=\K_{ZY}$ evaluates $\K_{ZY}\widetilde{\mathbf R}_{XX}^{-1}$ without forming $\widetilde{\mathbf R}_{XX}^{-1}$.

The trace of the inverse is
\begin{alignat}{3}
    \operatorname{Tr}
    \left(
        \widetilde{\mathbf R}_{XX}^{-1}
    \right)
    &=
    \frac{N-M}{\sigma_Y^2}
    +
    \sum_{j=1}^{M}
    \frac1{\mathbf{\Sigma}_{X,j}^2+\sigma_Y^2}.
    \label{eq: kZ efficient inverse trace}
\end{alignat}
Moreover,
\begin{alignat}{3}
    \operatorname{Tr}
    \left(
        \mathbf R_{XX}
    \right)
    &=
    \operatorname{Tr}
    \left(
        \K_{ZX}^{\top}
        \widetilde{\K}_{ZZ}^{-1}
        \K_{ZX}
    \right)
    =
    \left\|
        \mathbf\Psi_{ZX}
    \right\|_{\mathrm F}^{2}
    =
    \sum_{j=1}^{M}
    \mathbf{\Sigma}_{X,j}^2,
    \label{eq: kZ efficient Nystrom trace}
\end{alignat}
so that
\begin{alignat}{3}
    \operatorname{Tr}
    \left(
        \K_{XX}
        -
        \mathbf R_{XX}
    \right)
    &=
    \operatorname{Tr}
    \left(
        \K_{XX}
    \right)
    -
    \sum_{j=1}^{M}
    \mathbf{\Sigma}_{X,j}^2.
    \label{eq: kZ efficient sparsity trace}
\end{alignat}

\subsection{Efficient Gradient Evaluation}
The gradient expressions above never require the dense $N\times N$ matrix $\widetilde{\mathbf R}_{XX}^{-1}$ to be formed, and neither $\widetilde{\C}_{XX}$ nor $\widetilde{\C}_{YY}$ is factorized. The Cholesky factorization is retained only for $\widetilde{\K}_{ZZ}$; all endpoint inverse actions are evaluated from the singular value decompositions of $\mathbf\Psi_{ZX}$ and $\mathbf\Psi_{ZY}$. Within each objective-and-gradient evaluation the following are computed once and reused across all scalar gradient components:

\begin{enumerate}[label=(\roman*), leftmargin=2em]
    \item the Cholesky factor $\Lb_{ZZ}$ of $\widetilde{\K}_{ZZ}$, the whitened matrices $\mathbf\Psi_{ZX}=\Lb_{ZZ}^{-1}\K_{ZX}$ and $\mathbf\Psi_{ZY}=\Lb_{ZZ}^{-1}\K_{ZY}$, and their thin singular value decompositions $\mathbf\Psi_{ZX}=\mathbf{M}_X\mathbf{\Sigma}_X\mathbf{N}_{X}^{\top}$ and $\mathbf\Psi_{ZY}=\mathbf{M}_Y\mathbf{\Sigma}_Y\mathbf{N}_{Y}^{\top}$;
    \item the endpoint inverse-square-root actions of~\eqref{eq: kZ SVD endpoint actions}, the whitened Koopman matrix $\U'$, and $\mathcal S=\lVert\U'\rVert_F^2$ from~\eqref{eq: kZ whitened score}; the singular values of $\mathbf\Psi_{ZX}$ also supply $\mathcal L_{\mathrm c}$ through~\eqref{eq: kZ efficient complexity} and $\mathcal L_{\mathrm s}$ through~\eqref{eq: efficient sparsity error};
    \item the matrices $\bm\Gamma_X$ and $\bm\Gamma_Y$ from~\eqref{eq: kZ Gamma XY} and $\mathbf A_X$, $\mathbf A_Y$ from~\eqref{eq: kZ SVD AX AY}; $\mathbf T_{XY}$ is an algebraic device for the differential and is not formed numerically;
    \item the sparse weights $\W_{ZX}=\widetilde{\K}_{ZZ}^{-1}\K_{ZX}$, required by the master Nyström differential~\eqref{eq: kZ master Nystrom differential} and by the $\sigma_Z$ derivatives~\eqref{eq: kZ sigmaZ term derivatives}.
\end{enumerate}

The total cost of this preparation remains $\mathcal O(M^2N+M^3)$. Each scalar gradient component is then a contraction of $\mathbf A_Y$, $\mathbf A_X$, $\bm\Gamma_X$, and $\bm\Gamma_Y$ with the corresponding kernel derivatives, at $\mathcal O(MN)$ apiece. All of the above must be recomputed whenever the optimized kernel parameters, pseudo-inputs, or $\sigma_Z$ change; the flow-map noise estimates $\sigma_Y$ and $\sigma_X$ remain fixed throughout the lifted optimization stage.

\end{appendices}
\end{document}